\documentclass[BTech]{SRM_Thesis}
\usepackage{t1enc}
\usepackage{tikz}
\usepackage{subfigure}
\usepackage{pgfplots}
\usepackage{setspace} 
\usepackage{geometry}
\usepackage{graphicx} 
\usepackage[section]{placeins}
\graphicspath{{./Figures/}} 
\usepackage{xcolor} 
\usepackage{epstopdf}
\usepackage{lscape}
\usepackage{fancyhdr}
\usepackage{natbib} 
\usepackage[hyphens]{url} 
\usepackage{hyperref} 
\usepackage{amsmath} 
\usepackage[ruled,vlined]{algorithm2e} 
\usepackage{listings} 
\usepackage{amssymb}
\usepackage{wasysym}
\usepackage{titlesec}
\usepackage{textcomp}
\usepackage{pifont}
\usepackage{appendix}
\pdfmapfile{=tfrupee.map} 
\usepackage{tfrupee} 
\usepackage{indentfirst} 
\usepackage[printonlyused]{acronym} 
\usepackage{multirow} 
\usetikzlibrary{decorations.pathmorphing}
\usetikzlibrary{shapes,arrows,shadows,patterns}
\DeclareMathOperator*{\argmin}{\arg\!\min}
\DeclareMathOperator*{\argmax}{\arg\!\max}

\begin{document}

\pagenumbering{roman}


\title{AutoDRIVE -- An Integrated Platform for Autonomous Driving Research and Education} 
\firstauthor{Tanmay Alias Manjeet Vilas Samak}
\firstauthorregno{[RA1711018010101]} 
\secondauthor{Chinmay Alias Manmeet Vilas Samak}
\secondauthorregno{[RA1711018010102]}
\thirdauthor{}
\thirdauthorregno{}
\fourthauthor{}
\fourthauthorregno{}
\fifthauthor{}
\fifthauthorregno{}
\guide{Mr. K. Sivanathan, M.E., (Ph.D.)} 
\designation{Assistant Professor (Sr.G.)} 
\guidedepartment{Mechatronics Engineering} 
\hod{Dr. G. Murali, M.E., Ph.D.} 
\department{Mechatronics Engineering} 
\date{MAY 2021} 
\maketitle
\clearpage


\certificate
\clearpage


\abstract
{\normalsize
	
This work presents AutoDRIVE, a comprehensive research and education platform for implementing and validating intelligent transportation algorithms pertaining to vehicular autonomy as well as smart city management. It is an openly accessible platform featuring a 1:14 scale car with realistic drive and steering actuators, redundant sensing modalities, high-performance computational resources, and standard vehicular lighting system. Additionally, the platform also offers a range of modules for rapid design and development of the infrastructure. The AutoDRIVE platform encompasses Devkit, Simulator and Testbed, a harmonious trio to develop, simulate and deploy autonomy algorithms. It is compatible with a variety of software development packages, and supports single as well as multi-agent paradigms through local and distributed computing. AutoDRIVE is a product-level implementation, with a vast scope for commercialization. This versatile platform has numerous applications, and they are bound to keep increasing as new features are added. This work demonstrates four such applications including autonomous parking, behavioural cloning, intersection traversal and smart city management, each exploiting distinct features of the platform.
}

\pagebreak
\clearpage


\acknowledgement
{\normalsize

At the commencement of this report, we would like to add a few words of appreciation for all those who have been a part of this project, directly or indirectly.

We thank the Department of Mechatronics Engineering, SRM Institute of Science and Technology for providing all facilities and support to meet our project requirements.

Our sincere thanks also go to our Head of the Department, Dr. G. Murali for giving us a big helping hand and providing us with an excellent atmosphere for working on this project.

We would also like to express our deep and sincere gratitude to our project guide Mr. K. Sivanathan for his valuable guidance, consistent encouragement, personal caring and timely help. All through the work, in spite of his busy schedule, he has extended cheerful and cordial support to us for completing this work and making this project a success. His guidance helped us in all the time of project and writing of this report. We could not have imagined having a better advisor and mentor for this project.

We are also thankful to the review panel for taking out their precious time to evaluate this project. Their critical feedback and remarks throughout the reviews have definitely helped in enriching the quality of this project.

Finally, we are grateful to all the sources of information without which this project would be incomplete. It is due to their efforts and research that our report is more accurate and convincing.

\begin{flushright}
{\bf Authors}
\end{flushright}

}

\clearpage


\begin{spacing}{1.15}
\tableofcontents
\addtocontents{toc}{\protect\thispagestyle{empty}}
\thispagestyle{empty}
\end{spacing}
\clearpage


\listoftables
\addcontentsline{toc}{chapter}{LIST OF TABLES}
\clearpage


\listoffigures
\addcontentsline{toc}{chapter}{LIST OF FIGURES}
\clearpage


\abbreviations
\begin{spacing}{1.15}
	\begin{tabbing}
		xxxxxxxxxxxxx \= xxxxxxxxxxxxxxxxxxxxxxxxxxxxxxxxxxxxxxxxxxxxxxxx \kill 
		\textbf{e.g.} \> For example\\
		\textbf{i.e.} \> That is to say\\
		\textbf{viz.} \> Namely\\
		\textbf{w.r.t.} \> With respect to\\
		\textbf{a.k.a.} \> Also known as\\
		\textbf{ADAS} \> Advanced Driver-Assistance Systems\\
		\textbf{ODD} \> Operational Design Domain\\
		\textbf{LIDAR} \> Light Detection and Ranging Unit\\
		\textbf{RADAR} \> Radio Detection and Ranging Unit\\
		\textbf{SONAR} \> Sound Detection and Ranging Unit\\
		\textbf{GNSS} \> Global Navigation Satellite System\\
		\textbf{IMU} \> Inertial Measurement Unit\\
		\textbf{V2X} \> Vehicle to Everything Communication\\
		\textbf{ADSS} \> Autonomous Driving Software Stack\\
		\textbf{SCSS} \> Smart City Software Stack\\
		\textbf{ROS} \> Robot Operating System\\
		\textbf{RViz} \> ROS Visualization Tool\\
		\textbf{RQT Graph} \> ROS Computation Graph\\
		\textbf{API} \> Application Programming Interface\\
		\textbf{PID} \> Proportional-Integral-Derivative control\\
		\textbf{MPC} \> Model Predictive Control\\
		\textbf{CPU} \> Central Processing Unit\\
		\textbf{GPU} \> Graphics Processing Unit\\
		\textbf{MCU} \> Microcontroller Unit\\
		\textbf{RAM} \> Random Access Memory\\
		\textbf{HDRP} \> High-Definition Render Pipeline\\
		\textbf{GUI} \> Graphical User Interface\\
		\textbf{SCM} \> Smart City Manager\\
		\textbf{IoT} \> Internet of Things\\
		\textbf{V2I} \> Vehicle to Infrastructure Communication\\
		\textbf{RPM} \> Revolutions Per Minute\\
		\textbf{SLAM} \> Simultaneous Localization and Mapping\\
		\textbf{IRLS} \> Iteratively Reweighted Least Squares\\
		\textbf{AMCL} \> Adaptive Monte Carlo Localization\\
		\textbf{TEB} \> Timed-Elastic-Band\\
		\textbf{NLP} \> Non-Linear Program\\
		\textbf{NLPP} \> Non-Linear Programming Problem\\
		\textbf{ODE} \> Ordinary Differential Equation\\
		\textbf{CV} \> Computer Vision\\
		\textbf{RGB} \> Red, Green, Blue\\
		\textbf{ROI} \> Region of Interest\\
		\textbf{LUT} \> Look-Up Table\\
		\textbf{ML} \> Machine Learning\\
		\textbf{DIL} \> Deep Imitation Learning\\
		\textbf{CNN} \> Convolutional Neural Network\\
		\textbf{CONV} \> Convolutional Layer\\
		\textbf{FC} \> Fully Connected Layer\\
		\textbf{Sim2Real} \> Simulation to Real-World Transfer\\
		\textbf{PWM} \> Pulse Width Modulation\\
		\textbf{V2V} \> Vehicle to Vehicle Communication\\
		\textbf{DRL} \> Deep Reinforcement Learning\\
		\textbf{POMDP} \> Partially Observable Markov Decision Process\\
		\textbf{FCNN} \> Fully Connected Neural Network\\
		\textbf{PPO} \> Proximal Policy Optimization\\
		\textbf{BOM} \> Bill of Materials\\
	\end{tabbing}
\end{spacing}

\pagebreak
\clearpage


\chapter*{\centerline{LIST OF SYMBOLS}}
\addcontentsline{toc}{chapter}{LIST OF SYMBOLS}

\begin{spacing}{1.15}
	\begin{tabbing}
		xxxxxxxxxxxxxxxxx \= xxxxxxxxxxxxxxxxxxxxxxxxxxxxxxxxxxxxxxxxxxxxxxxx \kill 
		\textbf{Section \ref{Sub-Section: Odometry}}\\
		\textbf{$t$} \> Time instant\\
		\textbf{$\Delta t$} \> Time interval\\
		\textbf{$L$} \> Registered laser scan\\
		\textbf{$L_i$} \> Indexed laser scan w.r.t. $t$\\
		\textbf{$L_i^w$} \> Warped indexed laser scan\\
		\textbf{$\alpha$} \> Scan coordinate\\
		\textbf{$N$} \> Scan size\\
		\textbf{$P$} \> Arbitrary point observed in laser scan\\
		\textbf{$P'$} \> Apparently displaced point $P$ as observed in consecutive laser scan\\
		\textbf{$r$} \> Range coordinate of point $P$ from LIDAR reference frame\\
		\textbf{$\theta$} \> Angular coordinate of point $P$ from LIDAR reference frame\\
		\textbf{$[\dot{r},\dot{\theta}]$} \> Velocity of point $P$ in polar coordinates\\
		\textbf{$[\dot{x},\dot{y}]$} \> Cartesian velocity of point $P$\\
		\textbf{$f$} \> LIDAR field of view\\
		\textbf{$\xi_i=[v_{x,i},v_{y,i},\omega_i]$} \> 2D twist representing velocity of object $i$\\
		\textbf{$\rho(\xi)$} \> Geometric residual for given twist $\xi$\\
		\textbf{$k$} \> Tunable parameter\\
		\textbf{$F(\rho)$} \> Cauchy M-estimator\\
		\textbf{$w(\rho)$} \> Associated weights of $F(\rho)$\\
		\\
		\textbf{Section \ref{Sub-Section: SLAM}}\\
		\textbf{$P_m$} \> Map coordinate\\
		\textbf{$M(P_m)$} \> Occupancy probability\\
		\textbf{$\nabla M(P_m)$} \> Spatial gradient in x and y direction\\
		\textbf{$\xi=[p_x,p_y,\psi]^T$} \> Vehicle pose vector\\
		\textbf{$\Delta\xi$} \> Vehicle pose change\\
		\textbf{$S_i(\xi)$} \> Coordinates of a scanned point $s_i=[s_{i,x},s_{i,y}]^T$\\
		\\
		\textbf{Section \ref{Sub-Section: Localization}}\\
		\textbf{$t$} \> Time instant\\
		\textbf{$S$} \> Sample (particle) set\\
		\textbf{$n$} \> Number of samples\\
		\textbf{$w$} \> Importance weights of samples\\
		\textbf{$\alpha$} \> Weight normalization factor\\
		\textbf{$x$} \> Vehicle state\\
		\textbf{$Bel(x)$} \> Belief over vehicle state\\
		\textbf{$z$} \> Measurement/observation\\
		\textbf{$u$} \> Control inputs\\
		\textbf{$\delta$} \> Upper quantile bound\\
		\textbf{$\varepsilon$} \> K-L distance bound\\
		\textbf{$b$} \> Bin\\
		\textbf{$\Delta$} \> Bin size\\
		\textbf{$k$} \> Number of bins with non-zero probability\\
		\textbf{$\chi^2_k$} \> Chi-square distribution\\
		\\
		\textbf{Section \ref{Sub-Section: Navigation}}\\
		\textbf{$G(V,E)$} \> Graph (map representation)\\
		\textbf{$V$} \> Vertices of the graph\\
		\textbf{$E$} \> Edges of the graph\\
		\textbf{$start$} \> Source node on graph (current pose)\\
		\textbf{$end$} \> Goal node on graph (parking pose)\\
		\textbf{$open$\_$list$} \> List of nodes to be traversed\\
		\textbf{$closed$\_$list$} \> List of nodes already traversed\\
		\textbf{$g(n)$} \> Cost to $n$-th node (from start node)\\
		\textbf{$h(n)$} \> Cost estimate from $n$-th node (to terminal node)\\
		\textbf{$f(n)$} \> Current cost of optimal path through node $n$\\
		\textbf{$t$} \> Continuous time instant\\
		\textbf{$k$} \> Discrete time instant\\
		\textbf{$\Delta T_k$} \> Time discretization\\
		\textbf{$L$} \> Vehicle wheelbase\\
		\textbf{$x$} \> Positional x-coordinate of vehicle\\
		\textbf{$y$} \> Positional y-coordinate of vehicle\\
		\textbf{$\psi$} \> Vehicle yaw\\
		\textbf{$s_k$} \> Vehicle pose at $k^{th}$ instant\\
		\textbf{$u_k$} \> Vehicle controls at $k^{th}$ instant\\
		\textbf{$u^*_k$} \> Optimal controls for vehicle at $k^{th}$ instant\\
		\textbf{$\mathcal{B}$} \> Path solution\\
		\textbf{$\mathcal{B^*}$} \> Optimal path solution\\
		\textbf{$\mathcal{O}$} \> Obstacle set\\
		\textbf{$v$} \> Longitudinal control command for vehicle\\
		\textbf{$\delta$} \> Lateral control command for vehicle\\
		\\
		\textbf{Section \ref{Sub-Section: Data Augmentation}}\\
		$\texttt{src}$ \> Source image\\
		$\texttt{dst}$ \> Destination (processed) image\\
		\textbf{$i$} \> Image pixel x-coordinate\\
		\textbf{$j$} \> Image pixel y-coordinate\\
		\textbf{$\theta$} \> Vehicle steering angle (rad)\\
		\textbf{$M_T$} \> Image translation matrix for panning operation\\
		\textbf{$M_R$} \> Image rotation matrix for tilting operation\\
		\\
		\textbf{Section \ref{Sub-Section: Data Preprocessing}}\\
		\textbf{$\delta$} \> Normalized steering command\\
		\textbf{$\varphi$} \> Vehicle steering limit (rad)\\
		\\
		\textbf{Section \ref{Sub-Section: DIL Deployment}}\\
		\textbf{$\eta$} \> Vehicle autonomy metric (\%)\\
		\textbf{$t_{lap}$} \> Lap time\\
		\textbf{$t_{int}$} \> Intervention time\\
		\textbf{$n_{int}$} \> Intervention count\\
		\\
		\textbf{Section \ref{Sub-Section: Problem Formulation}}\\
		\textbf{$S$} \> State space\\
		\textbf{$A$} \> Action space\\
		\textbf{$T$} \> Conditional state transition probability\\
		\textbf{$R$} \> Expected immediate reward\\
		\textbf{$\Omega$} \> Conditional observation probability\\
		\textbf{$O$} \> Observation space\\
		\textbf{$\gamma$} \> Discount factor\\
		\textbf{$t$} \> Time instant\\
		\textbf{$s_t$} \> State at $t$-th time instant\\
		\textbf{$o_t$} \> Observation at $t$-th time instant\\
		\textbf{$a_t$} \> Action at $t$-th time instant\\
		\textbf{$r_{t}$} \> Extrinsic reward at $t$-th time instant\\
		\textbf{$\pi_\theta$} \> Policy $\pi$ with parameters $\theta$\\
		\\
		\textbf{Section \ref{Sub-Section: DRL Training}}\\
		\textbf{$\alpha$} \> Learning rate\\
		\textbf{$\beta$} \> Entropy regularization strength\\
		\textbf{$\epsilon$} \> Policy update hyperparameter\\
		\textbf{$\lambda$} \> Regularization parameter\\
		\\
		\textbf{Section \ref{Sub-Section: Planning}}\\
		\textbf{$\mathcal{M}$} \> HD map\\
		\textbf{$\mathcal{X}$} \> Vehicle(s) states\\
		\textbf{$\mathcal{L}$} \> Traffic light set\\
		\textbf{$\mathcal{S}$} \> Traffic sign set\\
		\textbf{$\tau$} \> Normalized throttle command\\
		\textbf{$\delta$} \> Normalized steering command
	\end{tabbing}
\end{spacing}

\pagebreak
\clearpage


\pagenumbering{arabic}


\chapter{INTRODUCTION}
\label{Chapter: Introduction}

\section{Autonomous Driving}
\label{Section: Autonomous Driving}
	
	Autonomous driving \cite{AutonomousVehicles2020} is a rather complex engineering problem requiring multidisciplinary expertise, and is not something that can be declared as \textit{``solved''} any day soon; especially since the ideal operational design domain (ODD) for an autonomous vehicle is quite diverse, ranging from freely drivable highways, to densely crowded urban traffic, or even unstructured environments with unmarked roads. An immaculate realization of this technology shall, therefore, mark a significant step in the field of engineering and technology.
	
	\subsection{Modular Approach}
	\label{Sub-Section: Modular Approach}
	
		\begin{figure}[htpb]
		\centering
		\includegraphics[width=\textwidth]{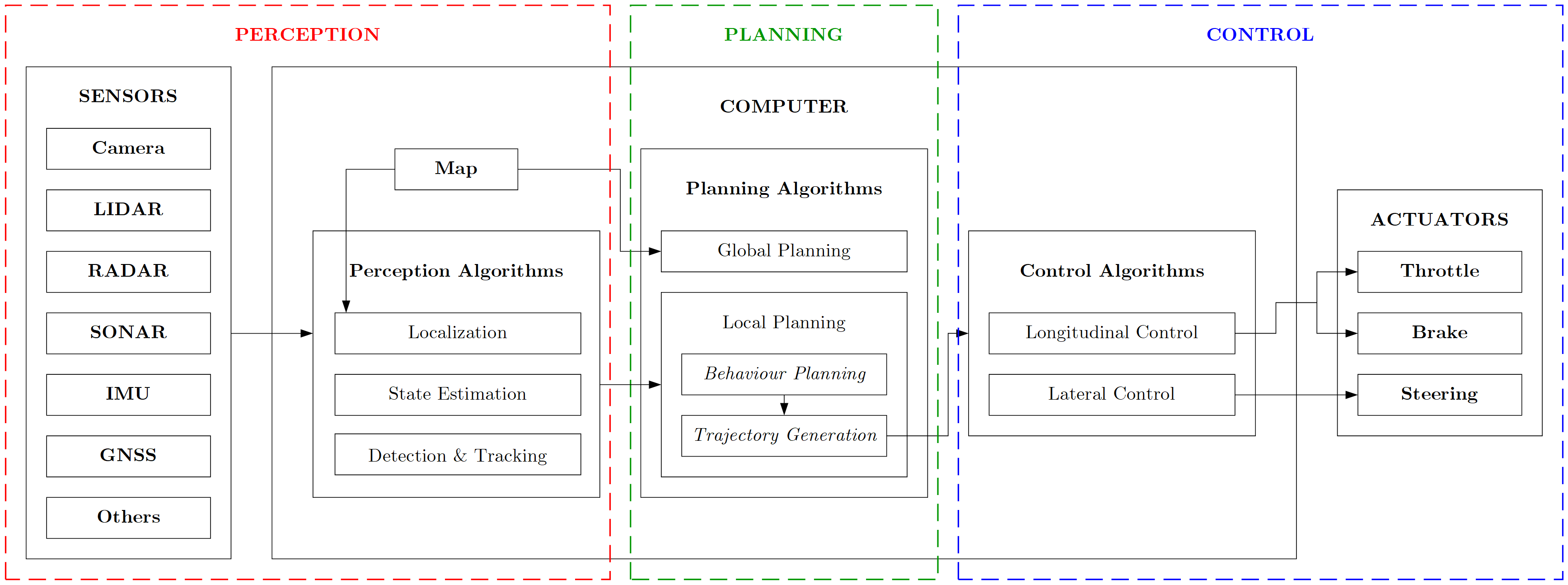}
		\caption{Modular Autonomous Driving Architecture}
		\label{Figure: Modular Autonomous Driving Architecture}
		\end{figure}
		
		Traditionally, a typical autonomous vehicle is considered as a steered non-holonomic mobile robot. Consequently, it uses most of the technology employed for autonomous mobile robots, with appropriate modifications to suite its kinodynamic constraints. This is known as the \textbf{modular} approach, which comprises of the \textit{perception}, \textit{planning} and \textit{control} modules stacked together, as depicted in Figure \ref{Figure: Modular Autonomous Driving Architecture}.
			
		The very first step in modular approach is \textit{perception}, wherein the vehicle observes the environment along with its own intrinsic parameters using exteroceptive sensors (like camera, LIDAR, RADAR, SONAR, etc.) and proprioceptive sensors (like GNSS, IMU, speedometer, odometer, tachometer, etc.) respectively. This is followed by processing the raw sensory data, which may include amplification, thresholding, filtering, and other algorithms, so as to reduce the noise and uncertainty associated with them, and then interpret state of the vehicle and/or environment using the extracted information (known as state estimation). This step of information extraction and interpretation often requires knowledge (in form of a database) of the sensor calibration matrices, previous vehicle states and control inputs (if any). One of the key challenges in perception module is localization, which is the notion of estimating pose of the vehicle w.r.t its environment (often probabilistically), given a map of that environment, vehicle’s previous (or initial) state and control inputs, wherein the map (occupancy grid, feature map, sparse/dense landmark map etc.), previous vehicle states and control inputs are stored in a knowledge database. Lastly, the vehicle also detects and tracks any static/dynamic objects in the scene, which is extremely important for dynamic re-planning and online decision making.
		
		The next step is to use some sort of cognition so as to plan a safe and efficient path form the source to destination. This is termed as \textit{planning}, which includes two steps viz. global planning (i.e. planning a route from source to destination w.r.t. map as per the mission plan, generally using shortest-path algorithms based on search heuristics) and local planning (i.e. generating an optimal trajectory online considering a predefined cost function accounting for static/dynamic obstacles, motion constraints, actuator saturation limits, acceleration and jerk limits, etc.). The motion planner ultimately generates a reference trajectory over a certain time-horizon to be tracked by the vehicle.
		
		The final step in modular autonomous driving architecture is \textit{motion control} \cite{ControlStrategies2020}. This mainly requires generating adequate control commands for the actuators (throttle, brake, steering) for executing the planned trajectory (path execution). Longitudinal (throttle and brake) and lateral (steering) controllers may be implemented in a coupled or de-coupled fashion. Several traditional controllers such as PID, Pure-Pursuit, Stanley, MPC, etc. can be adopted in order to accomplish this task. The controller drives the actuators so that the vehicle executes a controlled motion within the environment.
	
	\subsection{End-to-End Approach}
	\label{Sub-Section: End-to-End Approach}
	
		\begin{figure}[htpb]
			\centering
			\includegraphics[width=\textwidth]{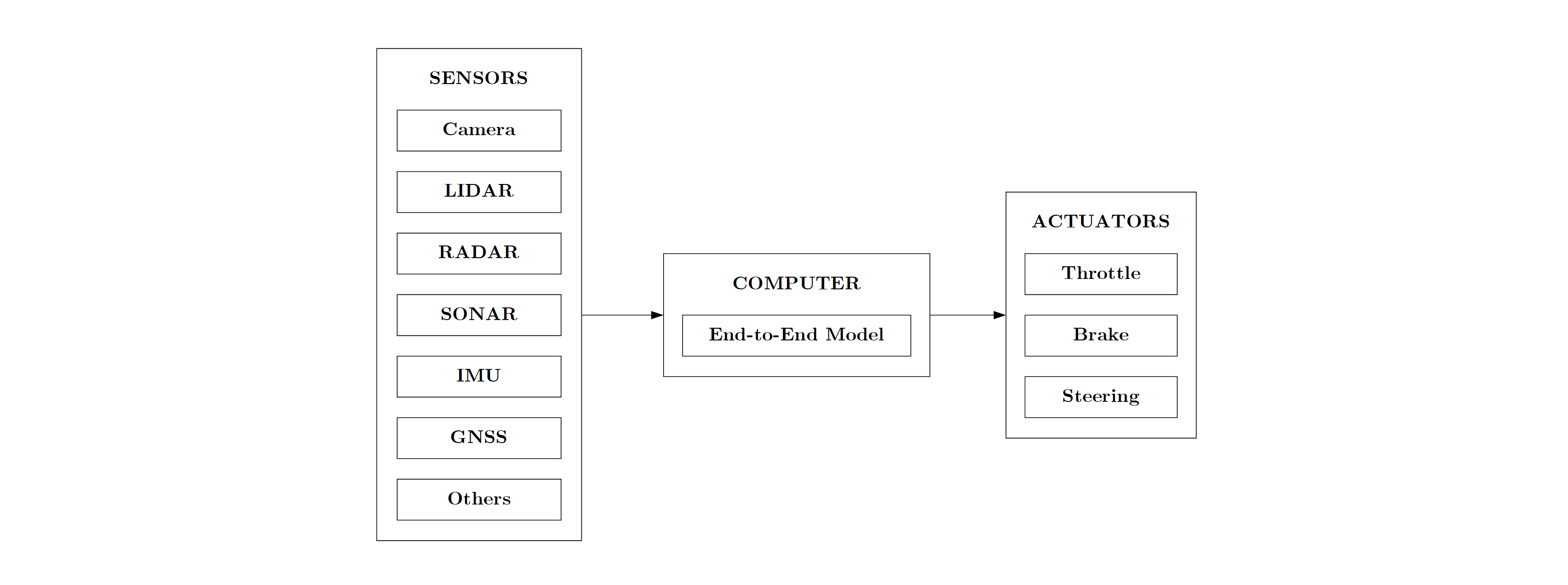}
			\caption{End-to-End Autonomous Driving Architecture}
			\label{Figure: End-to-End Autonomous Driving Architecture}
		\end{figure}
		
		More recently, \textbf{end-to-end} strategies are being developed to generate motion control commands directly based on the sensory inputs, as depicted in Figure \ref{Figure: End-to-End Autonomous Driving Architecture}. Compared to the modular pipeline, this approach significantly reduces the latency of an autonomous driving software stack and yields an enhanced real-time performance. However, as of today, the only known way to implement such end-to-end algorithms is by exploiting deep learning techniques to train a neural network model to directly map the sensory measurements to control commands by approximating the traditional perception, planning and control modules, which cannot be theoretically guaranteed to work flawlessly all the time. Furthermore, a lot of effort is required to fine tune the training parameters in order to achieve near-perfect non-linear approximation and impart sufficient robustness to the models.

\section{Presenting AutoDRIVE}
\label{Section: Presenting AutoDRIVE}

\begin{figure}[htpb]
		\centering
		\includegraphics[width=0.75\textwidth]{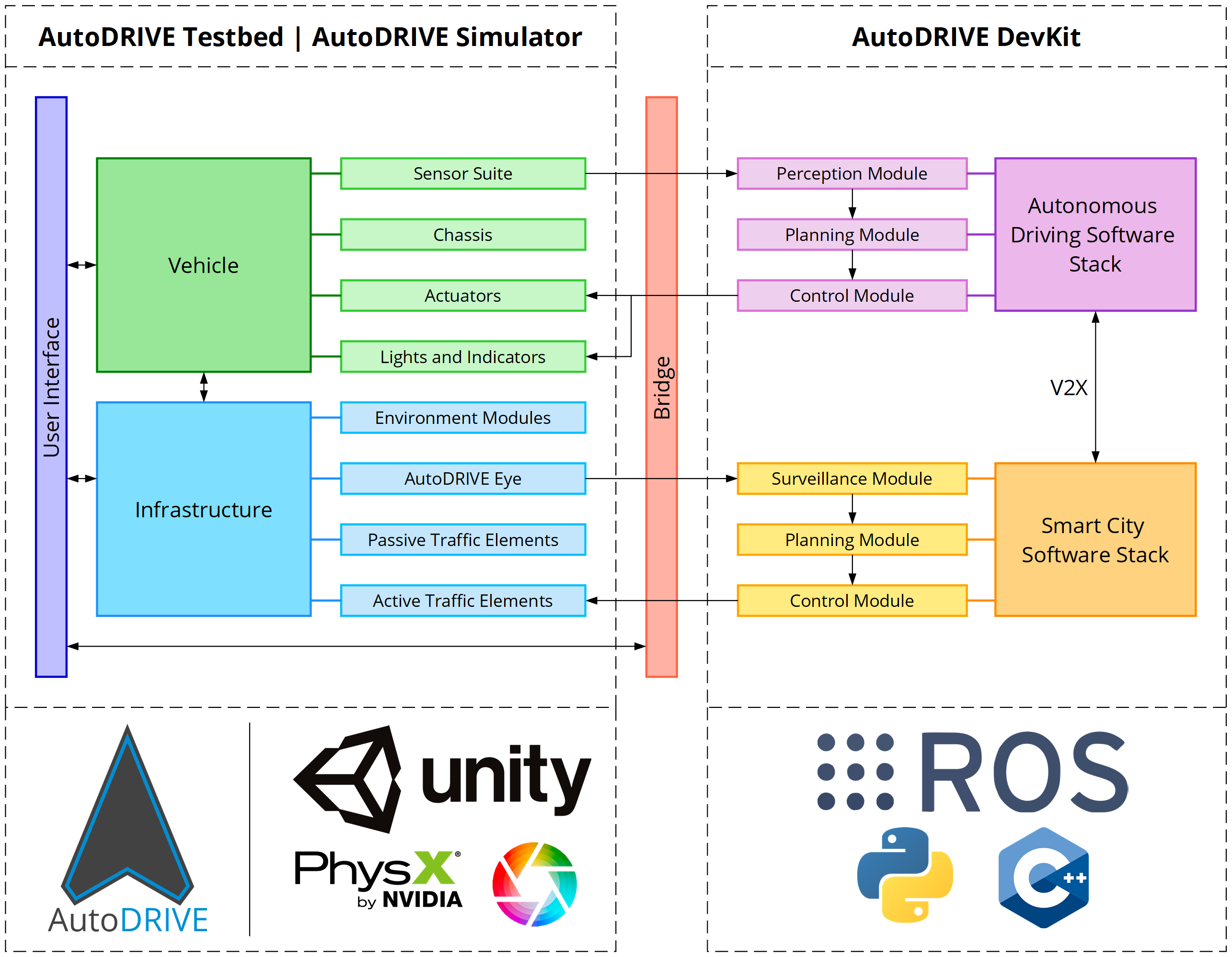}
		\caption{Proposed System Architecture of the AutoDRIVE Platform}
		\label{Figure: Proposed System Architecture}
	\end{figure}

	AutoDRIVE is developed to be a comprehensive platform for research and education pertaining to scaled self-driving cars. As depicted in Figure \ref{Figure: Proposed System Architecture}, the platform is primarily targeted towards autonomous driving, while also supporting development of smart-city solutions. It provides AutoDRIVE Simulator and Testbed, a well-suited duo to bridge the gap between software simulation and hardware deployment, and offers AutoDRIVE Devkit for flexible development of intelligent transportation algorithms.
	
	AutoDRIVE Testbed is a hardware platform for prototyping and testing autonomous driving solutions. It comprises of a 1:14 scale car model equipped with realistic drive and steering actuators, redundant sensing modalities, high-performance computational resources, and standard vehicular lighting system. The vehicle is designed to suit both manual as well as autonomous driving applications. Additionally, the testbed provides a modular infrastructure development kit, comprising of various terrain modules, road kits, obstruction modules and traffic elements, which can be used to design a variety of driving scenarios. It also offers the AutoDRIVE Eye to monitor the entire scene in real-time.
	
	AutoDRIVE Simulator can be considered as a digital twin of the AutoDRIVE Testbed. It is primarily meant for virtually prototyping autonomy algorithms as a part of the recursive simulation-validation cycle, or due to any limitations prohibiting the use of hardware testbed. It is developed using Unity game engine, thereby offering accurate physics simulation at the back-end by exploiting NVIDIA’s PhysX Engine, and realistic graphics rendering at the front-end by exploiting Unity’s High-Definition Render Pipeline and Post-Processing Stack. It provides an interactive user interface for ease of access to various functionalities, and offers a bi-directional communication bridge to link with autonomy algorithms developed by the users. The standalone simulator application offers cross-platform support (Windows, macOS and Linux), enabling convenient installation.
	
	AutoDRIVE Devkit provides a variety of software packages and tools, for rapid development of autonomy algorithms pertaining to autonomous driving and/or smart city management, thereby allowing the users to flexibly exploit AutoDRIVE Testbed and/or AutoDRIVE Simulator. It supports both local and distributed computing, thereby offering the flexibility to develop both centralized and decentralized autonomy algorithms. It is primarily compatible with the Robot Operating System (ROS), which supports development of modular autonomy algorithms. Additionally, it also offers scripting application programming interfaces (APIs) for Python and C++, which can be exploited to directly develop embedded scripts targeted towards autonomous driving. Finally, the Devkit also features a webapp for developing smart city solutions.
	
\section{Motivation}
\label{Section: Motivation}

	Development of autonomous systems is difficult and expensive beyond imagination. Specifically, working on full-scale autonomous vehicles is not feasible due to various factors including (but not limited to) monetary, spatial and safety constraints. As a result, almost all practitioners from academia (and some from industry) exploit down-scaled vehicles for hardware deployment and validation of their autonomy algorithms -- most of the leading technological research institutes can be seen developing their own scaled autonomous vehicles in the past decade. However, developing countries especially lack the resources and tools to develop such vehicles. Most of these vehicles, if not all, are derived from commercially available radio-controlled (RC) cars, which are not even available in developing countries, not to mention that they are very expensive.
	
	Additionally, most of the existing platforms available for such applications are observed to lack integrity in terms of hardware-software co-development, and hence neither follow nor foster the principles of mechatronics approach of system design. Some only offer software simulation tools, while others only provide physical scaled vehicles to test autonomy algorithms. Furthermore, some of these platforms are highly application specific with a limited sensor suite and stringent design requirements; some even lack an on-board computation device and are merely teleoperated from a remote computer to execute the intended mission.
	
	Another bottleneck is associated with the common practice of testing autonomy algorithms in simulation prior to the hardware deployment (mostly on scaled vehicles as discussed earlier) for validation. It originates from the fact that current research community lacks simulation tools for scaled vehicles, due to which, most students and researchers make use of full-scale autonomous vehicle simulators to validate their autonomy algorithms, following which they are deployed onto scaled autonomous vehicles. It is observed that transferring from simulation to real-world (sim2real) under such circumstances calls for significant additional effort in re-tuning the autonomy algorithms due to the fact that dynamic behaviour of a scaled vehicle is much different than its full-scale counterpart and simulators designed for the later fail to capture it successfully. Additionally, implementations pertaining to visual perception are extremely sensitive to variation in sensory data and often fail if transferred directly from simulation to real-world.
	
	AutoDRIVE takes all the above factors into consideration, and provides a comprehensive and integrated solution for implementing and testing autonomy algorithms in an expedited yet convenient way.

\section{Objectives}
\label{Section: Objectives}

	The primary objective of this project was to develop a comprehensive platform for furthering research and education in the field of autonomous driving. As described earlier, several factors were taken into account while designing AutoDRIVE in order to make it an openly accessible, integrated platform for scaled self-driving cars and related intelligent transportation applications. This called for the development of AutoDRIVE Devkit -- to flexibly develop autonomy algorithms, AutoDRIVE Simulator -- to virtually prototype them, and AutoDRIVE Testbed -- to deploy and validate them on physical hardware.
	
	The secondary objective of this project was to exploit the developed platform for implementing various intelligent transportation algorithms in order to demonstrate some of its prominent features and functionalities. This called for carefully designing several driving scenarios, and implementing autonomy algorithms employing distinct technologies; each targeted towards exploiting one or more unique features of the platform.

\clearpage


\chapter{LITERATURE SURVEY}
\label{Chapter: Literature Survey}

\section{Hardware Testbeds}
\label{Section: Hardware Testbeds}

	Self-driving car companies and startups like Waymo \cite{Waymo2021}, Zoox \cite{Zoox2021}, Tesla \cite{Tesla2021}, Cruise \cite{Cruise2021}, Aptiv \cite{Aptiv2021} and Baidu \cite{Apollo2021}, for instance, along with automotive giants moving into this field including Audi \cite{Audi2021}, BMW \cite{BMW2021}, Mercedes-Benz \cite{Mercedes-Benz2021}, Ford \cite{Ford2021}, Toyota \cite{Toyota2021}, Volvo \cite{Volvo2021} and Volkswagen \cite{Volkswagen2021}, and their partner companies like NVIDIA \cite{NVIDIA2021}, Uber ATG (now a subsidiary of Aurora \cite{Aurora2021}) and Lyft \cite{Lyft2021}, to name a few, have developed their own full-scale autonomous test vehicles. Some of these firms even have dedicated test facilities spread over several acers of land to mimic realistic traffic infrastructure in a controlled environment. However, only a handful like Udacity \cite{Udacity2021} and CETRAN \cite{CETRAN2021} allow third-party developers to access their vehicles and facilities; the former allowing a capstone project only to their nanodegree subscribers and the later providing services only to its partner organizations. In other words, these facilities are by no means openly accessible to students and researchers, not to mention they would charge a hefty sum to have autonomy algorithms tested on their vehicles.
	
	As a result, most of the educational and research institutes have started developing their own scaled autonomous vehicles since the past decade. Such vehicles include the MIT Racecar \cite{MIT-Racecar2021} by Massachusetts Institute of Technology (MIT), AutoRally \cite{AutoRally2021} by Georgia Institute of Technology (Georgia Tech), F1/10 \cite{F1102019} by University of Pennsylvania (UPenn), Delft Scaled Vehicle (DSV) \cite{DSV2017} by Delft University of Technology (TU Delft) and Multi-agent System for non-Holonomic Racing (MuSHR) \cite{MuSHR2019} by University of Washington (UW), to name a few. However, as described earlier, most of these, if not all, use commercially available RC cars as their foundation, which are very expensive and probably not even available in developing countries. Additionally, some of these vehicles, like the MIT Racecar, employ a really expensive sensor suite, which is generally an overkill for scaled autonomy applications. Some of the other small-scale autonomous vehicles include HyphaROS RaceCar \cite{HyphaROS-Racecar2021} and Donkey Car \cite{DonkeyCar2021}; both being application-specific vehicles for autonomous driving using map-based navigation and imitation learning, respectively. Again, both these vehicles are also based on commercially available RC cars.
	
	Another scaled platform for autonomy research, Duckietown \cite{Duckietown2017}, was primarily designed to be openly accessible and inexpensive. Its principal notion was making use of minimal hardware resources to achieve autonomy, thus reducing the cost and setup time. However, the fact that it utilizes differential-drive robots called Duckiebots, which in no way account for the kinodynamic constraints of a steered car-like vehicle, does not even qualify it as a self-driving car research platform, as they claim in their later versions. Nevertheless, it is a really good platform for teaching fundamentals of mobile robot autonomy, much like the TurtleBot \cite{Turtlebot2021}.
	
\section{Software Simulators}
\label{Section: Software Simulators}

	For several years, the automotive industry has made use of simulators like Ansys Automotive \cite{AnsysAutomotive2021} and Adams Car \cite{AdamsCar2021} to simulate vehicular dynamics at different levels, thereby accelerating the development of its end-products. Since the past few years, however, owing to the increasing popularity of advanced driver-assistance systems (ADAS) and autonomous vehicles, most of the traditional automotive simulators, such as Ansys Autonomy \cite{AnsysAutonomy2021}, CarSim \cite{CarSim2021} and CarMaker \cite{CarMaker2021}, have started releasing vehicular autonomy features in their updated versions.
	
	Apart from these, there are several commercial simulators that specifically target the field of autonomous driving. These include NVIDIA's Drive Constellation \cite{DRIVEConstellation2021}, Cognata \cite{Cognata2021}, rFpro \cite{rFpro2021}, dSPACE \cite{dSPACE2021} and PreScan \cite{PreScan2021}, to name a few. In the recent past, several research projects have also tried adopting computer games like GTA V \cite{Richter2016, Richter2017, Johnson-Roberson2017} in order to virtually simulate self-driving cars, but they were quickly shut down by the game's publisher.
	
	Nevertheless, the open-source community has also developed several simulators for such applications. Gazebo \cite{Gazebo2004}, a simulator natively adopted by ROS \cite{ROS2009}, has been used extensively by most of the scaled autonomous vehicles described earlier in Section \ref{Section: Hardware Testbeds}. TORCS \cite{TORCS2021}, another open-source simulator widely known in the self-driving community, is probably one of the earliest to specifically target manual and autonomous racing problems. Other prominent examples include CARLA \cite{CARLA2017}, AirSim \cite{AirSim2018} and Deepdrive \cite{Deepdrive2021} developed using the Unreal \cite{Unreal2021} game engine along with Apollo GameSim \cite{ApolloGameSim2021} and LGSVL Simulator \cite{LGSVLSimulator2020} developed using the Unity \cite{Unity2021} game engine.
	
	It is to be noted, however, that none of the aforementioned simulators, at their full capacity, provide support for scaled-down vehicles and/or infrastructure, which creates unpredictable problems and causes unnecessary delays in sim2real transfer onto scaled autonomous vehicles, as described in Section \ref{Section: Motivation}.

\clearpage


\chapter{PROJECT MANAGEMENT}
\label{Chapter: Project Management}

\section{Work Breakdown Structure}
\label{Section: Work Breakdown Structure}

\begin{figure}[htpb]
	\centering
	\includegraphics[width=\textwidth]{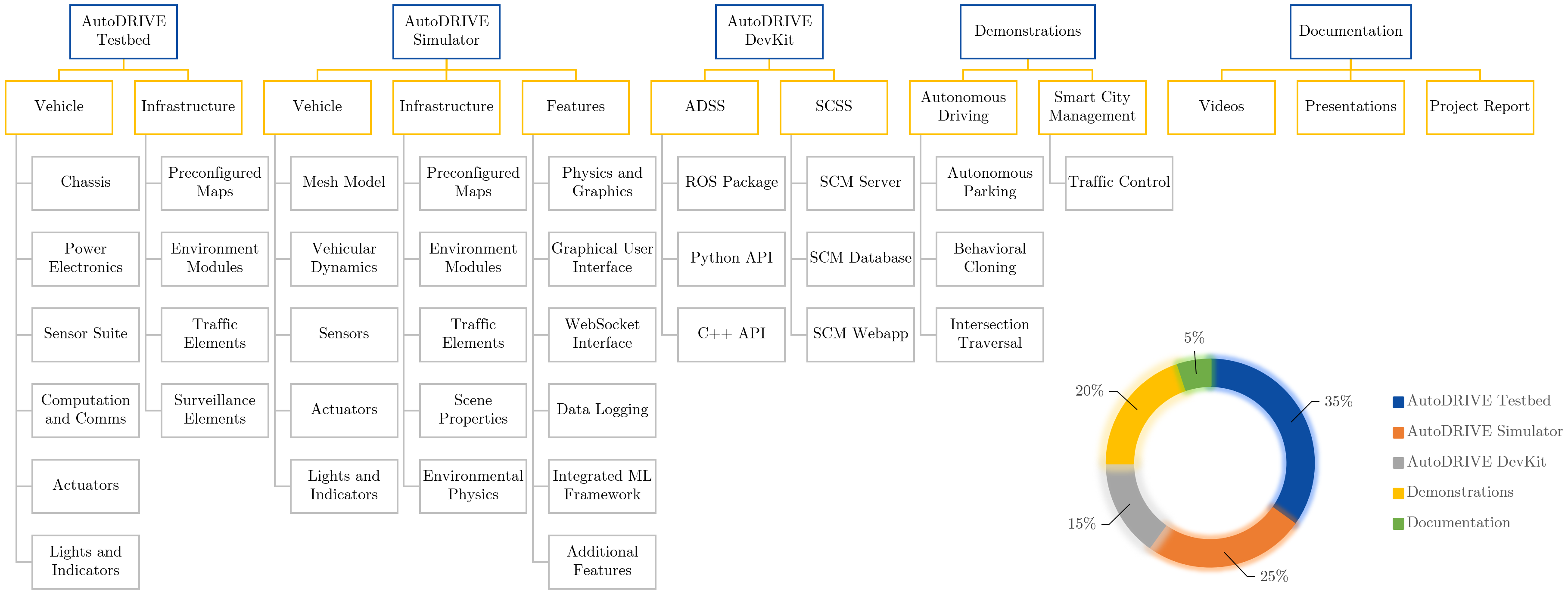}
	\caption{Work Breakdown Structure}
	\label{Figure: Work Breakdown Structure}
\end{figure}

As discussed in Section \ref{Section: Objectives}, we defined the project as a set of objectives (primary and secondary) to be accomplished. This followed dividing the project into smaller segments to be addressed as the project progressed. Figure \ref{Figure: Work Breakdown Structure} elucidates the work breakdown structure of this project.

The project initiated with platform development (primary objective), which was a major share ($\sim$75\%) of the entire project. Chapter \ref{Chapter: Platform Development} describes the platform development phase in detail, individually concentrating on development of AutoDRIVE Testbed (refer Section \ref{Section: AutoDRIVE Testbed}), AutoDRIVE Simulator (refer Section \ref{Section: AutoDRIVE Simulator}) and AutoDRIVE Devkit (refer Section \ref{Section: AutoDRIVE Devkit}). The developed platform was to be further exploited for the implementation of intelligent transportation algorithms pertaining to both vehicular autonomy as well as smart city management. Chapter \ref{Chapter: Platform Exploitation} describes the four applications being demonstrated as a part of this project. These include autonomous parking (refer Section \ref{Section: Autonomous Parking}), behavioural cloning (refer Section \ref{Section: Behavioural Cloning}), intersection traversal (refer Section \ref{Section: Intersection Traversal}) and smart city management (refer Section \ref{Section: Smart City Management}), each exploiting distinct features of the platform.

Finally, the project had to be documented at each phase. This included recording and editing demonstration videos, preparing and delivering review presentations taking place throughout the semester, and drafting of this final project report.

\section{Responsibility Assignment}
\label{Section: Responsibility Assignment}

Based on the work breakdown structure described in Section \ref{Section: Work Breakdown Structure}, each element of the project was assigned to either of the authors as a primary or secondary responsibility. Table \ref{Table: Responsibility Assignment Matrix} describes the responsibilities of each author.

\begin{table}[htpb]
	\centering
	\caption{Responsibility Assignment Matrix}
	\label{Table: Responsibility Assignment Matrix}
	\resizebox{0.70\textwidth}{!}{%
		\begin{tabular}{llcc}
			\hline
			\multicolumn{1}{c}{\multirow{2}{*}{\textbf{WBS ITEM}}} & \multicolumn{1}{c}{\multirow{2}{*}{\textbf{PROJECT COMPONENT}}} & \multicolumn{2}{c}{\textbf{RESPONSIBILITY$^{\mathrm{*}}$}} \\ \cline{3-4} 
			\multicolumn{1}{c}{}      & \multicolumn{1}{c}{}                                                            & \textbf{Tanmay} & \textbf{Chinmay}   \\ \hline
			\textbf{1}                                             & \multicolumn{3}{l}{\textbf{AutoDRIVE Testbed}}                                            \\ \hline
			1.1                                                    & Vehicle                                                  & P                   & S        \\
			1.2                                                    & Infrastructure                                           & S                   & P        \\ \hline
			\textbf{2}                                             & \multicolumn{3}{l}{\textbf{AutoDRIVE Simulator}}                                          \\ \hline
			2.1                                                    & Vehicle                                                  & P                   & S        \\
			2.2                                                    & Infrastructure                                           & S                   & P        \\
			2.3                                                    & Features                                                 & P                   & S        \\ \hline
			\textbf{3}                                             & \multicolumn{3}{l}{\textbf{AutoDRIVE Devkit}}                                             \\ \hline
			3.1                                                    & ADSS                                                     & S                   & P        \\
			3.2                                                    & SCSS                                                     & P                   & S        \\ \hline
			\textbf{4}                                             & \multicolumn{3}{l}{\textbf{Demonstrations}}                                               \\ \hline
			4.1                                                    & Autonomous Driving                                       & S                   & P        \\
			4.2                                                    & Smart City Management                                    & P                   & S        \\ \hline
			\textbf{5}                                             & \multicolumn{3}{l}{\textbf{Documentation}}                                                \\ \hline
			5.1                                                    & Videos                                                   & P                   & S        \\
			5.2                                                    & Presentations                                            & S                   & P        \\
			5.3                                                    & Project Report                                           & S                   & P        \\ \hline
			\multicolumn{4}{l}{$^{\mathrm{*}}$\textit{P} and \textit{S} denote primary and secondary responsibility respectively.}
		\end{tabular}%
	}
\end{table}

It is to be noted that both the authors contributed equally and worked on all aspects of the project collectively. The responsibility assignment only indicates the person in charge of a particular task, and that the authors have no conflict of interests.

\section{Project Execution Plan}
\label{Section: Project Execution Plan}

This project was academically associated with the course 15MH496L Major Project for the eighth semester of academic year 2020--21 at the Department of Mechatronics Engineering, School of Mechanical Engineering, SRM Institute of Science and Technology.

\begin{figure}[htpb]
	\centering
	\includegraphics[width=\textwidth]{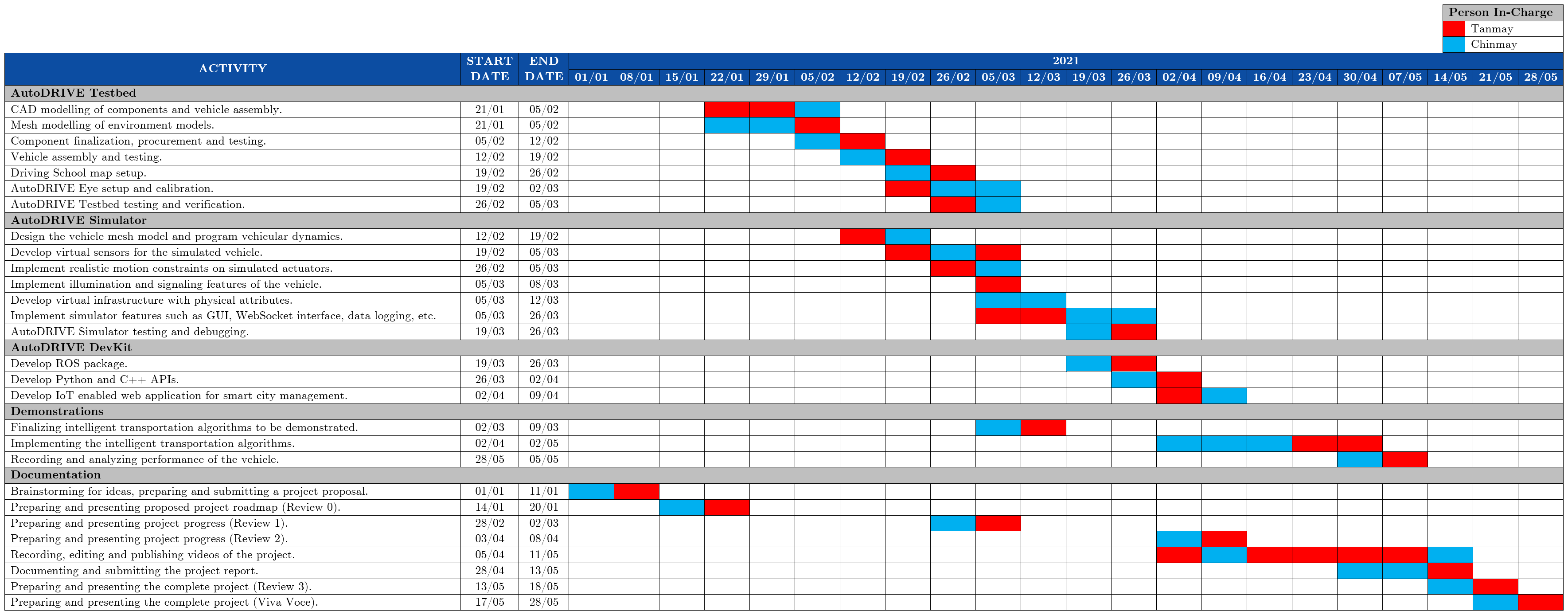}
	\caption{Project Execution Plan}
	\label{Figure: Project Execution Plan}
\end{figure}

Figure \ref{Figure: Project Execution Plan} depicts a Gantt chart describing  the project schedule throughout the semester in relation to the responsibility assignment (refer Section \ref{Section: Responsibility Assignment}). The project commenced in January 2021 and was completed in May 2021. We followed the road-map described in Section \ref{Section: Work Breakdown Structure} as far as possible, and at times even parallelized the work in order to enhance the project progress.

\clearpage


\chapter{PLATFORM DEVELOPMENT}
\label{Chapter: Platform Development}

\section{AutoDRIVE Testbed}
\label{Section: AutoDRIVE Testbed}

	As described earlier, AutoDRIVE Testbed is a physical setup for deploying and validating autonomy algorithms on a scaled vehicle (named Nigel) within a scaled infrastructural setup. Consequently, its development required designing, manufacturing, assembling and calibrating various hardware and software subsystems of the vehicle and infrastructure. The entire testbed was developed following the mechatronics approach of system design, paying key attention to the synergistic integration of mechanical, electrical and software components. The following sections describe vehicle and infrastructure development in a greater detail.
	
	\subsection{Vehicle Development}
	\label{Sub-Section: Vehicle Development}
	
		\begin{figure}[htpb]
			\centering
			\subfigure[]{\includegraphics[width=0.51\textwidth]{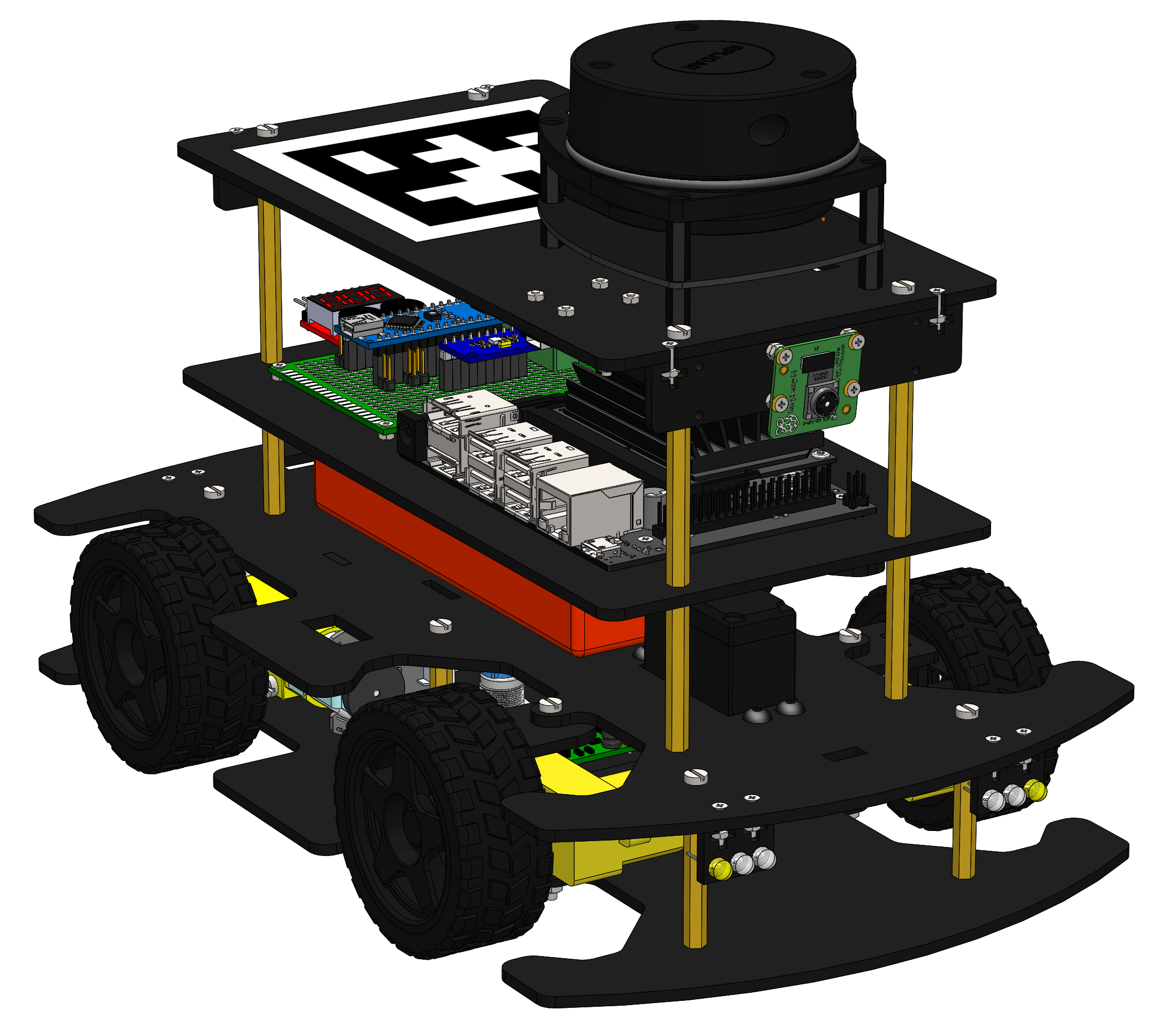}}
			\subfigure[]{\includegraphics[width=0.47\textwidth]{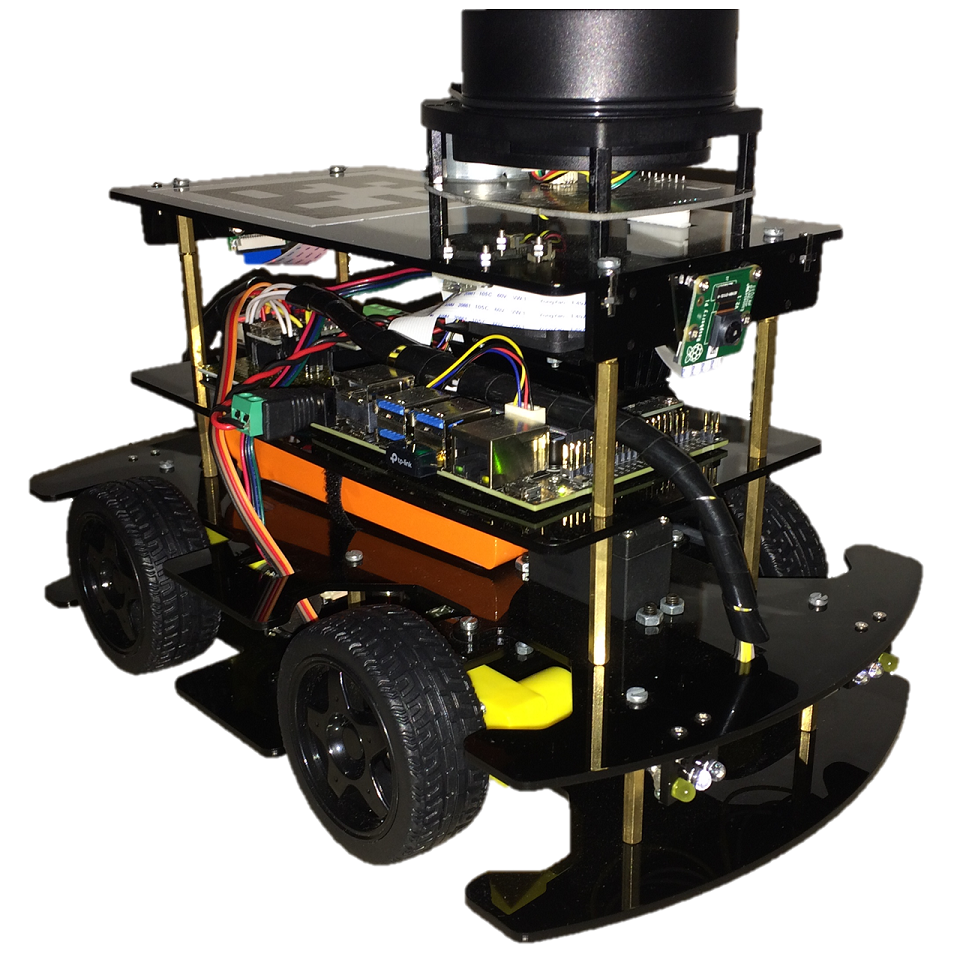}}
			\caption{Nigel: (a) CAD Assembly, and (b) Physical Prototype}
			\label{Figure: Nigel}
		\end{figure}
		
		The designed vehicle, Nigel (refer Figure \ref{Figure: Nigel}), comprises of four modular platforms, each housing distinct components and subsystems of the vehicle. The individual platforms are fabricated by laser cutting 3 mm thick acrylic sheets, which are stacked at the designed offsets employing brass spacers of appropriate length.
		
\begin{figure}[htpb]
			\centering
			\includegraphics[width=\textwidth]{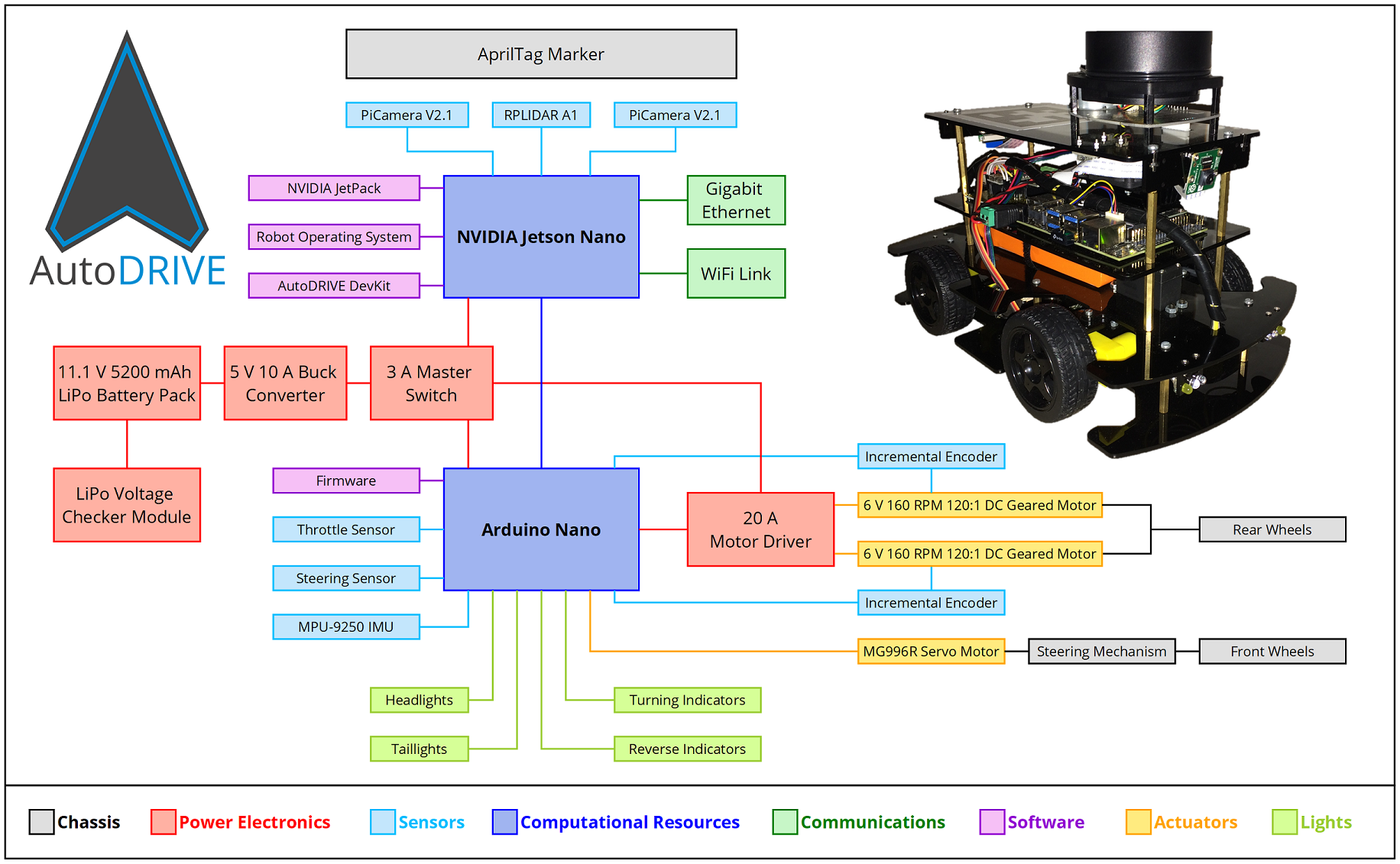}
			\caption{Hardware-Software Architecture of Nigel}
			\label{Figure: Vehicle System Architecture}
		\end{figure}

		The first (i.e. bottom-most) platform hosts the steering mechanism along with power electronics components including a master switch, buck converter and motor driver with inbuilt electrical braking functionality. The second platform mounts a servo actuator coupled with the steering mechanism on first platform, along with front and rear lighting units. It also has a provision for strapping a lithium polymer (LiPo) battery pack. Two DC geared motors driving the rear wheels and two free gearboxes coupled with the front wheels are fastened to their respective acrylic mounts fixed between the first and second platforms. The third platform accommodates an onboard computer, along with a custom auxiliary board and LiPo voltage checker module. The auxiliary board routes connections to the low-level microcontroller unit and takes care of power distribution as well. It also hosts a 9-axis IMU. Finally, the fourth platform mounts two CSI cameras and a planar LIDAR unit. It also has vacant space to paste an AprilTag marker so as to uniquely identify the vehicle. Figure \ref{Figure: Vehicle System Architecture} illustrates detailed system architecture of the vehicle.
		
		\subsubsection{Chassis}
		\label{Sub-Sub-Section: Vehicle Chassis}
		
\begin{figure}[htpb]
				\centering
				\includegraphics[width=\textwidth]{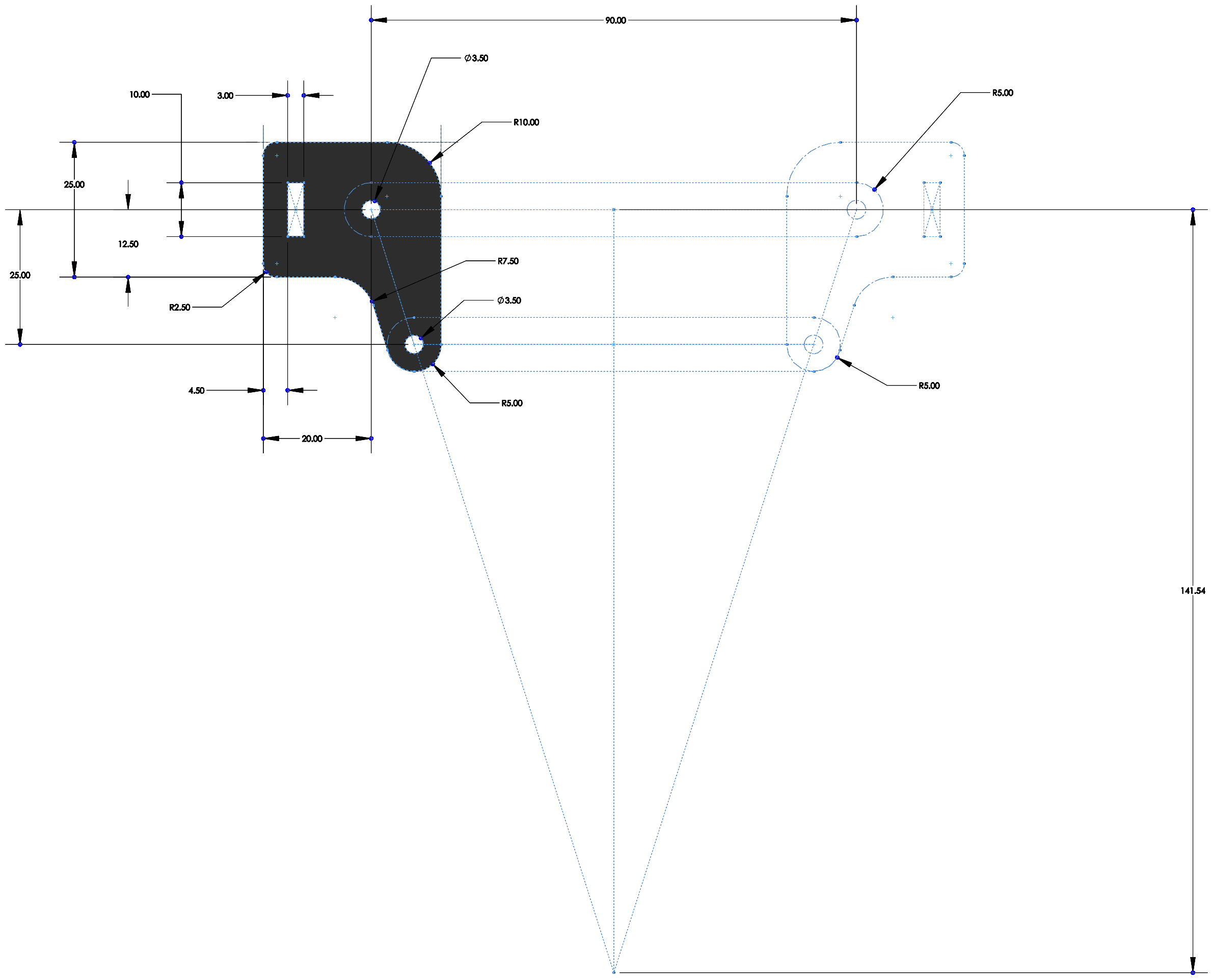}
				\caption{Nigel's Ackermann Steering Geometry}
				\label{Figure: Ackermann Steering Geometry}
			\end{figure}

			Nigel is a 1:14 scale model vehicle resembling an actual car in terms of kinodynamic constraints. It has two DC geared motors driving its rear wheels and an RC servo motor steering its front wheels. The vehicle adopts an Ackermann steering mechanism (refer Figure \ref{Figure: Ackermann Steering Geometry}) in order to avoid side-slippage of the tires while turning at nominal speeds for which the vehicle was designed.
			
		\subsubsection{Power Electronics}
		\label{Sub-Sub-Section: Vehicle Power Electronics}
		
			The vehicle is energized by an 11.1 V 5200 mAh LiPo battery pack. However, as with any other robot, this power needs to be managed effectively in order to ensure safe operation of all the electrical subsystems of the vehicle. First, the voltage is stepped down to 5 V using a 24 V 10 A rated constant current (CC), constant voltage (CV) buck converter. This regulated power is supplied, via a 12 V 3 A rated master switch, to all the electrical subsystems including the onboard computer (Jetson Nano Developer Kit), the auxiliary board and the 24 V 20 A rated motor driver module. The auxiliary board helps distribute this power efficiently.
			
			\begin{figure}[htpb]
				\centering
				\includegraphics[width=\textwidth]{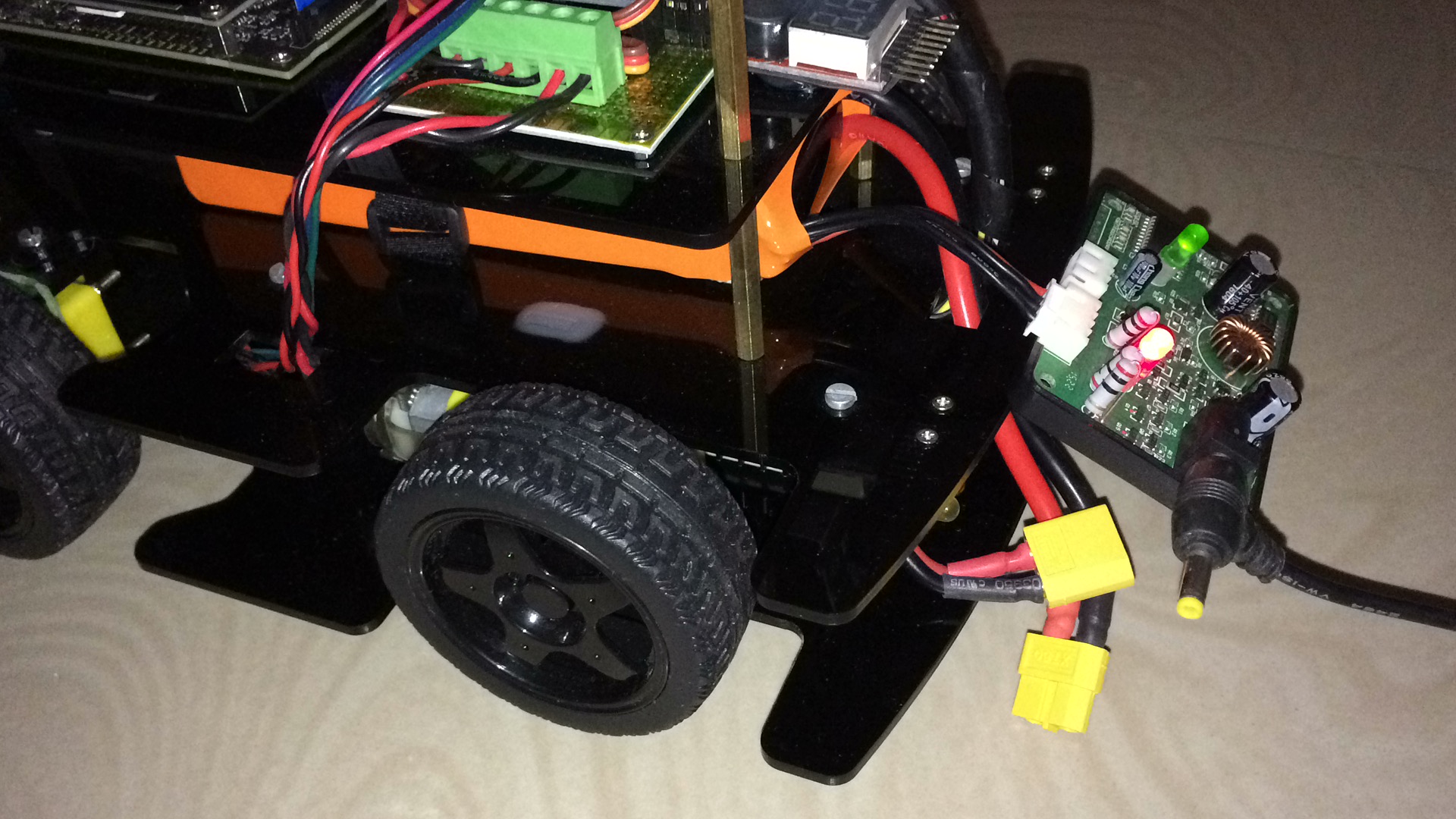}
				\caption{Nigel's Battery Charging Setup}
				\label{Figure: Vehicle Battery Charging}
			\end{figure}
			
			It is to be noted that working with LiPo battery packs, especially ones with capacity as high as that employed for this vehicle, is quite dangerous -- they can catch fire. Absolute care must be taken to avoid shorting of power and ground rails, and appropriate ventilation must be provided to all the power distribution components so as to regulate their temperatures at nominal values; passive and/or active thermal management systems must be installed wherever required. Additionally, charging and discharging cycles of the battery must be carefully monitored; a LiPo voltage checker module is provided for this purpose. Figure \ref{Figure: Vehicle Battery Charging} demonstrates the recommended battery-charging setup, using a balance charger module, with the battery pack disconnected.
		
		\subsubsection{Sensor Suite}
		\label{Sub-Sub-Section: Vehicle Sensor Suite}
		
			Nigel is provided with a comprehensive sessor suite comprising of throttle and steering feedbacks (virtual sensors), high resolution incremental encoders, 3-axis indoor positioning system (IPS), 9-axis inertial measurement unit (IMU), 360$^\circ$ planar light detection and ranging (LIDAR) unit as well as front and rear-view CSI cameras. Following is a brief summary of each sensor aboard the vehicle:
			
			\begin{itemize}
				\item \textbf{Throttle Feedback:} It is a feedback link that essentially records and keeps track of the latest drive command published to the vehicle. It is implemented as a functional block within the vehicle's firmware.
				\item \textbf{Steering Feedback:} It is a feedback link that essentially records and keeps track of the latest steering command published to the vehicle. It is also implemented as a functional block within the vehicle's firmware.
				\item \textbf{Wheel Encoders:} Both the rear-wheel drive motors are installed with hall-effect based magnetic encoders, each with a 16 pulses per revolution (PPR) resolution. They measure angular displacement of the respective motor shaft prior to gear reduction (120:1), and can therefore determine the angular displacement of the respective actuator's output shaft with a resolution as high as 1920 counts per revolution (CPR).
				\item \textbf{IPS:} The 95.25$\times$95.25 mm AprilTag markers present on the map as well as the vehicle can be detected and tracked using the AutoDRIVE Eye (refer Section \ref{Sub-Sub-Section: Surveillance Elements}). The AutoDRIVE Eye is also capable of estimating vehicle's 2D pose within the map by solving relative transformations between the vehicle-camera and map-camera subsystems. Unlike the large-scale GNSS, such pose estimates are sufficiently accurate even for indoor scenarios, but may suffer from coverage loss at times, similar to the GNSS counterparts. Consequently, this setup is termed as indoor positioning system (IPS).
				\item \textbf{IMU:} The auxiliary board hosts MPU-9250, a 9-axis inertial measurement unit (IMU). It is a microelectromechanical system (MEMS) based sensor comprising of MPU-6050, a combo of 3-axis accelerometer, 3-axis gyroscope and digital temperature sensor, along with AK8963, a 3-axis magnetometer. The vehicle's firmware is capable of parameterizing the sensor's register addresses, initializing it, and extracting properly scaled raw data from the accelerometer, gyroscope, and magnetometer. The said firmware can also directly provide a quaternion-based estimate of the absolute vehicle orientation; however, this requires appropriate calibration of the IMU.
				\item \textbf{LIDAR:} The fourth platform of the vehicle mounts RPLIDAR A1, a planar radial laser scanning unit. The said LIDAR unit can perform 360$^\circ$ scans around the vehicle at over 7 Hz update rate to detect objects within a span of 0.15 m and 12 m, with an angular resolution of 1$^\circ$.
				\item \textbf{Cameras:} The vehicle hosts two CSI camera modules (PiCamera V2.1), one for viewing the environment ahead of the vehicle, and other, that behind it. Alternatively, for stereo-vision and related applications, both the camera modules can be installed on the same front mount. The said camera modules are natively supported by the vehicle's onboard computer (Jetson Nano Developer Kit), thereby allowing users to customize the data acquisition parameters such as frame resolution and update rate. It is to be noted that both the camera modules are installed inverted, and as a result, the frames acquired from either of them need to be flipped vertically, at least for the ease of manual interpretation.
			\end{itemize}
		
		\subsubsection{Computational Resources, Communications and Software}
		\label{Sub-Sub-Section: Vehicle CRCS}
		
			Most of the vehicle's high-level computation is handled by an NVIDIA Jetson Nano Developer Kit - B01. In terms of communication ports and interfaces, it features a Micro-USB port to enter into Device Mode, a Gigabit Ethernet port for connecting to the local area network (LAN), four USB 3.0 ports (one each for WiFi link, LIDAR and MCU interfaces, leaving one spare), two MIPI CSI-2 connectors to interface camera modules, along with an HDMI output port and a DisplayPort connector to interface with external digital displays. The computer hosts NVIDIA JetPack SDK comprising of a Linux based operating system (OS) named Linux4Tegra (an Ubuntu 18.04 image specifically designed to run on NVIDIA's hardware platforms), along with GPU accelerated tools and libraries such as Compute Unified Device Architecture (CUDA) toolkit, CUDA Deep Neural Network (cuDNN) library and Tensor Runtime (TensorRT) software development kit (SDK), to name a few. Additional installations include ROS Melodic and AutoDRIVE Devkit.
			
			Apart from this, Nigel also hosts Arduino Nano, a microcontroller unit (MCU) to interact with low-level sensors, actuators and lighting modules. It extends an interface to the MPU-9250 IMU through inter-integrated circuit (I2C) bus and links with all other peripherals through its general-purpose input/output (GPIO) pins; interrupt enabled pins for the encoders. It is interfaced with Jetson Nano Developer Kit through a universal serial bus (USB) so as to setup a bidirectional serial communication link between the two. It runs the vehicle's firmware, which is primarily responsible for acquiring and preprocessing sensory data, controlling the actuators and commanding the lights and indicators. This way, not only does the MCU offload a series of redundant operations, but it also helps the users exploiting this platform focus on developing high-level applications.
		
		\subsubsection{Actuators}
		\label{Sub-Sub-Section: Vehicle Actuators}
		
			Nigel is provided with two 6 V 160 RPM rated 120:1 DC geared motors for driving its rear wheels, and a 9.4 kgf.cm MG996R servo for steering its front wheels; the steering actuator is saturated at $\pm$ 30$^\circ$ w.r.t. its zero-steer value. All the actuators are operated at 5 V, which translates to a maximum speed of $\sim$130 RPM for the drive actuators and $\sim$0.19 s/60$^\circ$ for the steering actuator.
		
		\subsubsection{Lights and Indicators}
		\label{Sub-Sub-Section: Vehicle Lights and Indicators}
		
			Nigel has a fully functional lighting system comprising of headlights, taillights, turning indicators and reverse indicators. The headlights can be switched off or set to low-beam (single-beam) or high-beam (dual-beam). The turning indicators can be switched off or set as left, right or hazard indicators. The taillights are automatically illuminated partially when headlights are enabled, fully when brakes are applied, and disabled otherwise. Finally, the reverse indicators are also automatically enabled when vehicle is driving in reverse direction, and disabled otherwise.
	
	\subsection{Infrastructure Development}
	\label{Sub-Section: Infrastructure Development}
	
		AutoDRIVE Testbed offers a modular and reconfigurable infrastructure development kit (IDK) for rapidly designing custom driving scenarios. This kit includes a range of environment modules, passive and active traffic elements, along with a surveillance element called AutoDRIVE Eye. Besides this, the testbed also provides two preconfigured maps, viz. Driving School and Parking School.
		
		\subsubsection{Environment Modules}
		\label{Sub-Sub-Section: Environment Modules}
			
			AutoDRIVE IDK's environment modules are static objects for designing custom scenes and environments. They primarily consist of terrain modules, road kits, and obstruction modules.
			
			\begin{figure}[htpb]
				\centering
				\subfigure[]{\includegraphics[width=0.19\textwidth]{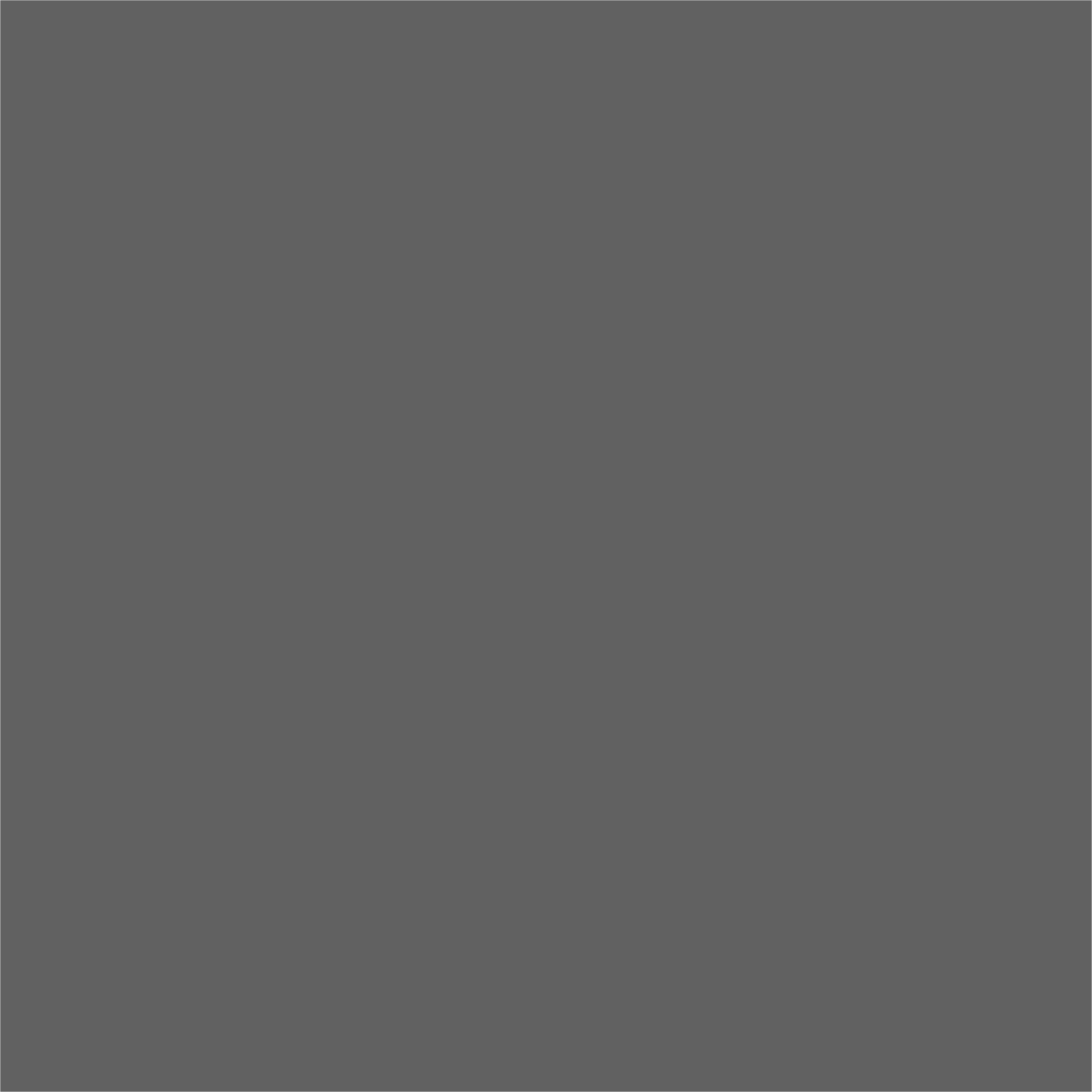}}
				\subfigure[]{\includegraphics[width=0.19\textwidth]{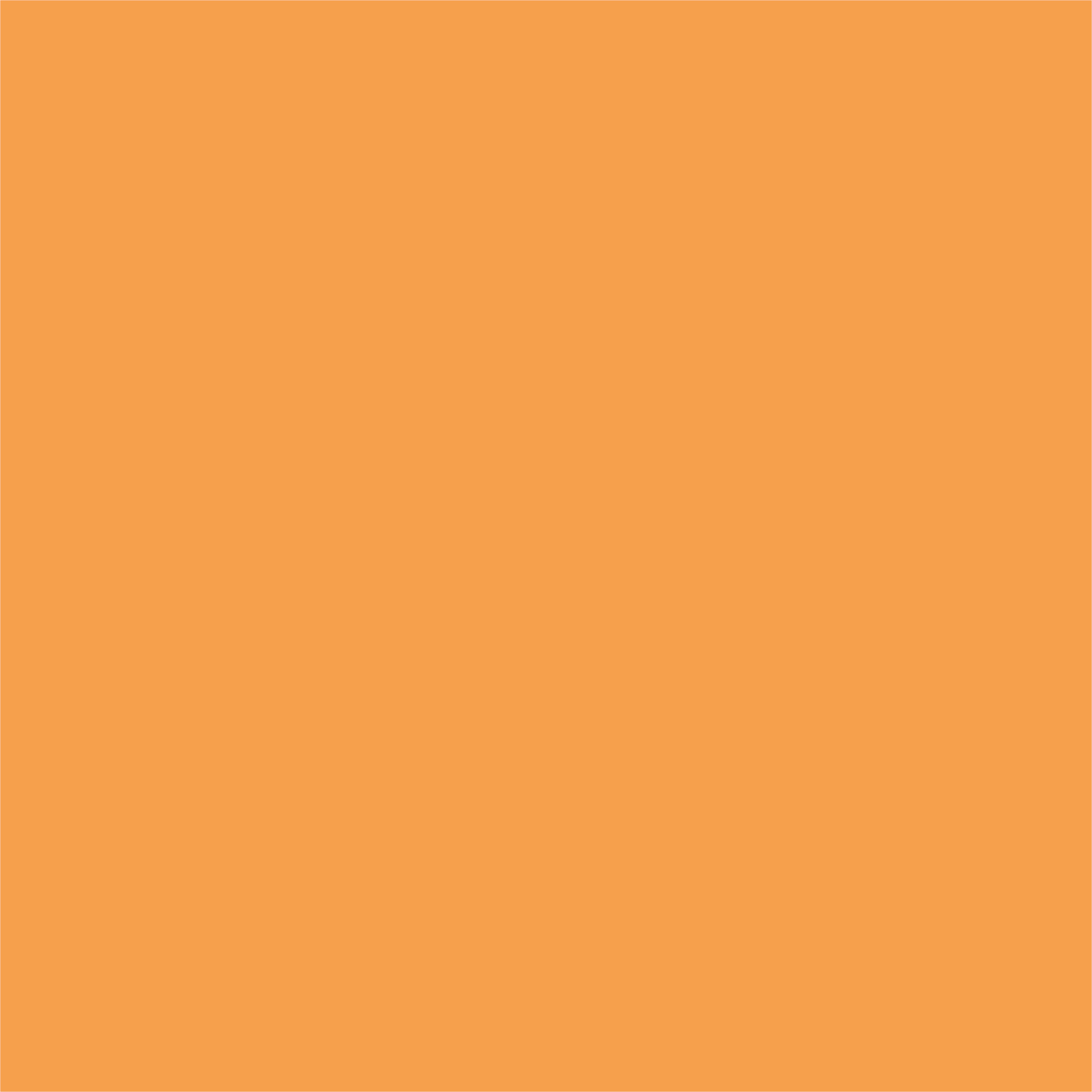}}
				\subfigure[]{\includegraphics[width=0.19\textwidth]{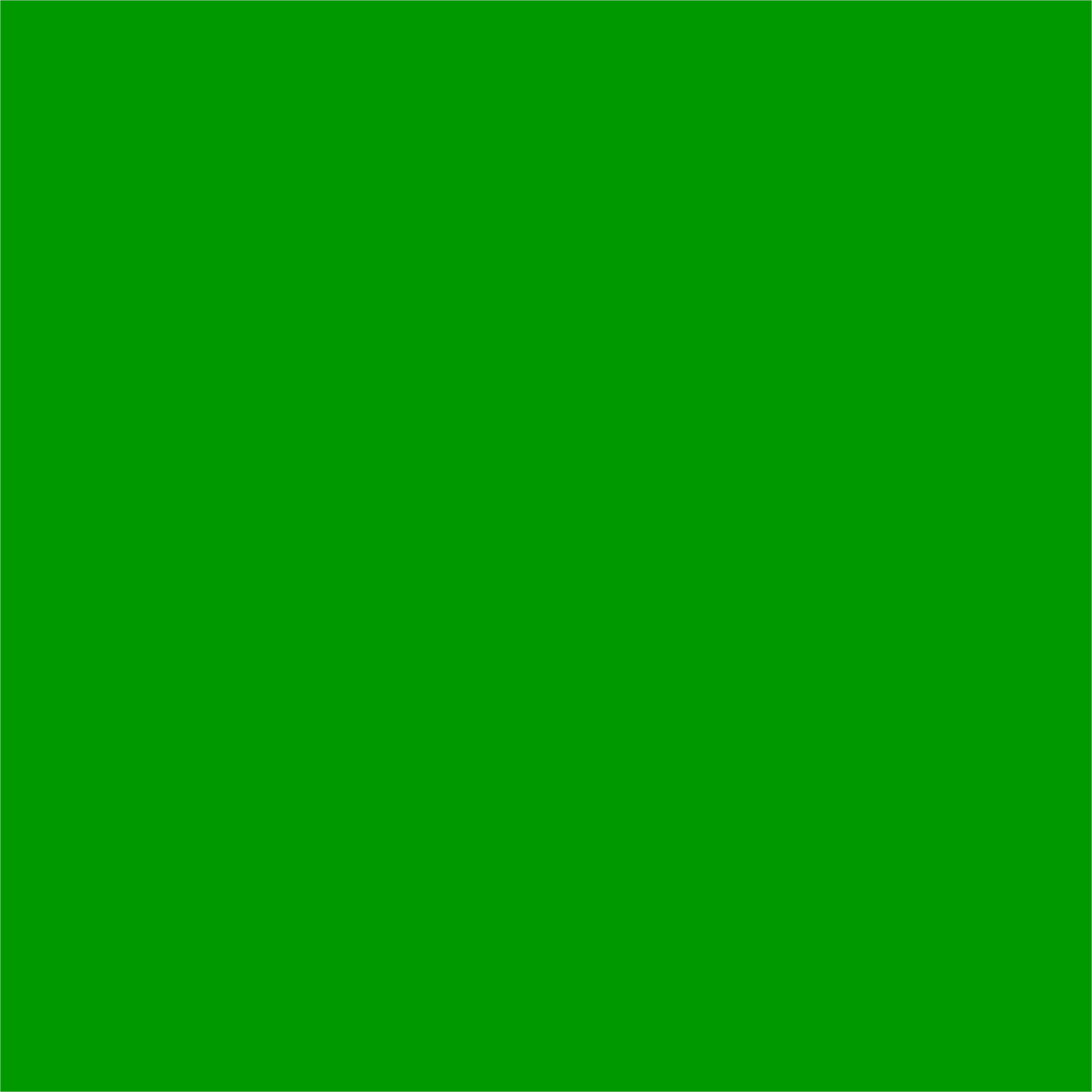}}
				\subfigure[]{\includegraphics[width=0.19\textwidth]{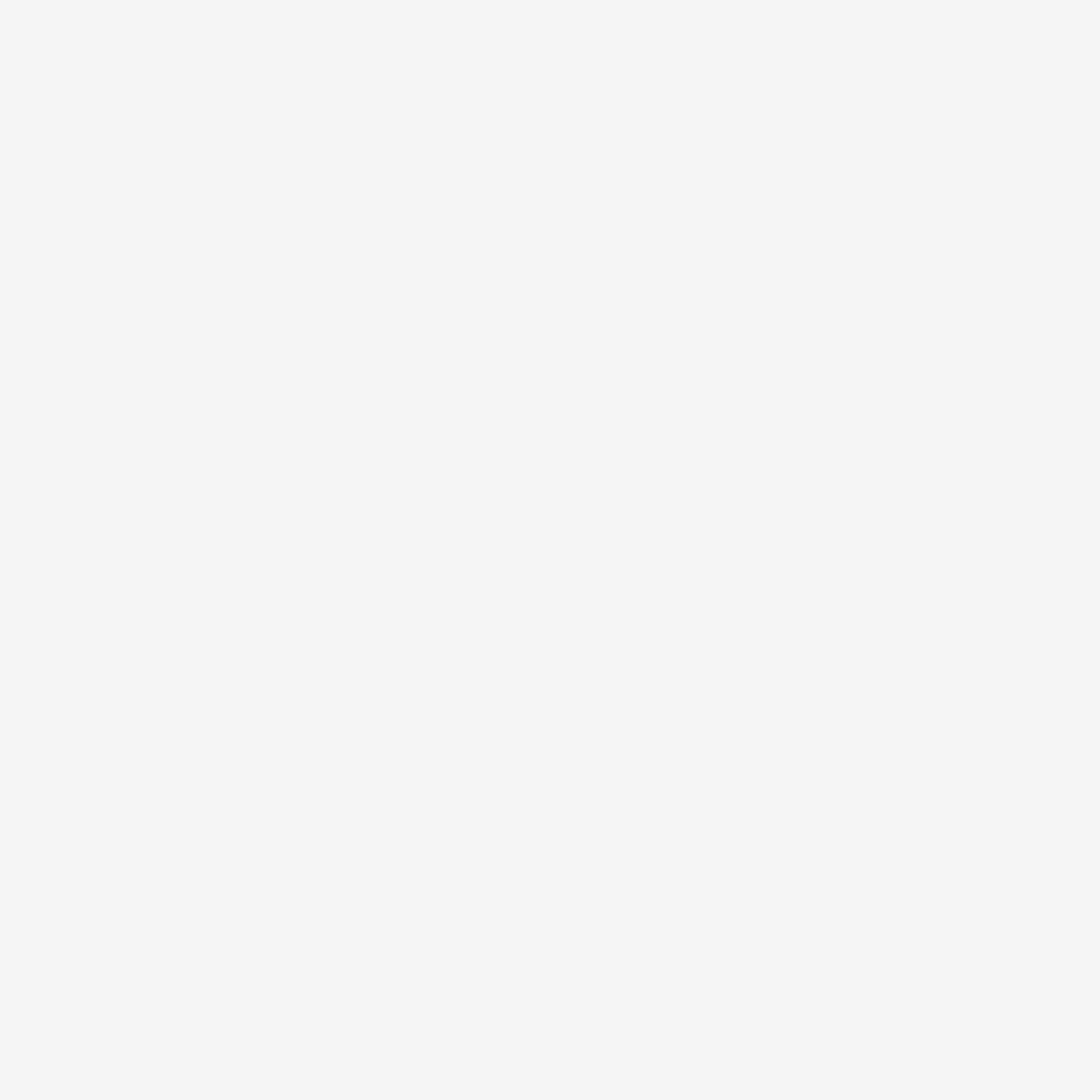}}
				\subfigure[]{\includegraphics[width=0.19\textwidth]{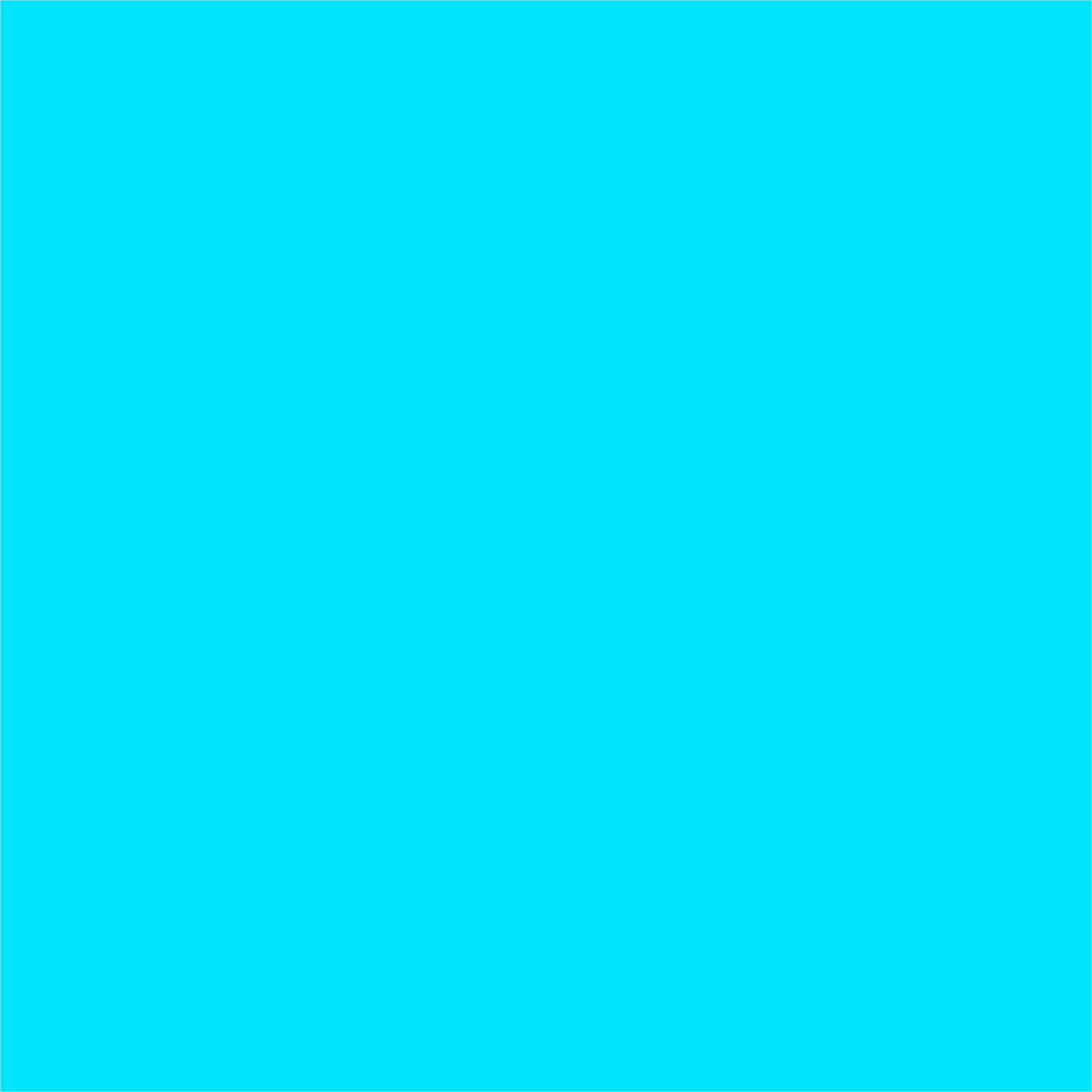}}
				\caption{Terrain Modules: (a) Asphalt, (b) Dirt, (c) Lawn, (d) Snow, and (e) Water}
				\label{Figure: Terrain Modules}
			\end{figure}
			
			The terrain modules, except for asphalt, which is designed for roadless driving scenarios, define non-drivable segments of the environment; the vehicles can semantically segment such zones, using simple computer vision techniques or advanced deep learning classification, and avoid driving over them. AutoDRIVE currently supports terrains including asphalt, dirt, lawn, snow and water (refer Figure \ref{Figure: Terrain Modules}). It is to be noted, however, that these modules are meant to be printed, and as a result, do not imitate physical conditions of the respective terrains.
			
			\begin{figure}[htpb]
				\centering
				\subfigure[]{\includegraphics[width=0.24\textwidth]{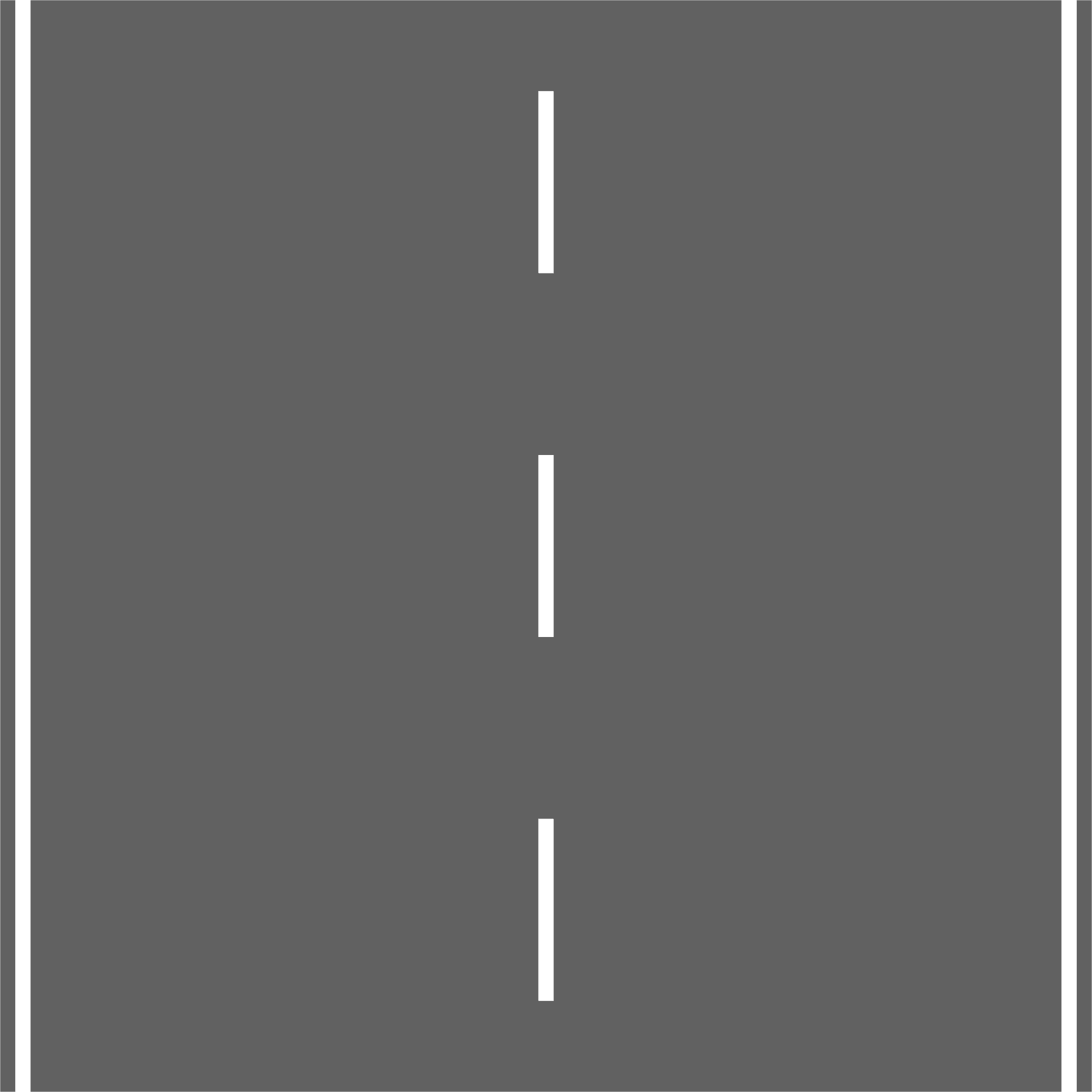}}
				\subfigure[]{\includegraphics[width=0.24\textwidth]{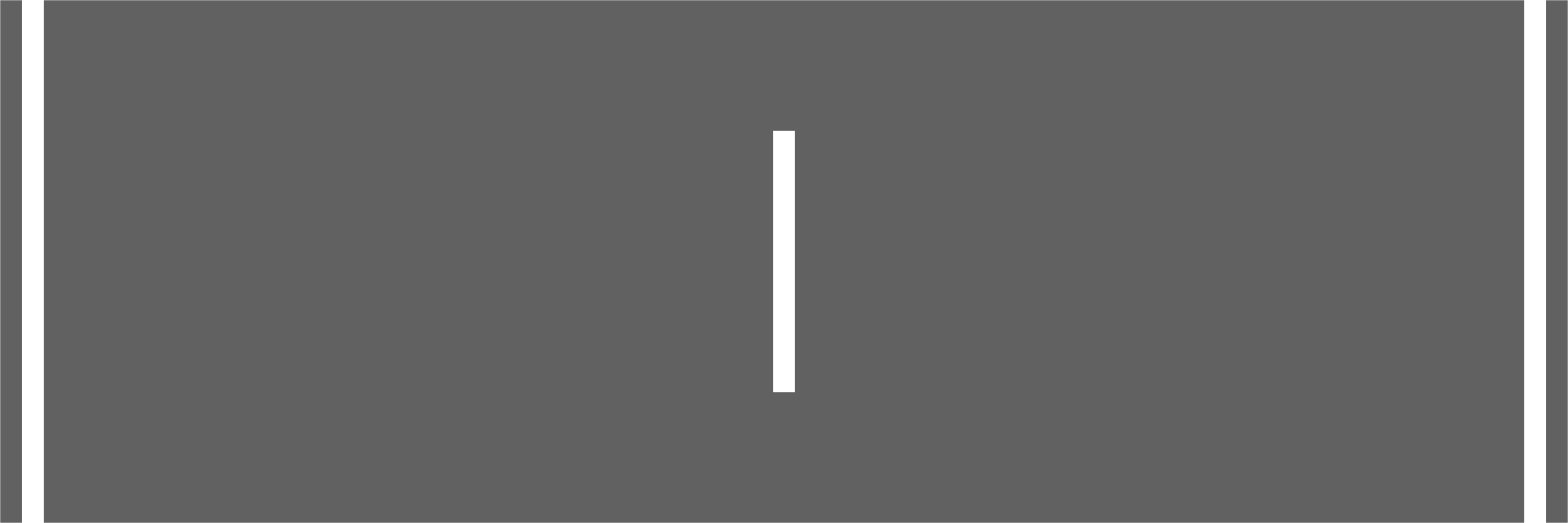}}
				\subfigure[]{\includegraphics[width=0.24\textwidth]{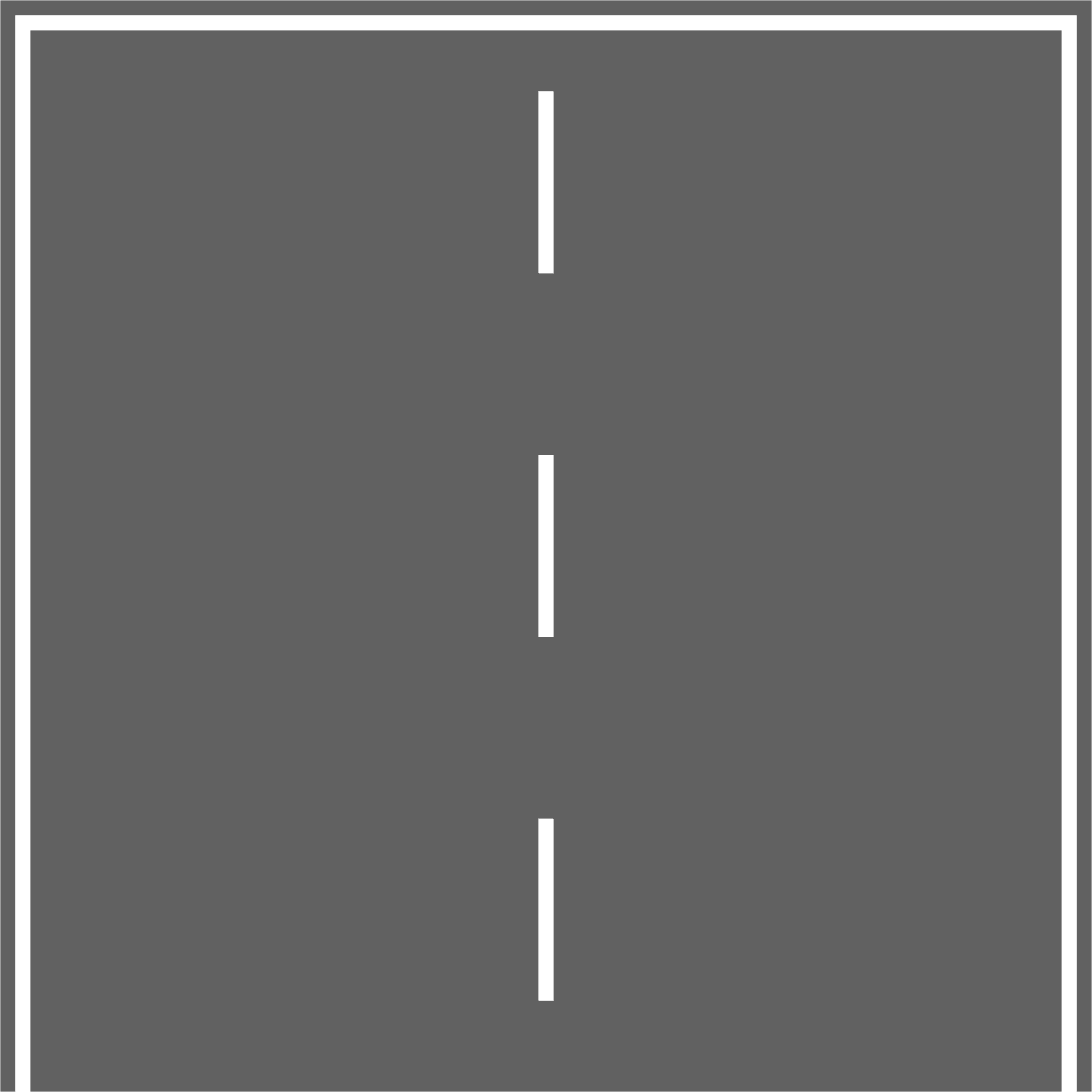}}
				\subfigure[]{\includegraphics[width=0.24\textwidth]{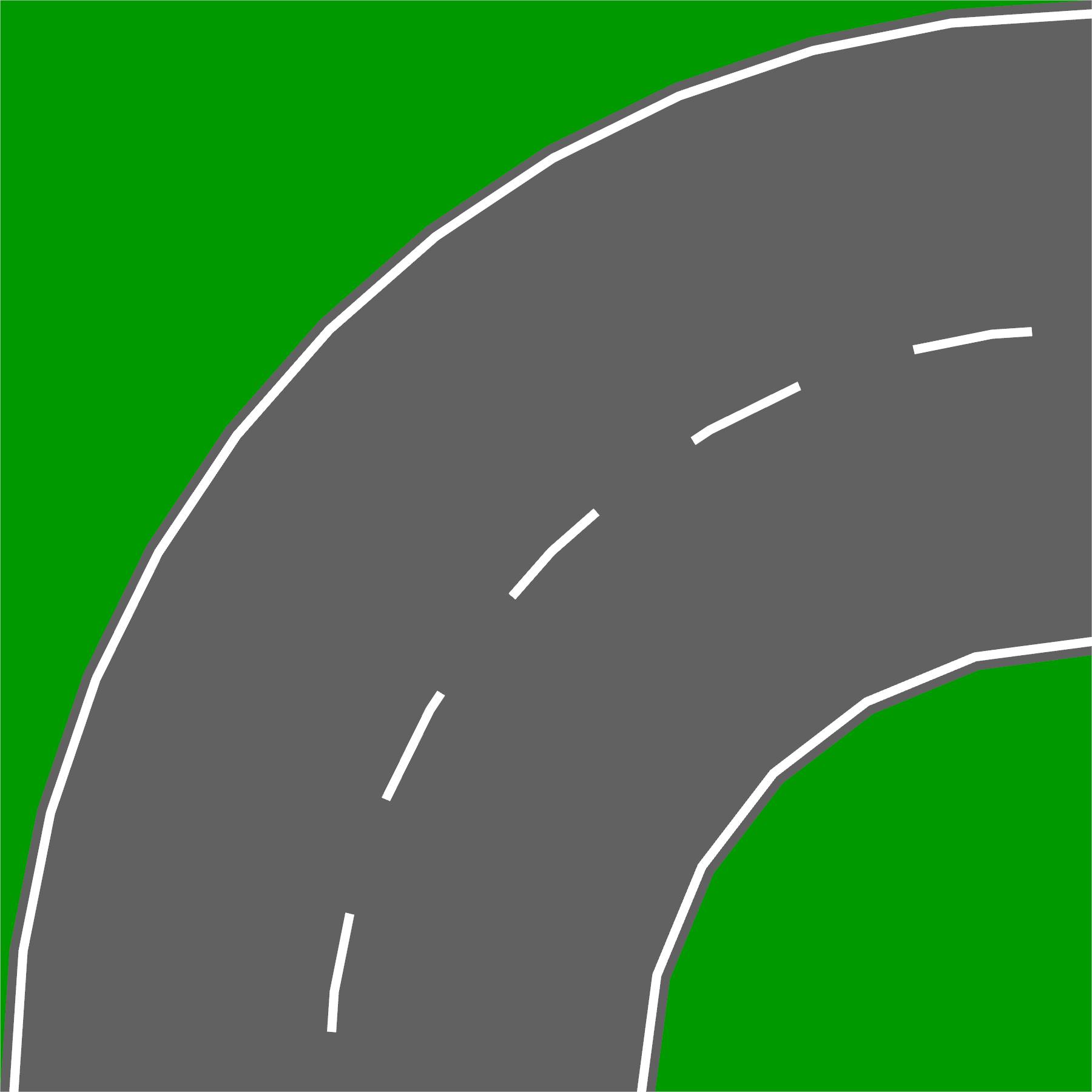}}
				\subfigure[]{\includegraphics[width=0.24\textwidth]{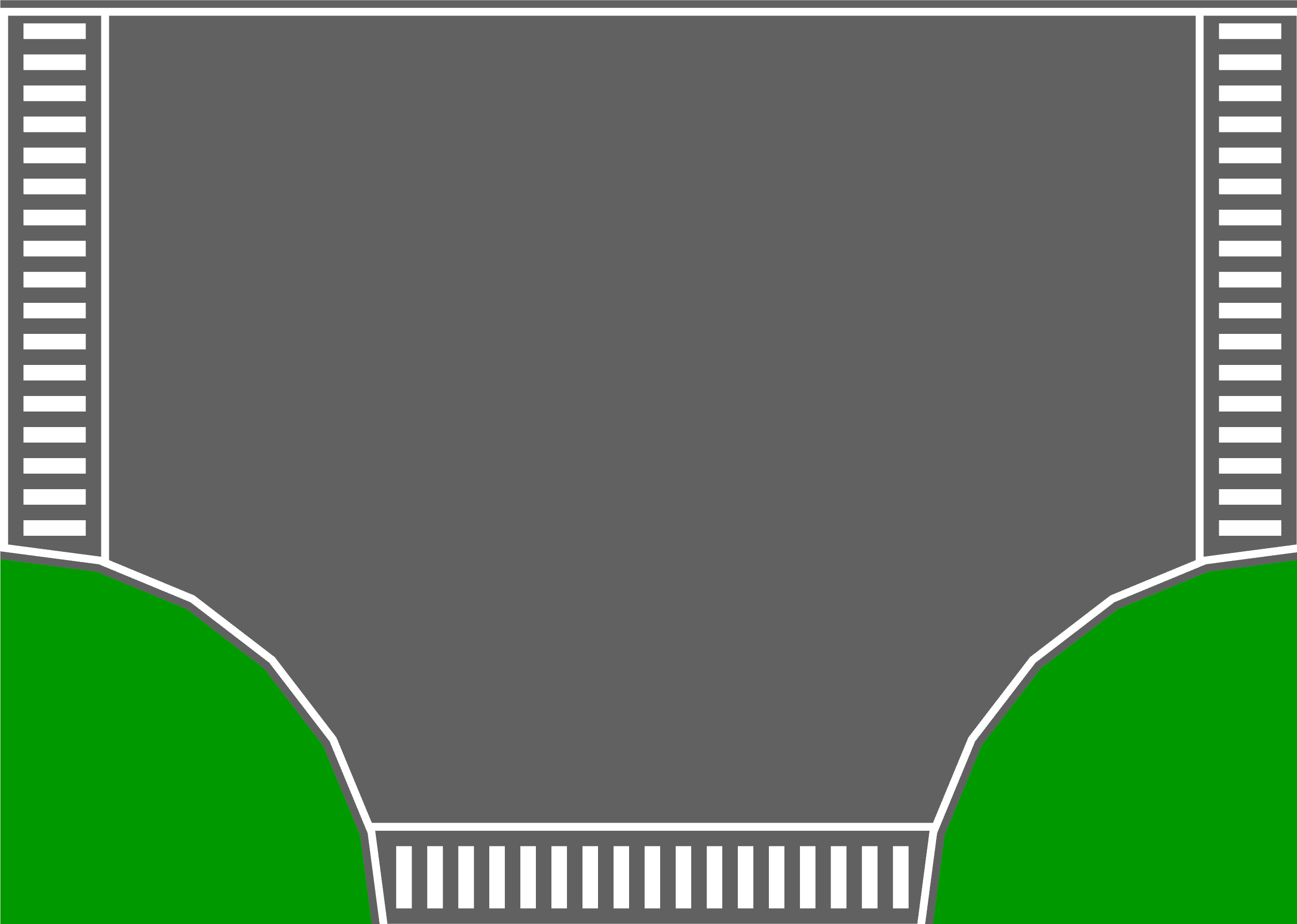}}
				\subfigure[]{\includegraphics[width=0.24\textwidth]{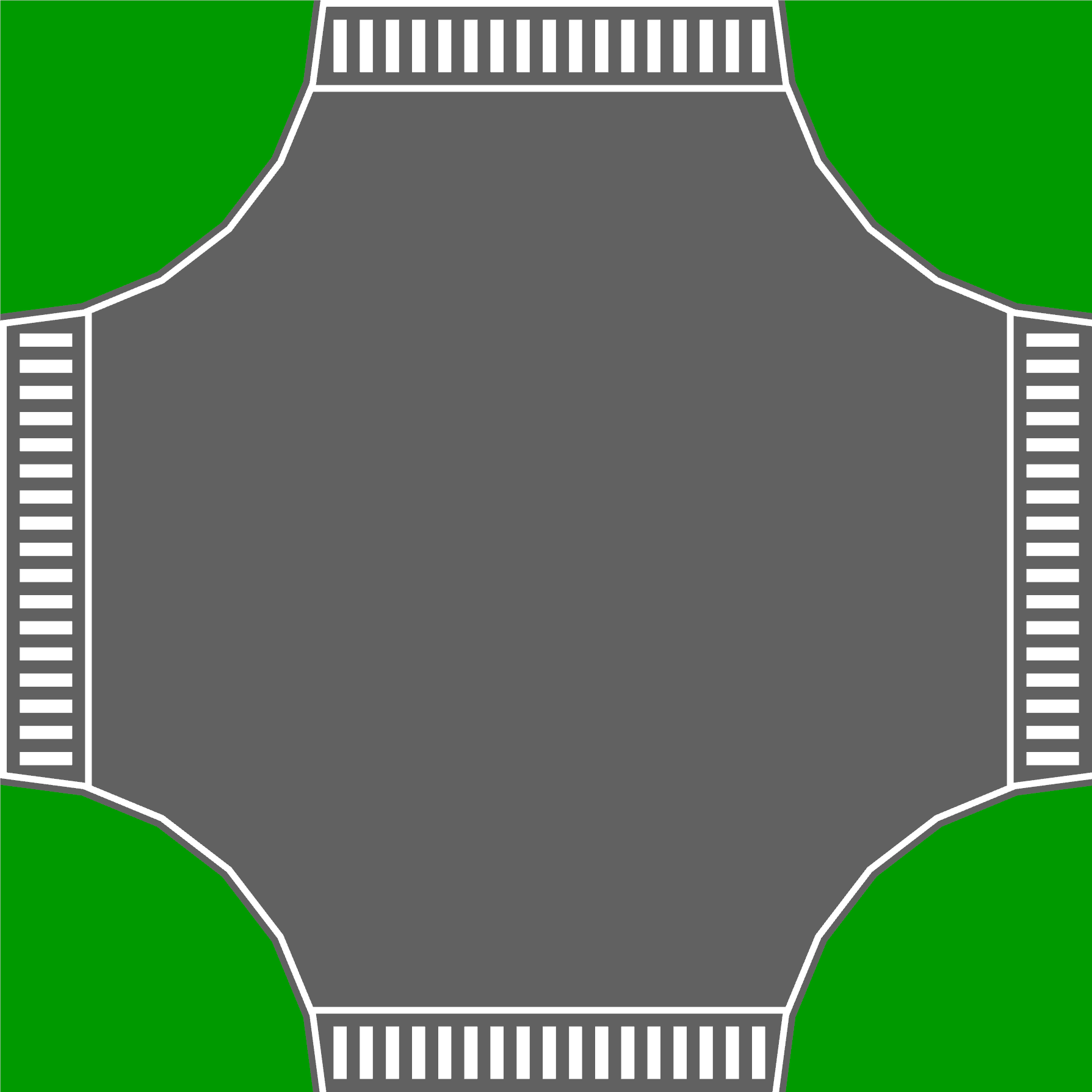}}
				\subfigure[]{\includegraphics[width=0.24\textwidth]{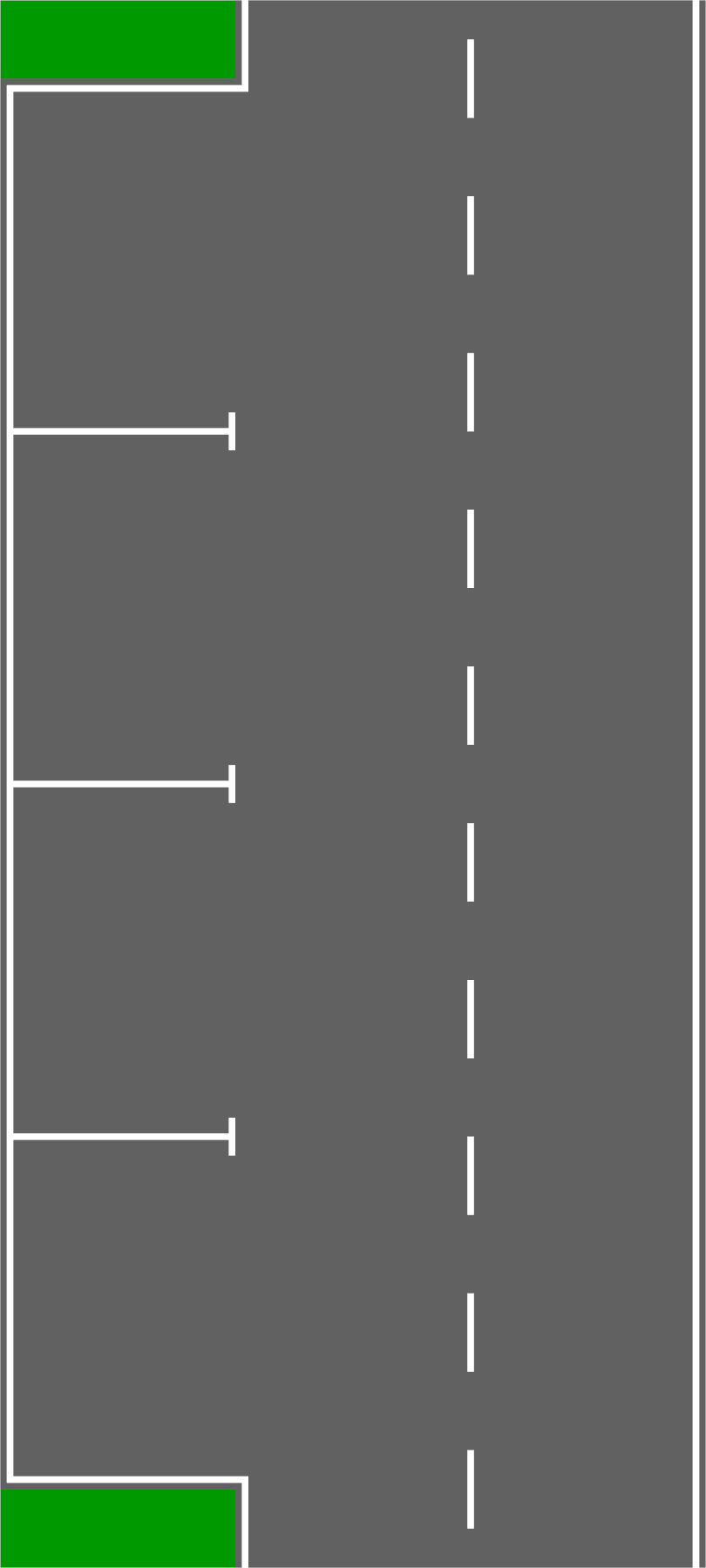}}
				\subfigure[]{\includegraphics[width=0.24\textwidth]{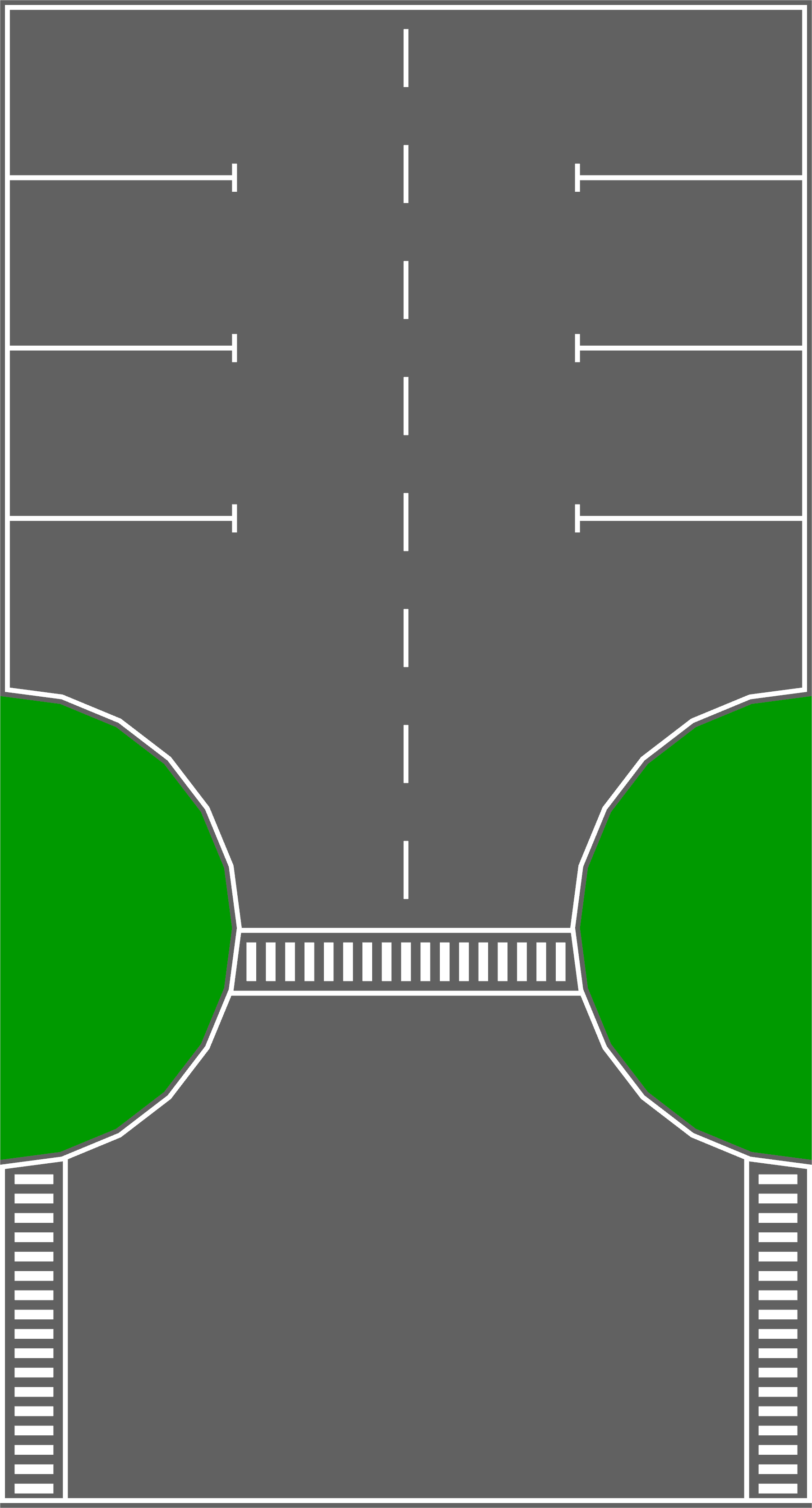}}
				\caption{Road Kit: (a) Straight Road, (b) Road Patch, (c) Dead End, (d) Curved Road, (e) 3-Way Intersection, (f) 4-Way Intersection, (g) Roadside Parking, and (h) Parking Lot}
				\label{Figure: Road Kit}
			\end{figure}
			
			The road kits help plan driving routes for testing specific manoeuvres within the operational design domain of a particular autonomy algorithm; nevertheless, they also support generalized scene development for hobbyists and enthusiasts. AutoDRIVE currently supports 1, 2, 4 and 6 lane road kits with each having straight road, road patch, dead end, curved road, 3-way intersection, 4-way intersection, roadside parking and parking lot modules. Figure \ref{Figure: Road Kit} depicts various modules available in the dual-lane road kit. Apart from this, experts may also design customized road modules using professional graphic editing tools and add them to their map.
			
			\begin{figure}[htpb]
				\centering
				\subfigure[]{\includegraphics[width=0.25\textwidth]{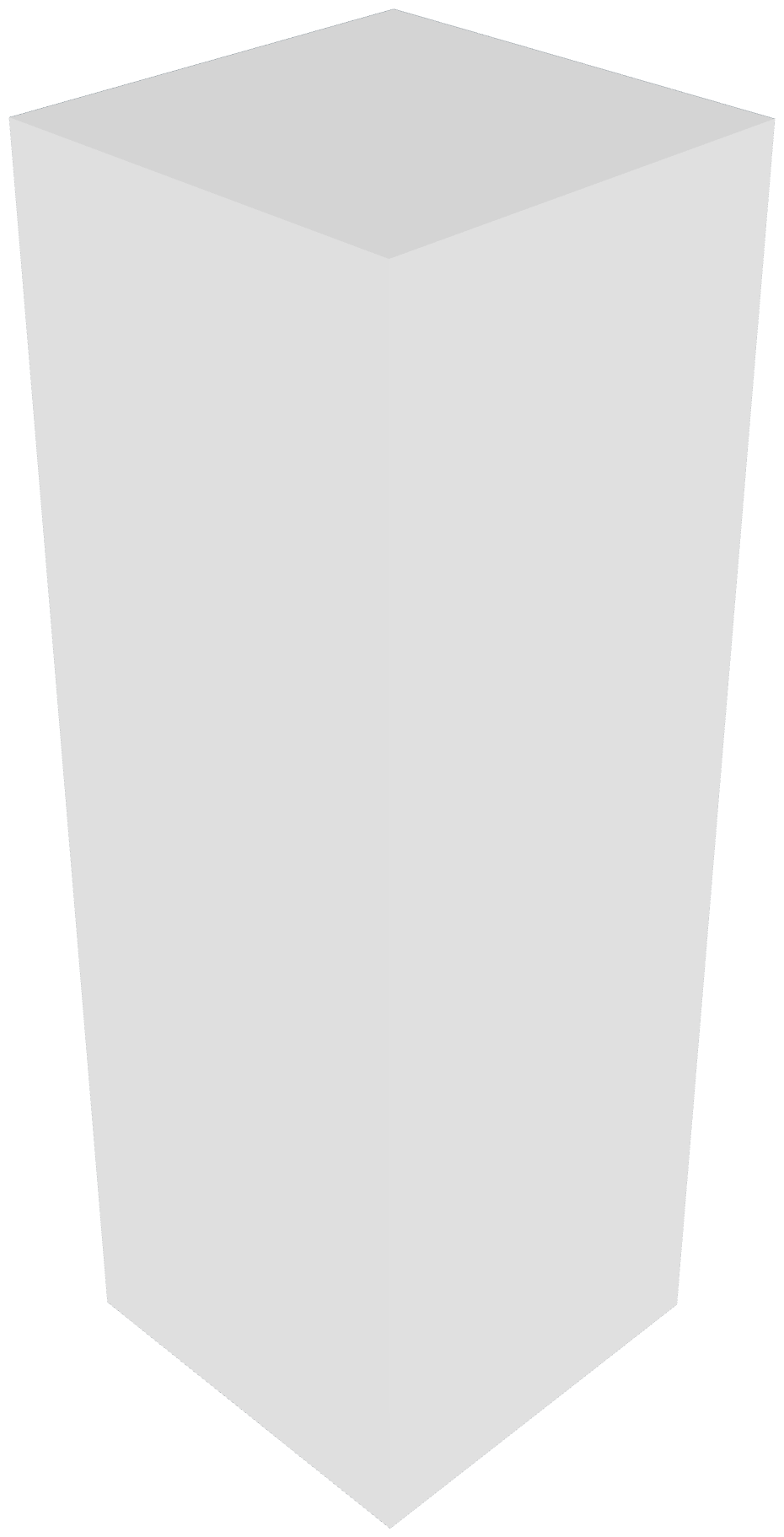}}
				\subfigure[]{\includegraphics[width=0.10\textwidth]{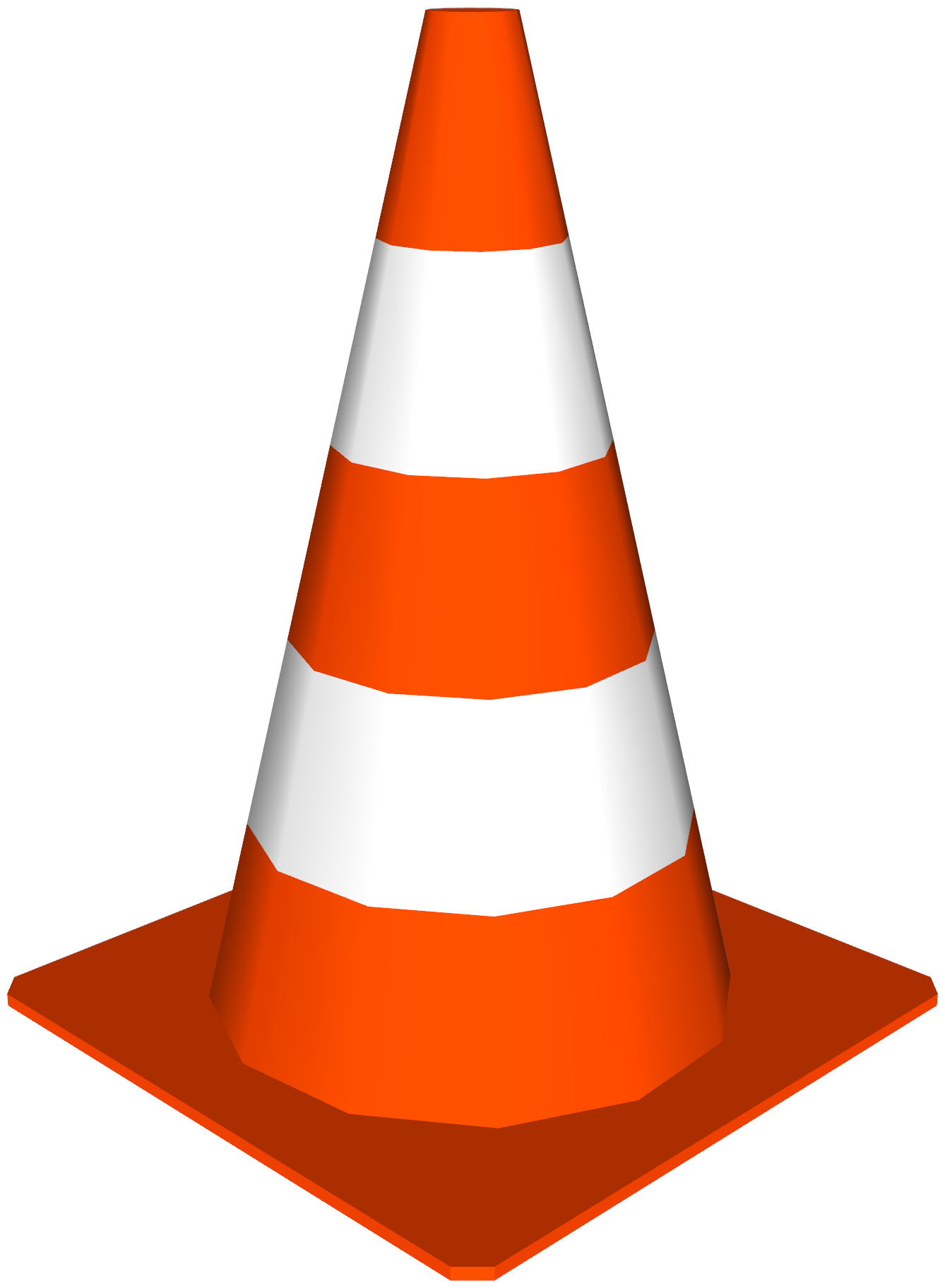}}
				\caption{Obstruction Modules: (a) Construction Box, and (b) Traffic Cone}
				\label{Figure: Obstruction Modules}
			\end{figure}
			
			The obstruction modules are 3D environmental objects, which are regarded as static obstacles within the scene. AutoDRIVE currently supports two such modules, viz. construction box and traffic cone (refer Figure \ref{Figure: Obstruction Modules}). The former is intended to be used for mimicking large-scale constructions on non-drivable segments of the environment, and is tall enough (300 mm) to be captured by the vehicle's cameras and LIDAR unit. The later, on the other hand, is intended to be used on or besides drivable segments but is comparatively shorter (70 mm), and can thus be sensed using the vehicle's cameras alone.
			
		\subsubsection{Traffic Elements}
		\label{Sub-Sub-Section: Traffic Elements}
			
			AutoDRIVE IDK's traffic elements are secondary infrastructure modules. They include passive traffic elements like road markings and traffic signs, along with active ones like the traffic light. These elements define traffic laws within a particular driving scenario, and thereby govern the traffic flow. Additionally, these modules support vehicle to infrastructure (V2I) communication, and can be integrated with the AutoDRIVE Smart City Manager (SCM) server.
			
			\begin{figure}[htpb]
				\centering
				\subfigure[]{\includegraphics[width=0.15\textwidth]{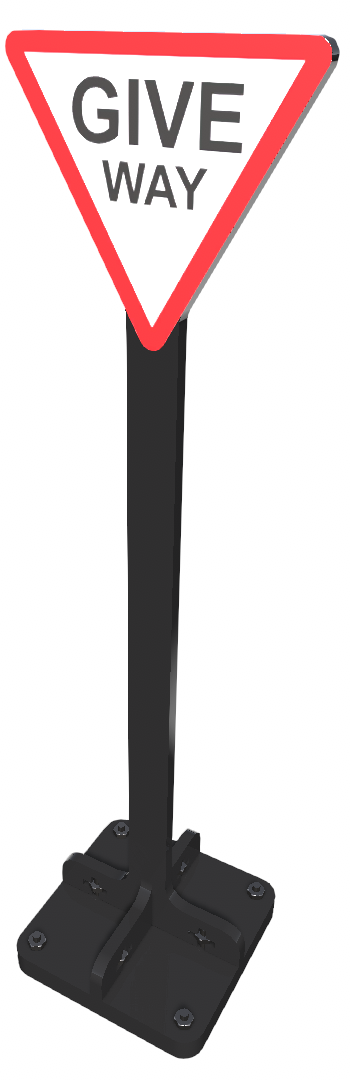}}
				\subfigure[]{\includegraphics[width=0.15\textwidth]{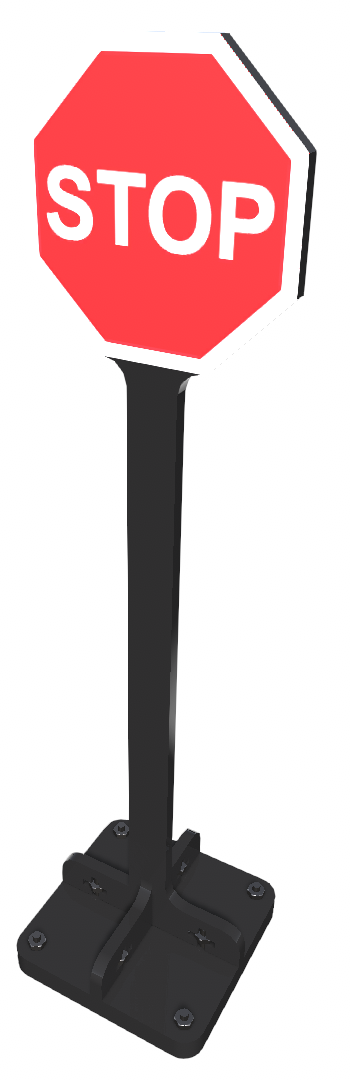}}
				\subfigure[]{\includegraphics[width=0.15\textwidth]{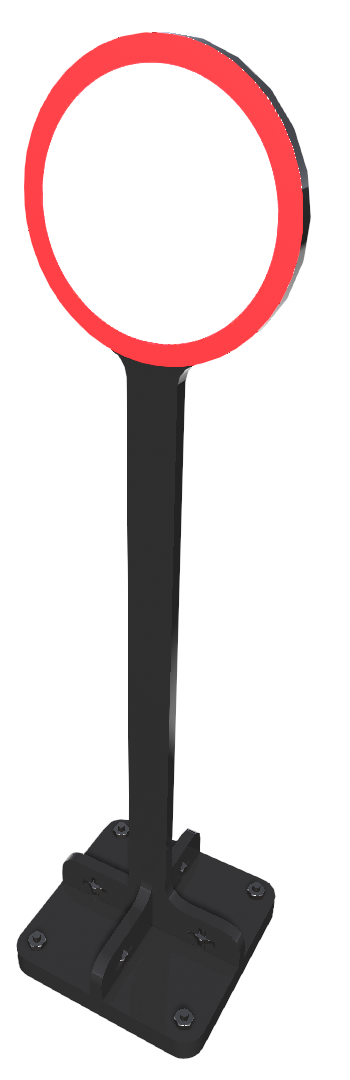}}
				\subfigure[]{\includegraphics[width=0.15\textwidth]{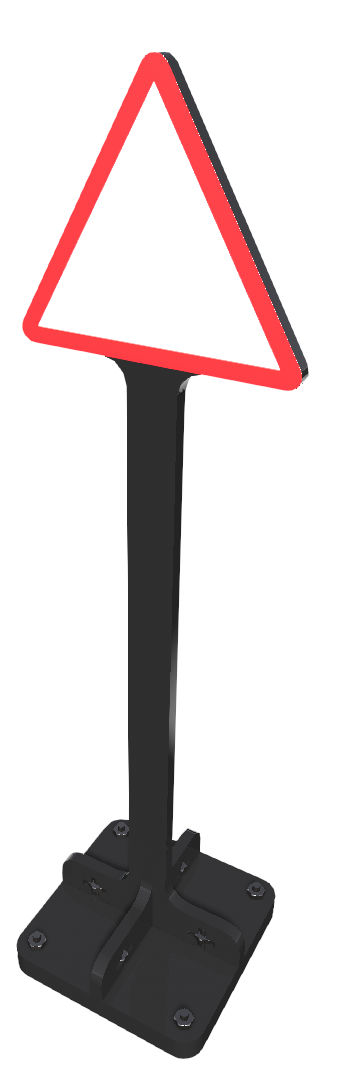}}
				\subfigure[]{\includegraphics[width=0.15\textwidth]{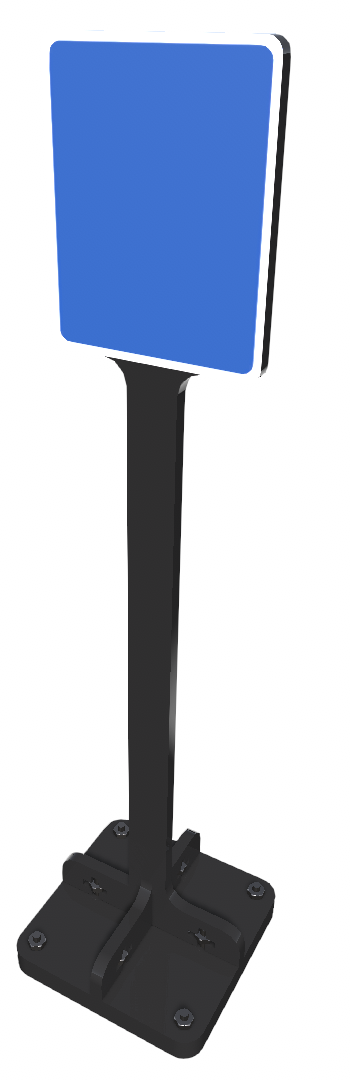}}
				\subfigure[]{\includegraphics[width=0.15\textwidth]{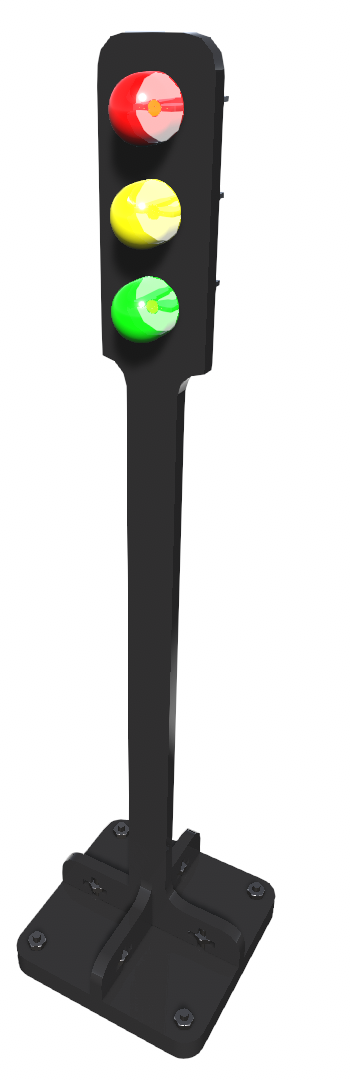}}				
				\caption{Traffic Elements: (a) Give Way Sign, (b) Stop Sign, (c) Regulatory Sign, (d) Cautionary Sign, (e) Informatory Sign, and (f) Traffic Light}
				\label{Figure: Traffic Elements}
			\end{figure}
			
			AutoDRIVE follows the convention of white road markings (refer Figure \ref{Figure: Road Kit}). The roads are marked with solid lines at the boundaries and dashed lines at each individual lane separation. The intersections are marked with white pedestrian crossings and stop-lines. Even the parking spaces are marked in white, with appropriate cushioning area.
			
			AutoDRIVE's traffic signs are classified under three major categories, viz. regulatory signs depicted in circles indicating rules and regulations, cautionary signs depicted in triangles indicating warnings pertaining to the upcoming scenario, and informatory signs depicted in rectangles indication useful information and cues. Apart from these primary classes, two signs, viz. give way sign and stop sign, both belonging to the regulatory class, have their unique shapes; the former resembling an inverted triangle, and the later, an octagon. Figures \ref{Figure: Traffic Elements}(a)-(e) depict various traffic sign templates.
			
			For active traffic control, AutoDRIVE supports a standard three-stage traffic light; refer Figure \ref{Figure: Traffic Elements}(f). A red signal mandates an immediate stop, green gives the right to drive through freely, and yellow allows traversal only if it is safe to do so. The designed traffic light supports internet of things (IoT), and can be controlled in a variety of ways including manual operation, timed operation, as well as cooperative operation using the AutoDRIVE SCM server. This way, not only can the users actively control traffic, but they can also develop smart city solutions for managing traffic flow autonomously.
			
		\subsubsection{Surveillance Elements}
		\label{Sub-Sub-Section: Surveillance Elements}
			
\begin{figure}[htpb]
				\centering
				\includegraphics[width=\textwidth]{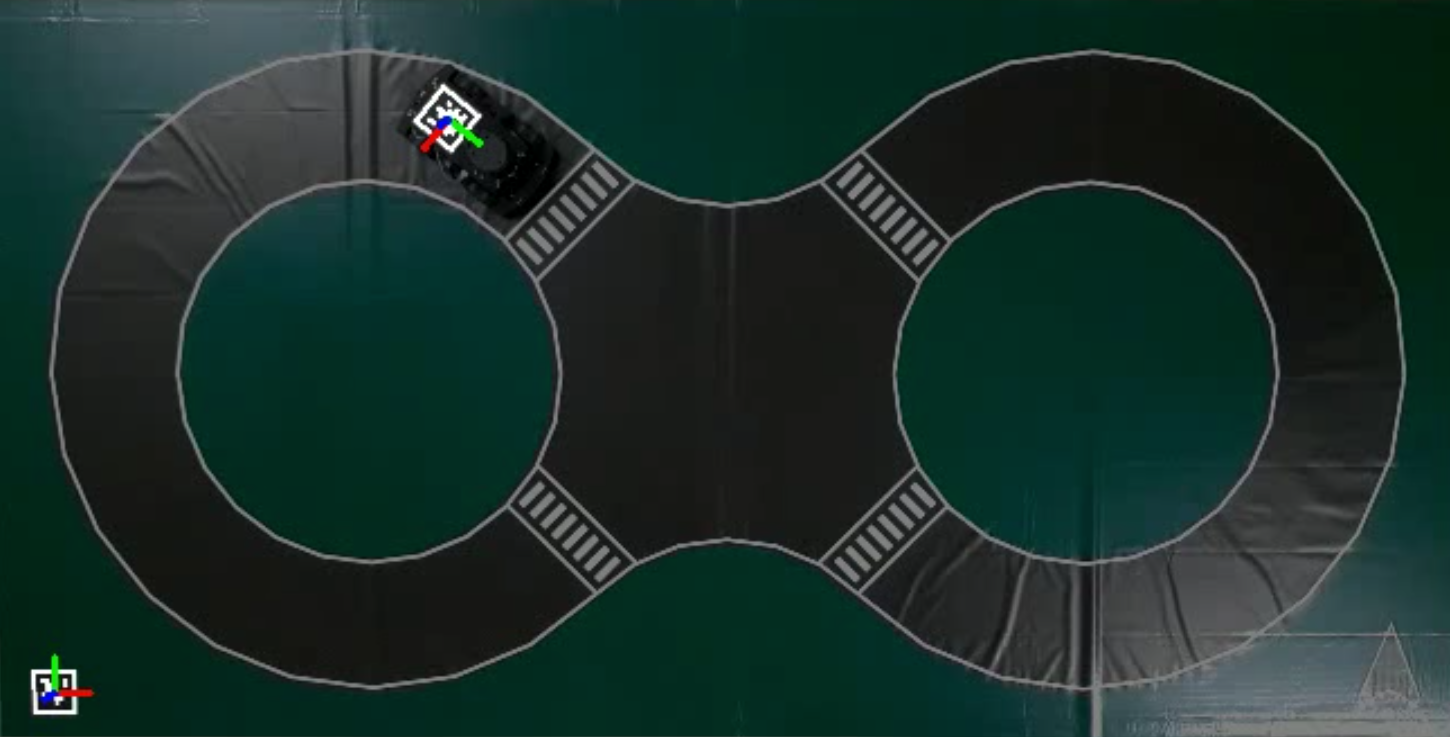}
				\caption{Vehicle Localization using the AutoDRIVE Eye}
				\label{Figure: AutoDRIVE Eye}
			\end{figure}

			As described earlier, AutoDRIVE IDK's traffic elements support V2I communication technology, and can be integrated with the AutoDRIVE SCM server for developing real-time monitoring and control applications. Additionally, the testbed features AutoDRIVE Eye, a surveillance element to view the entire scene from bird's eye view. The said element is also integrated with the AutoDRIVE SCM server, and, upon appropriate calibration of its intrinsic parameters, is capable of estimating vehicle's 2D pose within the map by detecting and tracking the AprilTag markers attached to them (refer Section \ref{Sub-Sub-Section: Vehicle Sensor Suite}); this functionality of the AutoDRIVE Eye is illustrated in Figure \ref{Figure: AutoDRIVE Eye}.
			
		\subsubsection{Preconfigured Maps}
		\label{Sub-Sub-Section: Preconfigured Maps}
			
\begin{figure}[htpb]
				\centering
				\subfigure[]{\includegraphics[width=0.49\textwidth]{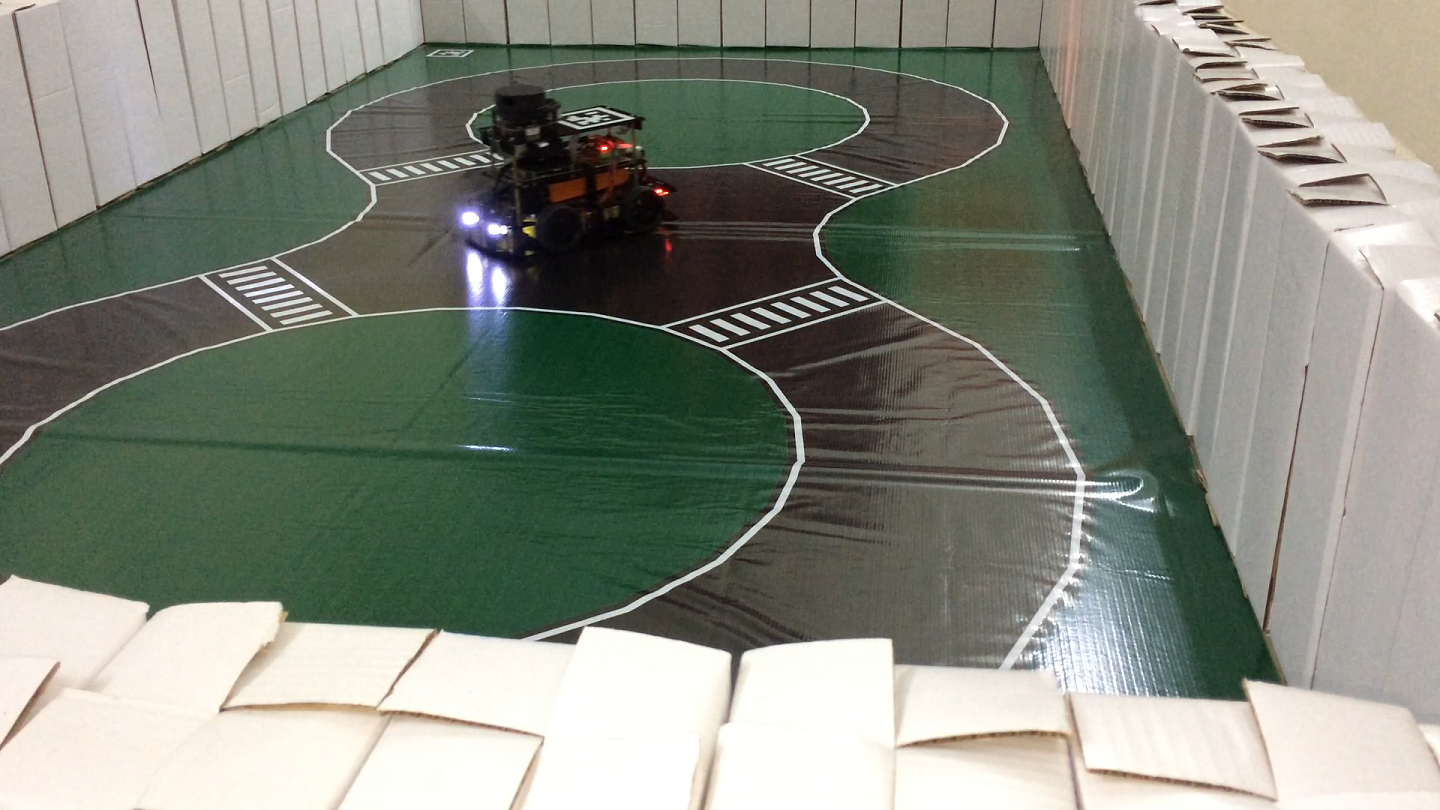}}
				\subfigure[]{\includegraphics[width=0.49\textwidth]{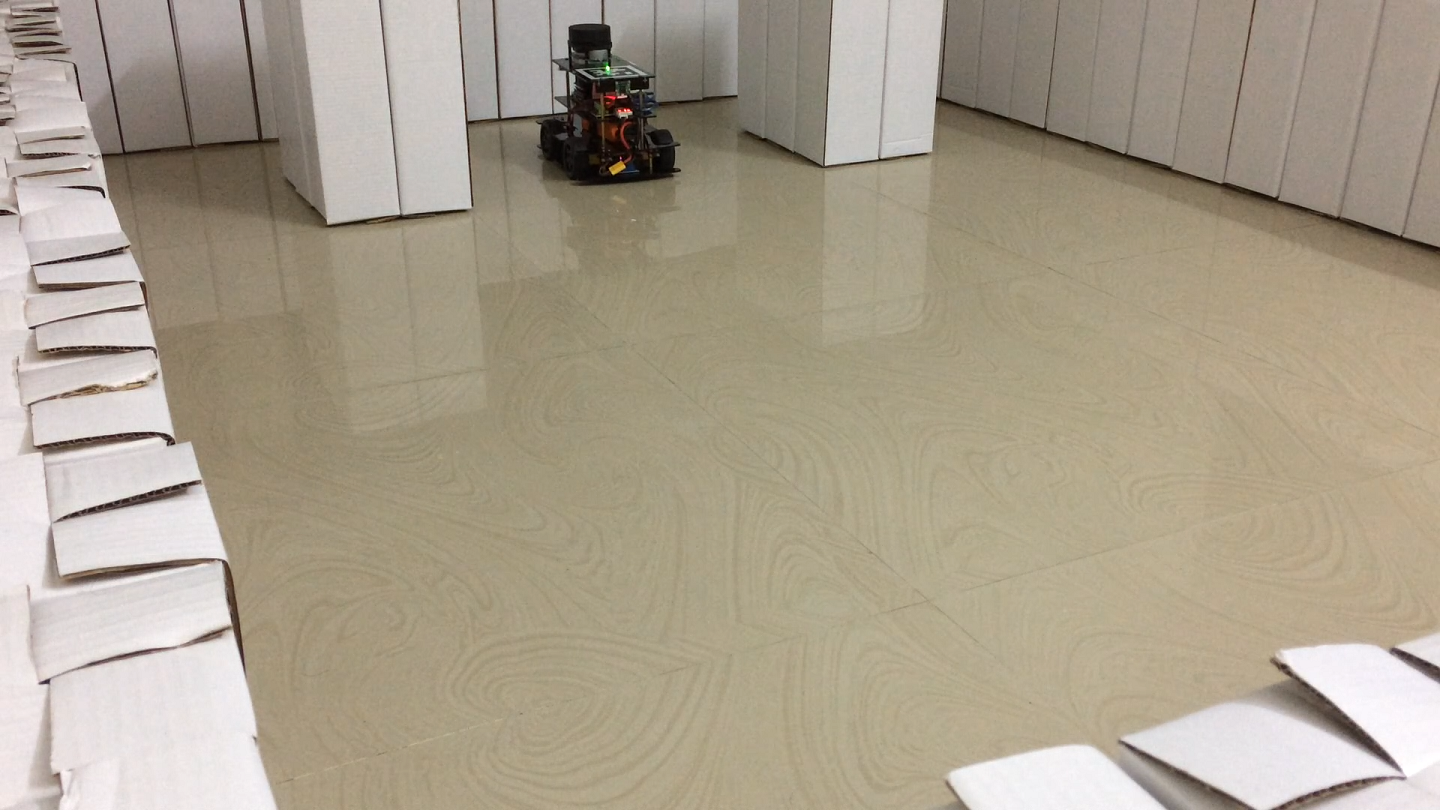}}
				\caption{Testbed Maps: (a) Driving School, and (b) Parking School}
				\label{Figure: Testbed Maps}
			\end{figure}

			As described earlier, AutoDRIVE Testbed currently provides two preconfigured maps, viz. Driving School and Parking School. The former is designed with an objective of supporting various driving manoeuvres including straight drives, along with left and right turns of various intensities. Additionally, it also features a 4-way intersection, which can be used for developing autonomy algorithms pertaining to intersection traversal and traffic control. The later, on the contrary, is designed specifically for autonomous parking applications. It uses construction boxes to define static objects such as walls and pillars of the parking lot, along with parked peer vehicles and does not provide any distinction between road surfaces (it treats all the available free space as drivable). Figure \ref{Figure: Testbed Maps} depicts the two preconfigured maps.
			
\section{AutoDRIVE Simulator}
\label{Section: AutoDRIVE Simulator}

	As described earlier, AutoDRIVE Simulator \cite{AutoDRIVESimulatorPaper2021, AutoDRIVESimulatorReport2020} can be considered as a digital twin of the AutoDRIVE Testbed. It is primarily targeted towards virtual prototyping of autonomy algorithms, either as a part of the recursive simulation-deployment workflow, or due to possible constraints limiting/prohibiting the use of hardware testbed. It can also be employed for developing standardized datasets by collecting and labelling the simulated driving data. Consequently, its development required modelling and simulating the vehicle (vehicular dynamics, sensors, actuators and lights) and the infrastructural setup (environmental physics, terrain modules, road kits, obstruction modules and traffic elements). The following sections describe vehicle and infrastructure simulation along with some of the prominent features of AutoDRIVE Simulator.
	
	\subsection{Vehicle Simulation}
	\label{Sub-Section: Vehicle Simulation}
	
		\begin{figure}[htpb]
			\centering
			\includegraphics[width=\textwidth]{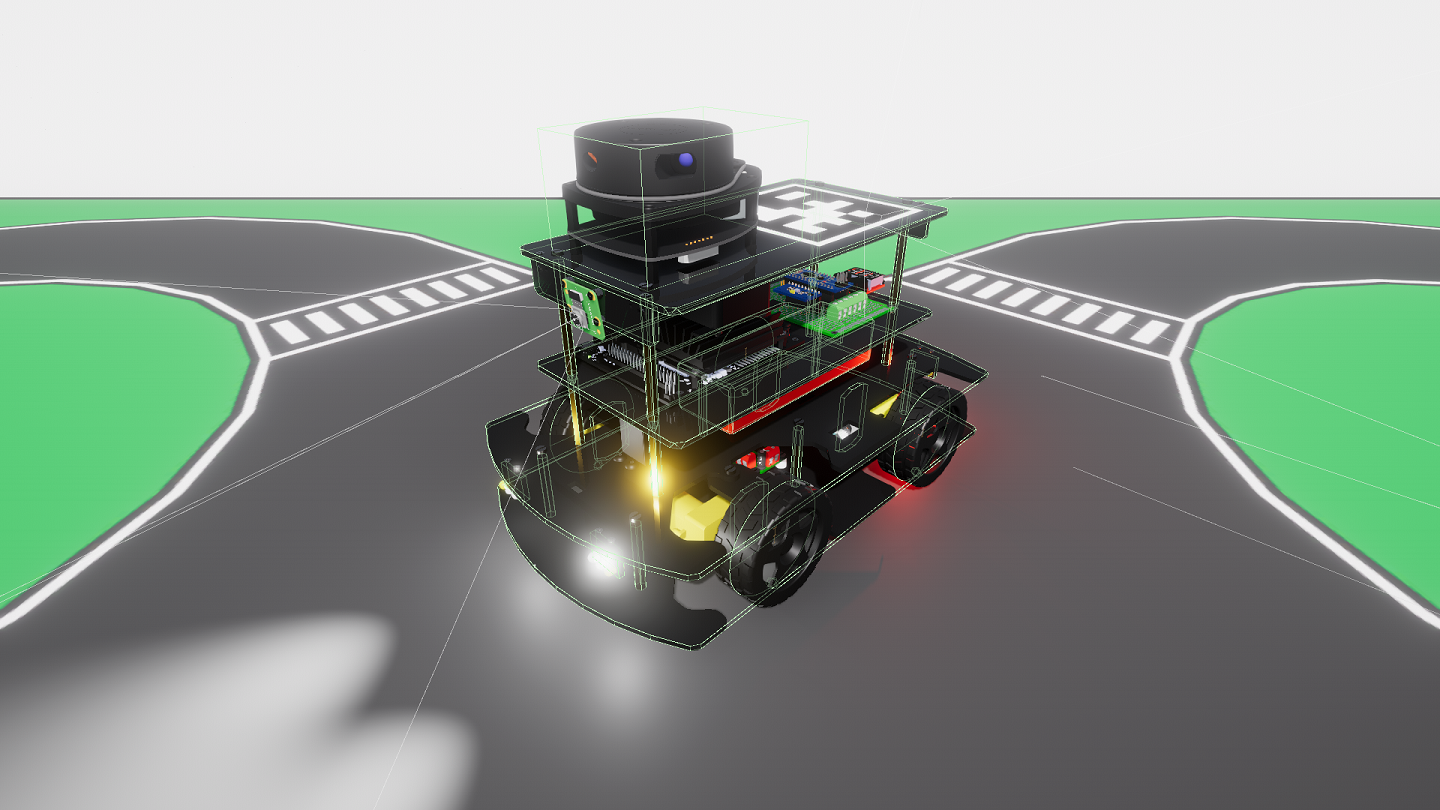}
			\caption{Vehicle Simulation}
			\label{Figure: Vehicle Simulation}
		\end{figure}
		
		The computer-aided design (CAD) model of the vehicle was converted into a mesh model, which was imported into the Unity Editor. Consequently, the vehicle was physically modelled within the Unity Editor in order to simulate realistic system dynamics. Figure \ref{Figure: Vehicle Simulation} depicts the simulated vehicle, with Unity Gizmos enabled to visualize its physical attributes. In addition to this, the vehicle was also enhanced aesthetically so as to visually resemble its real-world counterpart. Following is a brief summary of various aesthetic and physical attributes of the simulated vehicle:
		
		\begin{itemize}
			\item \textbf{Materials:} The material properties of various components of the vehicle were tuned to match their real-world counterparts, thereby producing a photo-realistic rendering of the vehicle on the simulator front-end, while also realistically simulating its various interacting elements.
			\item \textbf{Vehicular Dynamics:} The vehicle was assigned various physical attributes including mesh colliders for detecting body collisions, wheel colliders for simulating realistic tyre forces and interactions with the road, and rigid body component for simulating various forces acting on the vehicle (including gravity). Additionally, system parameters of the simulated vehicle like mass, friction limits and air drag, for instance, were also tuned close to its physical counterpart.
			\item \textbf{Sensors:} The entire sensor suite of the vehicle was simulated through back-end scripting of each individual sensor module. All the simulated sensors were developed to closely resemble their real-world counterparts in terms of static and dynamic characteristics.
			\item \textbf{Actuators:} The implemented wheel colliders were employed to simulate realistic actuator dynamics by accurately modelling the saturation limits along with appropriate response delays. Additionally, the motion control limits imposed on actuators of the real vehicle were also taken into account.
			\item \textbf{Lights and Indicators:} The vehicular elements used for illumination and signalling were provided with physical light objects for accurately simulating the raycasting and light diffusion.
		\end{itemize}
	
	\subsection{Infrastructure Simulation}
	\label{Sub-Section: Infrastructure Simulation}
	
		\begin{figure}[htpb]
			\centering
			\includegraphics[width=\textwidth]{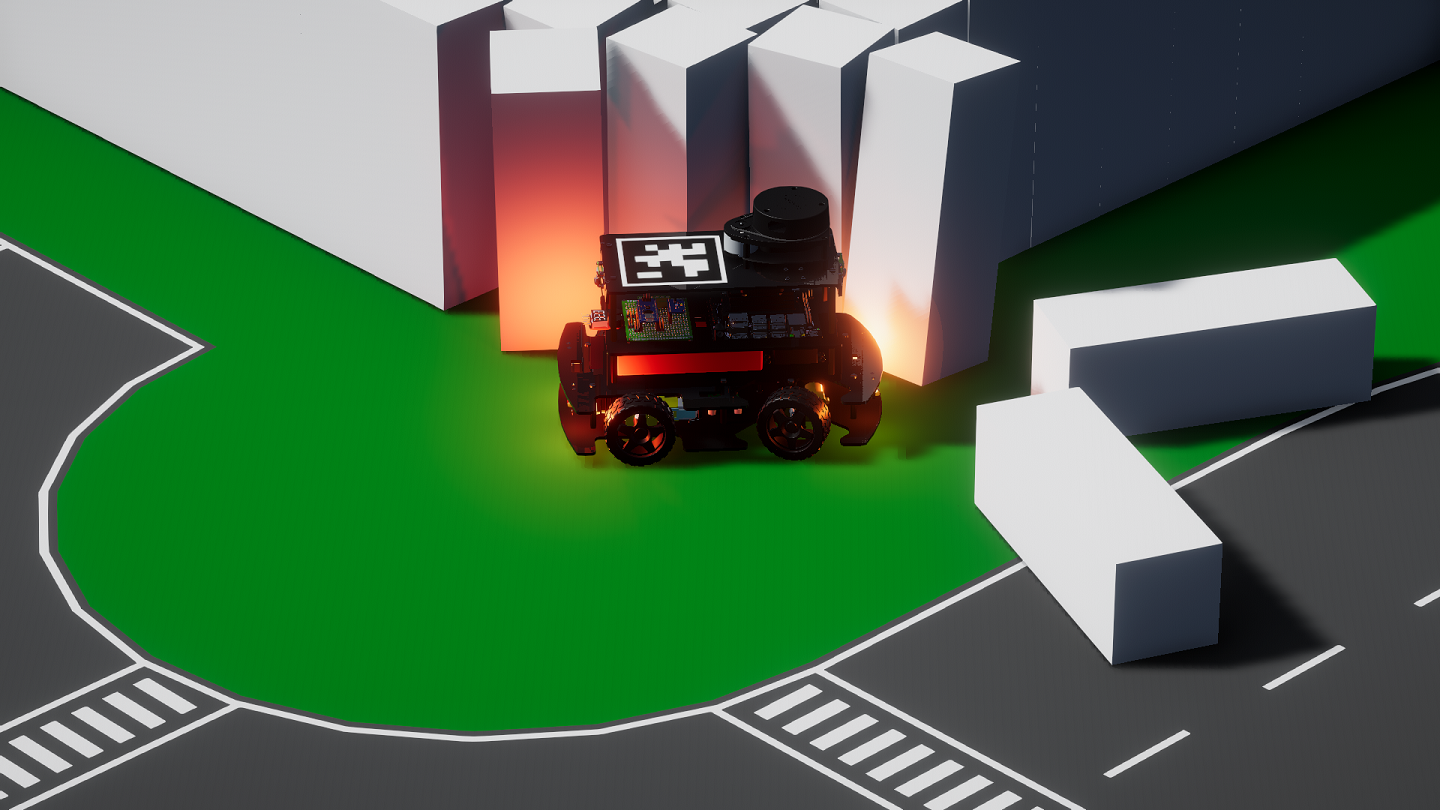}
			\caption{Infrastructure Simulation}
			\label{Figure: Infrastructure Simulation}
		\end{figure}
	
\begin{figure}[htpb]
			\centering
			\subfigure[]{\includegraphics[width=0.24\textwidth]{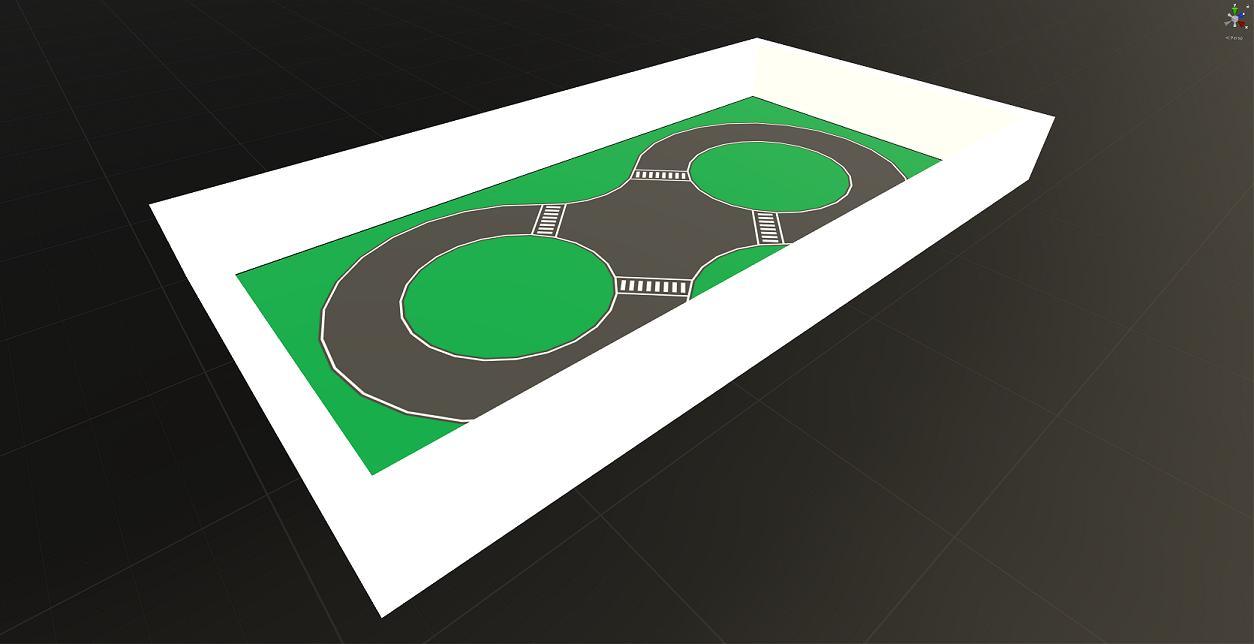}}
			\subfigure[]{\includegraphics[width=0.24\textwidth]{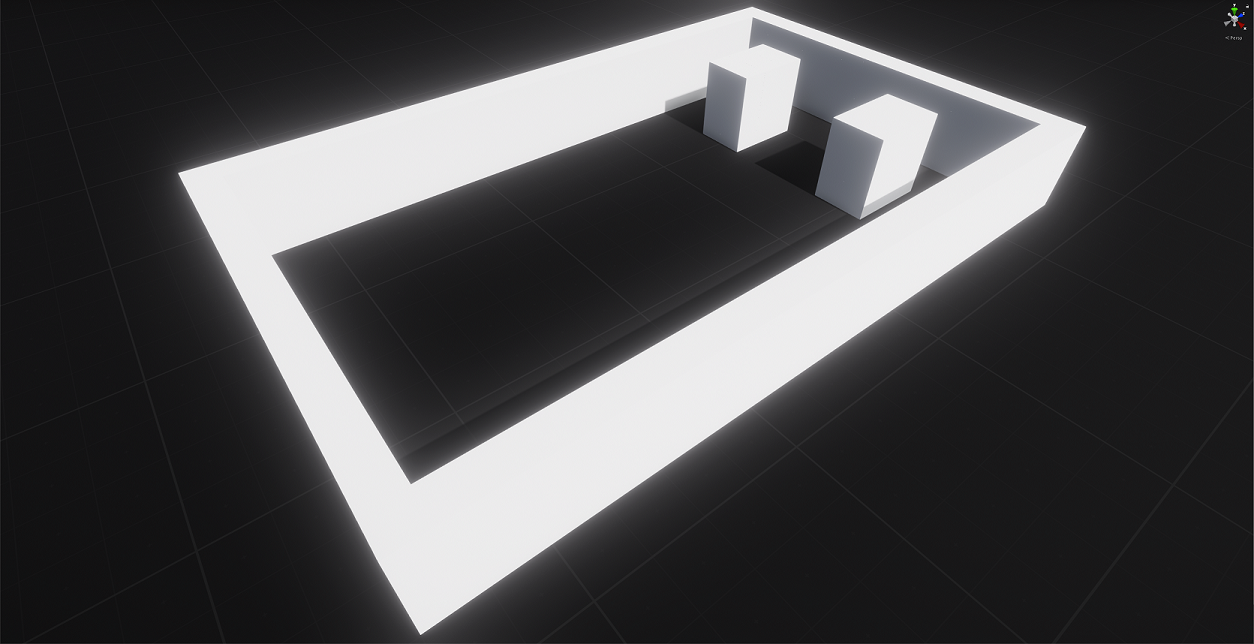}}
			\subfigure[]{\includegraphics[width=0.24\textwidth]{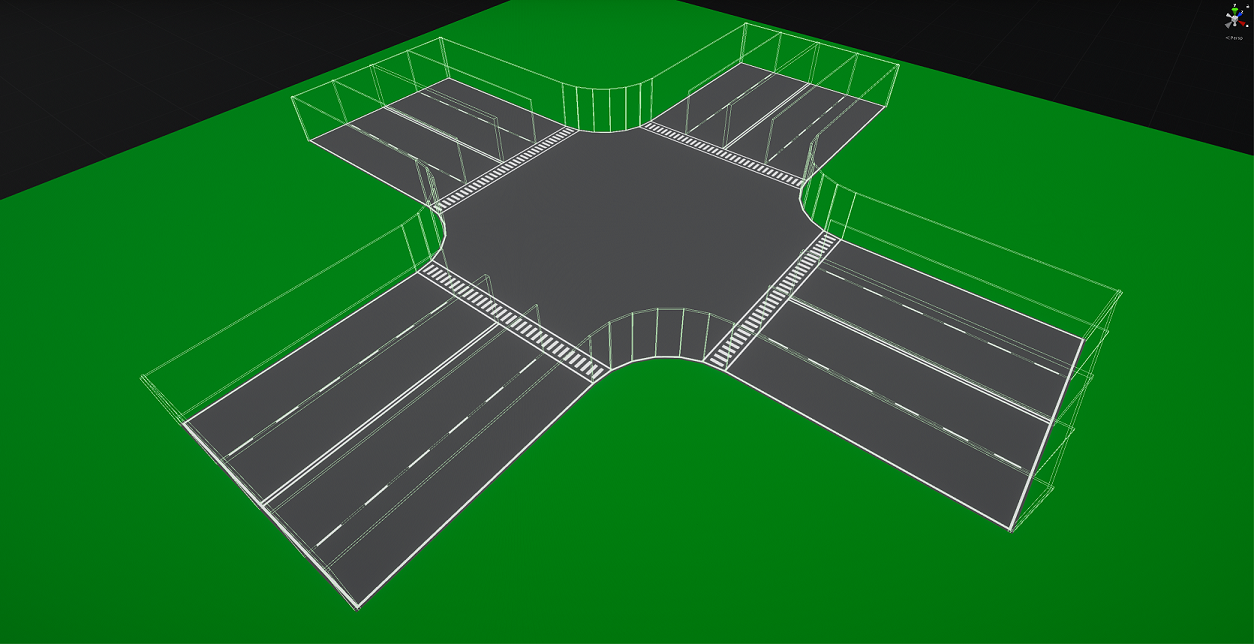}}
			\subfigure[]{\includegraphics[width=0.24\textwidth]{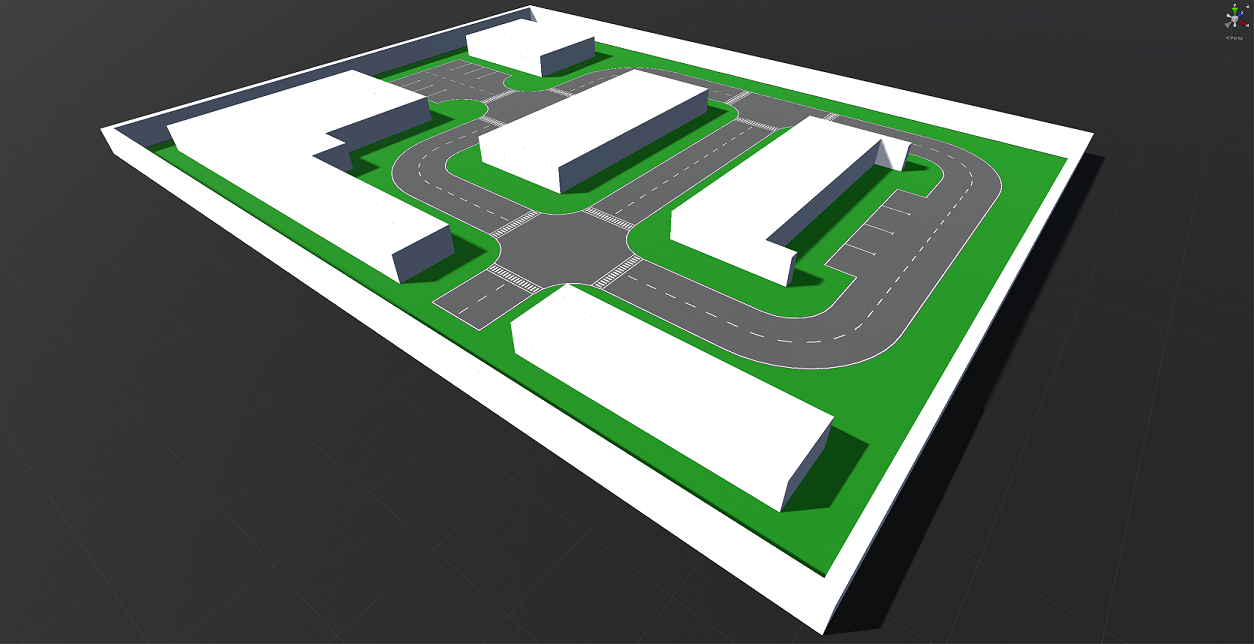}}
			\caption{Simulator Maps: (a) Driving School, (b) Parking School, (c) Intersection School, and (d) Tiny Town}
			\label{Figure: Simulator Maps}
		\end{figure}

		The CAD models of traffic elements were converted into mesh models, which, along with the environment modules were imported into Unity Editor. Similar to the simulated vehicle, their material properties were also tuned to generate photorealistic visual appearance. Each module was assigned with physical attributes such as mesh colliders for detecting interactions, and rigid body component for simulating gravity and other forces acting on the modules (refer Figure \ref{Figure: Infrastructure Simulation}). Additionally, the traffic light module was provided with physical light objects to simulate realistic raycasting and diffusion. Finally, these simulated infrastructure modules were used to design various driving scenarios; besides the two preconfigured maps supported by AutoDRIVE Testbed, the simulator provides additional two, viz. Intersection School and Tiny Town (refer Figure \ref{Figure: Simulator Maps}).
		
	\subsection{Simulator Features}
	\label{Sub-Section: Simulator Features}
	
		AutoDRIVE Simulator exploits NVIDIA's multi-threaded PhysX engine to simulate kinematics and dynamics of all the rigid bodies present within a particular scene. It also exploits Unity's High-Definition Render Pipeline (HDRP) and Post-Processing Stack to render enhanced photorealistic graphics. This way, AutoDRIVE bridges the gap between software simulation and hardware deployment.
		
		The simulator offers a full-duplex communication bridge so as to establish a connection with the externally developed autonomy algorithms. The bridge was implemented using WebSocket, which allows event-driven responses over a single transmission control protocol (TCP) connection, while also minimizing the data overhead, thereby making it an apt communication protocol for the simulator. The bridge parameters default to \texttt{127.0.0.1} (i.e. localhost) internet protocol (IP) address with \texttt{4567} port number, and can be reconfigured to setup the bridge in local or distributed computing mode.
		
		\begin{figure}[htpb]
			\centering
			\subfigure[]{\includegraphics[width=0.32\textwidth]{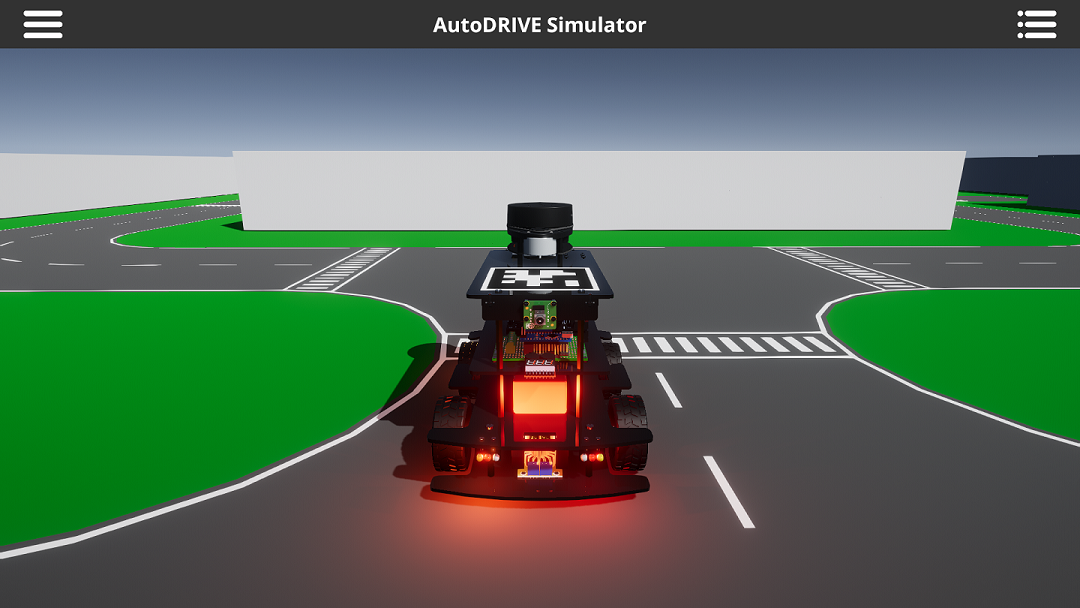}}
			\subfigure[]{\includegraphics[width=0.32\textwidth]{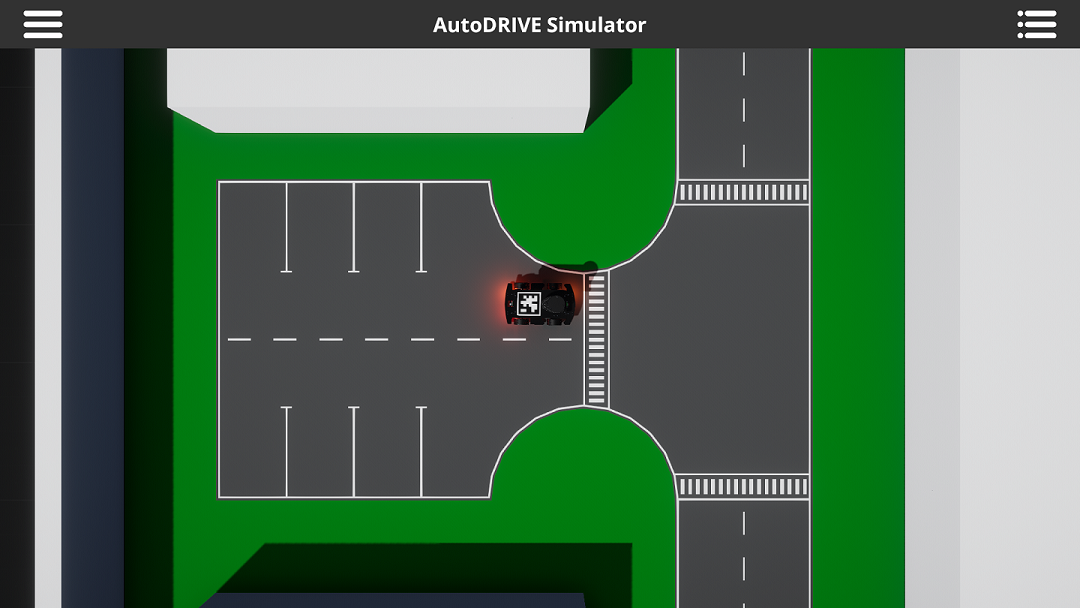}}
			\subfigure[]{\includegraphics[width=0.32\textwidth]{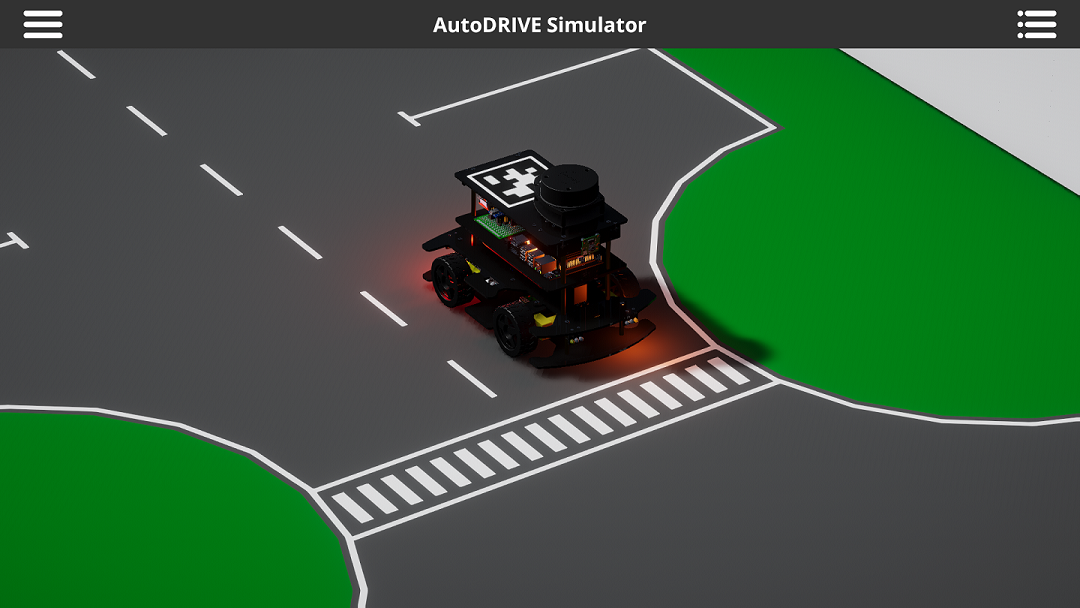}}
			\caption{Simulator Views: (a) Driver's Eye, (b) Bird's Eye, and (c) God's Eye}
			\label{Figure: Simulator Views}
		\end{figure}
		
		The simulator features three cameras, each providing a unique view (refer Figure \ref{Figure: Simulator Views}):
		
		\begin{itemize}
			\item \textbf{Driver's Eye:} Renders a vehicle-centric view from behind it, at a constant offset. It is equally recommended for comfortably visualizing the vehicle in manual and/or autonomous driving modes.
			\item \textbf{Bird's Eye:} Renders a vehicle-centric view from above it, which can be zoomed in/out (variable offset) depending on the required region of interest (ROI). This view is recommended for reverse driving and parking in manual mode, or getting a better judgement of the vehicle motion in autonomous mode.
			\item \textbf{God's Eye:} Renders a vehicle-centric perspective view from a statically positioned camera, which is able to track the vehicle by pivoting about its location, while also adjusting its focus for effective rendering. This view is recommended for autonomous driving, since it generates exceptional scenic visualization.
		\end{itemize}
	
		\begin{figure}[htpb]
			\centering
			\subfigure[]{\includegraphics[width=0.32\textwidth]{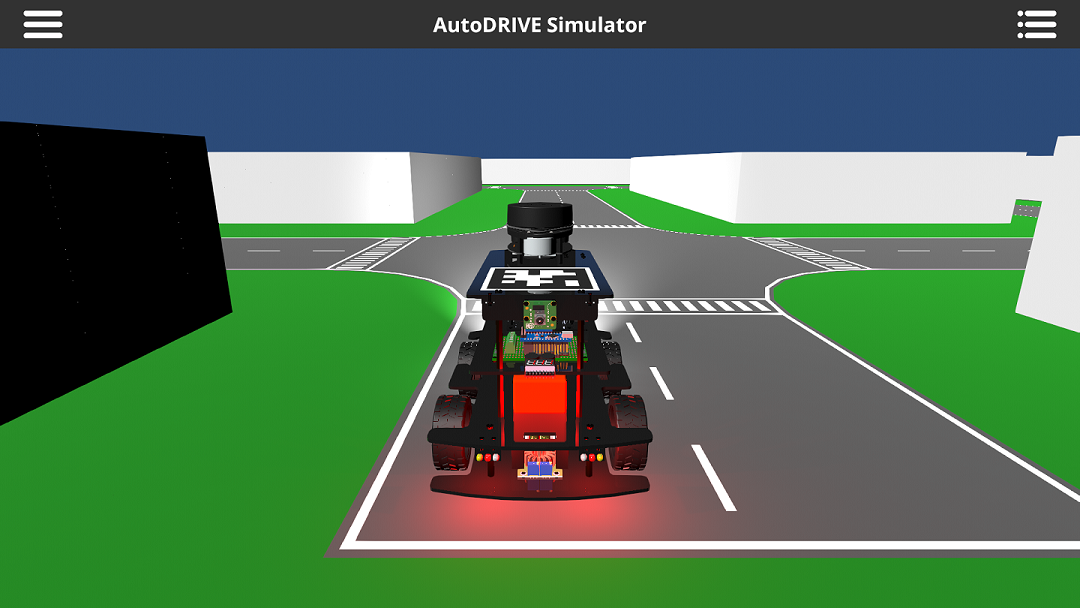}}
			\subfigure[]{\includegraphics[width=0.32\textwidth]{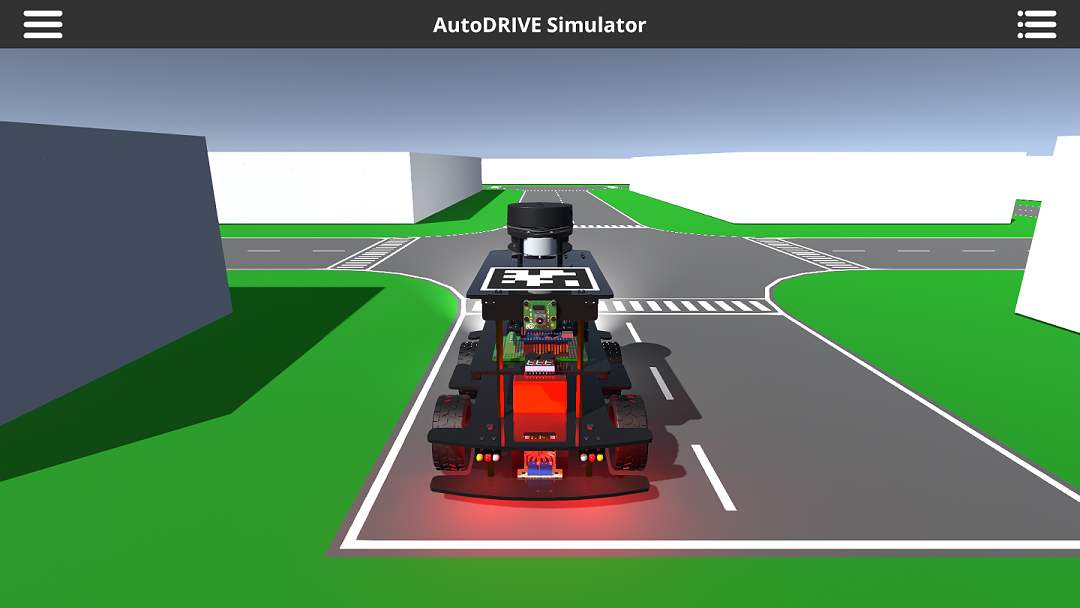}}
			\subfigure[]{\includegraphics[width=0.32\textwidth]{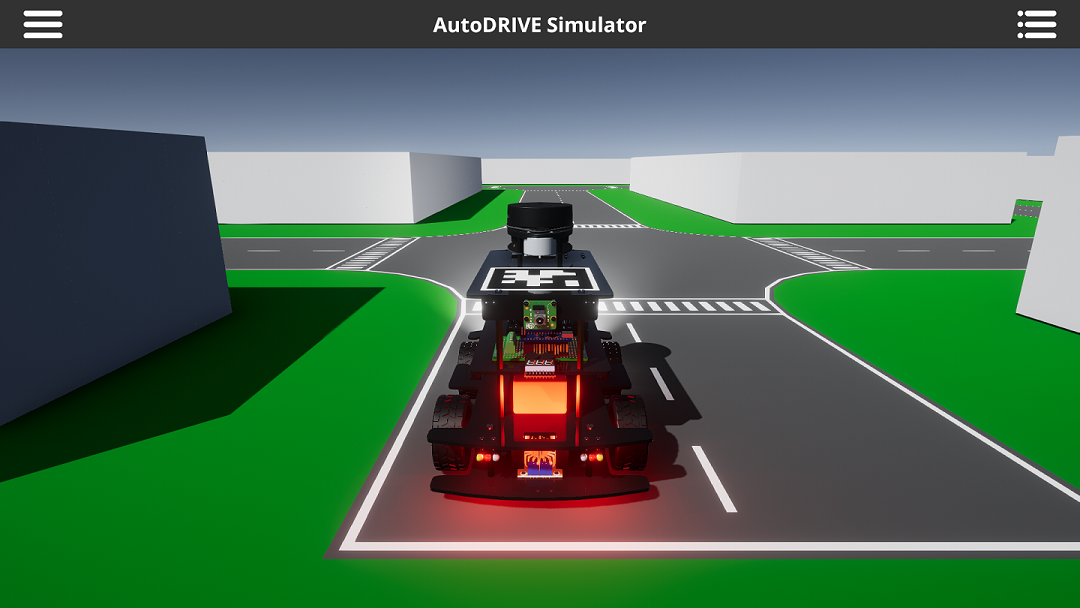}}
			\caption{Simulator Graphics Quality: (a) Low, (b) High, and (c) Ultra}
			\label{Figure: Simulator Graphics}
		\end{figure}
	
		Additionally, the simulator allows altering the graphics quality for smoother performance (refer Figure \ref{Figure: Simulator Graphics}). The rendering quality can be set based on the specifications of host machine and the application being targeted.
		
		\begin{itemize}
			\item \textbf{Low Quality:} This setting disables HDRP as well as Post-Processing Stack, thereby drastically reducing the graphical overhead. Additionally, it also disables shadows and skybox, which further relaxes the rendering process. This setting is particularly suitable for low-end systems, but is not recommended for computer vision applications.
			\item \textbf{High Quality:} This setting enables only HDRP along with the shadows and skybox, which generates moderate visual effects, and is generally a sweet-spot between quality and performance. It is recommended for most platforms and applications, with the exception of those targeting vision-based sim2real transfer.
			\item \textbf{Ultra Quality:} This setting enables HDRP and Post-Processing Stack along with the shadows and skybox. It is computationally very intensive and, as a result, is only recommended for high-end machines. Nevertheless, it is specifically useful when targeting vision-based sim2real applications as it generates photorealistic graphics (adequate domain randomization may still be required).
		\end{itemize}
	
		\begin{figure}[htpb]
			\centering
			\subfigure[]{\includegraphics[width=0.49\textwidth]{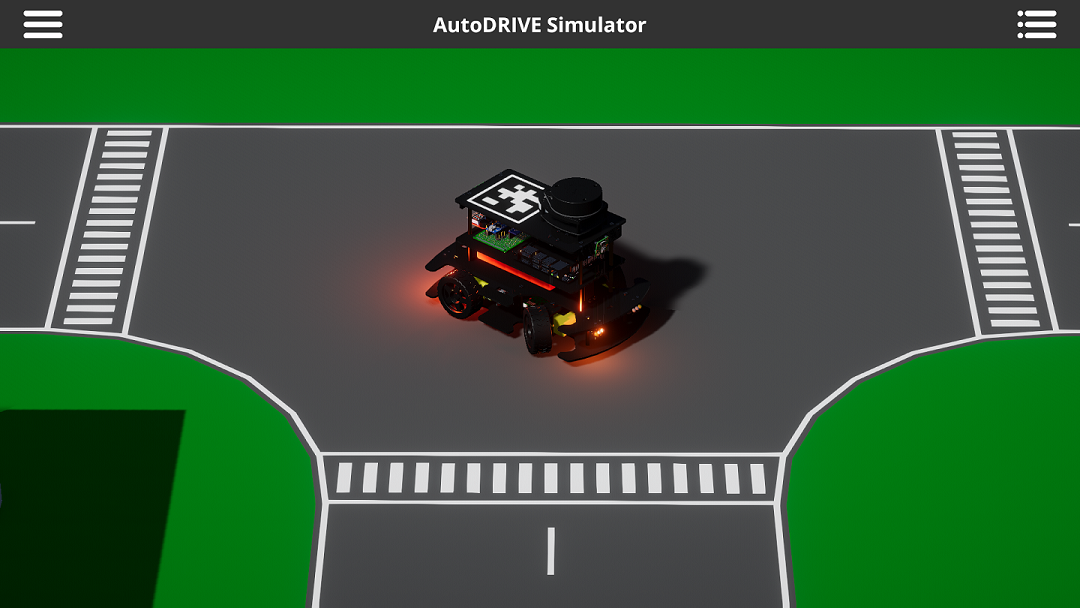}}
			\subfigure[]{\includegraphics[width=0.49\textwidth]{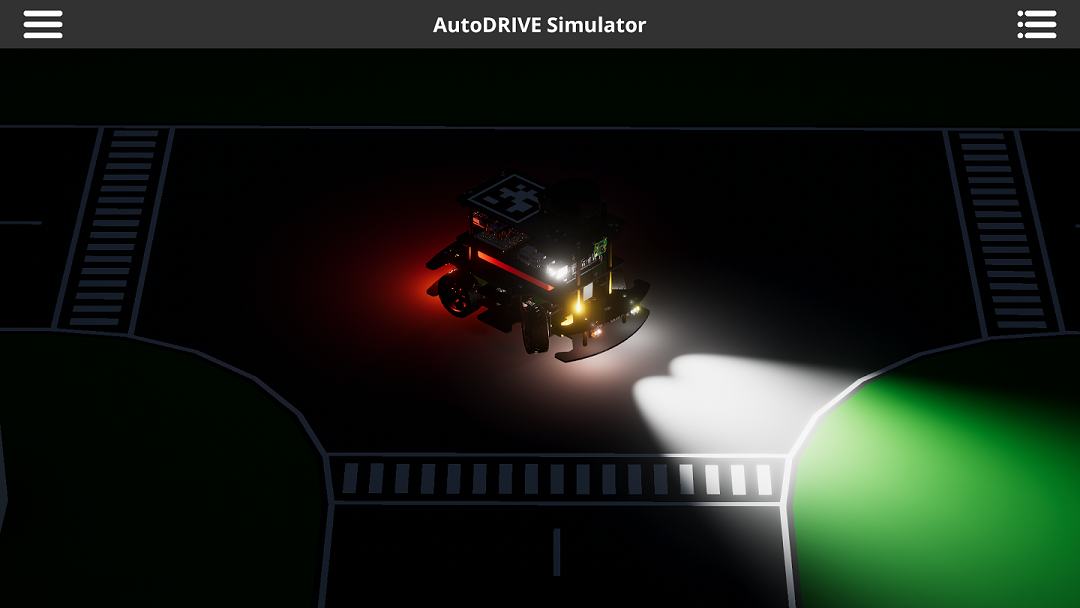}}
			\caption{Simulator Scene Light: (a) Enabled, and (b) Disabled}
			\label{Figure: Simulator Scene Light}
		\end{figure}
	
		The simulator also allows toggling the scene light on or off to simulate day and night driving conditions, respectively (refer Figure \ref{Figure: Simulator Scene Light}).
		
		\begin{figure}[htpb]
			\centering
			\includegraphics[width=\textwidth]{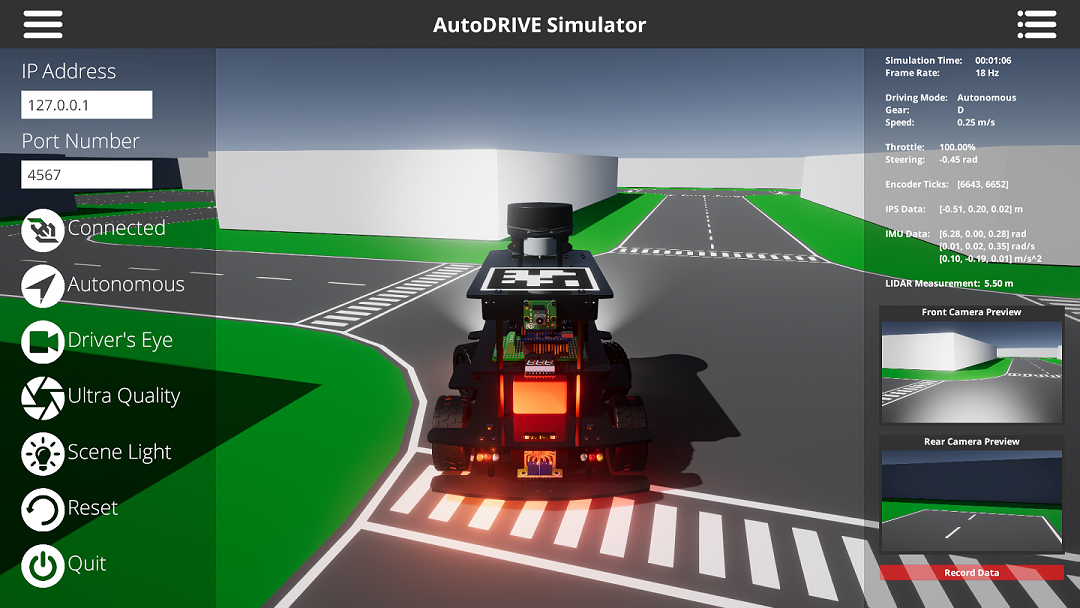}
			\caption{Graphical User Interface of AutoDRIVE Simulator}
			\label{Figure: GUI}
		\end{figure}
	
		The simulator features an interactive graphical user interface (GUI) comprising of a menu panel, on left-hand side and a heads up display (HUD), on right-hand side. The former hosts input fields for controlling various features of the simulator. The later, on the other hand, displays prominent simulation parameters along with vehicular stats and sensory data in real-time. It also hosts the data recording functionality, which can be used to export sensory data of the vehicle for a specific run. Figure \ref{Figure: GUI} depicts the simulator's GUI, with both menu and HUD panels enabled.
		
		Finally, it is worth mentioning that the simulator, in its source form, natively supports C\# scripting. Additionally, it has also been integrated with Unity ML Agents Toolkit \cite{ML-Agents2018}, a machine learning framework for developing learning-based applications directly from within the simulator.
	
\section{AutoDRIVE Devkit}
\label{Section: AutoDRIVE Devkit}

	As described earlier, AutoDRIVE Devkit is a collection of frameworks, tools and libraries aiding in rapid and flexible development of autonomy algorithms targeted towards the hardware testbed and/or software simulator. Consequently, its development required writing software stacks for autonomous driving as well as smart city management and providing the necessary communication interfaces. The following sections describe the software development in a greater detail.
	
	\subsection{Autonomous Driving Software Stack}
	\label{Sub-Section: ADSS}
	
		\begin{figure}[htpb]
			\centering
			\includegraphics[width=\textwidth]{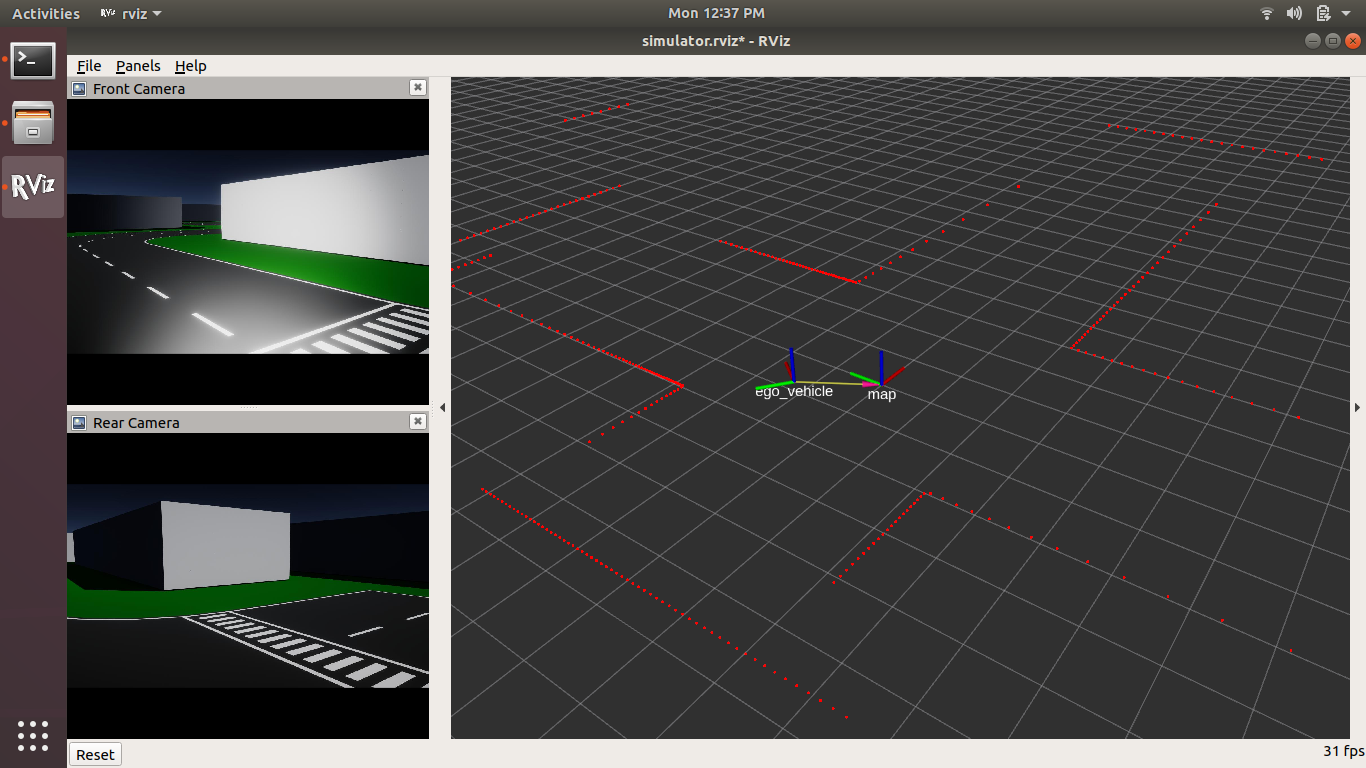}
			\caption{ADSS: A Sample RViz Application Instance}
			\label{Figure: ADSS}
		\end{figure}
	
		The autonomous driving software stack (ADSS) aids in development of autonomy algorithms specifically targeting the vehicle. It offers a modular software package compatible with ROS, along with C++ and Python APIs for convenience.
		
		The ROS package supports modular and flexible algorithm development targeted towards autonomous driving. It can be installed on a Linux based workstation for interfacing with the simulator (locally/remotely), and can be natively installed on Nigel's onboard computer for hardware deployment. Figure \ref{Figure: ADSS} depicts a sample use case of the ROS package for visualizing vehicle state and sensory data on the server-side.
		
		The scripting APIs for Python and C++, on the other hand, can be exploited to develop high-performance autonomous driving algorithms, without ROS as an intermediary. Such source codes can be interfaced with the simulator using a third-party WebSocket library, or can be directly deployed on Nigel's onboard computer for hardware validation.
	
	\subsection{Smart City Software Stack}
	\label{Sub-Section: SCSS}
	
		\begin{figure}[htpb]
			\centering
			\includegraphics[width=\textwidth]{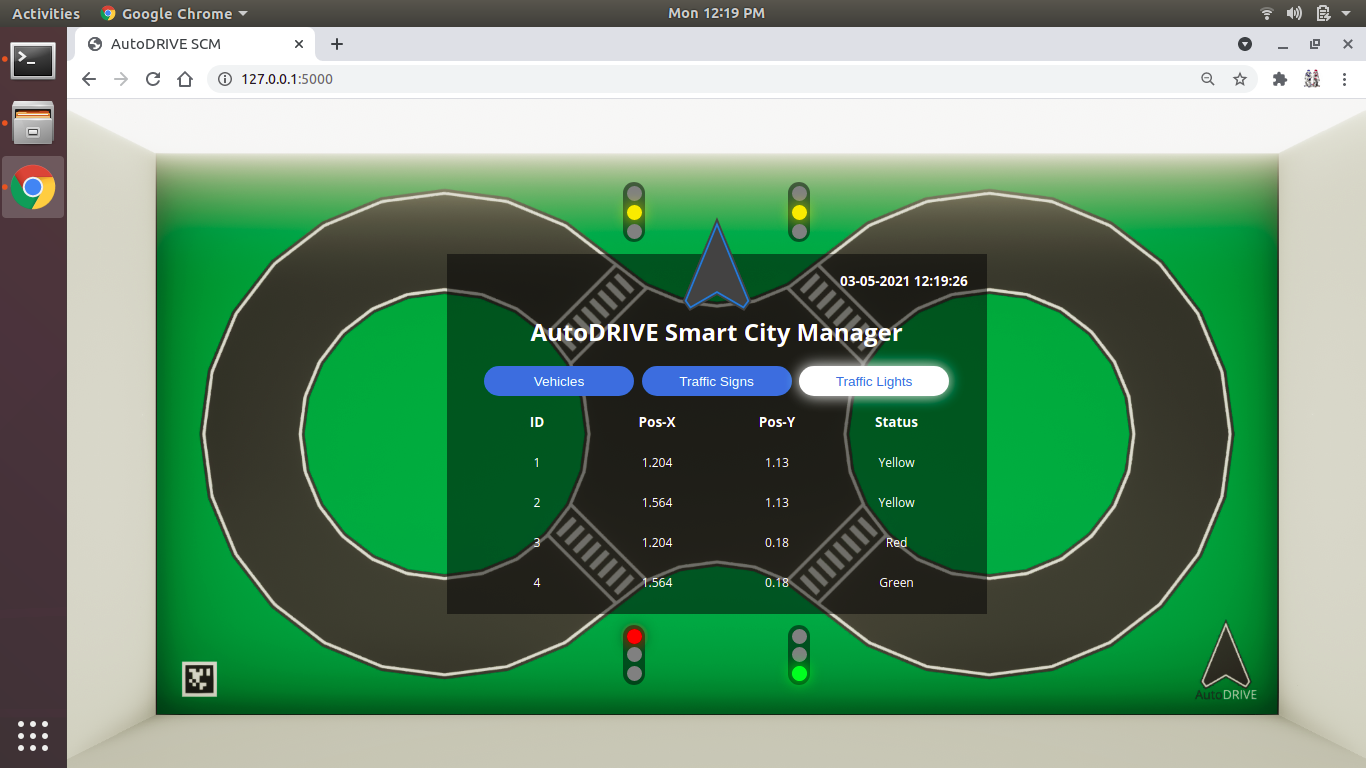}
			\caption{SCSS: A Sample SCM Webapp Instance}
			\label{Figure: SCSS}
		\end{figure}
	
		The smart city software stack (SCSS) aids in development of autonomy algorithms specifically targeting the infrastructure. It can also work in tandem with ADSS to develop smart city applications pertaining to autonomous traffic management.
		
		AutoDRIVE Devkit offers a centralized smart city manager (SCM) server to monitor and control various \textit{``smart''} elements. The server hosts a database to keep track of all the vehicles along with active and passive traffic elements present within a particular scene. It is currently capable of identifying individual vehicles and tracking their 2D pose estimates in real-time. It also stores the positional coordinates of traffic signs and lights, along with their states (sign names and light status).
		
		Additionally, the SCM server also provides an interactive webapp, allowing the users to connect with the database for monitoring and controlling the traffic flow in real-time. Figure \ref{Figure: SCSS} depicts a sample use case of the AutoDRIVE SCM webapp for controlling the traffic lights manually.

\clearpage


\chapter{PLATFORM EXPLOITATION}
\label{Chapter: Platform Exploitation}

\section{Autonomous Parking}
\label{Section: Autonomous Parking}

The application of autonomous parking was chosen to demonstrate AutoDRIVE's ability to support development and validation of modular autonomy algorithms (refer Section \ref{Sub-Section: Modular Approach}). For this particular demonstration, the probabilistic robotics approach was followed involving the implementation of following component algorithms:

\begin{itemize}
	\item Odometry
	\item Simultaneous localization and mapping (SLAM)
	\item Probabilistic map-based localization
	\item Autonomous navigation (path planning and motion control)
\end{itemize}

Although AutoDRIVE Simulator offers extremely high fidelity simulation, for implementation of these algorithms, AutoDRIVE Testbed was chosen, instead, to demonstrate hardware capabilities and, more importantly, to tackle the real-world uncertainty associated with perception, planning and control. The entire software stack was developed exploiting the ROS package from AutoDRIVE Devkit.

This application demonstrates a single-agent scenario, wherein the vehicle's objective was to map an unknown environment, localize within that map and navigate safely from a particular location (initial pose) to a specified destination (parking pose). The only source of perception for the vehicle was a planar radial LIDAR unit mounted on top of it. Based on the odometry and localization information extracted from laser scans (point cloud data) and depending on the parking pose specified, the path planner had to generate global and local trajectories in order to avoid collision with any static or dynamic obstacles in the scene. Finally, the control system had to generate appropriate actuation commands for the final control elements (throttle/brake and steering) in order to achieve a complete two-dimensional control for tracking the local reference trajectory.

\subsection{Odometry}
\label{Sub-Section: Odometry}
Since LIDAR was the only sensing modality used for this implementation (see Section \ref{Section: Autonomous Parking}), conventional odometry based on wheel motion (using incremental encoders) was not an option. Therefore, inspired by optical flow principle used in visual odometry applications, we implemented range flow based odometry estimation algorithm \cite{RF2O2016} by correlating consecutive laser scans obtained from the planar radial LIDAR unit onboard.

Odometry requires initial position estimate of the vehicle (referenced to zero) and recursive velocity estimation to track the position with respect to the said initial reference. Since LIDAR and vehicle reference coordinate systems share a static rigid body transform, it is hereinafter assumed that estimating LIDAR odometry is sufficient to compute vehicle odometry using a single coordinate transformation.

\begin{figure}[htpb]
	\centering
	\includegraphics[width=\textwidth]{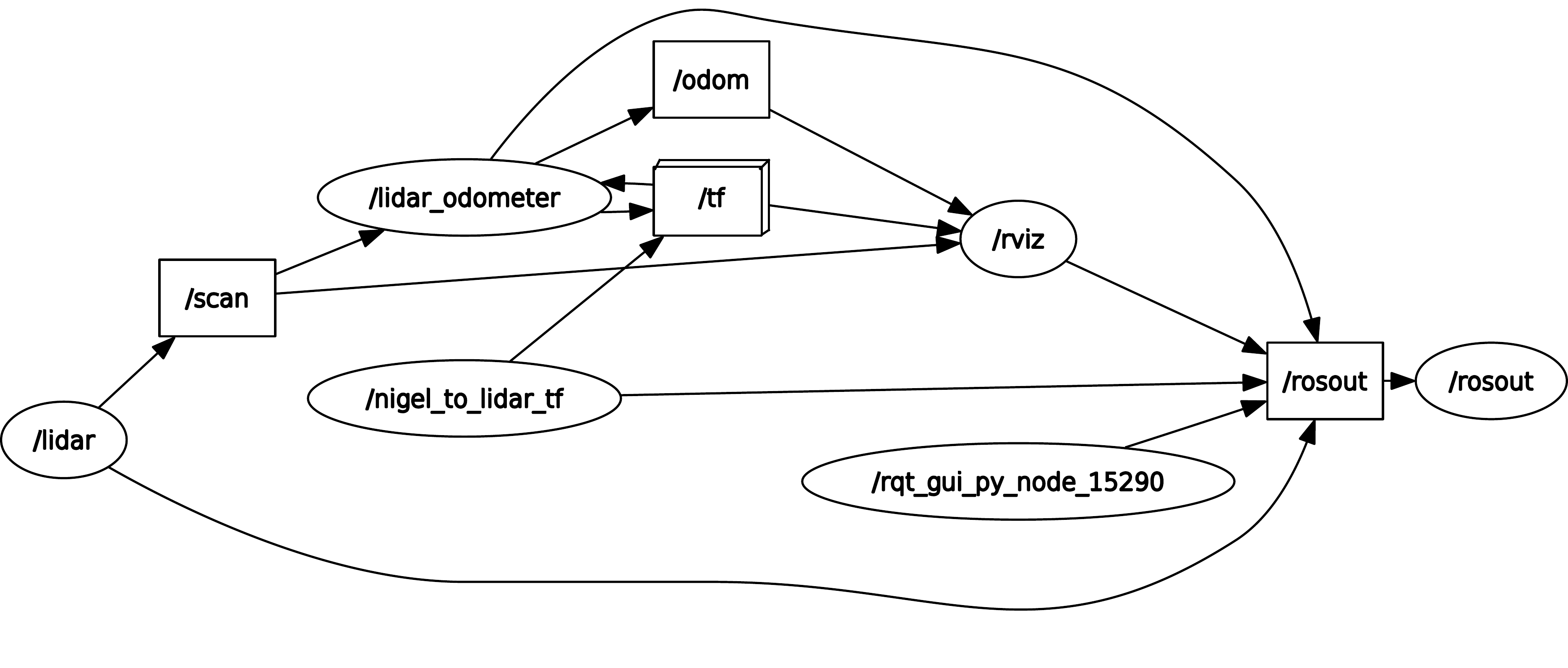}
	\caption{ROS Computation Graph for Odometry}
	\label{Figure: Odometry RQT Graph}
\end{figure}

Figure \ref{Figure: Odometry RQT Graph} depicts the computation graph illustrating the underlying processes during LIDAR based range flow odometry. The entire odometry algorithm, including the mathematical details is elucidated in the following text.

Initially, assuming that the surrounding is static, the velocity of LIDAR can be estimated based on the apparent motion that it observes. If $L\left(t,\alpha\right)$ is the laser scan at time $t$ with scan coordinate $\alpha\in[0,N)$, where $N$ is the size of scan, then position of an arbitrary point $P$ in reference to LIDAR's coordinate frame can be represented using polar coordinates $r$ and $\theta$.

Thus, the relationship between scan coordinate ($\alpha$) and angular coordinate of the point ($\theta$) is given by
\begin{equation}
\label{Equation: 5.1}
\alpha=\left(\frac{N-1}{f}\right)\theta+\left(\frac{N-1}{2}\right)=k_\alpha\theta+\left(\frac{N-1}{2}\right)
\end{equation}

where $k_\alpha=\left(\frac{N-1}{f}\right)$ and $f$ is the field of view of the LIDAR unit.

By the virtue of relative motion, point $P$ moves to $P'$ w.r.t. LIDAR in time interval $\Delta t$ corresponding to a new laser scan. Thus, a linear range flow constraint can be derived based on geometric resemblance of the two consecutive laser scans. Now, if we assume $L$ is differentiable, then range of a point in the recent laser scan could be represented by the following expression
\begin{equation}
\label{Equation: 5.2}
L(t+\Delta t,\alpha+\Delta\alpha)=L(t,\alpha)+\frac{\partial L}{\partial t}(t,\alpha)\Delta t+\frac{\partial L}{\partial \alpha}(t,\alpha)\Delta \alpha+O(\Delta t^2,\Delta \alpha^2)
\end{equation}

Neglecting the higher order terms in the above expression (Equation \ref{Equation: 5.2}) a simplified equation relating scan gradients with $r$ and $\alpha$ of a point over time interval of $\Delta t$ can be obtained as follows
\begin{equation}
\label{Equation: 5.3}
\frac{\Delta L}{\Delta t}\simeq L_t+L_\alpha\frac{\Delta \alpha}{\Delta t}
\end{equation}

where $\Delta L=L(t+\Delta t,\alpha+\Delta\alpha)-L(t,\alpha)$, $L_t=\frac{\partial L}{\partial t}(t,\alpha)$ and $L_\alpha=\frac{\partial L}{\partial \alpha}(t,\alpha)$.

Considering $\dot{r}=\frac{\Delta L}{\Delta t}$ and $\dot{\alpha}=\frac{\Delta \alpha}{\Delta t}$, Equation \ref{Equation: 5.3} can be rewritten using relationship defined in Equation \ref{Equation: 5.1} to obtain the range flow constraint equation (Equation \ref{Equation: 5.4}) as follows
\begin{equation}
\label{Equation: 5.4}
\dot{r}\simeq L_t+L_\alpha=L_t+L_\alpha k_\alpha\dot{\theta}
\end{equation}

Since all the rigid body transforms are specified in Cartesian coordinate system, we transform the velocities $\dot{r}$ and $\dot{\theta}$ of all the points in the laser scan to Cartesian representation $(\dot{x},\dot{y})$ using the following set of equations
\begin{equation}
\label{Equation: 5.5}
\left\{\begin{matrix}
\dot{r}=\dot{x}\cos\theta+\dot{y}\sin\theta\\ 
r\dot{\theta}=\dot{y}\cos\theta-\dot{x}\sin\theta
\end{matrix}\right.
\end{equation}

Next, we impose the rigidity hypothesis i.e. assuming every apparent motion observed by LIDAR is due to its own translation or rotation, we obtain
\begin{equation}
\label{Equation: 5.6}
\begin{bmatrix}\dot{x}\\ \dot{y}\end{bmatrix}=\begin{bmatrix}-v_{x,s}+y\omega_s\\ -v_{y,s}-x\omega_s\end{bmatrix}
\end{equation}

where $\xi_s=[v_{x,s},v_{y,s},\omega_s]$ is a 2D twist representing velocity of sensor (i.e. LIDAR) and $[\dot{x},\dot{y}]$ is the velocity of the point $P$ in Cartesian coordinate system.

Finally, substituting Equation \ref{Equation: 5.5} into Equation \ref{Equation: 5.4} and imposing Equation \ref{Equation: 5.6}, we can transform Equation \ref{Equation: 5.4} into sensor velocity constraint as follows
\begin{equation}
\label{Equation: 5.7}
\left(\cos\theta+\frac{L_\alpha k_\alpha\sin\theta}{r}\right)v_{x,s}+\left(\sin\theta-\frac{L_\alpha k_\alpha\cos\theta}{r}\right)v_{y,s}+(x\sin\theta-y\cos\theta-L_\alpha k_\alpha)\omega_s+L_t=0
\end{equation}

Thus, as evident from Equation \ref{Equation: 5.7}, a single point in the laser scan imposes 3 linearly independent restrictions to the sensor motion and would theoretically suffice to estimate the LIDAR odometry.

In practice however, several factors such as presence of dynamic objects, noisy range measurements and linear approximations in Equation \ref{Equation: 5.3} make it nearly impossible to estimate sensor motion with just 3 restrictions. We therefore use all points in the laser scan to estimate LIDAR odometry. For a given sensor twist $\xi$, we define a geometric residual $\rho(\xi)$ for every point by evaluating Equation \ref{Equation: 5.7} as follows
\begin{equation}
\label{Equation: 5.8}
\rho(\xi)=L_t+(x\sin\theta-y\cos\theta-L_\alpha k_\alpha)\omega+\left(\cos\theta+\frac{L_\alpha k_\alpha\sin\theta}{r}\right )v_x+\left(\sin\theta-\frac{L_\alpha k_\alpha\cos\theta}{r}\right)v_y
\end{equation}

Next, we define the optimization problem of minimizing all the geometric residuals in order to accurately estimate sensor motion (odometry) as follows
\begin{align}
\label{Equation: 5.9}
\xi_M=\argmin_{\xi}\sum_{i=1}^{N}F(\rho_i(\xi))
\end{align}

where $k$ is a tunable constant and $F(\rho)=\frac{k^2}{2}\ln\left(1+\left(\frac{\rho}{k}\right)^2\right)$ is the Cauchy M-estimator with associated weights $w(\rho)=\frac{1}{1+\left(\frac{\rho}{k}\right)^2}$.

The system is solved with IRLS method until convergence is achieved, wherein the geometric residuals $\rho(\xi)$ are iteratively computed and the associated weights $w(\rho)$ are consecutively updated.

At this stage there is only one limiting factor, which is the simplified linearization of Equation \ref{Equation: 5.2} into Equation \ref{Equation: 5.3}, since this approximation is only valid for constant range gradients and small inter-scan displacement. In order to overcome this, we estimate odometry using coarse to fine approach, wherein coarser (initial) levels give a crude estimate and the finer (deeper) levels improve it subsequently.

Let us assume that $L_1$ and $L_2$ are two consecutive laser scans. We begin by constructing two Gaussian pyramids by successively downsampling these original scans using a bilateral filter. Next, the motion estimation problem as defined in Equation \ref{Equation: 5.9} is iteratively solved using coarse-to-fine scheme. Every time we proceed to a subsequent (finer) level, the second scan $L_2$ is warped against the first scan $L_1$ based on the velocity (twist) estimated in the previous level ($\xi_p$).

In order to achieve this, a rigid body transformation governed by $\xi_p$ is first applied to every point $P$ in scan $L_2$ as follows
\begin{equation}
\label{Equation: 5.10}
\begin{bmatrix}x^w\\ y^w\\ 1\end{bmatrix}=e^{\hat{\xi_p}}\begin{bmatrix}x\\ y\\ 1\end{bmatrix},
\hat{\xi_p}=\Delta t\begin{bmatrix}0 & -\omega_p & v_{x,p}\\ \omega_p & 0 & v_{y,p}\\ 0 & 0 & 0\end{bmatrix}
\end{equation}

Next, the transformed points $[x^w,y^w]$ are re-projected on the original second scan $L_2$ so as to obtain its warped scan $L_2^w$ as follows
\begin{equation}
\label{Equation: 5.11}
L_2^w(\alpha^w)=\sqrt{(x^w)^2+(y^w)^2}, \alpha^w=k_\alpha \arctan\left(\frac{y^w}{x^w}\right)+\frac{N-1}{2}
\end{equation}

Thus, as the estimate $\xi_p$ approaches true sensor velocity, the warped second scan $L_2^w$ moves closer to first original scan $L_1$ thereby reducing the inter-scan distance and allowing us to linearize Equation \ref{Equation: 5.2} for estimating the odometry.

Lastly, the static transform of vehicle reference frame w.r.t. LIDAR reference frame is used to estimate vehicle odometry from LIDAR odometry (as discussed earlier).

\begin{figure}[htpb]
	\centering
	\includegraphics[width=\textwidth]{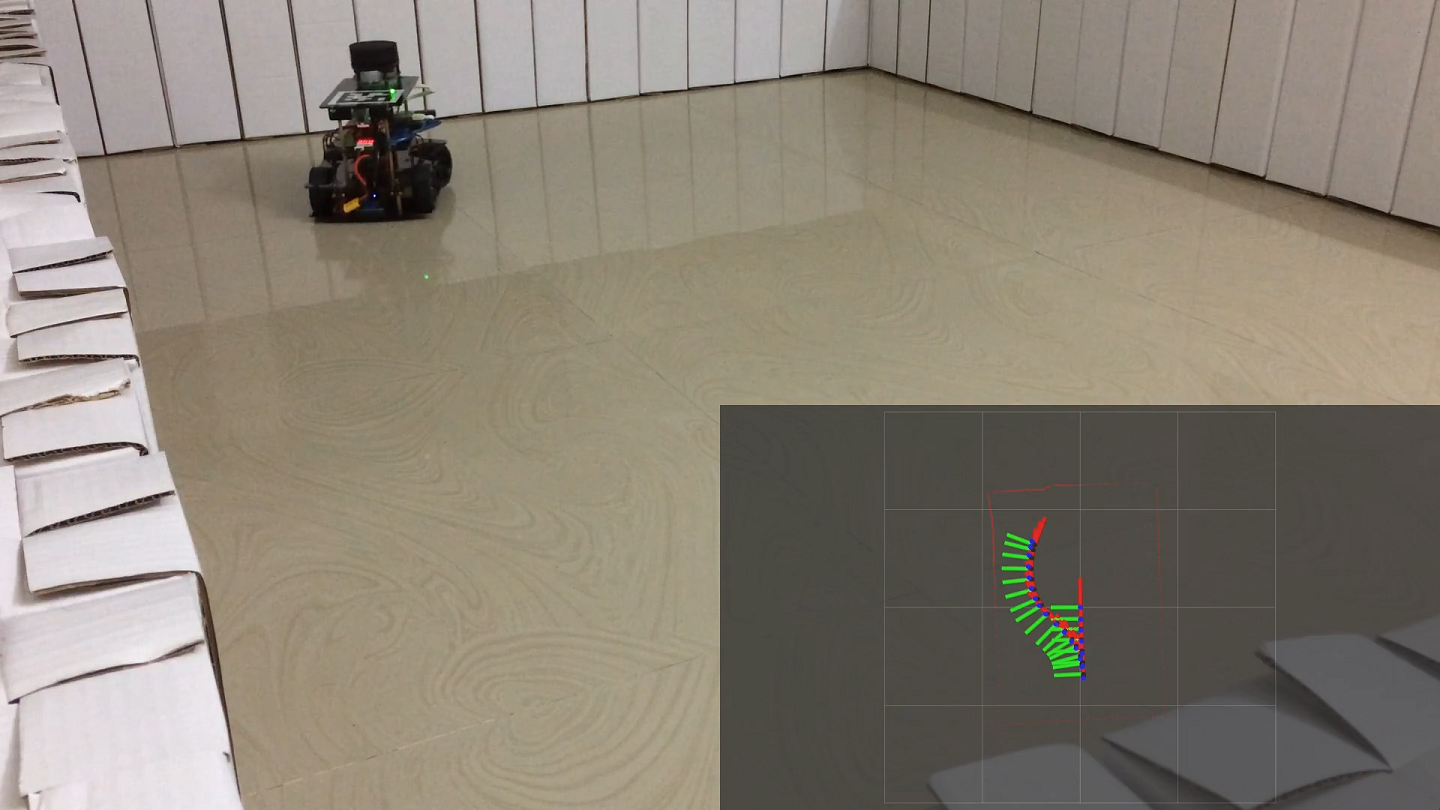}
	\caption{Odometry Demonstration using AutoDRIVE Testbed}
	\label{Figure: Odometry Demo}
\end{figure}

Figure \ref{Figure: Odometry Demo} illustrates Nigel being teleoperated in the scene with range flow odometry in action. The time-synchronized RViz window in the bottom-right corner shows the current laser scan and pose estimate of the vehicle being updated based on LIDAR odometry.

\subsection{Simultaneous Localization and Mapping (SLAM)}
\label{Sub-Section: SLAM}
SLAM is an excellent approach for mapping an unknown environment online with no other resources than observations from a (moving) vehicle exploring the environment. It is fundamentally based on the principle of building and updating a map while simultaneously tracking the vehicle pose w.r.t. it. Herein, a map is built from the initial observation and localization estimate (generally referenced to zero), the localization estimate is updated w.r.t. generated map and then the map is recursively updated based on new observations.

For this particular application, we implemented and configured the Hector SLAM algorithm \cite{HectorSLAM2011} in order to perform simultaneous localization and mapping purely based on the laser scans obtained from LIDAR unit onboard the vehicle.

\begin{figure}[htpb]
	\centering
	\includegraphics[width=\textwidth]{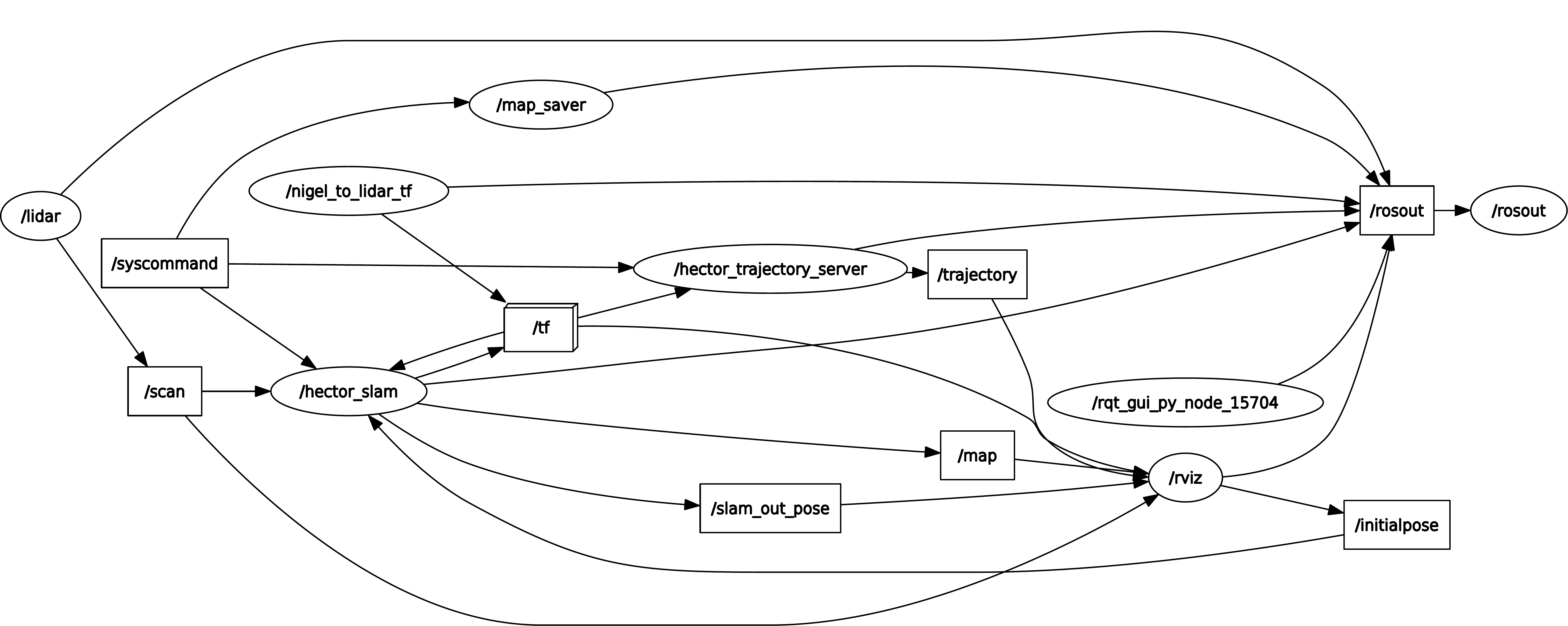}
	\caption{ROS Computation Graph for SLAM}
	\label{Figure: SLAM RQT Graph}
\end{figure}

Figure \ref{Figure: SLAM RQT Graph} depicts the computation graph illustrating the underlying processes during simultaneous localization and mapping. The process starts with first laser scan being registered as initial map. Then, based on the LIDAR scan rate and time elapsed since the previous scan, the vehicle pose change is estimated from transformation of updated laser scan. Later, scan matching is performed in order to align the laser scan with map built thus far and as a result, the map is updated based on new pose estimate. This entire process is continued until successful mapping of the environment is accomplished. The further details are elucidated below.

Based on the initial laser scan, the occupancy probabilities (of the occupancy grid map) and the spatial derivatives are estimated using bilinear filtering. Here, each cell of the grid map can be thought of as a sample of the occupancy probability distribution. Thus, for a particular map coordinate $P_m$ the probability of occupancy can be expressed as $M(P_m)$ and the spatial gradient as $\nabla M(P_m)=\left(\frac{\partial M}{\partial x}(P_m),\frac{\partial M}{\partial y}(P_m)\right)$.

Next, we have to optimally align the current laser scan with the map built so far. Hence, we need to find the transformation $\xi=[p_x,p_y,\psi]^T$ corresponding to the vehicle pose in the map (localization) which minimizes the following criterion
\begin{align}
\label{Equation: 5.12}
\xi^*=\argmin_{\xi}\sum_{i=1}^{n}[1-M(S_i(\xi))]^2
\end{align}
where $S_i(\xi)$ represent coordinates of a scanned point expressed as $s_i=[s_{i,x},s_{i,y}]^T$. That is
\begin{equation}
\label{Equation: 5.13}
S_i(\xi)=\begin{bmatrix}\cos\psi & -\sin\psi\\ \sin\psi & \cos\psi\end{bmatrix}\begin{bmatrix}s_{i,x}\\ s_{i,y}\end{bmatrix}+\begin{bmatrix}p_x\\ p_y\end{bmatrix}
\end{equation}

\begin{equation}
\label{Equation: 5.14}
\therefore \frac{\partial S_i(\xi)}{\partial \xi}=\begin{bmatrix}1 & 0 & -\sin\psi\cdot s_{i,x}-\cos\psi\cdot s_{i,y}\\ 0 & 1 & \cos\psi\cdot s_{i,x}-\sin\psi\cdot s_{i,y}\end{bmatrix}
\end{equation}

Thus, the idea is to find $\Delta\xi$ which optimizes the scan matching error, given the initial pose estimate of the vehicle $\xi$ according to the following law
\begin{equation}
\label{Equation: 5.15}
\sum_{i=1}^{n}[1-M(S_i(\xi+\Delta\xi))]^2\rightarrow 0
\end{equation}

First order expansion (Taylor series) of the term $M(S_i(\xi+\Delta\xi))$ yields
\begin{equation}
\label{Equation: 5.16}
\sum_{i=1}^{n}\left[1-M(S_i(\xi))-\nabla M(S_i(\xi))\frac{\partial S_i(\xi)}{\partial \xi}\Delta\xi\right]^2\rightarrow 0
\end{equation}

To minimize the above criterion, we have to set the partial derivative w.r.t. variable $\Delta\xi$ to $0$, thus yielding
\begin{equation}
\label{Equation: 5.17}
2\sum_{i=1}^{n}\left[\nabla M(S_i(\xi))\frac{\partial S_i(\xi)}{\partial \xi}\right]^T\left[1-M(S_i(\xi))-\nabla M(S_i(\xi))\frac{\partial S_i(\xi)}{\partial \xi}\Delta\xi\right]=0
\end{equation}

Finally, solving for $\Delta\xi$ gives
\begin{equation}
\label{Equation: 5.18}
\Delta\xi=\textup{H}^{-1}\sum_{i=1}^{n}\left[\nabla M(S_i(\xi))\frac{\partial S_i(\xi)}{\partial \xi}\right]^T[1-M(S_i(\xi))]
\end{equation}
where $\textup{H}=\left[\nabla M(S_i(\xi))\frac{\partial S_i(\xi)}{\partial \xi}\right]^T\left[\nabla M(S_i(\xi))\frac{\partial S_i(\xi)}{\partial \xi}\right]$.\\

Using $\frac{\partial S_i(\xi)}{\partial \xi}$ from Equation \ref{Equation: 5.14} and map gradient $\nabla M(S_i(\xi))$ from bilinear filter approximation (as discussed earlier), Equation \ref{Equation: 5.18}) can be evaluated to descend towards the minimum by a stepping amount of $\Delta\xi$.

In other words, this operation gives us the vehicle pose change $\Delta\xi$ w.r.t. each time stamp. This can be used to update the vehicle pose in the map built so far (simultaneous localization) and transform the laser scan to this newly updated pose. The map is then updated based on the transformed laser scan (mapping).

\begin{table}[htpb]
	\centering
	\caption{Prominent Parameters for SLAM}
	\label{Table: SLAM Parameters}
	\resizebox{\textwidth}{!}{%
		\begin{tabular}{lll}
			\hline
			\textbf{Parameter}                         & \textbf{Value}                     & \textbf{Remarks}                    \\ \hline
			\tt map\_size                              & 80.0                               & Number of grid cells                                \\
			\tt map\_resolution                        & 0.05                               & Map resolution (m)                                  \\
			\tt map\_start\_x                          & 0.5                                & Origin of map reference frame on x-axis (middle)    \\
			\tt map\_start\_y                          & 0.5                                & Origin of map reference frame on y-axis (middle)    \\
			\tt map\_multi\_res\_levels                & 2.0                                & Multi-resolution map levels                         \\
			\tt free\_cell\_prob\_sat                  & 0.4                                & Free cell probability saturation limit (lower)      \\
			\tt ocpd\_cell\_prob\_sat                  & 0.9                                & Occupied cell probability saturation limit (upper)  \\
			\tt linear\_distance\_thresh               & 0.4                                & Linear displacement threshold for map update (m)    \\
			\tt angular\_distance\_thresh              & 0.06                               & Angular displacement threshold for map update (rad) \\
			\tt lidar\_min\_thresh                     & 0.15                               & Minimum detection threshold of LIDAR unit (m)       \\
			\tt lidar\_max\_range                      & 12.0                               & Maximum detection range of LIDAR unit (m)           \\
			\tt trajectory\_update\_rate               & 4.0                                & Update rate for the saved trajectory (Hz)           \\
			\tt trajectory\_publish\_rate              & 0.25                               & Publish rate for the visualized trajectory (Hz)     \\ \hline
		\end{tabular}%
	}
\end{table}

Although laser scans are pretty accurate, it is not a good idea to trust them 100\% while updating the map as discussed earlier. Therefore, the occupancy probabilities of map cells are saturated with upper and lower limits. 

\begin{figure}[htpb]
	\centering
	\includegraphics[width=\textwidth]{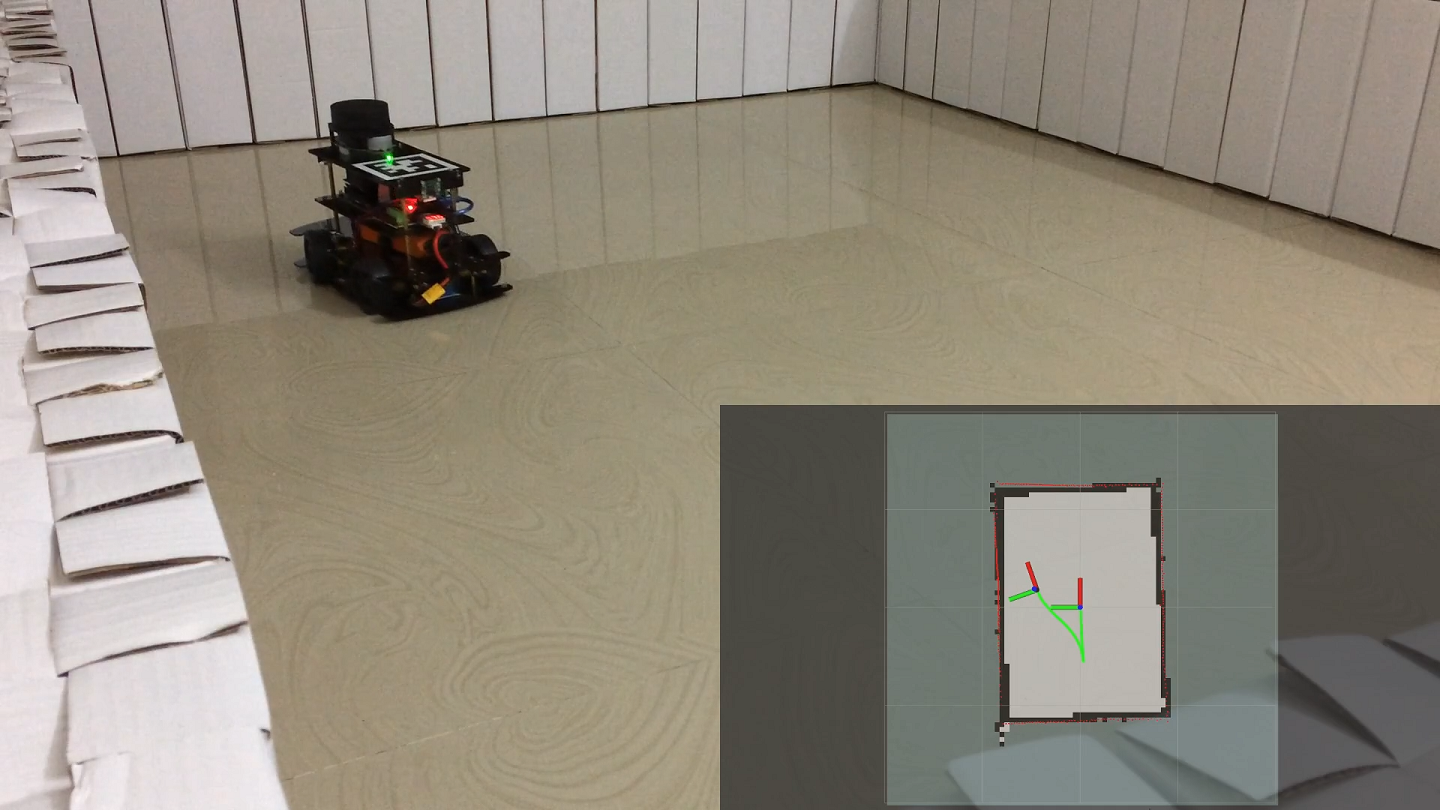}
	\caption{SLAM Demonstration using AutoDRIVE Testbed}
	\label{Figure: SLAM Demo}
\end{figure}

Additionally, in order to avoid getting stuck in local optima, multi-resolution map representation is used adhering to the coarse-to-fine scheme (similar to Section \ref{Sub-Section: Odometry}) wherein multiple occupancy grid maps (with each coarser level loosing resolution by 50\%) are simultaneously updated. The scan matching is started with the coarsest map level (which is available fastest in time) and the updated pose estimates of this level are used as base estimates for further levels. Table \ref{Table: SLAM Parameters} summarizes some of the prominent parameters of the SLAM system implemented for this particular application.

\begin{figure}[htpb]
	\centering
	\includegraphics[width=0.5\textwidth]{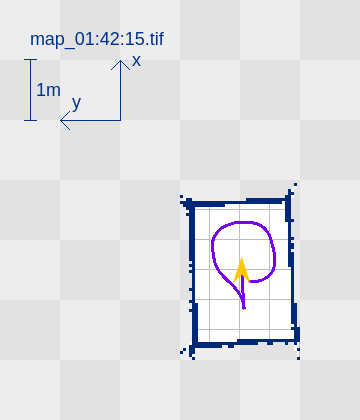}
	\caption{Final Map of the Autonomous Parking Scene}
	\label{Figure: SLAM Map}
\end{figure}

Figure \ref{Figure: SLAM Demo} illustrates Nigel being teleoperated in the scene with simultaneous localization and mapping in action. The time-synchronized RViz window in the bottom-right corner shows the occupancy grid map built so far, the current laser scan, map reference frame, vehicle reference frame (pose estimate) and trajectory followed by the vehicle so far. The final map (to the scale) of the scene obtained after performing SLAM is depicted in Figure \ref{Figure: SLAM Map}.

\subsection{Probabilistic Map-Based Localization}
\label{Sub-Section: Localization}
Once map of the environment was available (obtained using SLAM as discussed in Section \ref{Sub-Section: SLAM}), the next step was to implement a probabilistic map-based localization system that would later aid in the process of navigation for autonomous parking application.

\begin{figure}[htpb]
	\centering
	\includegraphics[width=\textwidth]{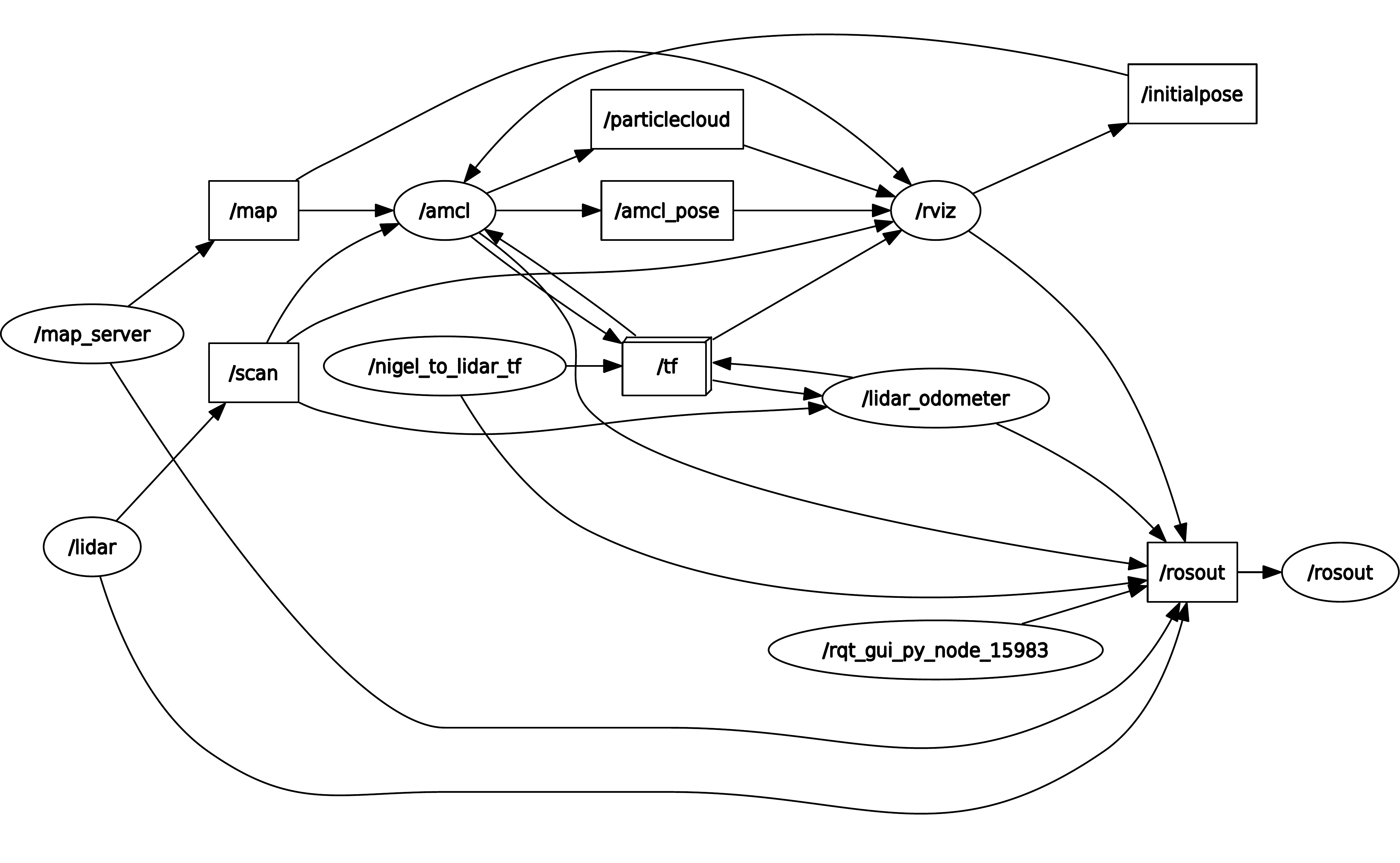}
	\caption{ROS Computation Graph for Localization}
	\label{Figure: Localization RQT Graph}
\end{figure}

For this particular application, we implemented Adaptive Monte Carlo Localization (AMCL) algorithm based on KLD-sampling particle filter \cite{AMCL2001}. Figure \ref{Figure: Localization RQT Graph} depicts the computation graph illustrating the high-level abstraction of the underlying processes during adaptive Monte Carlo localization.

\begin{algorithm}[]
	\SetAlgoLined
	\caption{Localization}
	\label{Algorithm: Localization}
	\KwIn{Samples $S_{t-1}=\left\{\left\langle x^{(i)}_{t-1},w^{(i)}_{t-1}\right\rangle|i=1,\cdots,n\right\}$ representing $Bel(x_{t-1})$, control inputs $u_{t-1}$, measurements $z_t$, bounds $\delta$ and $\varepsilon$, bin size $\Delta$}
	\KwOut{Samples $S_{t}$ representing $Bel(x_{t})$}
	\nl $S_{t}:=\emptyset$, $n:=0$, $k:=0$, $\alpha:=0$\\
	\nl \textbf{draw} $j(n)$ \textbf{with probability} $\propto w^{(i)}_{t-1}$\\
	\nl \textbf{sample} $x^{(i)}_{t}\sim p(x_t|x_{t-1},u_{t-1})$ \textbf{using} $x^{(j(n))}_{t-1}$ \textbf{and} $u_{t-1}$\\
	\nl $w^{(n)}_{t}:=p(z_t|x^{(n)}_{t})$\\
	\nl $\alpha:=\alpha+w^{(n)}_{t}$\\
	\nl $S_t:=S_t\cup \left\{\left\langle x^{(n)}_{t},w^{(n)}_{t}\right\rangle\right\}$\\
	\nl \If{($x^{(n)}_{t}\in$ \textup{empty bin} $b$)}{
		\nl $k:=k+1$\\
		\nl $b:=$ \textbf{non-empty}\\
	}
	\nl $n:=n+1$\\
	\nl \While{($n<\frac{1}{2\varepsilon}\chi^2_{k-1,1-\delta}$)}{
		\nl \For{($i\leftarrow 1,\cdots,n$)}{
			\nl $w^{(i)}_{t}:=\frac{w^{(i)}_{t}}{\alpha}$\\
		}
	}
	\nl \Return{$S_t$}\\
\end{algorithm}

Algorithm \ref{Algorithm: Localization} elucidates in-depth process of adaptive particle filter localization of the vehicle in a known map.

The very first line is the initialization step for the filter where we initialize the samples $S_t$ as a null set and all other variables to zero. Next, we generate samples until a pre-specified bound on the K-L distance $\varepsilon$ is reached. This involves the following steps.

We first perform re-sampling of previously available particles based on their normalized importance weights $w^{(i)}_{t-1}$ as a recursive measure to ensure convergence. Next, we predict the state $x_t$ based on previous state $x_{t-1}$ and control inputs $u_{t-1}$. The importance weights $w^{(n)}_{t}$ are then computed based on closeness of true and predicted measurements, and the normalization factor $\alpha$ is updated based on these new weights. We then insert the sample $\left\{\left\langle x^{(n)}_{t},w^{(n)}_{t}\right\rangle\right\}$ into the sample set $S_t$. Next, we determine the number of bins $k$ with non-zero probability and finally increment $n$.

We then check whether the K-L distance bound $\varepsilon$ is reached i.e. while $n<\frac{1}{2\varepsilon}\chi^2_{k-1,1-\delta}$, we normalize all the updated importance weights. Finally the new sample set $S_t$ is returned which holds state estimate of the vehicle at current time step $t$.

\begin{figure}[htpb]
	\centering
	\includegraphics[width=\textwidth]{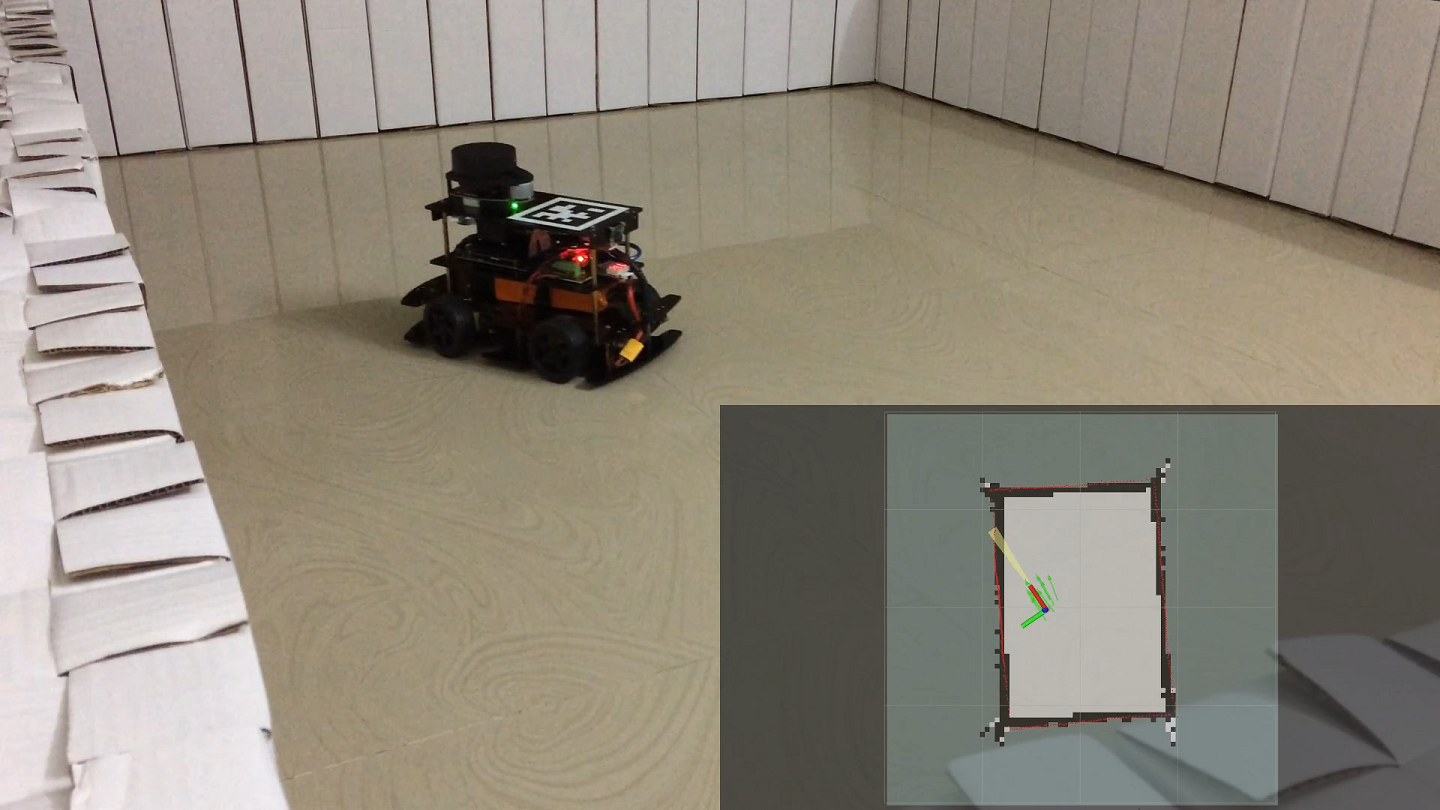}
	\caption{Localization Demonstration using AutoDRIVE Testbed}
	\label{Figure: Localization Demo}
\end{figure}

Figure \ref{Figure: Localization Demo} illustrates Nigel being teleoperated in the scene with adaptive Monte Carlo localization in action. The time-synchronized RViz window in the bottom-right corner shows the known (static) occupancy grid map, current laser scan, particle cloud (green arrows) and vehicle reference frame (pose estimate) with uncertainty in position (magenta) and orientation (pale yellow).

\begin{table}[htpb]
	\centering
	\caption{Prominent Parameters for Localization}
	\label{Table: Localization Parameters}
	\resizebox{\textwidth}{!}{%
		\begin{tabular}{lll}
			\hline
			\textbf{Parameter}                         & \textbf{Value}                     & \textbf{Remarks}                                      \\ \hline
			\tt min\_particles                         & 500                                & Minimum number of particles                           \\
			\tt max\_particles                         & 3000                               & Maximum number of particles                           \\
			\tt kld\_err                               & 0.02                               & Maximum error between true and predicted distribution \\
			\tt min\_dist\_update                      & 0.01                               & Minimum translation to perform filter update (m)      \\
			\tt min\_angle\_update                     & 0.20                               & Minimum rotation to perform filter update (rad)       \\
			\tt resample\_thresh                       & 1                                  & Number of filter updates before resampling            \\
			\tt initial\_x\_coord                      & 0.0                                & Initial estimate of positional x-coordinate           \\
			\tt initial\_y\_coord                      & 0.0                                & Initial estimate of positional y-coordinate           \\
			\tt initial\_orient                        & 0.0                                & Initial estimate of orientation (yaw)                 \\
			\tt laser\_min\_range                      & 0.15                               & Minimum detection threshold of LIDAR unit (m)         \\
			\tt laser\_max\_range                      & 12.0                               & Maximum detection range of LIDAR unit (m)             \\ \hline
		\end{tabular}%
	}
\end{table}

Table \ref{Table: Localization Parameters} summarizes some of the prominent parameters of the localization system implemented for this particular application.

\subsection{Autonomous Navigation}
\label{Sub-Section: Navigation}
Autonomous parking requires the vehicle to be able to navigate autonomously from its current location to the specified parking location. This in turn demands all the components discussed earlier viz. odometry (Section \ref{Sub-Section: Odometry}), occupancy grid map of the environment (obtained using SLAM, see Section \ref{Sub-Section: SLAM}) and localization system (Section \ref{Sub-Section: Localization}). Additionally, global and local path planners along with a controller for path tracking are required. Since odometry, mapping and localization are discussed in great detail earlier, we will focus majorly on path planning and control aspects of autonomous navigation in this section.

\begin{figure}[htpb]
	\centering
	\includegraphics[width=\textwidth]{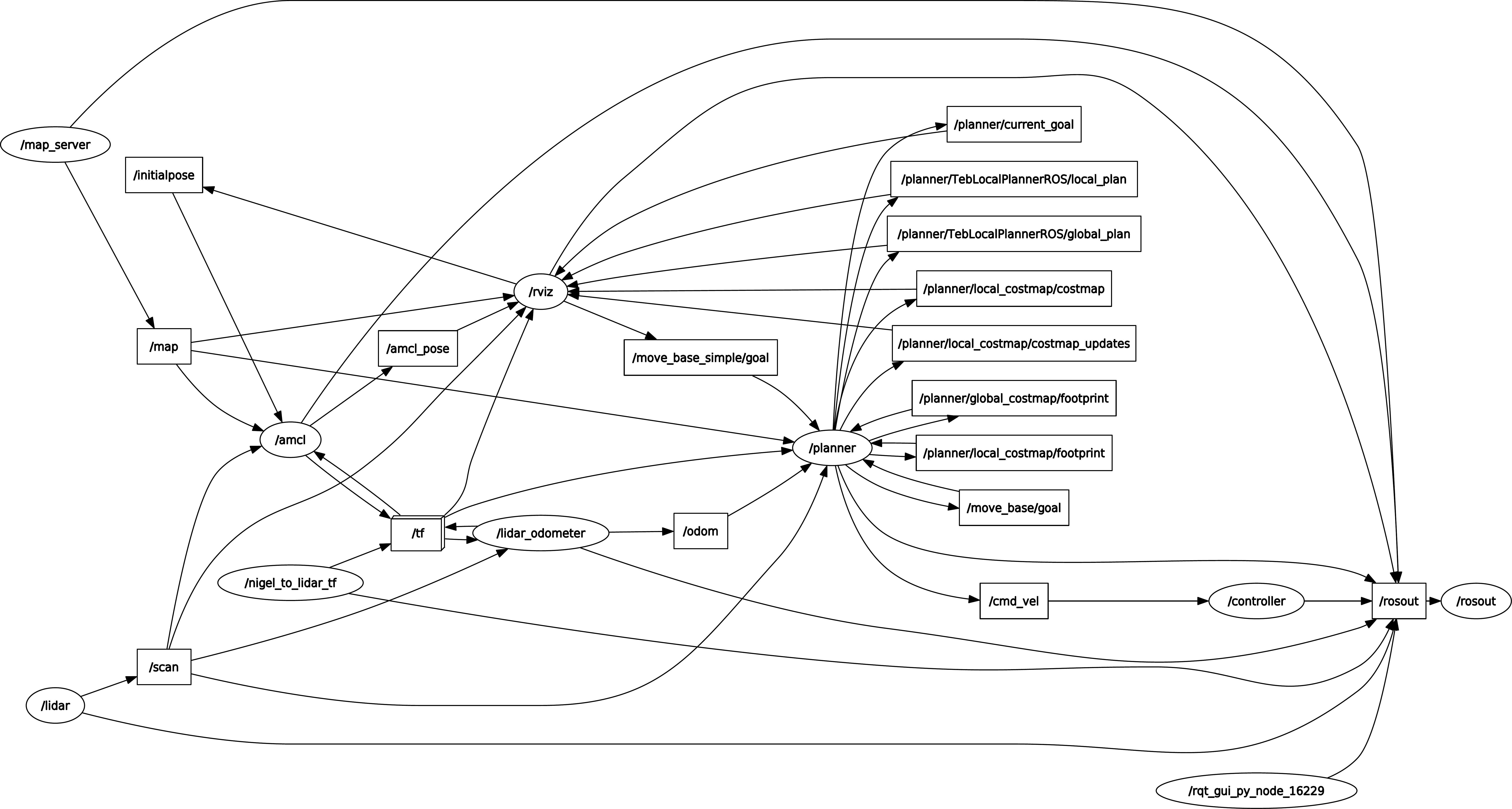}
	\caption{ROS Computation Graph for Autonomous Navigation}
	\label{Figure: Navigation RQT Graph}
\end{figure}

For this particular application, we implemented A* search algorithm \cite{AStar1968} for global path planning, applied timed-elastic-band (TEB) approach \cite{TEBPlanner2017} for real-time online trajectory re-planning and high-level predictive control (implicitly embedded in the optimized local trajectory), and adopted a low-level proportional controller for motion control of the vehicle. Figure \ref{Figure: Navigation RQT Graph} depicts the computation graph illustrating the high-level abstraction of the underlying processes during autonomous navigation required for the said application.

Assuming that map of the environment is available and that the odometry and localization systems are providing respective information; given a goal (parking) pose, the first step is to plan a global path from current location (obtained from localization system) to the goal using A* search algorithm operating on Manhattan distance heuristic (see Algorithm \ref{Algorithm: Global Path Planning}). It is to be noted that this path is planned using the costmap derived from static map of the environment (full map size and resolution) and thus, cannot account for dynamic obstacles or objects that weren't a part of the scene while mapping was done.

\begin{algorithm}[]
	\SetAlgoLined
	\caption{Global Path Planning}
	\label{Algorithm: Global Path Planning}
	\KwIn{Graph $G(V,E)$, source node $start$, goal node $end$}
	\KwOut{Optimal path $opt$\_$path$ from $start$ to $end$ (if one exists)}
	\nl $open$\_$list=\left\{start\right\}$\\
	\nl $closed$\_$list=\left\{ \right\}$\\
	\nl $g(start)=0$\\
	\nl $h(start)=heuristic$\_$function(start,end)$\\
	\nl $f(start)=g(start)+h(start)$\\
	\nl \While{($open$\_$list$ \textup{is not empty})}{
		\nl $m=$ \textbf{node with least} $f$, \textbf{on top of} $open$\_$list$\\
		\nl \If{($m==end$)}{
			\nl \Return{opt\_path}\\
		}
		\nl \textbf{remove} $m$ \textbf{from} $open$\_$list$\\
		\nl \textbf{add} $m$ \textbf{to} $closed$\_$list$\\
		\nl \For{(\textup{each} $n$ \textup{in} $child(m)$)}{
			\nl \If{($n$ \textup{in} $closed$\_$list$)}{
				\nl \textbf{continue}\\
			}
			\nl $cost=g(m)+distance(m,n)$\\
			\nl \If{($n$ \textup{in} $open$\_$list$ \textup{and} $cost<g(n)$)}{
				\nl \textbf{remove} $n$ \textbf{from} $open$\_$list$ as new path is better\\
			}
			\nl \If{($n$ \textup{in} $closed$\_$list$ \textup{and} $cost<g(n)$)}{
				\nl \textbf{remove} $n$ \textbf{from} $closed$\_$list$\\
			}
			\nl \If{($n$ \textup{not in} $open$\_$list$ \textup{and} $n$ \textup{not in} $closed$\_$list$)}{
				\nl \textbf{add} $n$ \textbf{to} $open$\_$list$\\
				\nl $g(n)=cost$\\
				\nl $h(n)=heuristic$\_$function(n,end)$\\
				\nl $f(n)=g(n)+h(n)$\\
			}
		}
	}
	\nl \Return{failure}\\
\end{algorithm}

Once the global plan is available, the next step is to use it along with local coastmap updated using most recent laser scans in order to plan local trajectory up to the farthest point available in the local costmap in the direction of goal. As mentioned earlier, we make use of TEB approach (see Algorithm \ref{Algorithm: Local Path Planning}) to achieve this.

\begin{algorithm}[]
	\SetAlgoLined
	\caption{Local Path Planning}
	\label{Algorithm: Local Path Planning}
	\KwIn{Current vehicle pose $s_c$, final vehicle pose $s_f$, initial path solution $\mathcal{B}$, obstacle set $\mathcal{O}$}
	\KwOut{High-level optimal control command $u^*_c$}
	\nl \tcp{Initialize or update path solution}
	\nl \For{(\textup{\textbf{iterations}} $1,\cdots,I_{teb}$)}{
		\nl \textbf{adjust} $n$\\
		\nl \textbf{update} obstacle constraints \textbf{from} set $\mathcal{O}$\\
		\nl $\mathcal{B}^*\leftarrow \textup{solveNLPP}(\mathcal{B})$\\
	}
	\nl \tcp{Check solution feasibility}
	\nl $u^*_c\leftarrow$ \textbf{map} $\mathcal{B^*}$ \textbf{to} $u^*_c$\\
	\nl \Return{$u^*_c$}\\
\end{algorithm}

Given the current vehicle pose $s_c$, parking pose $s_f$, initial path solution from global plan $\mathcal{B}$ and obstacle set $\mathcal{O}$ based on local costmap, we initialize (or update in subsequent iterations) the path solution. The outer loop with $I_{teb}$ iterations adjusts the number of vehicle configurations $n$ depending on time discretization $\Delta T_k$ to maintain trajectory length, updates obstacle constraints using local costmap, and solves a timed-elastic-band optimization problem (see Equation \ref{Equation: 5.19}) iteratively (inner loop) using the coarse global plan $\mathcal{B}$ (or previous local plan in subsequent iterations) to generate optimal local trajectory $\mathcal{B^*}$.

The timed-elastic-band optimization problem is formulated as a non-linear program subject to vehicle's kinodynamic constraints as follows
\begin{align}
\label{Equation: 5.19}
\argmin_{\mathcal{B}}\sum_{k=1}^{n-1}[\Delta T_k]^2
\end{align}

Next, feasibility of the local plan is verified by implicitly inferring the control command $u^*_c$ (comprising of throttle $v$ and steering $\delta$) from the optimal plan $\mathcal{B^*}$. In order to achieve this, we use vehicle's kinodynamic model (nonlinear ODE) given by
\begin{equation}
\label{Equation: 5.20}
\dot{s}(t)=\begin{bmatrix}\dot{x}(t)\\ \dot{y}(t)\\ \dot{\psi}(t)\end{bmatrix}=\begin{bmatrix}v(t)\cos\psi(t)\\ v(t)\sin\psi(t)\\ \frac{v(t)}{L}\tan\delta(t)\end{bmatrix}
\end{equation}
where, $t$ denotes time, $s(t)$ is pose of the vehicle with wheelbase $L$ and $u(t)=[v(t),\delta(t)]^T$ is the corresponding control vector.

We take the inverse of Equation \ref{Equation: 5.20} to infer control inputs $u(t)$ for the vehicle based on trajectory $\mathcal{B}:=\left\{s_1,\Delta T_1,s_2,\Delta T_2,\cdots,s_{n-1},\Delta T_{n-1},s_n\right\}$ consisting of subsequent vehicle pose configurations $(s_k)_{k=1,\cdots,n}$ w.r.t. time intervals $(\Delta T_k)_{k=1,\cdots,n-1}$ as follows
\begin{equation}
\label{Equation: 5.21}
u(t)=\begin{bmatrix}v(t)\\ \delta(t)\end{bmatrix}=\begin{bmatrix}\gamma(\cdot)\sqrt{\dot{x}^2(t)+\dot{y}^2(t)}\\ \arctan\left(Lv^{-1}(t)\dot\psi(t)\right)\end{bmatrix}
\end{equation}
where $\gamma(\cdot)$ function retains the velocity direction w.r.t. vehicle poses $s_k$ and $s_{k+1}$.

Finally, the high-level optimal control action $u^*_c=[v^*_c,\delta^*_c]^T$ is returned. This entire process is repeated online in order to achieve successful dynamic re-planning of the local trajectory.

Once the local trajectory and high-level optimal control commands are available, the next (and final) task is to implement a controller that is capable of actually tracking these commands. Since all the computation including odometry, localization, path planning (global and local) and control was supposed to be handled by a single-board computer onboard the vehicle, another implied necessity was to have a minimum latency control loop. We therefore implemented a proportional controller that would bind the high-level commands to actuator saturation limits and generate normalized control commands $\in [-1,1]$ for the vehicle.

\begin{table}[h]
	\centering
	\caption{Prominent Parameters for Autonomous Navigation}
	\label{Table: Navigation Parameters}
	\resizebox{\textwidth}{!}{%
		\begin{tabular}{lll}
			\hline
			\textbf{Parameter}                         & \textbf{Value}                     & \textbf{Remarks}                                            \\ \hline
			\tt global\_costmap\_size                  & 80                                 & Number of grid cells in static global costmap                                   \\
			\tt local\_costmap\_size                   & 1.5                                & Local costmap side length (m)                                                   \\
			\tt rolling\_window                        & true                               & Local costmap will be vehicle-centric                                           \\
			\tt range\_obstacle                        & 3.0                                & Update costmap for obstacles within this range (m)                              \\
			\tt range\_raytrace                        & 3.5                                & Raytrace free space upto this value (m)                                         \\
			\tt radius\_inflation                      & 0.025                              & Maximum cushioning distance from obstacle to assign obstacle cost (m)      \\
			\tt cost\_scaling\_factor                  & 10.0                               & Exponential rate at which obstacle costs drop off                               \\
			\tt lin\_vel\_max                          & 0.2                                & Linear velocity bound for vehicle (m/s)                                         \\
			\tt ang\_vel\_max                          & 0.5236                             & Angular velocity bound for vehicle (rad/s)                                      \\
			\tt lin\_acc\_max                          & 0.15                               & Linear acceleration bound for vehicle (m/s$^2$)                                 \\
			\tt ang\_acc\_max                          & 0.3927                             & Angular acceleration bound for vehicle (rad/s$^2$)                              \\
			\tt turning\_radius\_min                   & 0.24515                            & Minimum turning radius of the vehicle (m)                                       \\
			\tt vehicle\_wheelbase                     & 0.14154                            & Wheelbase ($L$) of the vehicle (m)                                              \\
			\tt vehicle\_footprint                     & line                               & Vehicle footprint model used for TEB optimization                               \\
			\tt xy\_goal\_tolerance                    & 0.1                                & Positional goal tolerance (m)                                                   \\
			\tt yaw\_goal\_tolerance                   & 0.1                                & Orientational goal tolerance (rad)                                              \\
			\tt min\_obstacle\_dist                    & 0.2                                & Minimum desired separation from obstacles w.r.t. vehicle's rear axle centre (m) \\
			\tt num\_inner\_iterations                 & 3                                  & Iterations for the inner TEB optimization loop                                  \\
			\tt num\_outer\_iterations                 & 3                                  & Iterations for the outer optimization loop ($I_{teb}$)                          \\ \hline
		\end{tabular}%
	}
\end{table}

\begin{figure}[!h]
	\centering
	\includegraphics[width=\textwidth]{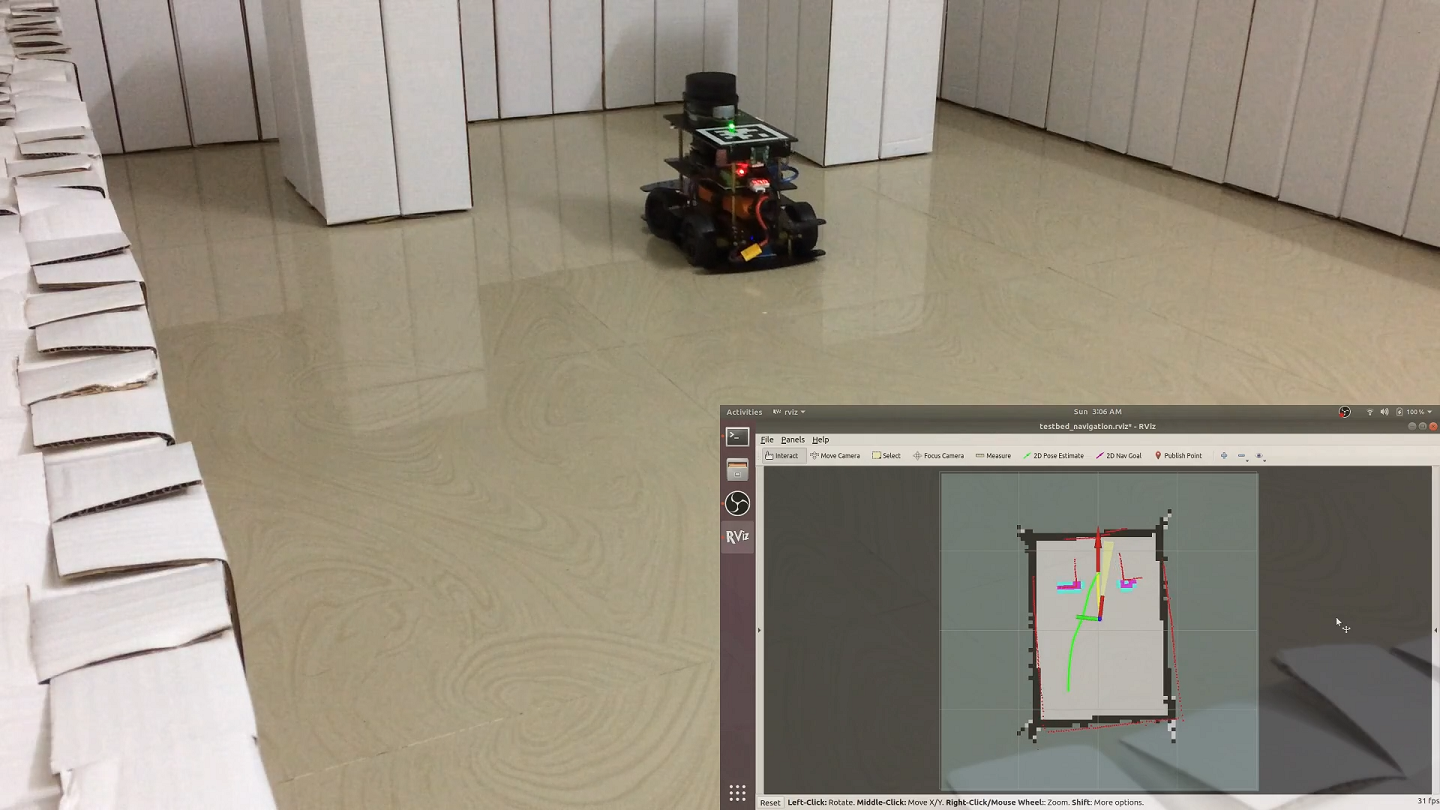}
	\caption{Autonomous Navigation Demonstration using AutoDRIVE Testbed}
	\label{Figure: Navigation Demo}
\end{figure}

Table \ref{Table: Navigation Parameters} summarizes some of the prominent parameters of the autonomous navigation system implemented for this particular application.

Figure \ref{Figure: Navigation Demo} depicts Nigel navigating autonomously from the source location to the specified parking destination. The time-synchronized RViz window in the bottom-right corner shows the known (static) occupancy grid map, global and local costmaps, current laser scan, vehicle pose estimate with uncertainty in position (magenta) and orientation (pale yellow), goal/parking pose (red arrow) as well as local (yellow) and global (green) paths. It is worth mentioning that the two occlusions in front of the vehicle were not a part of the environment while mapping was performed (see Figure \ref{Figure: SLAM Demo}). They were later introduced in the scene to increase the problem complexity resembling real-world conditions wherein peer vehicles in a parking space may or may not exist.

\section{Behavioural Cloning}
\label{Section: Behavioural Cloning}

The application of behavioural cloning \cite{RBC2021} was chosen to demonstrate AutoDRIVE's ability to support development of end-to-end autonomy algorithms (refer Section \ref{Sub-Section: End-to-End Approach}), along with its smooth sim2real transfer capability. This demonstration employed computer vision (CV) and deep imitation learning (DIL) techniques, with the following steps:

\begin{itemize}
	\item Data collection
	\item Data augmentation
	\item Data preprocessing
	\item Training
	\item Deployment
	\item Sim2Real transfer
\end{itemize}

This application demonstrates a single-agent scenario, wherein the vehicle's objective was to autonomously drive around the specified course, without overstepping any lane markings. The only source of perception for the vehicle was its front-view camera, providing real-time RGB images. A deep neural network was trained offline using labelled manual driving data, which was then deployed back onto the simulated vehicle in order to validate its autonomy. Finally, the same model was transferred onto the physical vehicle for testing its generalization capability. While the entire training pipeline was developed independently, instead of employing the simulator's integrated ML framework, the deployment pipeline was developed using AutoDRIVE Devkit's Python API.

\subsection{Data Collection}
\label{Sub-Section: Data Collection}

AutoDRIVE Simulator was exploited to record timestamped manual driving data for training the neural network. Table \ref{Table: Data Recording Configuration} presents the details pertaining to data collection.

\begin{table}[htpb]
	\centering
	\caption{Data Recording Configuration}
	\label{Table: Data Recording Configuration}
	\resizebox{0.5\textwidth}{!}{%
		\begin{tabular}{ll}
			\hline
			\textbf{Parameter}           & \textbf{Specification} \\ \hline
			Target simulation frame rate & 60 Hz                  \\
			Graphics quality             & Ultra                  \\
			Scene light                  & Enabled                \\
			Camera frame resolution      & 1280$\times$720 px     \\
			Target data collection rate  & 15 Hz                  \\
			Driving laps                 & 5                      \\
			Total data samples           & 1699                   \\
			Training data samples        & 1359                   \\
			Validation data samples      & 340                    \\ \hline
		\end{tabular}%
	}
\end{table}

The dataset comprised of vehicle states including its X and Y positional coordinates (m), yaw (rad) and rigid body velocity (m/s), along with sensory data including throttle (\%), steering angle (rad), left and right encoder ticks (count), front and rear camera frames (jpg) as well as LIDAR range array (m). Table \ref{Table: Sample Data Entries} holds ten sequential data entries from the dataset, for better understanding. However, not all the information was used for this application. Particularly, only the front camera frames and normalized steering commands were used as features and labels, respectively. Furthermore, the entire dataset was split into an 80:20 ratio for training and validation, as described in Table \ref{Table: Data Recording Configuration}.

\begin{table}[htpb]
	\centering
	\caption{Sample Data Entries}
	\label{Table: Sample Data Entries}
	\resizebox{\textwidth}{!}{%
		\begin{tabular}{lllllllllll}
			\hline
			\textbf{Pos-X} & \textbf{Pos-Y} & \textbf{Yaw} & \textbf{Vel} & \textbf{Throttle} & \textbf{Steering} & \textbf{LE Ticks} & \textbf{RE Ticks} & \textbf{Front Camera Frame}       & \textbf{Rear Camera Frame}        & \textbf{LIDAR Range Array}   \\ \hline
			0.304693       & -1.73417       & 3.929984     & 0.26         & 1.0000            & -0.0698132        & 5553              & 5553              & 2021\_04\_19\_00\_13\_49\_260.jpg & 2021\_04\_19\_00\_13\_49\_260.jpg & 0.5172184 $\cdots$ 0.5082338 \\
			0.290529       & -1.74953       & 3.935050     & 0.26         & 1.0000            & -0.1134464        & 5782              & 5782              & 2021\_04\_19\_00\_13\_49\_354.jpg & 2021\_04\_19\_00\_13\_49\_354.jpg & 0.5172184 $\cdots$ 0.5082338 \\
			0.273582       & -1.76931       & 3.946518     & 0.26         & 1.0000            & -0.1570796        & 6068              & 6069              & 2021\_04\_19\_00\_13\_49\_442.jpg & 2021\_04\_19\_00\_13\_49\_442.jpg & 0.4761972 $\cdots$ 0.4678237 \\
			0.260359       & -1.78543       & 3.955983     & 0.26         & 1.0000            & -0.1570796        & 6297              & 6298              & 2021\_04\_19\_00\_13\_49\_527.jpg & 2021\_04\_19\_00\_13\_49\_527.jpg & 0.4761972 $\cdots$ 0.4678237 \\
			0.247272       & -1.80166       & 3.965485     & 0.26         & 1.0000            & -0.1570796        & 6526              & 6527              & 2021\_04\_19\_00\_13\_49\_599.jpg & 2021\_04\_19\_00\_13\_49\_599.jpg & 0.4512160 $\cdots$ 0.4430620 \\
			0.234803       & -1.81821       & 3.979502     & 0.26         & 1.0000            & -0.2705260        & 6754              & 6756              & 2021\_04\_19\_00\_13\_49\_689.jpg & 2021\_04\_19\_00\_13\_49\_689.jpg & 0.4308175 $\cdots$ 0.4227354 \\
			0.223138       & -1.83527       & 3.996314     & 0.26         & 1.0000            & -0.3141593        & 6983              & 6986              & 2021\_04\_19\_00\_13\_49\_769.jpg & 2021\_04\_19\_00\_13\_49\_769.jpg & 0.4308175 $\cdots$ 0.4227354 \\
			0.212496       & -1.85272       & 4.019250     & 0.25         & 1.0000            & -0.3839724        & 7211              & 7215              & 2021\_04\_19\_00\_13\_49\_839.jpg & 2021\_04\_19\_00\_13\_49\_839.jpg & 0.4165480 $\cdots$ 0.4083009 \\
			0.202678       & -1.87062       & 4.043489     & 0.25         & 1.0000            & -0.3839724        & 7440              & 7444              & 2021\_04\_19\_00\_13\_49\_918.jpg & 2021\_04\_19\_00\_13\_49\_918.jpg & 0.4165480 $\cdots$ 0.4083009 \\
			0.195582       & -1.88422       & 4.061698     & 0.25         & 1.0000            & -0.3839724        & 7611              & 7616              & 2021\_04\_19\_00\_13\_49\_984.jpg & 2021\_04\_19\_00\_13\_49\_984.jpg & 0.4155301 $\cdots$ 0.4066035 \\ \hline
		\end{tabular}%
	}
\end{table}

It is to be noted that adequate care was taken so as to avoid any inherent biases within the dataset, including skewed-steering bias as well as excessive zero-steering bias, which would mislead the neural network being trained. Additionally, online data balancing was performed every time when looping through the dataset, during the core training phase. Not only did this immunize the network against any unnoticed biases, but it also reduced the chance of overfitting as a novel part of the unbalanced data was discarded each time.

\subsection{Data Augmentation}
\label{Sub-Section: Data Augmentation}

Probabilistic data augmentation was carried out online, in order to impart generalization capability to the neural network model being trained. This also helped in achieving adequate domain randomization for the sim2real transfer. A total of five augmentation techniques, viz. shadows, brightness, flip, pan and tilt were employed for this particular application. Table \ref{Table: Augmentation Probabilities} hosts the probabilities associated with each individual augmentation being applied to a particular data sample.

\begin{table}[htpb]
	\centering
	\caption{Augmentation Probabilities}
	\label{Table: Augmentation Probabilities}
	\resizebox{0.6\textwidth}{!}{%
		\begin{tabular}{ll}
			\hline
			\textbf{Augmentation Technique} & \textbf{Application Probability} \\ \hline
			Shadows                         & 0.3                              \\
			Brightness                      & 0.4                              \\
			Flip                            & 0.5                              \\
			Pan                             & 0.1                              \\
			Tilt                            & 0.05                             \\ \hline
		\end{tabular}%
	}
\end{table}

\subsubsection{Shadows}
\label{Sub-Sub-Section: Shadows}

\begin{figure}[htpb]
	\centering
	\includegraphics[width=\textwidth]{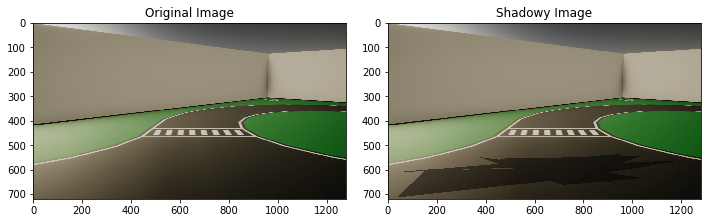}
	\caption{Shadow Augmentation}
	\label{Figure: Shadow Augmentation}
\end{figure}

This augmentation technique adds a number of polygonal shadows with specified darkness coefficient to a particular camera frame, within the given region of interest (ROI). For this application, a total of four quadrangular shadows with 0.5 darkness coefficient were added within the bottom ${1/3}^{rd}$ of a frame (refer Figure \ref{Figure: Shadow Augmentation}). Not only did this generalize the neural network against a variety of shadows, but adding random dark patterns onto the images also immunized it against a series of stray marks, creases and folds, which was especially helpful as a domain randomization technique for sim2real transfer.

\subsubsection{Brightness}
\label{Sub-Sub-Section: Brightness}

\begin{figure}[htpb]
	\centering
	\includegraphics[width=\textwidth]{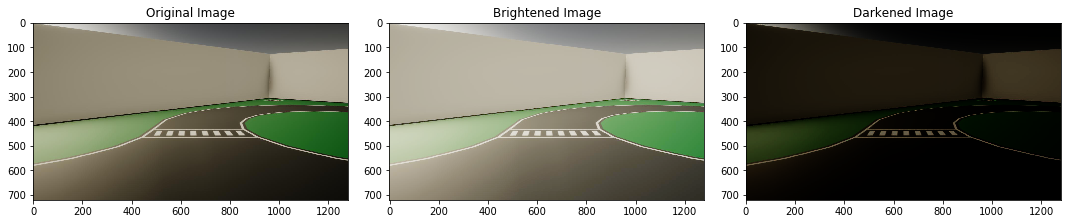}
	\caption{Brightness Augmentation}
	\label{Figure: Brightness Augmentation}
\end{figure}

This augmentation technique alters the brightness of a particular camera frame by a random coefficient within the specified range. For this application, the brightness of a frame was randomly altered between 0.25 (dark) and 1.75 (bright) using appropriate gamma correction by applying a look-up table (LUT) transform on the input image, as defined in \ref{Equation: 5.22}.
\begin{equation}
\label{Equation: 5.22}
\texttt{dst}_{i,j} = \texttt{lut}\left ( \texttt{src}_{i,j} \right )
\end{equation}
where, $\texttt{lut}$ is a 256-element array defined as follows:
\begin{equation}
\label{Equation: 5.23}
\texttt{floor}\left ( 255 * \left [ \frac{k}{255} \right ]^{1/\gamma} \; \texttt{for} \; k \; \texttt{in} \; \left [ 0, 1, \cdots, 254, 255 \right ]_{1 \times 256} \right )
\end{equation}

This augmentation helped the neural network generalize against a variety of lighting conditions, which was quite useful for transferring it to real world.

\subsubsection{Flip}
\label{Sub-Sub-Section: Flip}

\begin{figure}[htpb]
	\centering
	\includegraphics[width=\textwidth]{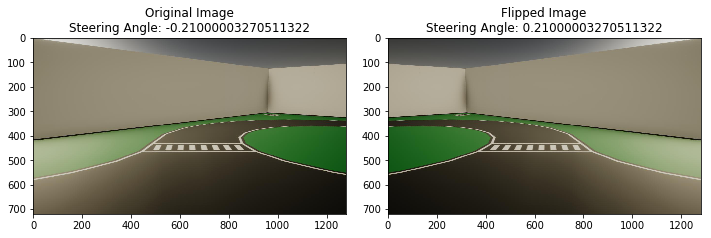}
	\caption{Flip Augmentation}
	\label{Figure: Flip Augmentation}
\end{figure}

This augmentation technique flips a particular camera frame horizontally (refer Figure \ref{Figure: Flip Augmentation}), and negates the corresponding steering angle in order to compensate for the inverted image. Equation \ref{Equation: 5.24} respectively describes the image flipping and steering negation functions.
\begin{align}
\label{Equation: 5.24}
\begin{cases}
\texttt{dst}_{i,j} = \texttt{src}_{w-i-1,j}\\
\theta_t = - \theta_t
\end{cases}
\end{align}
where $w$ is width of the image, and $\theta_t$ is the recorded steering angle at time $t$.

This technique helped reduce any unbalance present within the steering angle distribution, since each data sample had a 50\% change of being applied with this augmentation (refer Table \ref{Table: Augmentation Probabilities}).

\subsubsection{Pan}
\label{Sub-Sub-Section: Pan}

\begin{figure}[htpb]
	\centering
	\includegraphics[width=\textwidth]{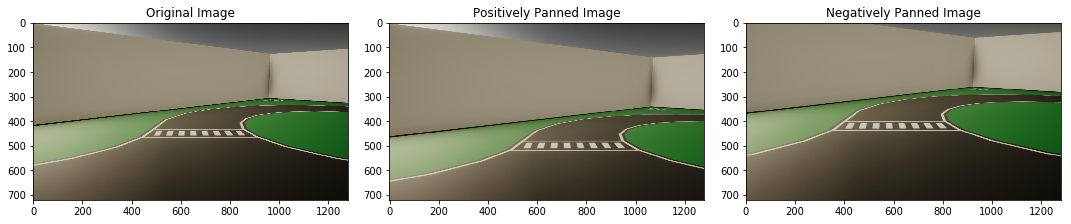}
	\caption{Pan Augmentation}
	\label{Figure: Pan Augmentation}
\end{figure}

This augmentation technique pans a particular camera frame by translating its pixels horizontally ($x$) and/or vertically ($y$) by random coefficients ($t_x$, $t_y$) sampled from respective specified ranges (in this case, $\pm$10\% w.r.t. the original image dimensions). This transformation is defined by the translation matrix $M_T$ described in Equation \ref{Equation: 5.25}. Finally, null area of the panned image was cropped out and it was resized to original dimensions (refer Figure \ref{Figure: Pan Augmentation}).
\begin{equation}
\label{Equation: 5.25}
M_T = \begin{bmatrix}
1 & 0 & t_x\\ 
0 & 1 & t_y
\end{bmatrix}
\end{equation}

This technique helped account for mechanical vibrations and misalignments (if any), especially in case of the physical vehicle.

\subsubsection{Tilt}
\label{Sub-Sub-Section: Tilt}

\begin{figure}[htpb]
	\centering
	\includegraphics[width=\textwidth]{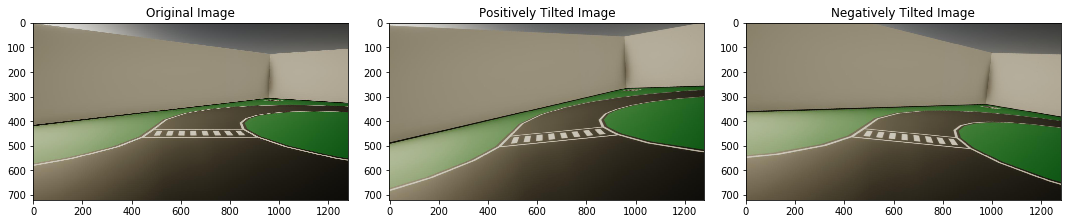}
	\caption{Tilt Augmentation}
	\label{Figure: Tilt Augmentation}
\end{figure}

This augmentation technique rotates a particular camera frame of size $w \times h$ through its centre by a random angle $\phi$ sampled from a specified range (in this case, $\pm$5$^\circ$). This transformation is defined by the rotation matrix $M_T$ described in Equation \ref{Equation: 5.26}. Finally, null area of the tilted image was cropped out by fitting a central, axis-aligned rectangular ROI of maximal area defined by size $w_{roi} \times h_{roi}$ (refer Equation \ref{Equation: 5.27}), and it was resized to original dimensions (refer Figure \ref{Figure: Tilt Augmentation}).
\begin{equation}
\label{Equation: 5.26}
M_R = \begin{bmatrix}
\cos\phi & \sin\phi & \frac{w}{2}(1-\cos\phi )-\frac{h}{2}\sin\phi \\ 
-\sin\phi & \cos\phi & \frac{w}{2}\sin\phi +\frac{h}{2}(1-\cos\phi )
\end{bmatrix}
\end{equation}
\begin{align}
\label{Equation: 5.27}
w_{roi}, h_{roi} = \begin{cases}
\frac{h}{2\sin\phi},\frac{h}{2\cos\phi}; \textup{\qquad\qquad\qquad\, \scriptsize{Half-constrained case}}\\
\frac{w\cos\phi-h\sin\phi}{\cos(2\phi)},\frac{h\cos\phi-w\sin\phi}{\cos(2\phi)}; \textup{\quad \scriptsize{Fully-constrained case}}
\end{cases}
\end{align}

Similar to the panning augmentation, this technique also helped account for mechanical vibrations and misalignments (if any), especially in case of the physical vehicle.

\subsection{Data Preprocessing}
\label{Sub-Section: Data Preprocessing}

\begin{figure}[htpb]
	\centering
	\includegraphics[width=0.8\textwidth]{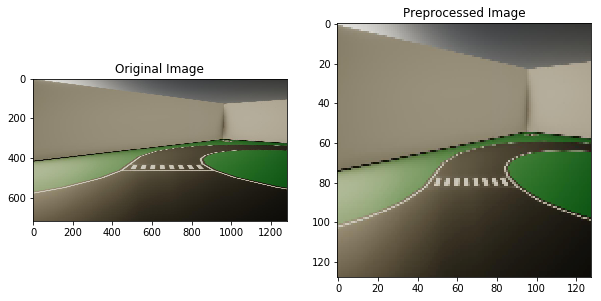}
	\caption{Data Preprocessing}
	\label{Figure: Preprocessing}
\end{figure}

Data preprocessing was employed primarily to reduce the amount of data being parsed to the neural network, and secondarily to normalize it in order to stabilize the training and achieve convergence faster. This followed rescaling the camera frames from their original size of 1280$\times$720 px, down to 128$\times$128 px, thereby also altering their aspect ratio, which enhanced feature extraction in horizontal direction through the use of square kernels. The images were also normalized and mean-centred as described in Equation \ref{Equation: 5.28}. Figure \ref{Figure: Preprocessing} depicts a sample frame before and after preprocessing.
\begin{equation}
\label{Equation: 5.28}
\texttt{dst}_{i,j} = \frac{\texttt{src}_{i,j}}{255} - 0.5
\end{equation}

Additionally, as discussed in Section \ref{Sub-Section: Data Collection}, the steering angle values were also normalized (refer Equation \ref{Equation: 5.29}) in the range of $[-1, 1]$ by dividing them with the steering limit $\varphi$ of the vehicle.
\begin{equation}
\label{Equation: 5.29}
\delta = \frac{\theta}{\varphi}
\end{equation}

\subsection{Training}
\label{Sub-Section: DIL Training}

Apart from core learning process, the generalized training phase comprised of all the aforementioned steps, some of which were performed only once (these included data collection and segregation), while others were performed recursively (these included data shuffling, balancing, augmentation and preprocessing).

The core learning process, on the other hand, was focused on training a deep neural network to predict steering commands for the vehicle in end-to-end fashion based on static RGB frames from its front camera (after preprocessing; refer Section \ref{Sub-Section: Data Preprocessing}).

\begin{figure}[htpb]
	\centering
	\includegraphics[width=\textwidth]{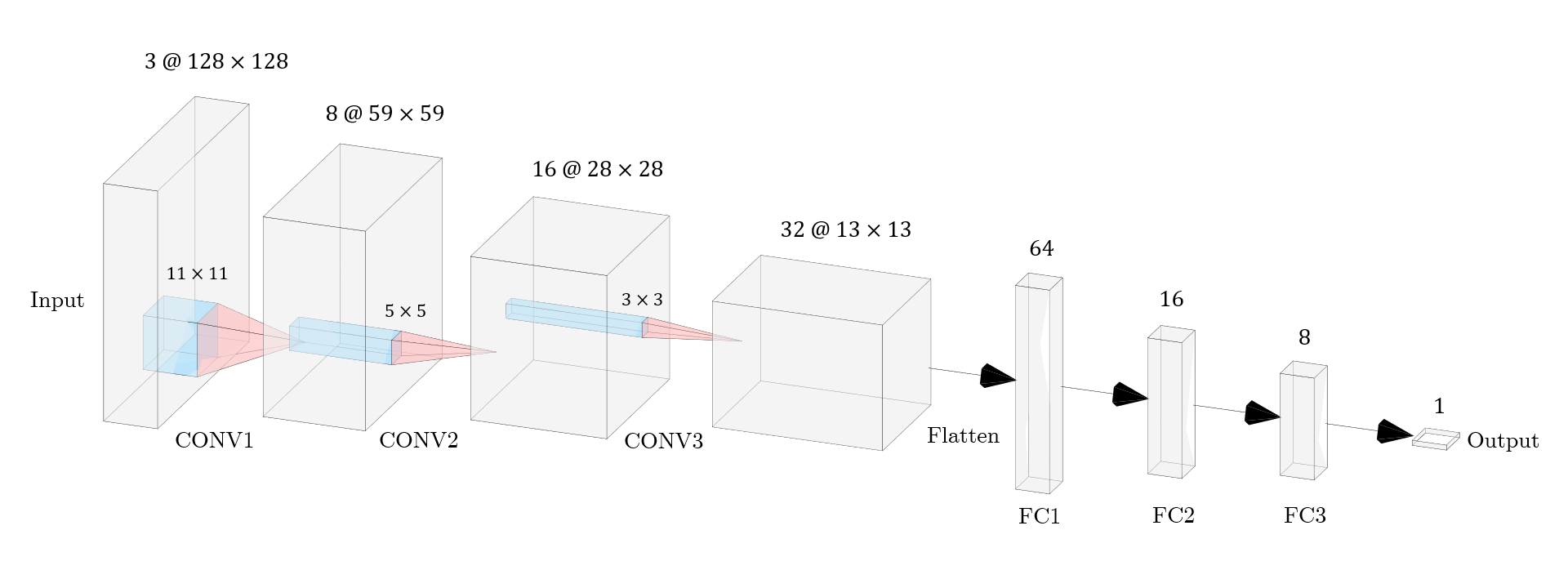}
	\caption{CNN Architecture}
	\label{Figure: CNN Architecture}
\end{figure}

This work adopted a 6-layer deep convolutional neural network (CNN) \cite{Krizhevsky2012} with 3 convolutional (CONV) and 3 fully connected (FC) layers, comprising a total of 358,129 trainable parameters (refer Figure \ref{Figure: CNN Architecture}). The network accepted 128$\times$128 px RGB frames and performed strided convolutions on them with a 2$\times$2 stride. The initial CONV layer was provided with 11$\times$11 kernels, followed by 5$\times$5 and 3$\times$3 kernels in the second and third layers, respectively. The output of third CONV layer was flattened and passed through the 3 FC layers, with progressively reducing sizes, each followed by a dropout \cite{Srivastava2014} layer, to finally output the predicted steering command. Table \ref{Table: DIL Hyperparameters} holds the hyperparameter values adopted for this demonstration.

\begin{table}[htpb]
	\centering
	\caption{Training Hyperparameters}
	\label{Table: DIL Hyperparameters}
	\resizebox{0.55\textwidth}{!}{%
		\begin{tabular}{ll}
			\hline
			\textbf{Hyperparameter}    & \textbf{Value} \\ \hline
			Data augmentation loops    & 64             \\
			Weight initialization      & Glorot Uniform \cite{Glorot2010} \\
			Bias initialization        & Zeros          \\
			Training steps per epoch   & 384            \\
			Validation steps per epoch & 2              \\
			Number of epochs           & 4              \\
			Batch size                 & 256            \\
			Learning rate              & 1.0e-3         \\
			Activation function        & ReLU \cite{Fukushima1975}        \\
			Dropout probability        & 0.25           \\
			Optimizer                  & Adam \cite{Kingma2014}           \\
			Loss metric                & MSE            \\ \hline
		\end{tabular}%
	}
\end{table}

\begin{figure}[htpb]
	\centering
	\includegraphics[width=0.5\textwidth]{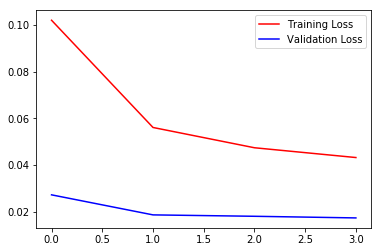}
	\caption{Training and Validation Loss}
	\label{Figure: DIL Training Results}
\end{figure}

Towards the end of training phase, the CNN had learnt to predict steering commands almost as good as humans. The training loss had exponentially decreased from 0.1021 (after the first epoch) to 0.0432 (after the fourth epoch) and nearly saturated in this neighbourhood. The validation loss had also reduced gradually from 0.0272 (after the first epoch) to 0.0173 (after the fourth epoch), which was an indicative of good training; neither had the model underfit to the data, nor had it overfit. Figure \ref{Figure: DIL Training Results} depicts the training and validation loss over the entire course of learning process.

\subsection{Deployment}
\label{Sub-Section: DIL Deployment}

\begin{figure}[htpb]
	\centering
	\subfigure[]{\includegraphics[width=0.49\textwidth]{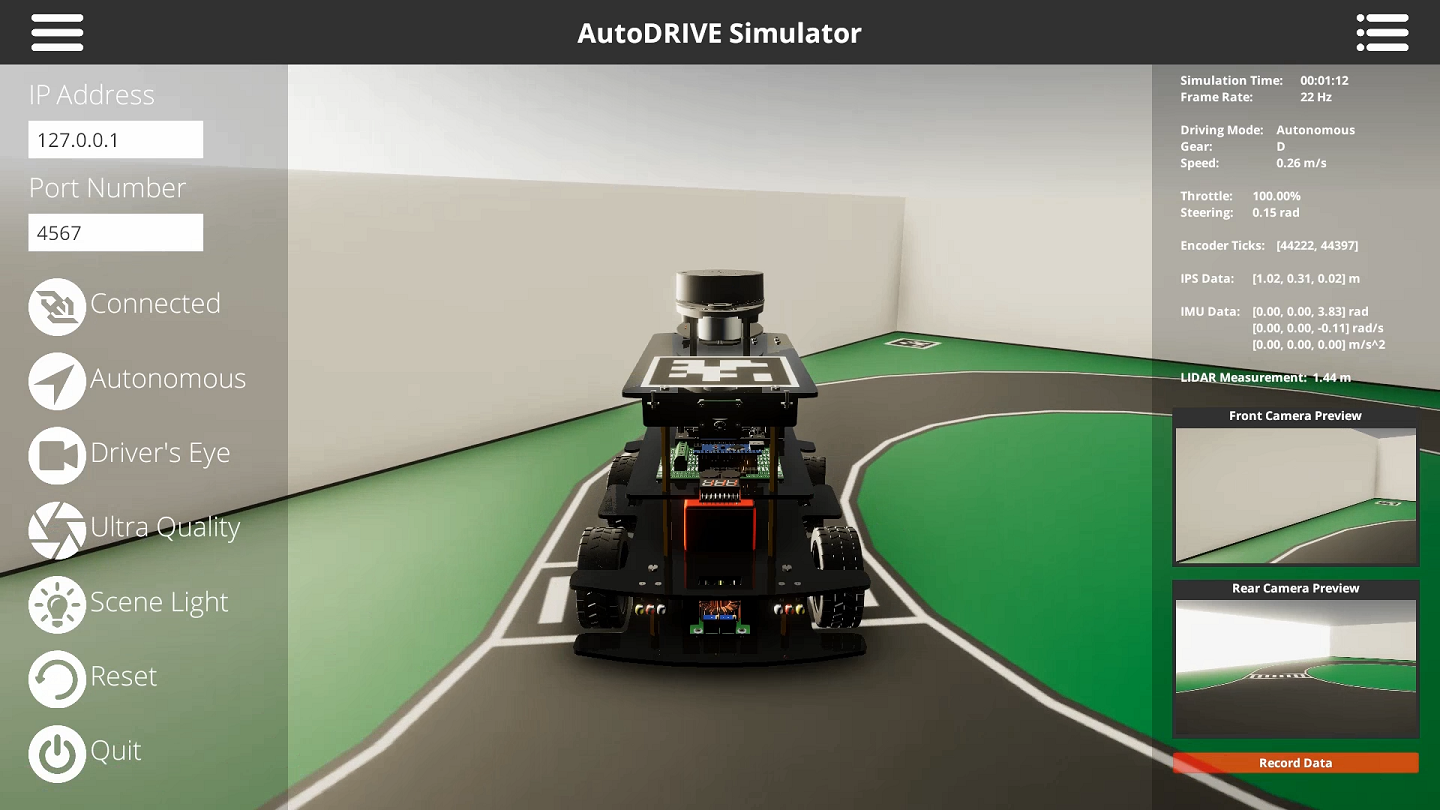}}
	\subfigure[]{\includegraphics[width=0.49\textwidth]{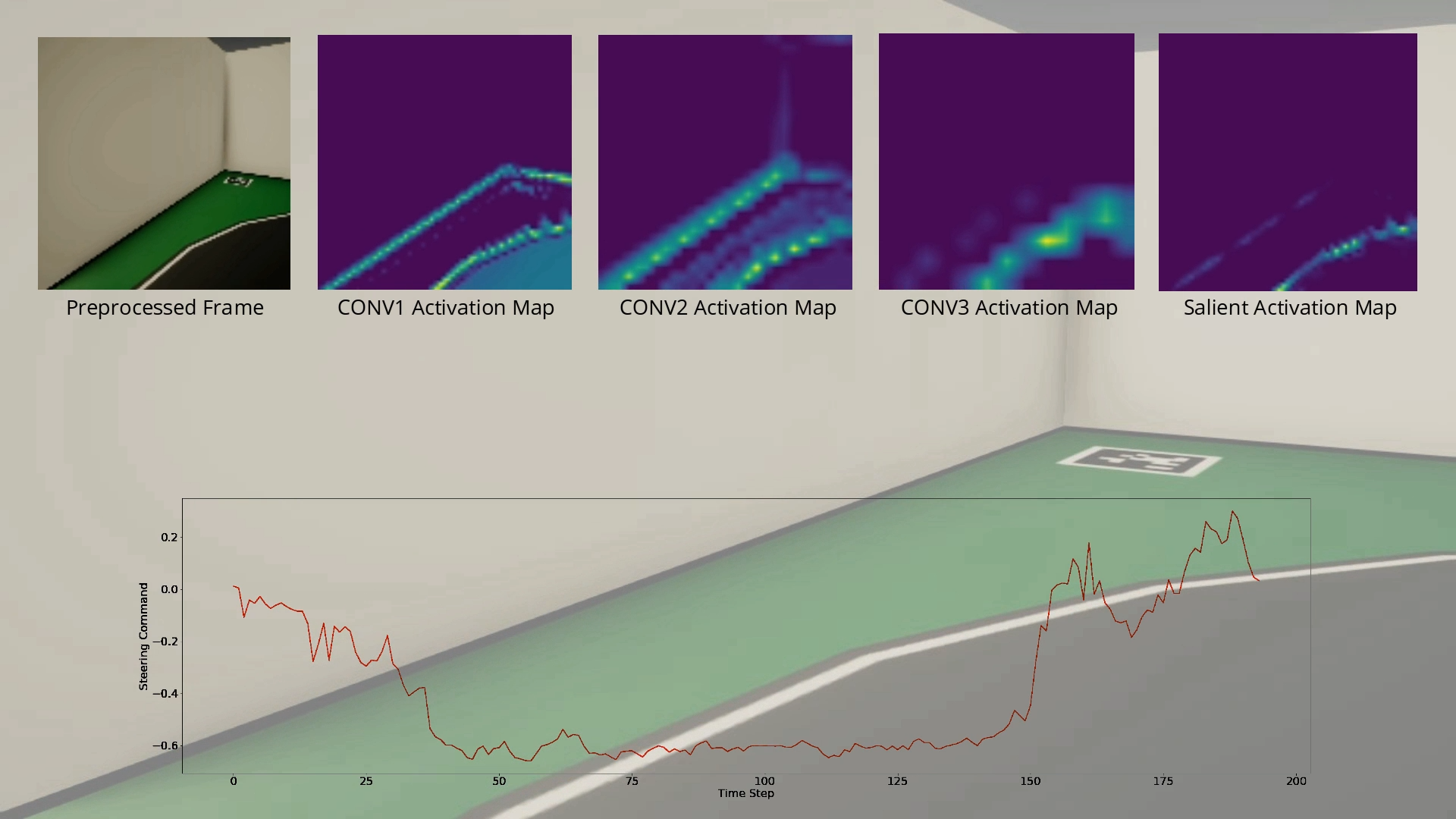}}
	\caption{Deployment Phase: (a) Autonomous Driving, and (b) Insight}
	\label{Figure: DIL Deployment}
\end{figure}

The trained model was first deployed back onto the simulated vehicle in order to validate its autonomy $\eta$ defined by the ratio of the required intervention time $t_{int}$ w.r.t. the total lap time $t_{lap}$ (refer Equation \ref{Equation: 5.30}). The vehicle exhibited 100\% autonomy, meaning it could complete several laps of the course without wandering off the drivable area.
\begin{equation}
\label{Equation: 5.30}
\eta\,(\%) = \left [ 1 - \frac{t_{int}}{t_{lap}} \right ] * 100
\end{equation}
where, the intervention time was assumed to be 6 seconds (standard time for an average unaware human to intervene) for every interference $n_{int}$; i.e. $t_{int} = 6 * n_{int}$.

Figure \ref{Figure: DIL Deployment} depicts the simulated vehicle driving autonomously; while steering command was predicted by the trained neural network model, throttle command was set constant at 1 (i.e. full throttle). The figure also gives an insight into the end-to-end pipeline by visualizing the original and preprocessed frames, along with the activation map of each convolutional layer. Additionally, a salient feature map is also developed by merging all the activation maps, thereby only visualizing the key activation zones. It can be observed how the trained model detected road lane markings, without any explicit CV algorithm. Finally, the predicted steering commands are also visualized.

\subsection{Sim2Real Transfer}
\label{Sub-Section: Sim2Real Transfer}

\begin{figure}[htpb]
	\centering
	\subfigure[]{\includegraphics[width=0.49\textwidth]{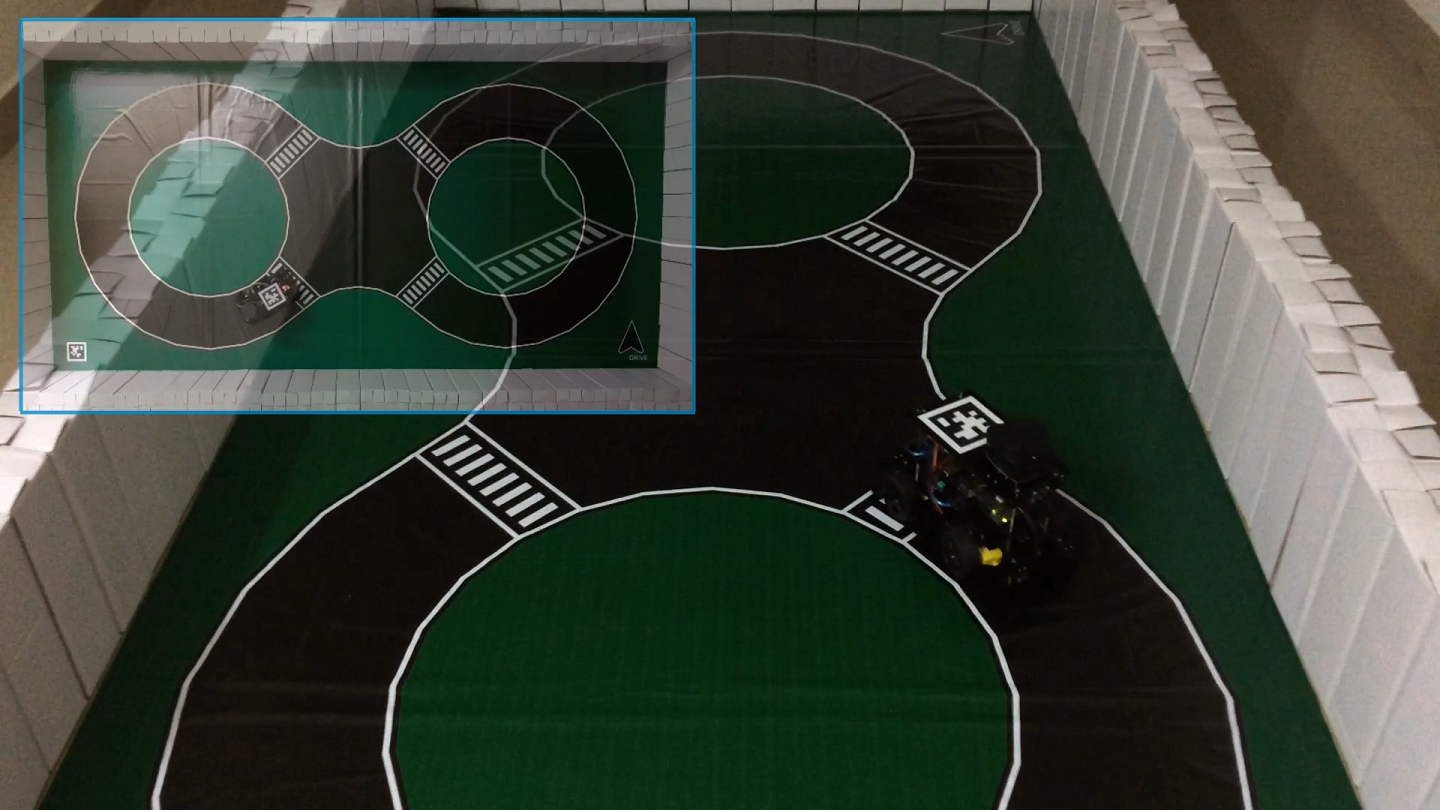}}
	\subfigure[]{\includegraphics[width=0.49\textwidth]{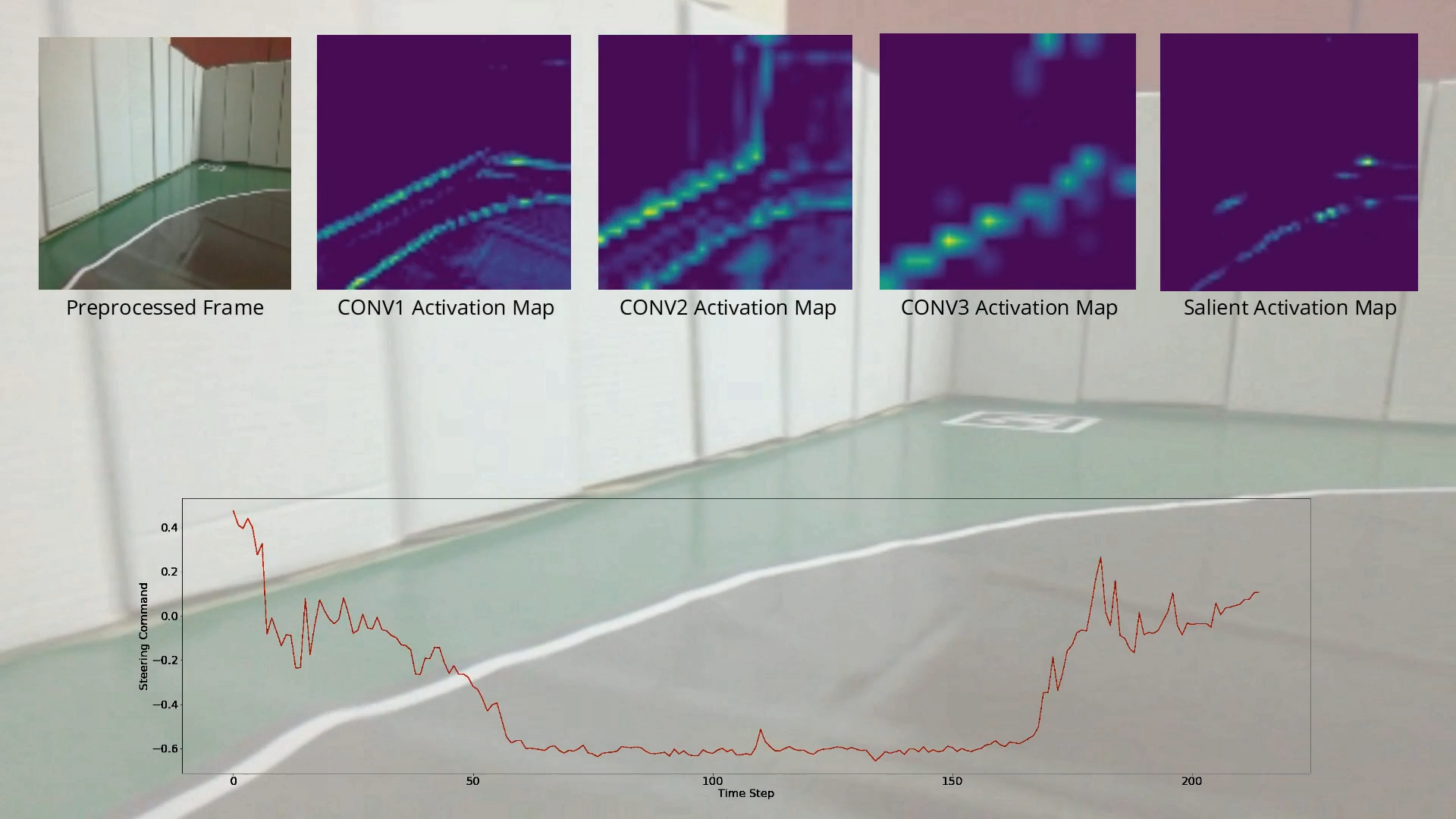}}
	\caption{Sim2Real Transfer: (a) Autonomous Driving, and (b) Insight}
	\label{Figure: DIL Sim2Real Transfer}
\end{figure}

Upon successfully validating autonomy of the trained model, it was finally deployed onto the physical vehicle for hardware validation. Nigel's onboard computer handled all the high-level computation, including real-time acquisition of front camera frames, preprocessing them, and running neural network inference in order to generate steering commands in an end-to-end manner. The throttle value was set constant at 0.8, so as to compensate for any time-lag present in the pipeline. Finally, the low-level MCU controlled the vehicle's actuators through appropriate pulse-width modulation (PWM) signals.

Figure \ref{Figure: DIL Sim2Real Transfer} depicts the physical vehicle driving autonomously, along with an insight into the end-to-end pipeline. It is worth mentioning that the model trained solely on simulated driving data (with adequate domain randomization) exhibited 100\% autonomy even in the real-world, which also indirectly verified the simulator's fidelity in terms of realistically representing real-world scenarios.

\section{Intersection Traversal}
\label{Section: Intersection Traversal}

The application of intersection traversal was chosen to demonstrate AutoDRIVE's ability to support development of end-to-end autonomy algorithms (refer Section \ref{Sub-Section: End-to-End Approach}), specifically employing the simulator's integrated ML framework. This demonstration employed vehicle to vehicle (V2V) communication and deep reinforcement learning (DRL) techniques, involving the following stages:

\begin{itemize}
	\item Problem formulation
	\item Training
	\item Deployment
\end{itemize}

Inspired from \cite{MARL2020}, this application demonstrates both single and multi-agent scenarios wherein each vehicle's objective was to autonomously traverse a 4-lane 4-way intersection, without colliding with the peer vehicles or overstepping any lane markings. Each vehicle could sense its own intrinsic states and was provided with limited state information about its peers; this application did not employ any exteroceptive sensing modalities. A deep neural network policy was trained online, independently for each scenario, in order to make the agents traverse the intersection safely. The entire application was developed using AutoDRIVE Simulator's integrated ML framework.

\subsection{Problem Formulation}
\label{Sub-Section: Problem Formulation}

\begin{figure}[htpb]
	\centering
	\subfigure[]{\includegraphics[width=0.49\textwidth]{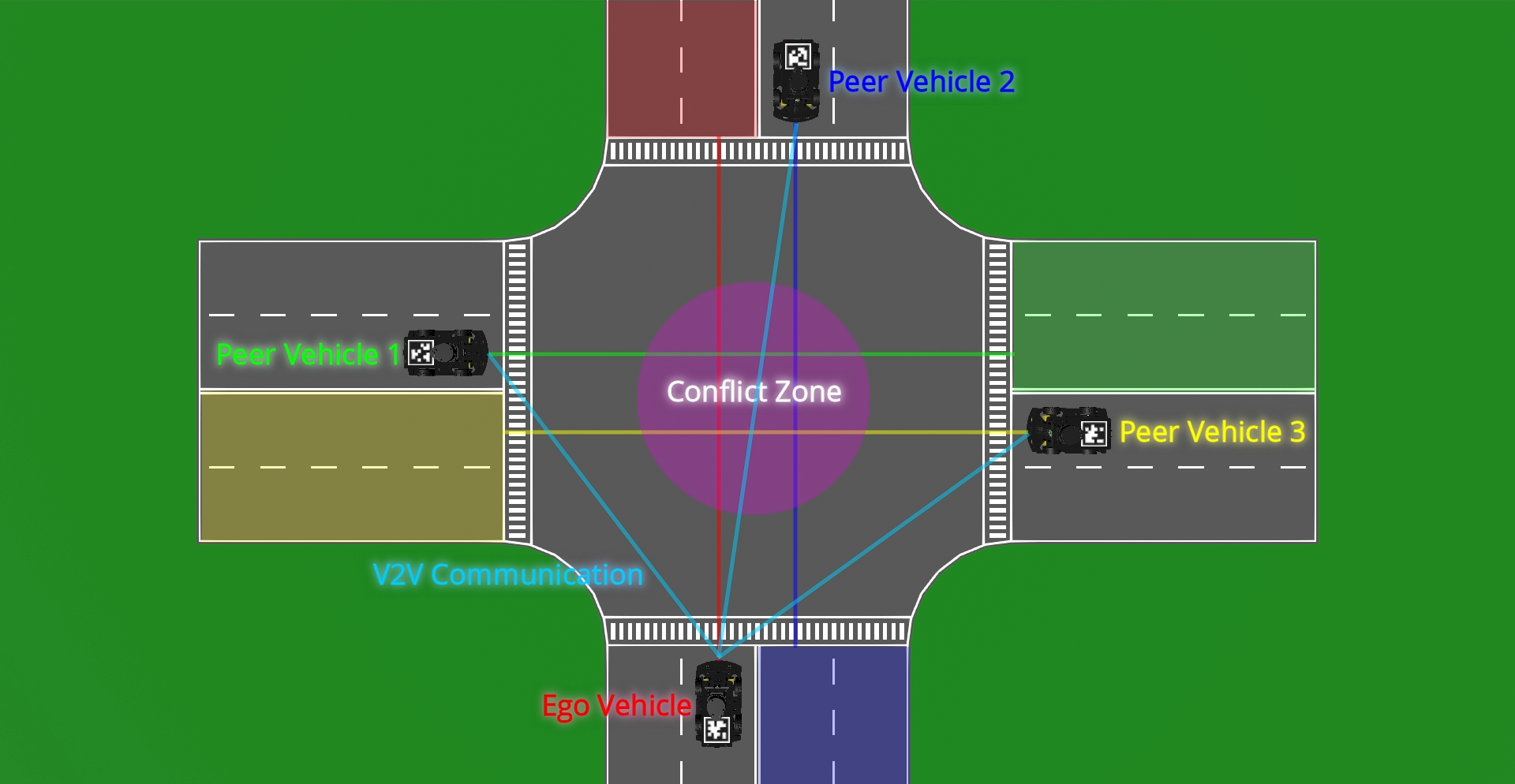}}
	\subfigure[]{\includegraphics[width=0.49\textwidth]{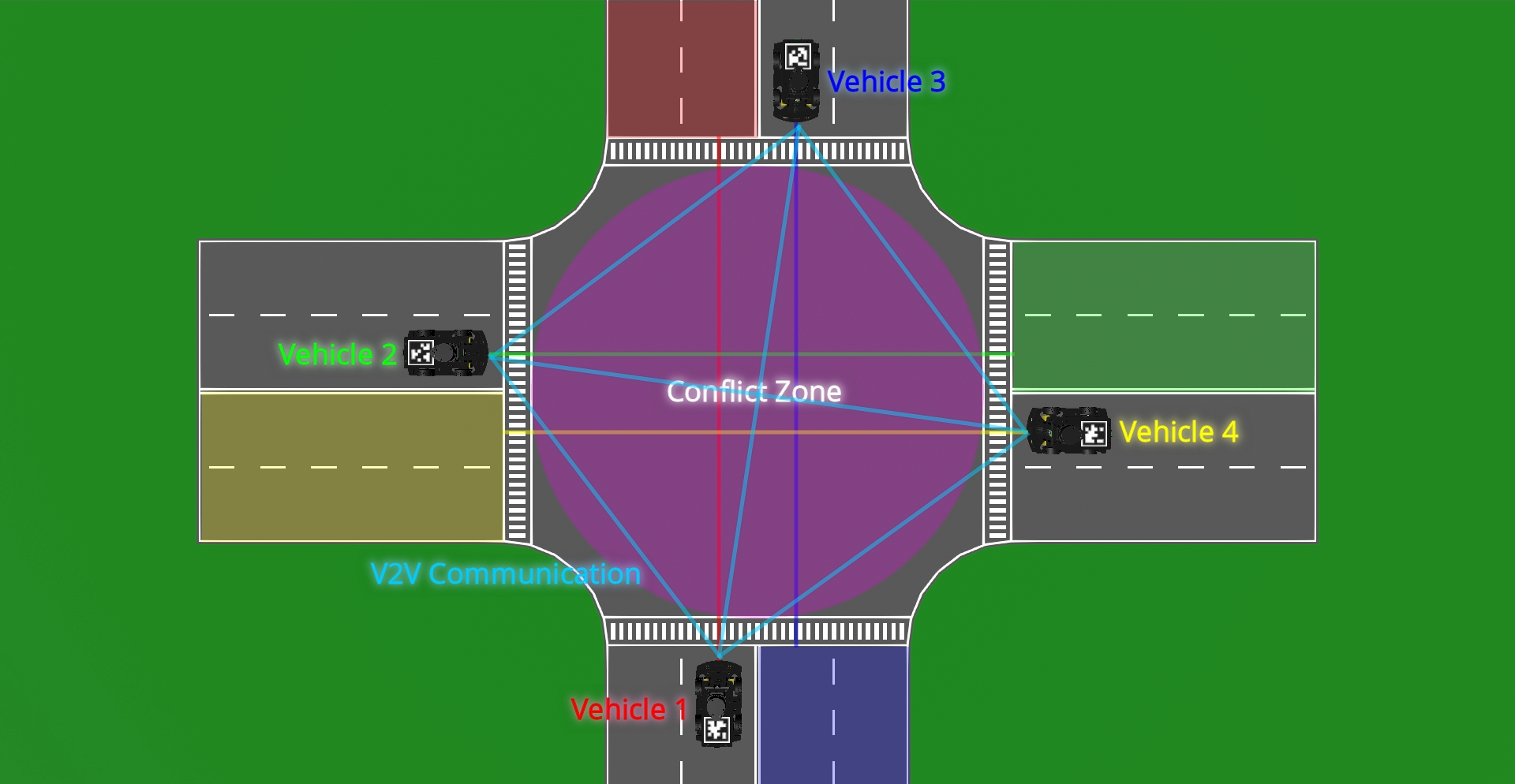}}
	\caption{Learning Scenarios: (a) Single-Agent, and (b) Multi-Agent}
	\label{Figure: DRL Scenarios}
\end{figure}

As discussed earlier, this particular application demonstrates both single and multi-agent scenarios for autonomous intersection traversal. Following is a brief summary of the two, with reference to Figure \ref{Figure: DRL Scenarios}:

\begin{itemize}
	\item \textbf{Single-Agent Learning Scenario:} Only the ego vehicle learnt to traverse the intersection, while the peer vehicles were controlled at different velocities using a simple heuristic. The peer vehicles shared their states with ego vehicle using V2V communication. All the vehicles were reset collectively, making this scenario quite deterministic.
	\item \textbf{Multi-Agent Learning Scenario:} All the vehicles learnt to traverse the intersection simultaneously, in a decentralized manner. The vehicles shared their states with each other using V2V communication, and were reset independently, making this scenario highly stochastic.
\end{itemize}

Nevertheless, in both of these scenarios, the problem being addressed is that of autonomous navigation in an unknown environment, wherein the exact structure of the environment is not known to any agent (no map or other data is provided, whatsoever), and consequently they cannot completely observe/obtain their state $s_t$. Thus, the problem of decision making for motion planning and control of a non-holonomic vehicle for intersection traversal can be defined as a Partially Observable Markov Decision Process (POMDP), which can capture hidden state information through the analysis of environmental observations.

\subsubsection{State Space}
\label{Sub-Sub-Section: State Space}

As described earlier, the state space $S$ for this particular problem of intersection traversal can be divided into observable $s^o \subset S$ and hidden $s^h \subset S$ components. While the former consists of vehicle's 2D pose and velocity, $s_{t}^{o}=\left [ p_{x}, p_{y}, \psi, v \right ]_{t} \in \mathbb{R}^{4}$, the later consists of the agent's goal coordinates, $s_{t}^{h}=\left [ g_{x}, g_{y} \right ]_{t} \in \mathbb{R}^{2}$. In other words, every agent can observe pose and velocity of its peers (through V2V communication), whereas its own goal location is only observable to itself (i.e. hidden from the others). Therefore, the complete state space of an agent participating in this problem is defined as a vector containing all the observable and hidden states:
\begin{equation}
\label{Equation: 5.31}
s_{t} = \left [ s_{t}^{o}, s_{t}^{h} \right ]
\end{equation}

\subsubsection{Observation Space}
\label{Sub-Sub-Section: Observation Space}

Based on the state space of an agent defined in Equation \ref{Equation: 5.31}), appropriate subset of vehicle's sensor suite was employed to collect the required observations (refer Equation \ref{Equation: 5.32}). This included IPS to get positional coordinates $\left [ p_{x}, p_{y} \right ]_{t} \in \mathbb{R}^{2}$, IMU to get yaw $\psi_{t} \in \mathbb{R}^{1}$, and incremental encoders to estimate the vehicle velocity $v_{t} \in \mathbb{R}^{1}$. However, instead of supplying raw sensor data, every agent $i$, where $0<i<N$, was provided with an observation vector of the form:
\begin{equation}
\label{Equation: 5.32}
o_{t}^{i} = \left [ g^{i}, \tilde{p}^{i}, \tilde{\psi}^{i}, \tilde{v}^{i} \right ]_{t} \in \mathbb{R}^{2+4(N-1)}
\end{equation}
where, $i$ represents ego-agent and $j \in \left [ 0, N-1 \right ]$ represents every other (peer) agent in the scene with a total of $N$ agents. This follows that $g_{t}^{i} = \left [ g_{x}^{i}-p_{x}^{i}, g_{y}^{i}-p_{y}^{i} \right ]_{t} \in \mathbb{R}^{2}$ is ego agent's goal location relative to itself, $\tilde{p}_{t}^{i} = \left [ p_{x}^{j}-p_{x}^{i}, p_{y}^{j}-p_{y}^{i} \right ]_{t} \in \mathbb{R}^{2(N-1)}$ is position of every peer agent relative to the ego agent, $\tilde{\psi}_{t}^{i} = \psi_{t}^{j}-\psi_{t}^{i} \in \mathbb{R}^{N-1}$ is yaw of every peer agent relative to the ego agent, and $\tilde{v}_{t}^{i} = v_{t}^{j} \in \mathbb{R}^{N-1}$ is velocity of every peer agent.

\subsubsection{Action Space}
\label{Sub-Sub-Section: Action Space}

The vehicles were designed as non-holonomic rear-wheel drive models with Ackermann steering mechanism. Consequently, the complete action space of an agent comprised of longitudinal (throttle/brake) and lateral (steering) motion control commands. For longitudinal control, the throttle command $\tau_t$ was set to 80\% of its upper saturation limit. The steering command $\delta_t$, on the other hand, was discretized as $\delta_t \in \left \{ -1, 0, 1 \right \}$, and was the only source of active control (refer Equation \ref{Equation: 5.33}) for the vehicles to safely navigate across the intersection:
\begin{equation}
\label{Equation: 5.33}
a_t = \delta_t \in \mathbb{R}^{1}
\end{equation}

\subsubsection{Reward Function}
\label{Sub-Sub-Section: Reward Function}

The extrinsic reward function (refer Equation \ref{Equation: 5.34}) was defined so as to reward each agent with $r_{goal}=+1$ for successfully traversing the intersection, or penalize it proportional to its distance from goal, i.e. $k_p * \left \| g_{t}^{i} \right \|_{2}$, for colliding or overstepping any lane markings. The penalty constant, $k_p$, was defined to be 0.425, which would yield the agent a maximum penalty of 1.
\begin{align}
\label{Equation: 5.34}
r_{t}^{i} = \begin{cases}
r_{goal}; \textup{\qquad\qquad if traversed the intersection safely}\\
-k_p * \left \| g_{t}^{i} \right \|_{2}; \textup{\;\, if collided or overstepped lane markings}
\end{cases}
\end{align}

This forced the agents to get closer to their respective goals, in order to reduce the penalty, ultimately leading them to the global optima -- a positive reward, $r_{goal}$. Not only did this result in a faster convergence, but it also severely restricted reward hacking.

\subsubsection{Optimization Problem}
\label{Sub-Sub-Section: Optimization Problem}

The task of intersection traversal, subject to collision avoidance and lane-keeping, is handled by the extrinsic reward function defined in Equation \ref{Equation: 5.34}, which encourages the agent to achieve its objective. The central idea for each individual agent is, therefore, to maximize the expected future discounted reward (refer Equation \ref{Equation: 5.35}) by learning a policy $\pi_\theta \left(a_t|o_t\right)$, which eventually transitions into the optimal policy $\pi^*$ as the training progresses.
\begin{align}
\label{Equation: 5.35}
\argmax_{\pi_\theta \left(a_t|o_t\right)} \quad &\mathbb{E}\left [ \sum_{t=0}^{\infty} \gamma^t r_t \right ]
\end{align}

\subsection{Training}
\label{Sub-Section: DRL Training}

\begin{figure}[htpb]
	\centering
	\includegraphics[width=0.59\textwidth]{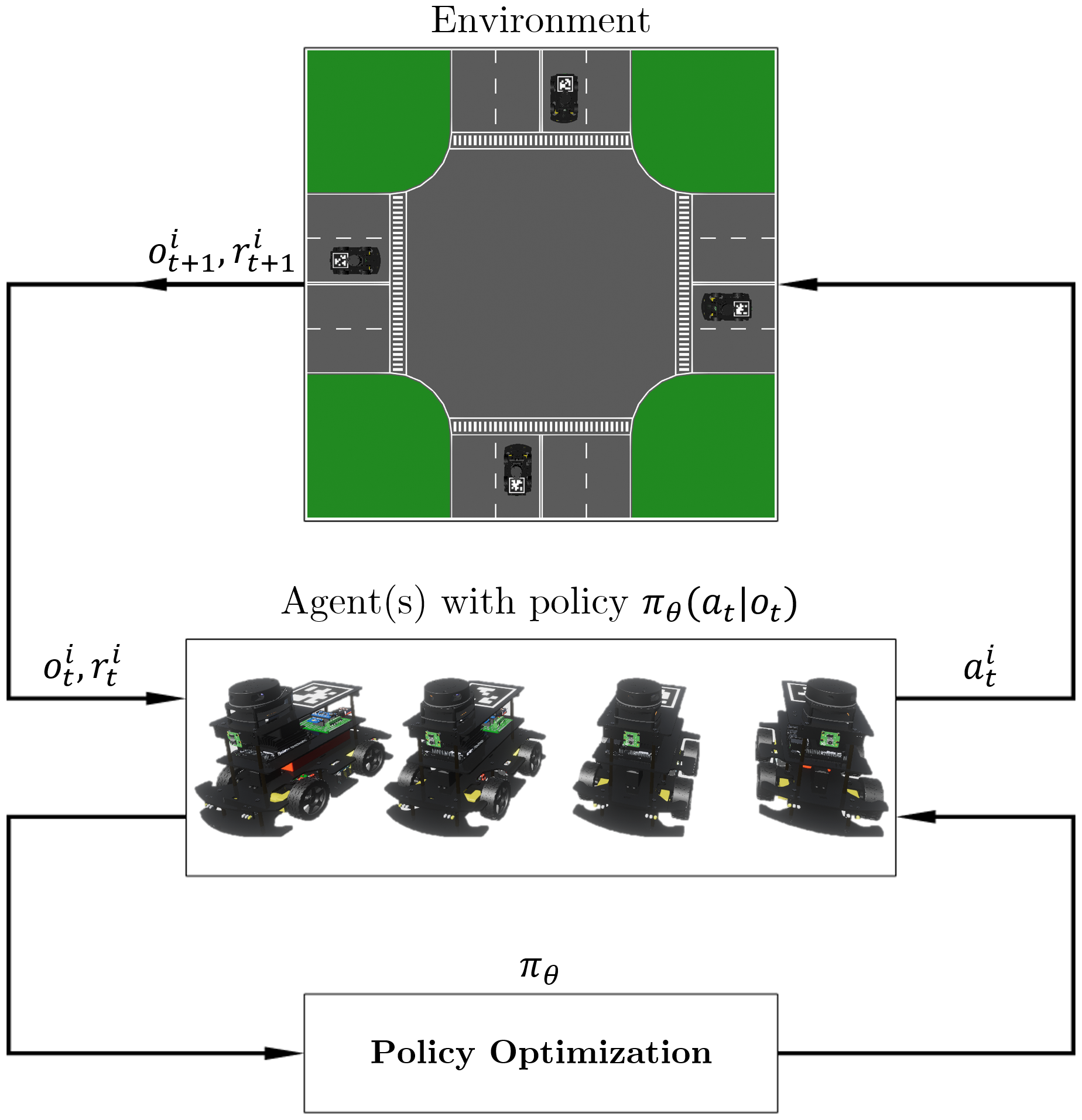}
	\caption{Deep Reinforcement Learning Architecture}
	\label{Figure: DRL Architecture}
\end{figure}

Figure \ref{Figure: DRL Architecture} illustrates the generalized learning architecture applicable to both single as well as multi-agent paradigms. At every $t$-th instant, each agent $i$ collects an observation vector $o_t^i$ and an extrinsic reward $r_t^i$, based on which, it takes an action $a_t^i$, governed by the policy $\pi_\theta$ updated thus far. Parallel to this observation-action loop, the policy $\pi_\theta$ is optimized online, with an aim of maximizing the expected future discounted reward (refer Section \ref{Sub-Section: Problem Formulation}).

This particular application adopted a fully connected neural network (FCNN) as a function approximator, $\pi_\theta \left ( a_t | o_t \right )$. The network comprised of $\mathbb{R}^{14}$ inputs (comparable to the size of observation vector), $\mathbb{R}^{1}$ outputs (comparable to the size of action vector), and three hidden layers with 128 neural units, each. Naturally, the policy parameters $\theta \in \mathbb{R}^d$ were defined in terms of the network parameters (i.e. weights and biases). This policy was trained to predict steering command for the vehicle directly based on the collected observations, using the proximal policy optimization (PPO) algorithm \cite{Schulman2017}. Table \ref{Table: DRL Training Configuration} hosts details pertaining to the training configuration adopted for this application.

\begin{table}[htpb]
	\centering
	\caption{Training Configuration}
	\label{Table: DRL Training Configuration}
	\resizebox{0.55\textwidth}{!}{%
		\begin{tabular}{ll}
			\hline
			\textbf{Hyperparameter}                   & \textbf{Value} \\ \hline
			Batch size                                & 1024           \\
			Buffer size                               & 10240          \\
			Learning rate ($\alpha$)                  & 3.0e-4         \\
			Learning rate schedule                    & Linear         \\
			Entropy regularization strength ($\beta$) & 5.0e-3         \\
			Policy update hyperparameter ($\epsilon$) & 0.2            \\
			Regularization parameter ($\lambda$)      & 0.97           \\
			Discount factor ($\gamma$)                & 0.99           \\
			Extrinsic reward strength                 & 1.0            \\
			Number of epochs                          & 3              \\
			Maximum steps [SA]                        & 1.0e6          \\
			Maximum steps [MA]                        & 2.5e6          \\ \hline
		\end{tabular}%
	}
\end{table}

The single and multi-agent paradigms were trained separately, and their learning progress was analysed. The two exhibited quite different behaviours, owing to the difference in difficulty of respective settings. Cumulative reward, episode length and policy entropy metrics were chosen for analysis, since they reveal the most significant information as to how the agents learn.

\begin{figure}[htpb]
	\centering
	\subfigure[]{\includegraphics[width=0.32\textwidth]{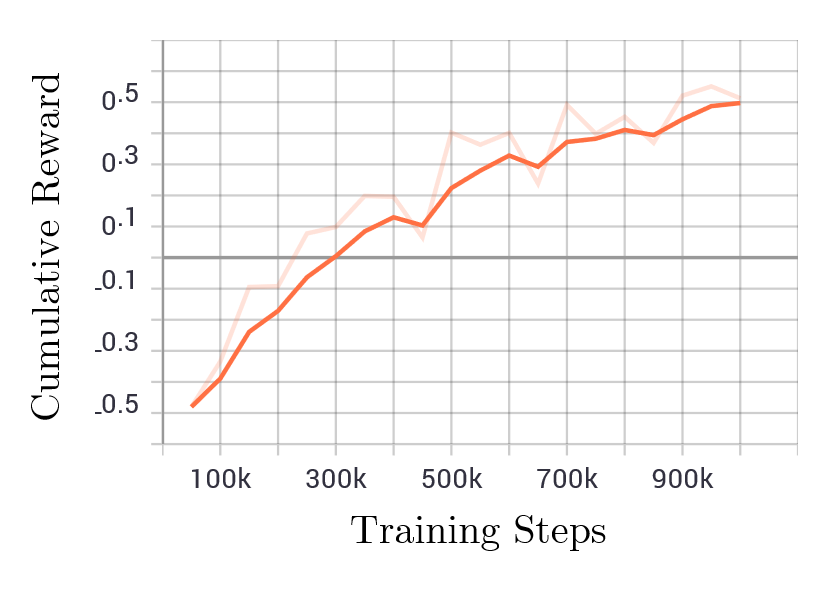}}
	\subfigure[]{\includegraphics[width=0.32\textwidth]{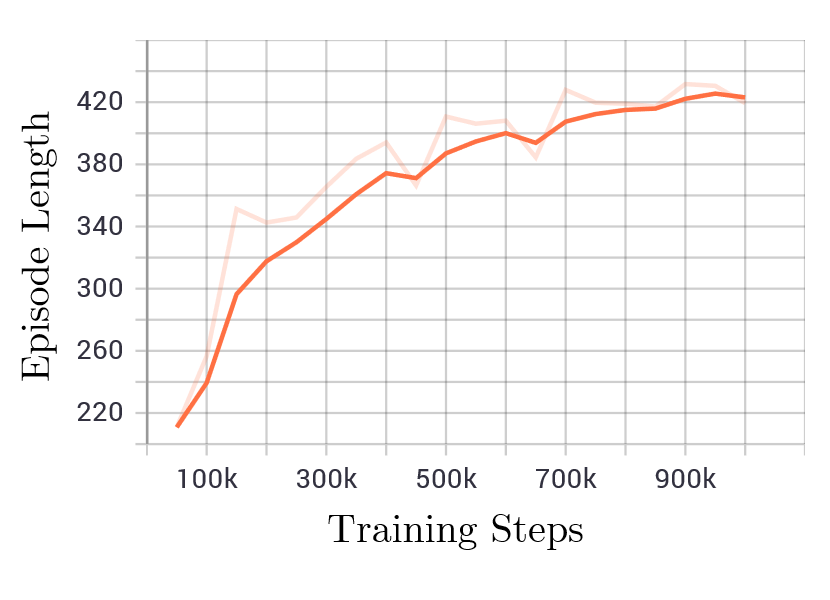}}
	\subfigure[]{\includegraphics[width=0.32\textwidth]{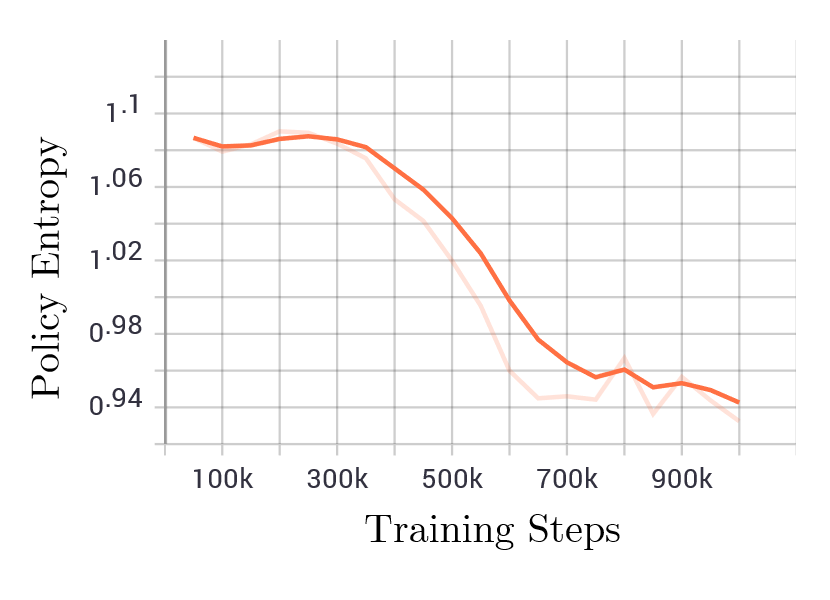}}
	\caption{Training Results for Single-Agent Paradigm: (a) Cumulative Reward, (b) Episode Length, and (c) Policy Entropy}
	\label{Figure: SATR}
\end{figure}

Learning analysis of the single-agent learning scenario (refer Figure \ref{Figure: SATR}) suggests that the training was quite stable. The cumulative reward initially rose and then settled (suggesting that agent had achieved global maxima, and exploited it), the episode length was stabilized at a reasonable value (suggesting that neither was the agent colliding before reaching its goal, nor was it wandering off in search of better reward), and the policy entropy had reduced gradually and stagnated at its minimal value (indicating that the policy had converged, and the agent was quite certain in choosing a particular action).

\begin{figure}[htpb]
	\centering
	\subfigure[]{\includegraphics[width=0.32\textwidth]{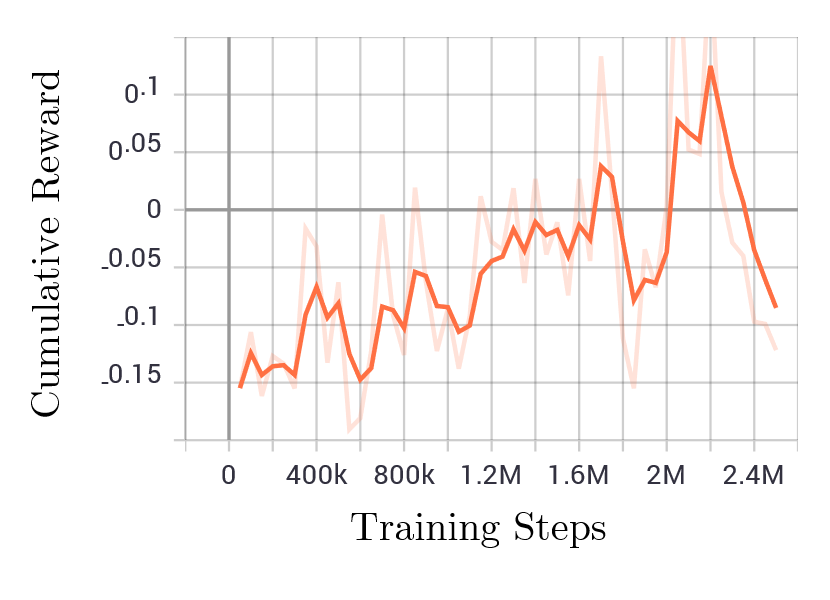}}
	\subfigure[]{\includegraphics[width=0.32\textwidth]{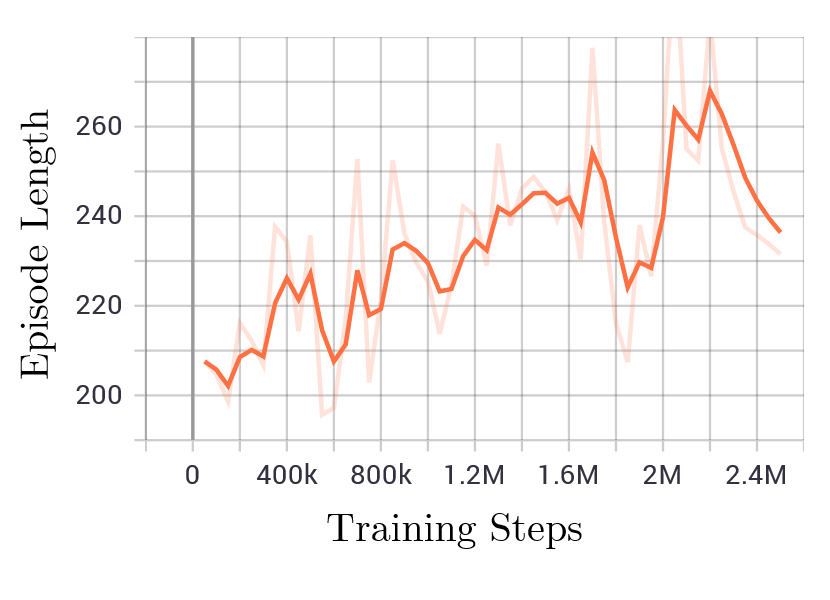}}
	\subfigure[]{\includegraphics[width=0.32\textwidth]{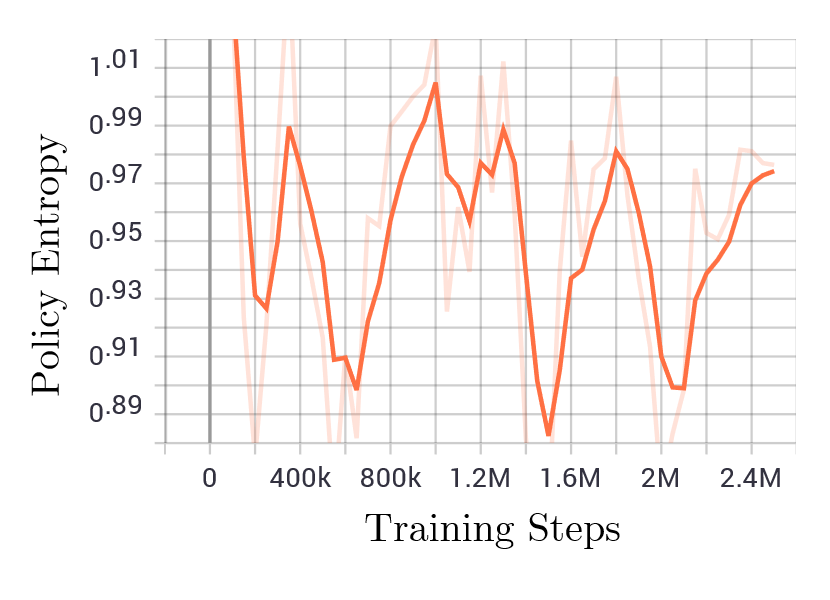}}
	\caption{Training Results for Multi-Agent Paradigm: (a) Cumulative Reward, (b) Episode Length, and (c) Policy Entropy}
	\label{Figure: MATR}
\end{figure}

Learning analysis of the multi-agent learning scenario (refer Figure \ref{Figure: MATR}), on the contrary, suggests that the training was rather unstable. This was primarily due to the highly stochastic nature of the designed scenario, which follows that there was no fixed solution to this problem. Additionally, there were several unjust instances throughout the training phase, wherein the vehicle(s) could not avoid colliding/overstepping lane markings in any case. Nevertheless, the agents ultimately learnt to traverse the intersection safely; although, their decisions seemed uncertain at times (e.g. the vehicles switched their actions rapidly).

\subsection{Deployment}
\label{Sub-Section: DRL Deployment}

The trained policies were deployed back onto the simulated vehicles, again independently for each of the two scenarios. As described earlier, the single-agent learning scenario was quite deterministic, and hence the ego vehicle was able to traverse the intersection safely almost every time. The multi-agent learning scenario, on the other hand, was highly stochastic, owing to which the vehicles had significantly low chance of success (especially that of all the vehicles being able to traverse the intersection safely, at once).

\begin{figure}[htpb]
	\centering
	\subfigure[]{\includegraphics[width=0.32\textwidth]{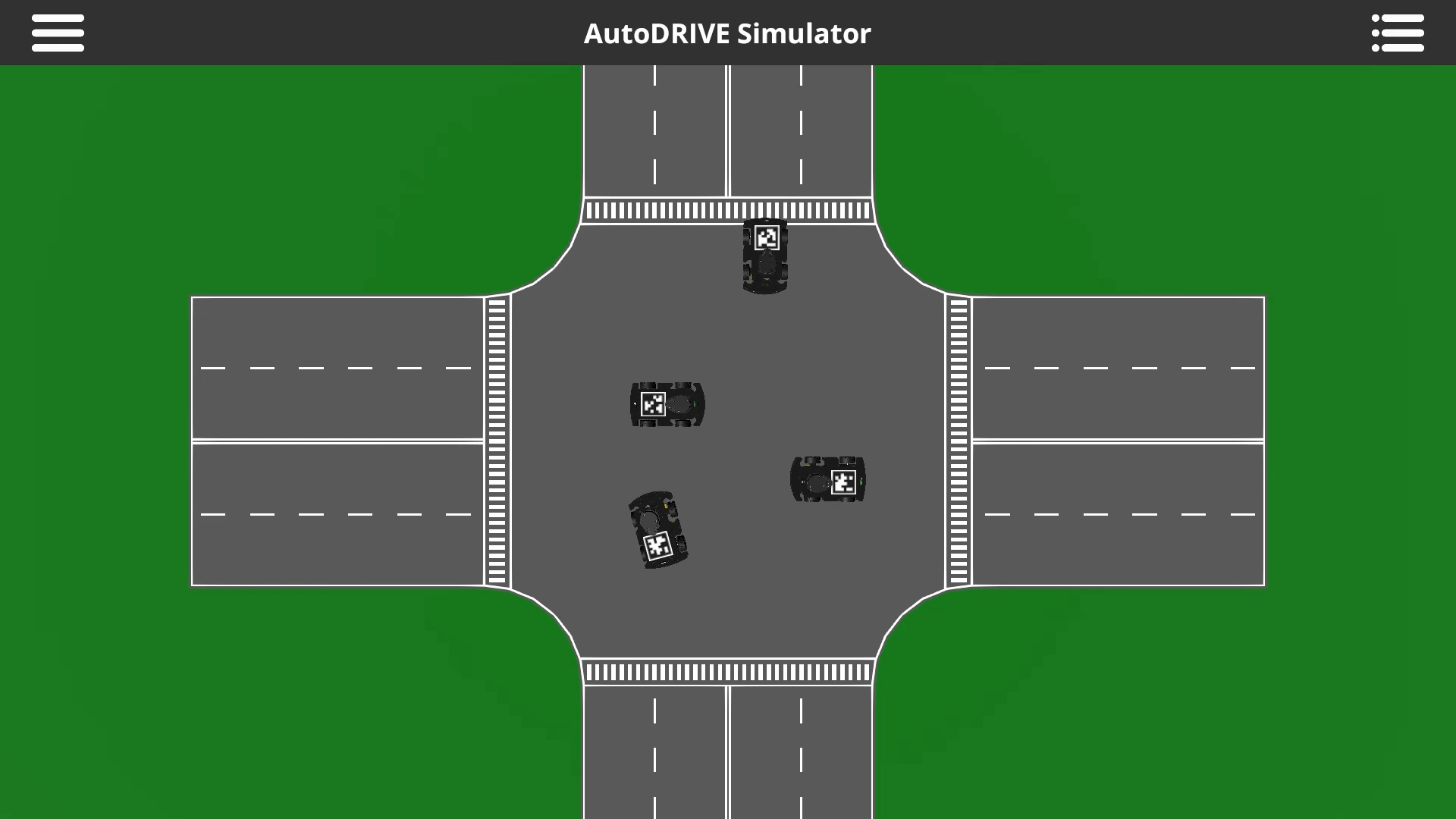}}
	\subfigure[]{\includegraphics[width=0.32\textwidth]{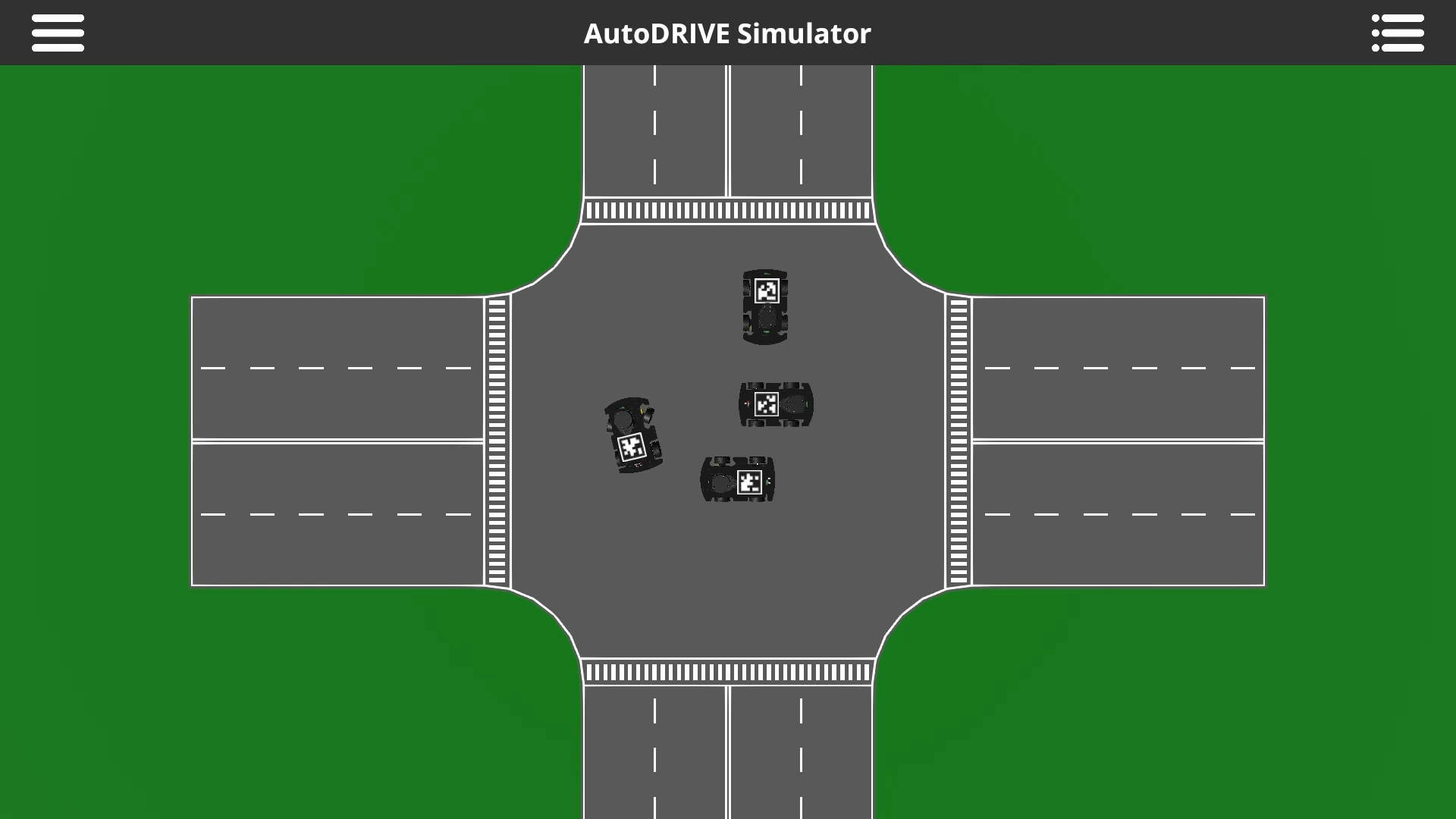}}
	\subfigure[]{\includegraphics[width=0.32\textwidth]{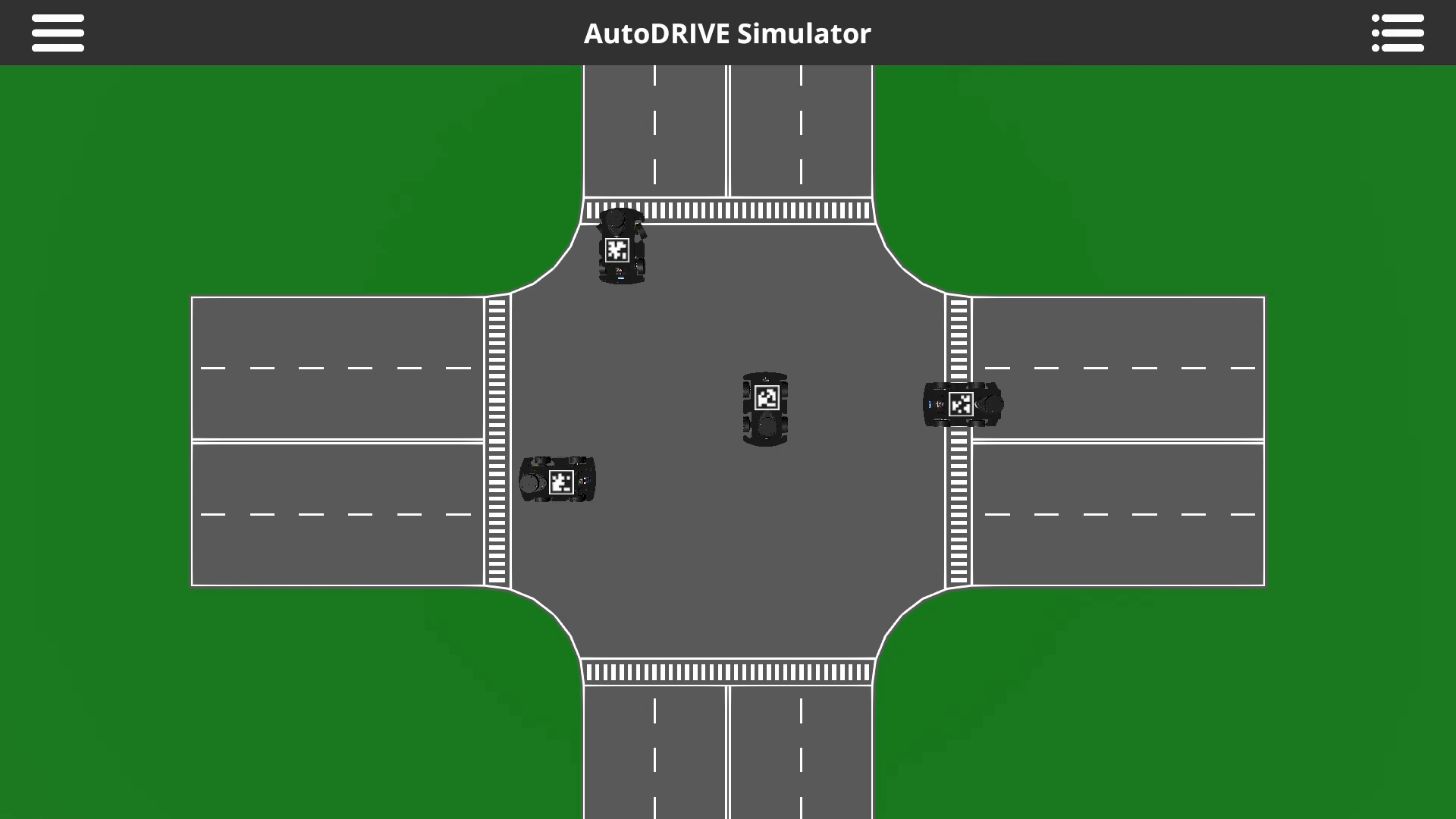}}
	\caption{Deployment Results for Single-Agent Paradigm: (a) Stage 1, (b) Stage 2, and (c) Stage 3}
	\label{Figure: SADR}
\end{figure}

Figure \ref{Figure: SADR} depicts three prominent stages of single-agent intersection traversal scenario. First, the ego vehicle approaches the conflict zone, where it could collide with either of its peers. Next, it performs a left-hand curve manoeuvre to avoid colliding with any of its peers. Finally, the vehicle performs a subtle right-hand curve manoeuvre to reach its goal. It was observed that the vehicle preferred approaching the left goal lane (since it was safer to do so); nevertheless, it also approached the right goal lane certain times.

\begin{figure}[htpb]
	\centering
	\subfigure[]{\includegraphics[width=0.32\textwidth]{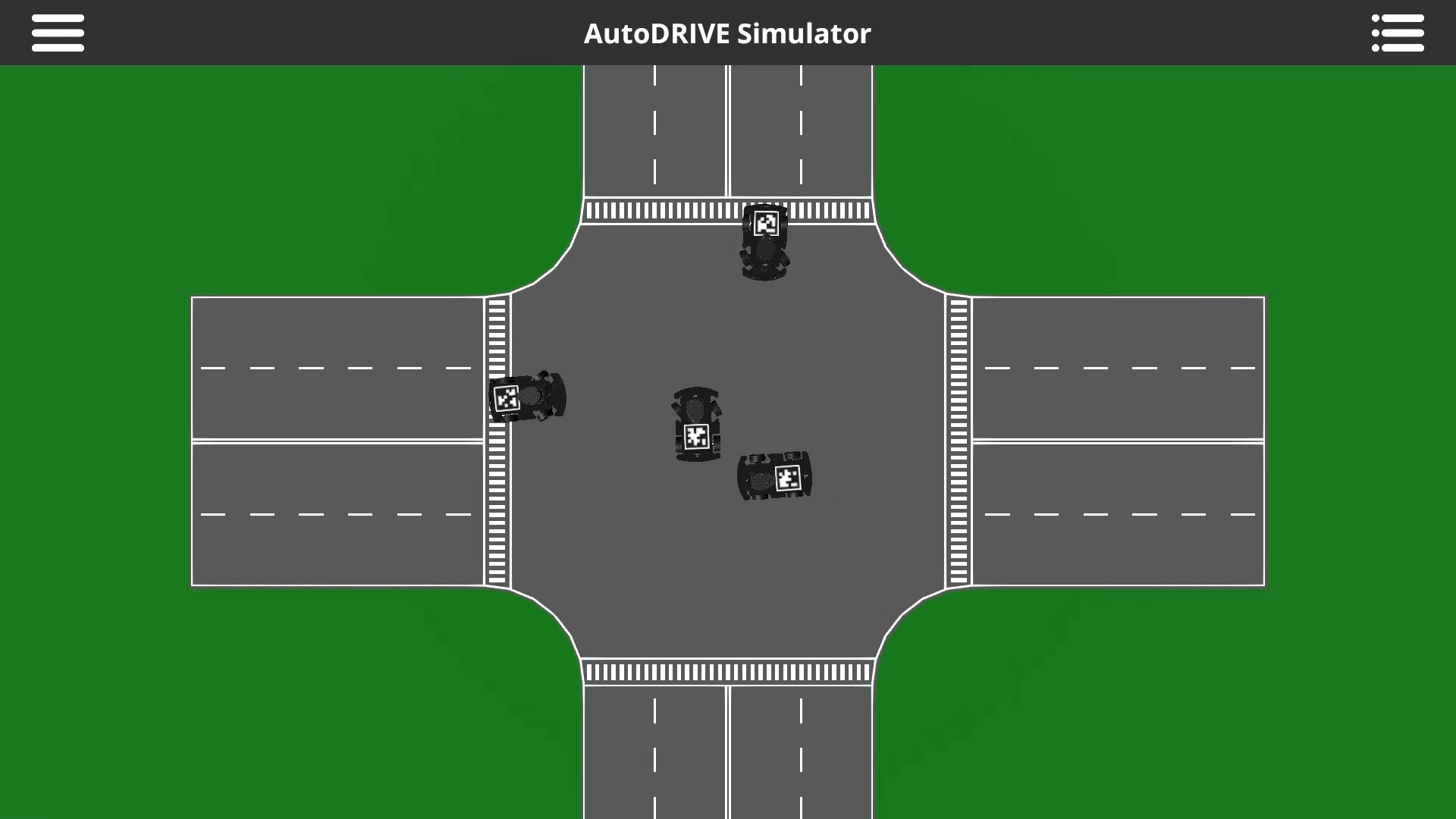}}
	\subfigure[]{\includegraphics[width=0.32\textwidth]{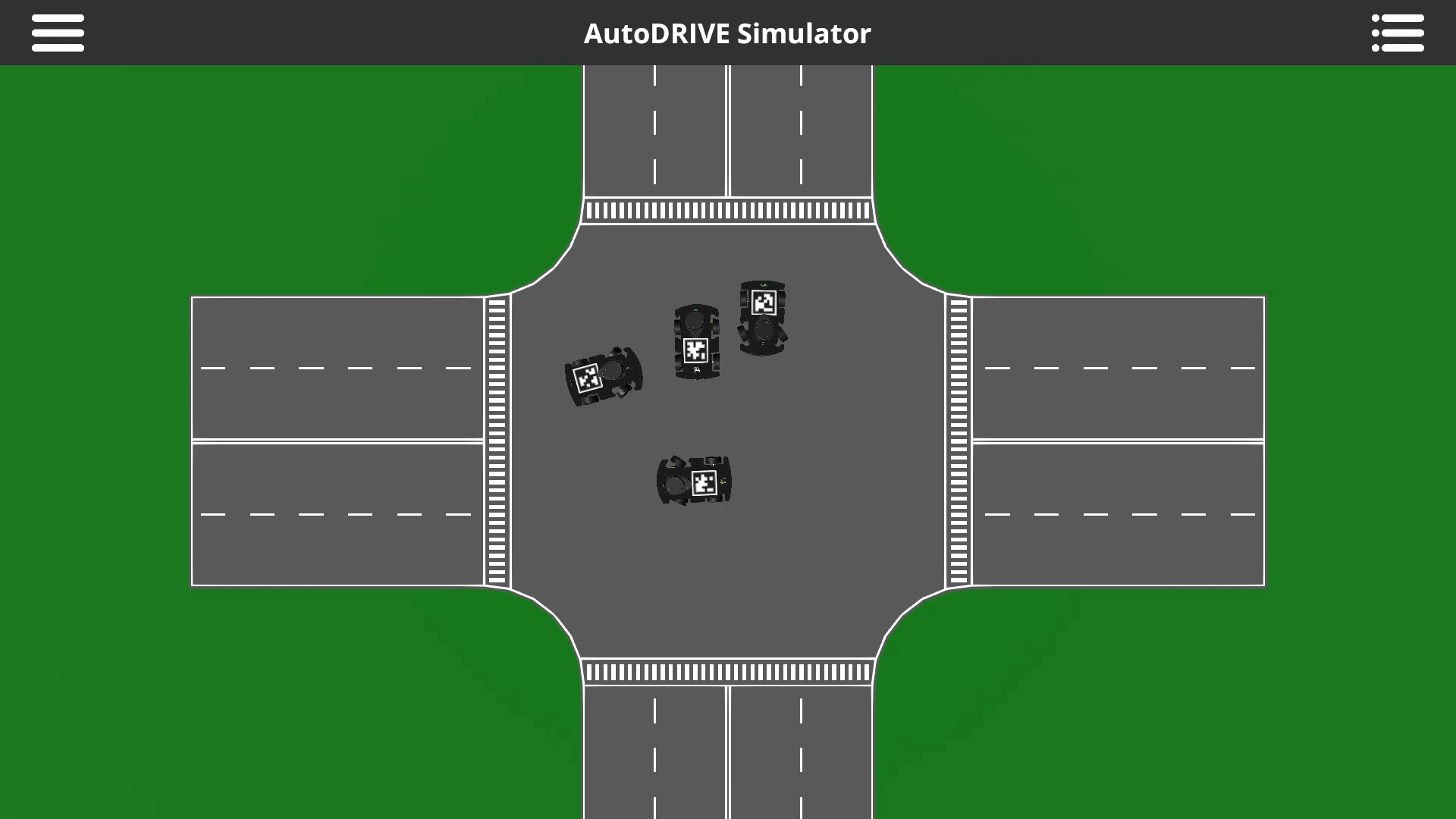}}
	\subfigure[]{\includegraphics[width=0.32\textwidth]{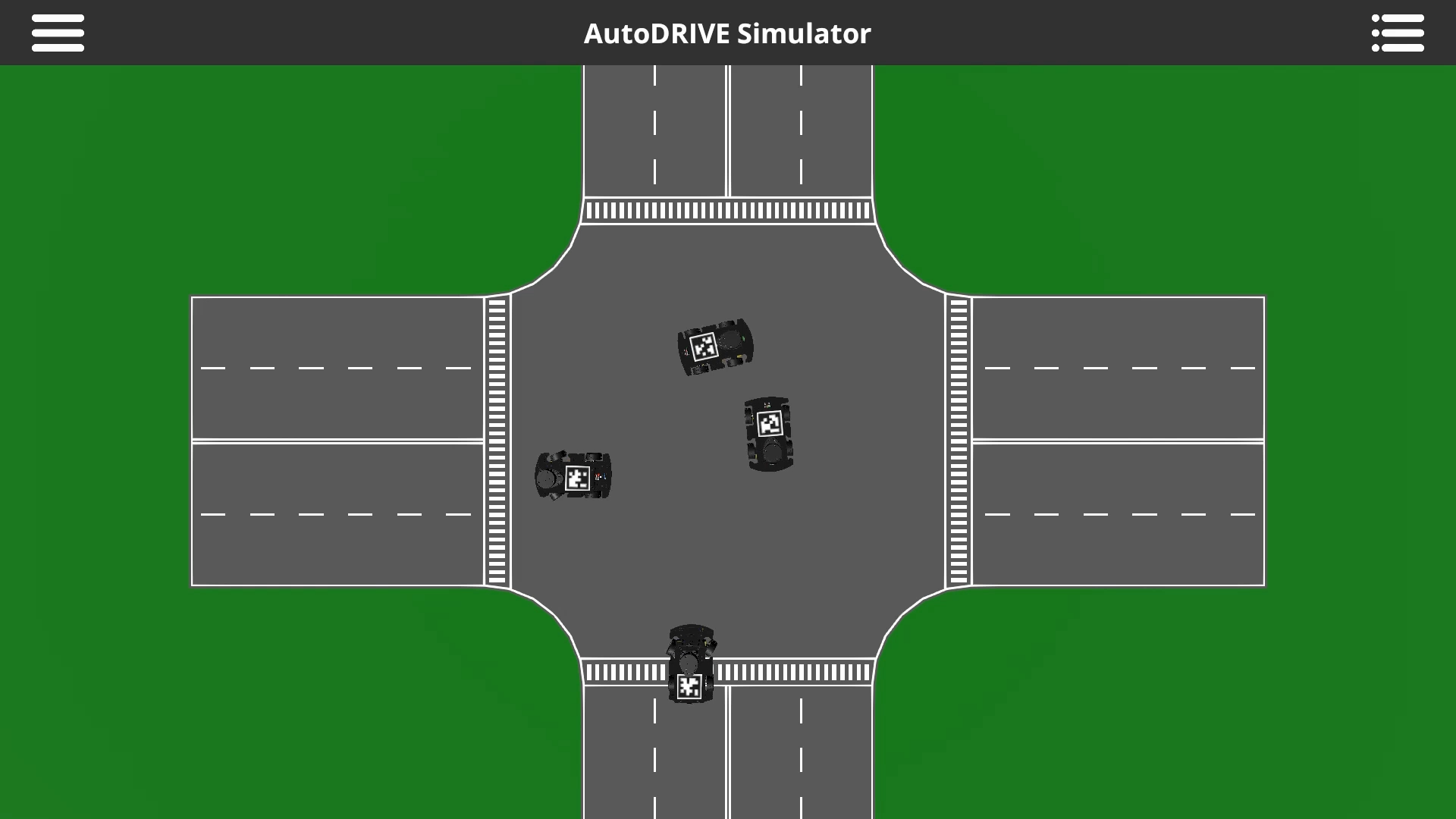}}
	\caption{Deployment Results for Multi-Agent Paradigm: (a) Stage 1, (b) Stage 2, and (c) Stage 3}
	\label{Figure: MADR}
\end{figure}

Figure \ref{Figure: MADR} depicts three prominent stages of multi-agent intersection traversal scenario. The first frame shows vehicles 1 and 4 avoiding collision, the second one shows vehicle 1 finding a gap between vehicles 2 and 3 in order to reach its goal, and the third one shows vehicles 2 and 3 avoiding collision, while vehicle 4 is approaching its goal and vehicle 1 is respawned.

\section{Smart City Management}
\label{Section: Smart City Management}

The application of smart city management was chosen to demonstrate AutoDRIVE's ability to support development of intelligent transportation algorithms interlinked with the infrastructural setup. This demonstration employed vehicle to infrastructure (V2I) communication and internet of things (IoT) techniques, involving the following stages:

\begin{itemize}
	\item Surveillance
	\item Planning
	\item Control
\end{itemize}

This application demonstrates a traffic management scenario, wherein the entire software stack governing all autonomy operations was executed by a centralized smart city manager (SCM) server. The server was provided with an HD map of the environment, which then kept track of the vehicle using V2I communication, along with the IoT-enabled active and passive traffic elements present in the scene. The vehicle was actively controlled to follow traffic rules set forth by the road markings, traffic signs, and traffic lights. The vehicle's only objective was to blindly follow motion control commands generated by the SCM server to drive autonomously. The entire application was developed using AutoDRIVE Devkit's SCSS toolkit.

\subsection{Surveillance}
\label{Sub-Section: Surveillance}

\begin{figure}[htpb]
	\centering
	\subfigure[]{\includegraphics[width=0.32\textwidth]{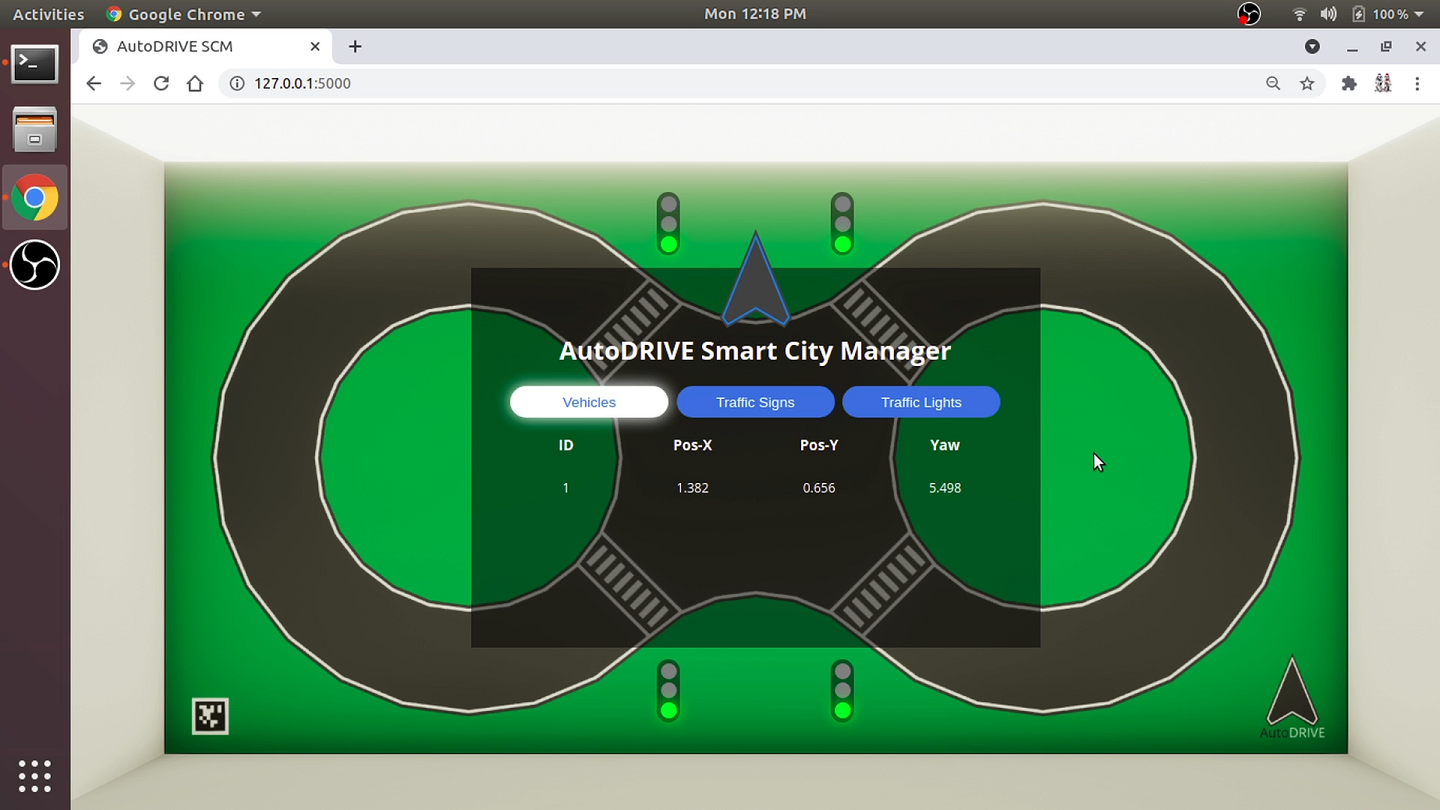}}
	\subfigure[]{\includegraphics[width=0.32\textwidth]{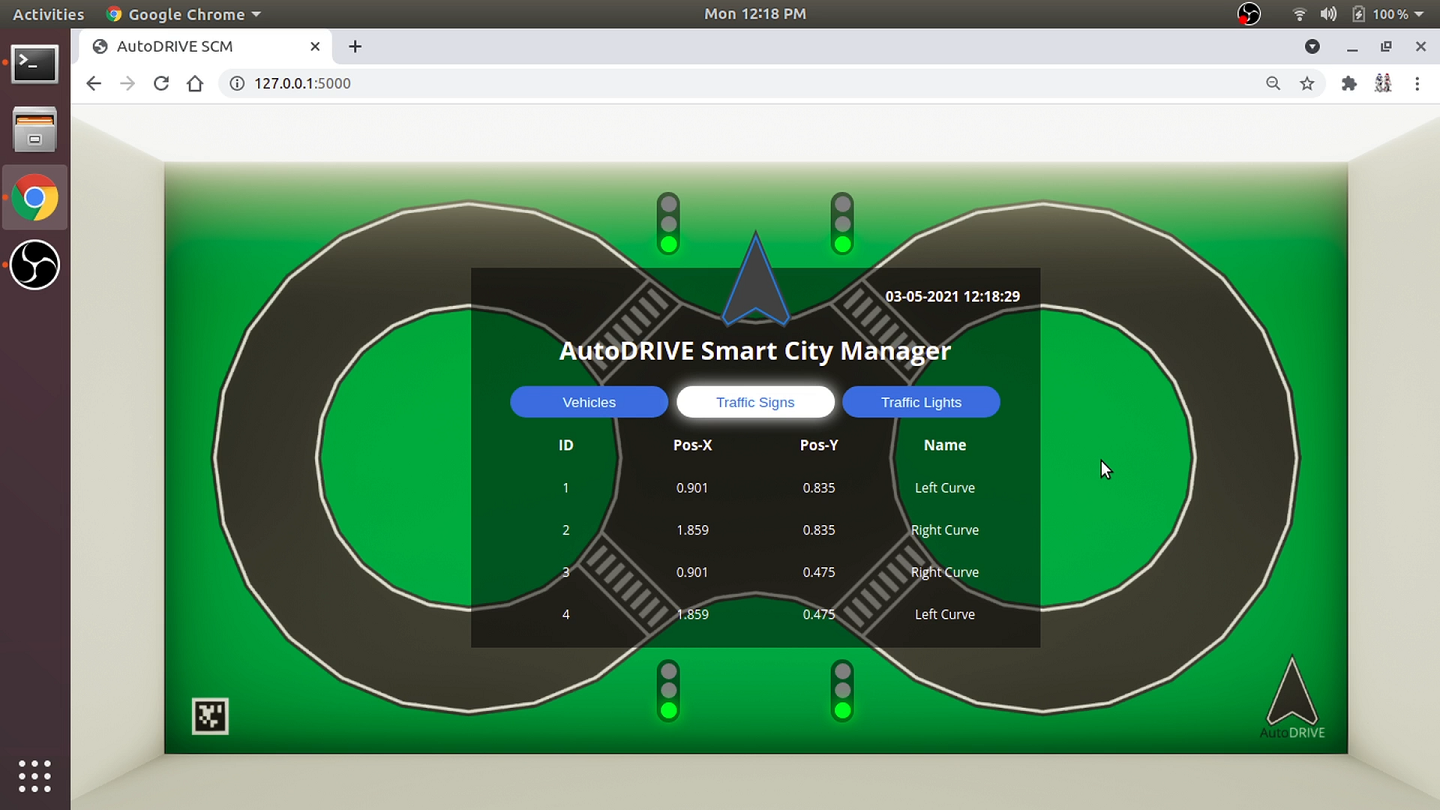}}
	\subfigure[]{\includegraphics[width=0.32\textwidth]{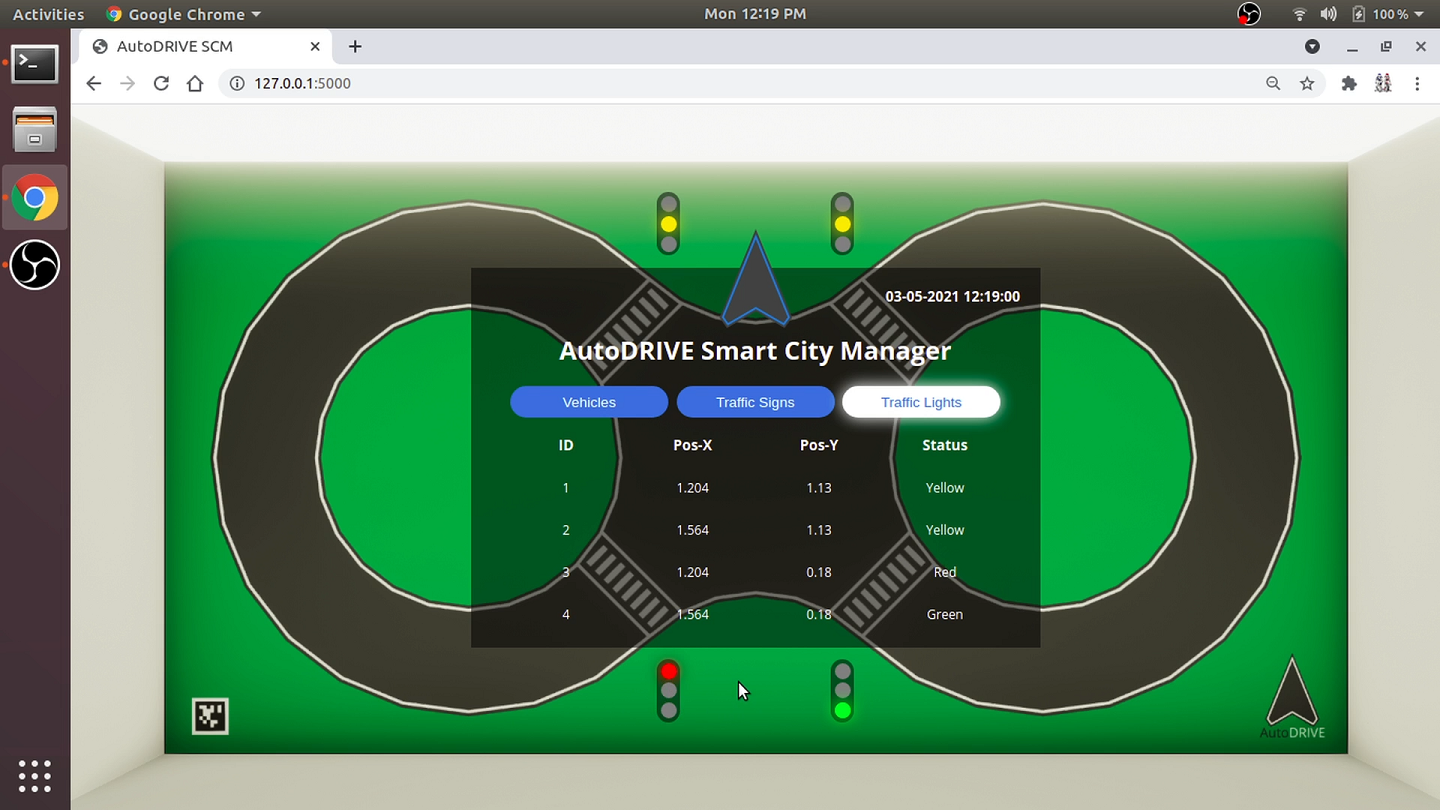}}
	\caption{SCM Database: (a) Vehicles, (b) Traffic Signs, and (c) Traffic Lights}
	\label{Figure: Database}
\end{figure}

The AutoDRIVE SCM server monitors all the key elements present within a particular scene. Particularly, it hosts a database to keep track of all the vehicles along with traffic signs and lights. It can identify individual vehicles and track their 2D pose estimates in real-time. It can also store the positional coordinates of all traffic signs and lights (as a part of the HD map), along with their states (sign names and light status). Figure \ref{Figure: Database} respectively illustrates the database tables including one each for vehicles, traffic signs and traffic lights.

\subsection{Planning}
\label{Sub-Section: Planning}

As described in Algorithm \ref{Algorithm: Smart City Management} the SCM server has knowledge of vehicle(s), traffic lights and traffic signs in the scene, since it has access to the centralized database. It initializes the throttle $\tau$, steering $\delta$ and steering trim \textbf{trim}($\delta$) to $0$.

\begin{algorithm}[h!]
	\SetAlgoLined
	\caption{Smart City Management}
	\label{Algorithm: Smart City Management}
	\KwIn{HD map $\mathcal{M}$, vehicle(s) states $\mathcal{X}$, traffic light set $\mathcal{L}$, traffic sign set $\mathcal{S}$}
	\KwOut{Vehicle control inputs $\tau$, $\delta$}
	\nl $\tau:=0$, $\delta:=0$, \textbf{trim}($\delta$) $:=0$\\
	\nl \While{(true)}{
		\nl \textbf{update}($\mathcal{X}$, $\mathcal{L}$, $\mathcal{S}$)\\
		\nl $\tau,\delta\leftarrow$ \textbf{generate}($\tau,\delta|\mathcal{X},\mathcal{M}$, \textbf{trim}($\delta$))\\
		\nl \uIf{(\textup{state of} $l^{(i)}\in\mathcal{L}|\mathcal{X},\mathcal{M}=$ \textup{red})}{
			\nl $\tau=0$\\
		}
		\nl \uElseIf{(\textup{state of} $l^{(i)}\in\mathcal{L}|\mathcal{X},\mathcal{M}=$ \textup{yellow})}{
			\nl $\tau=0$\\
		}
		\nl \uElse{
			\nl \textbf{continue}\\
		}
		\nl \uIf{(\textup{state of} $s^{(i)}\in\mathcal{S}|\mathcal{X},\mathcal{M}=$ \textup{left\_curve})}{
			\nl \textbf{trim}($\delta$) $=-0.7$\\
		}
		\nl \uElseIf{(\textup{state of} $s^{(i)}\in\mathcal{S}|\mathcal{X},\mathcal{M}=$ \textup{right\_curve})}{
			\nl \textbf{trim}($\delta$) $=0.7$\\
		}
		\nl \uElse{
			\nl \textbf{trim}($\delta$) $=0$\\
		}
		\nl \Return{$\tau,\delta$}\\
	}
\end{algorithm}

After this initialization step, an infinite loop begins by updating the states of vehicle $\mathcal{X}$, traffic lights $\mathcal{L}$ and traffic signs $\mathcal{S}$ to their most recent values. A high-level controller generates vehicle control inputs based on its current state w.r.t. the map $\mathcal{M}$ and steering trim in order to keep it driving over the roadway.

Next, the server checks status of traffic light $l^{(i)}\in\mathcal{L}$ currently of interest to the vehicle using its state information $\mathcal{X}$ w.r.t. map $\mathcal{M}$. If the status of this light is ``red'' or ``yellow'', throttle command of vehicle is set to zero to command a firm stop. Else, if the state of traffic light is ``green'', no high-priority signal is given to the vehicle and it keeps following the high-level controller.

Further, the server checks status of traffic sign $s^{(i)}\in\mathcal{S}$ currently of interest to the vehicle using its state information $\mathcal{X}$ w.r.t. map $\mathcal{M}$. If the traffic sign reads ``left curve'', the steering trim is set to $-0.7$ so as to advise the high-level controller to perform a left-hand curve manoeuvre. On the contrary, if the sign reads ``right curve'', the steering trim is set to $+0.7$ so as to advise the high-level controller to perform a right-hand curve manoeuvre. In case no valid sign is present in close proximity, e.g. while the vehicle arrives at pedestrian crossing (detected using $\mathcal{X}$ and $\mathcal{M}$ information), the steering trim is reset to zero advising the high-level controller to proceed straight.

This loop is recursively iterated by further updating the states of vehicle(s) and traffic elements, and controlling them to successfully achieve the mission objective.

\subsection{Control}
\label{Sub-Section: Control}

\begin{figure}[htpb]
	\centering
	\subfigure[]{\includegraphics[width=0.49\textwidth]{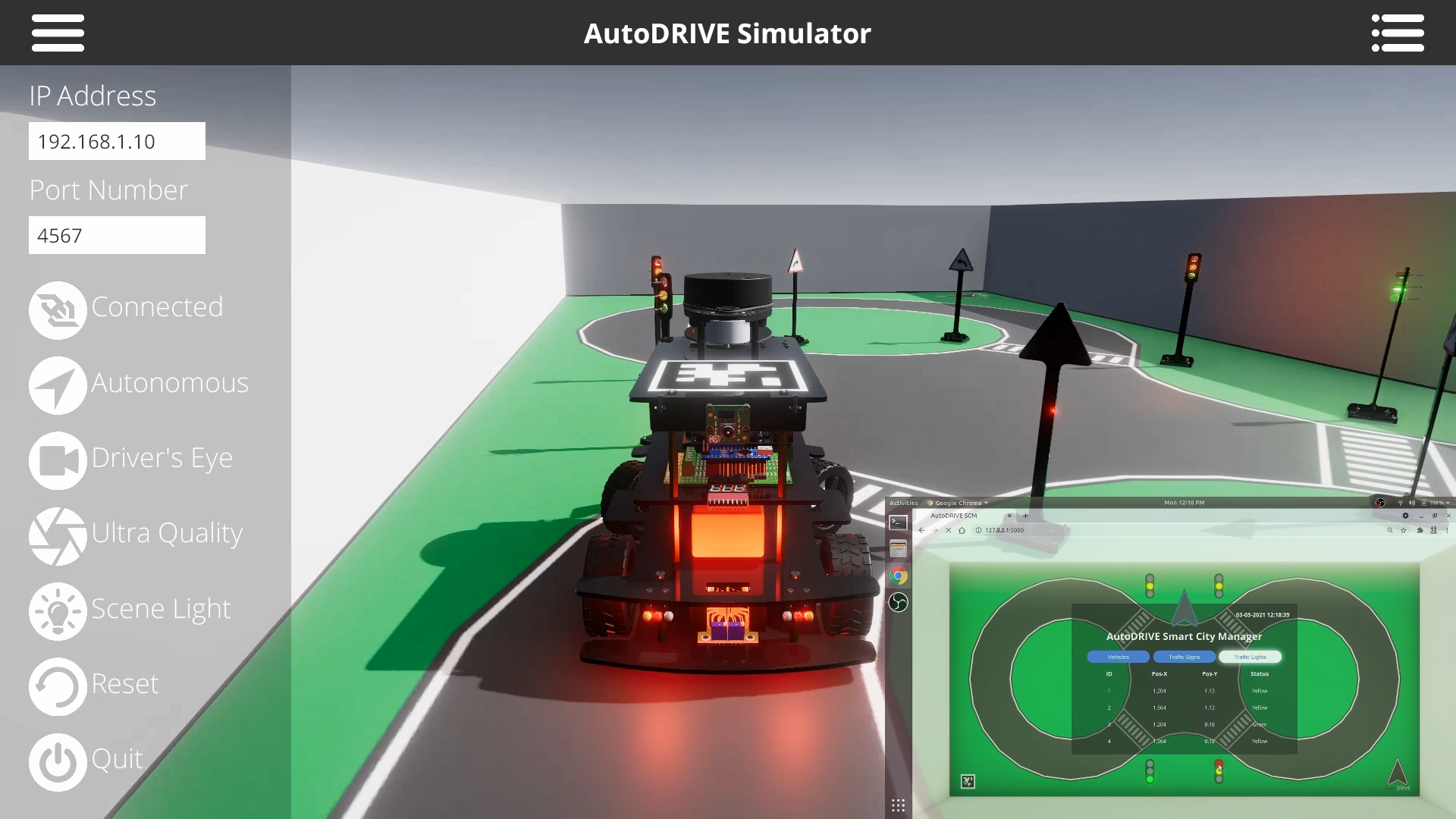}}
	\subfigure[]{\includegraphics[width=0.49\textwidth]{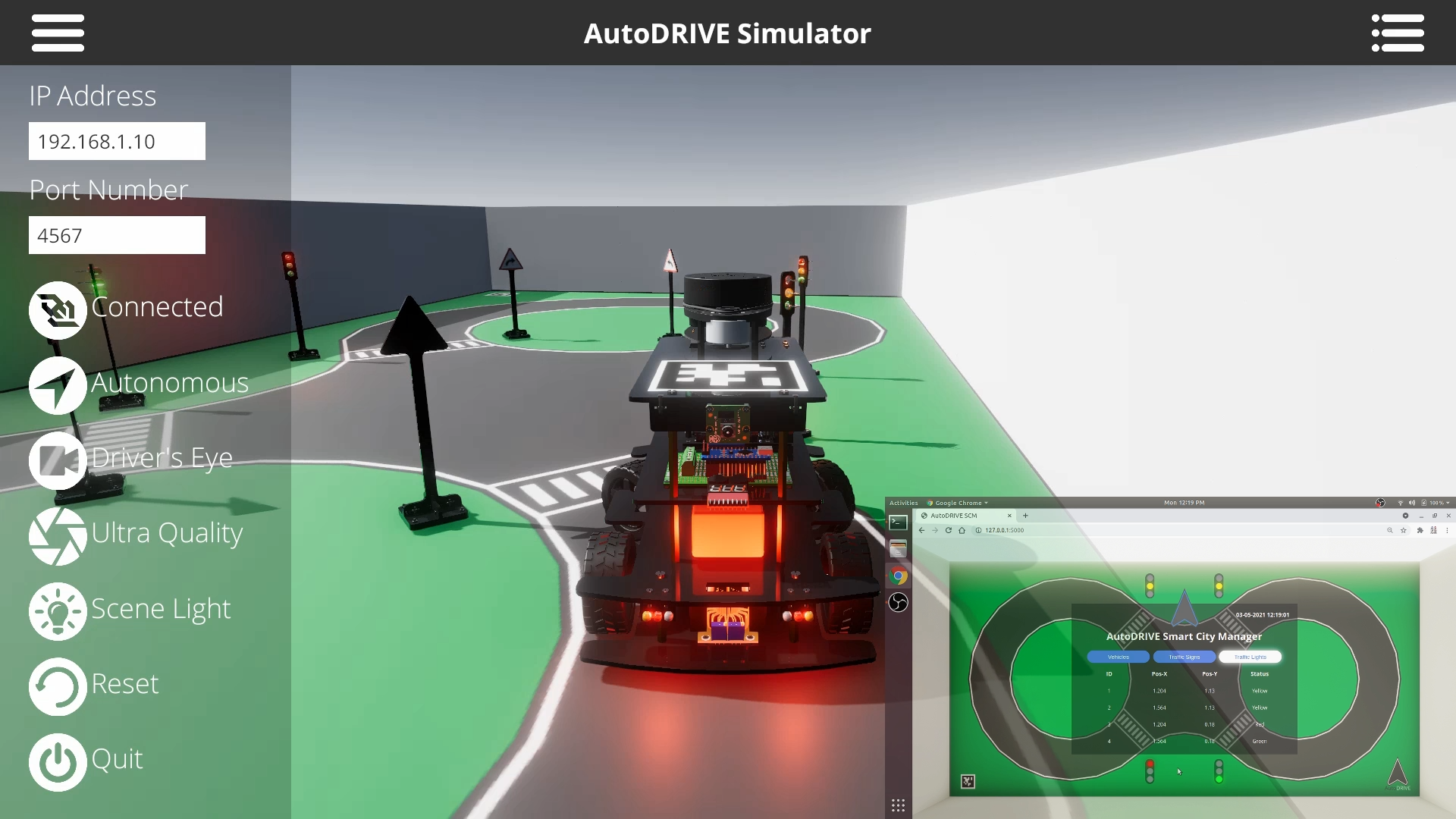}}
	\subfigure[]{\includegraphics[width=0.49\textwidth]{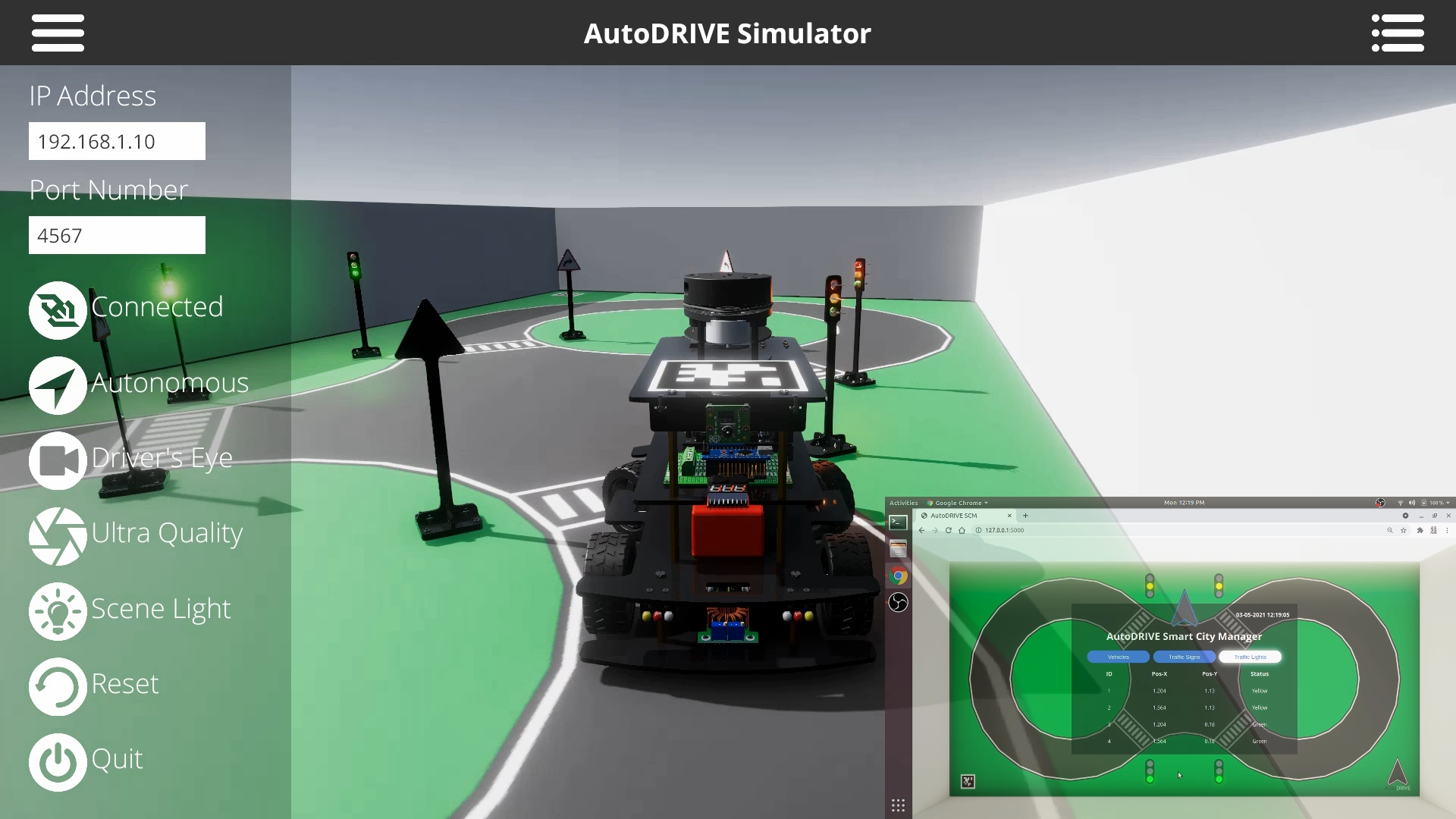}}
	\subfigure[]{\includegraphics[width=0.49\textwidth]{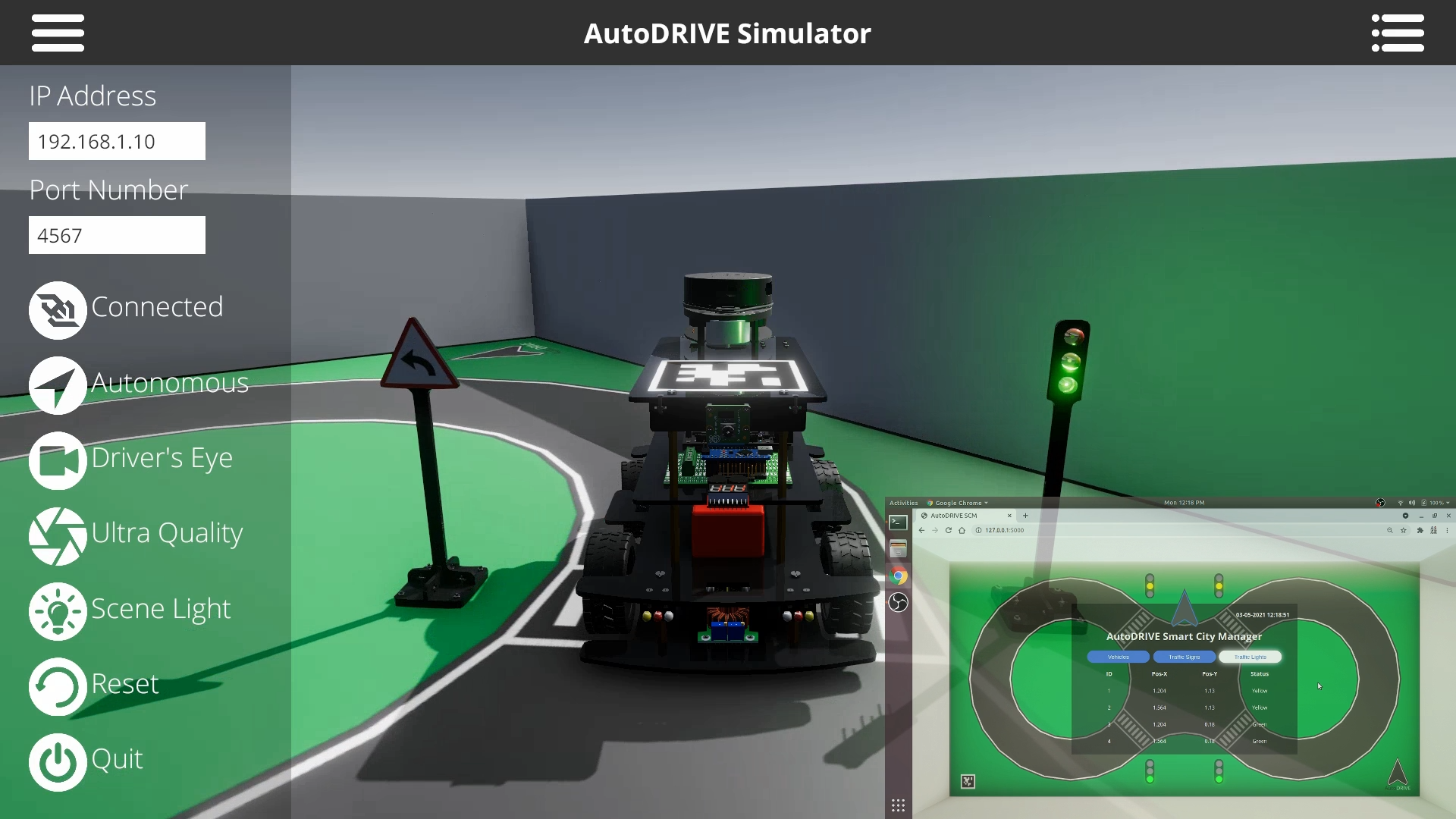}}
	\caption{Demonstration: (a) Vehicle Stops at Yellow Light, (b) Vehicle Stops at Red Light, (c) Vehicle Resumes upon Green Light, and (d) Vehicle Observes Left-Hand Curve Traffic Sign}
	\label{Figure: SCM Simulation}
\end{figure}

The control commands $\tau$ and $\delta$ are generated using Proximally Optimal Predictive (POP) Controller coupled with Adaptive Longitudinal Controller (ALC) \cite{POP2021}. These commands are communicated to the vehicle using AutoDRIVE Devkit's Python API, in order to make it drive autonomously. It is to be noted that the vehicle is solely controlled by the centralized SCM server, without any perception, planning or control module working aboard it. Figure \ref{Figure: SCM Simulation} illustrates various instances of the smart city management demonstration, wherein the vehicle autonomously navigates while observing all the traffic rules; the traffic lights are toggled manually.

\clearpage


\chapter{CONCLUSION AND FUTURE SCOPE}
\label{Chapter: Conclusion and Scope}

\section{Conclusion}

	We conclude the project by successfully accomplishing all the objectives that we had set at the commencement. This project has taught us a lot, especially pertaining to product development and scientific research. We are confident that we have undergone a significant skillset development throughout the course.
	
	Accomplishment of the primary objective of this project has yielded an integrated cyber-physical platform for autonomous driving research and education; we call this platform \textit{``AutoDRIVE''}. The prominent components of this platform include AutoDRIVE Testbed for hardware prototyping, AutoDRIVE Simulator for virtual prototyping and AutoDRIVE Devkit for convenient and flexible development of autonomy algorithms. All components of the platform are flexible enough to account for any additions included in the future. This is a product-level implementation, which can be used by everyone interested in the field, may it be to study, experiment and explore, teach, or carry out cutting-edge research. There are numerous applications of this platform and they are bound to keep increasing as new features are added.
	
	Merely accomplishing the platform development naturally raises the question \textit{``what is this platform really capable of?''}. The secondary objective of this project was therefore to demonstrate the key features and capabilities of this platform by exploiting it to implement various intelligent transportation algorithms. Four applications were shortlisted for this purpose, based on component of platform being exploited (testbed or simulator), nature of algorithms (model-based or data-driven, modular or end-to-end), technologies being demonstrated (SLAM, CV, DIL, DRL, V2V, V2I, SCM, etc.), sensors and actuators involved, and number of vehicles (single and multi-vehicle paradigms), to name a few. This secondary objective was also successfully accomplished despite the underlying complexities, which, we believe, will make the readers better appreciate this project.

\section{Future Scope}

	This project has only presented the first generation of AutoDRIVE, perhaps an introduction to the platform and its potential capabilities. As with any other product, AutoDRIVE is open to future upgrades in terms of both hardware and software. We say this because there are several future plans for this project, which only proves the fact that this project is going to be improved continuously. Expert engineers and researchers can contribute to the advancement of this platform by suggesting new, improved features.
	
	A potent addition to the platform would be that of heterogeneous vehicles and robotic pedestrians for setting up high-fidelity and challenging autonomous driving scenarios. Another very exciting aspect would be to extend support for large-scale vehicles and include them in the AutoDRIVE family as well.
	
	As mentioned earlier, this project was initiated with a vision of simplifying autonomous driving research and education by providing a powerful yet convenient platform, and we believe that an active community for this project would help achieve this overarching goal. Forming a community filled with individuals, both beginners and experts, collectively solving problems and learning from each other is thus one of the broader visions for this project; after all a great product is one with a vibrant community.
	
	Going further, keeping in mind that AutoDRIVE is a one-of-its-kind platform for autonomous driving research and education, we plan on promoting and/or commercializing this platform for academic and industrial use including (but not limited to) courses, workshops, competitions and research projects. This would, however, require development of crisp educational material and detailed documentation of various aspects of the platform, along with implementation of more and more autonomy algorithms targeted towards autonomous driving as well as smart city management. Consequently, these are some of the other potential areas to be focused on in the near future.

\clearpage


\begin{singlespace}
	\addcontentsline{toc}{chapter}{REFERENCES}
	\bibliography{References} 
\end{singlespace}

\clearpage


\appendix


\chapter{BILL OF MATERIALS}
\label{Appendix: A}

This appendix furbishes details pertaining to the hardware components (Section \ref{Section: Hardware BOM}) and software tools (Section \ref{Section: Software BOM}) utilized for realizing all the systems described in this project.

The net cost incurred for this project was \rupee 36115. It is to be noted that components contributing to non-recurring engineering (NRE) cost are not included in the bill of materials (BOM) for better clarity.

\section{Hardware Components}
\label{Section: Hardware BOM}

\begin{table}[htpb]
	\centering
	\caption{Vehicle Chassis BOM}
	\label{Table: Vehicle Chassis BOM}
	\resizebox{\textwidth}{!}{%
		\begin{tabular}{lllll}
			\hline
			\textbf{Component Name} & \textbf{Specifications}                               & \textbf{Rate (\rupee)} & \textbf{Qty.} & \textbf{Price (\rupee)} \\ \hline
			Acrylic cutouts         & 3 mm thick acrylic cutouts (laser cut)                & NA            & 21            & 790            \\
			Brass standoffs         & M3$\times$20 (M-F), M3$\times$40 (F-F, M-F) standoffs & NA            & NA            & 234            \\
			Nut-bolts               & M2, M2.5, M3, M4 nut-bolts                            & NA            & NA            & 400            \\
			Servo horn              & 25T horn for MG996R servo motor                       & 90            & 1             & 90             \\
			TT motor gearboxes      & 100 RPM TT motor gearbox                              & 65            & 2             & 130            \\
			Wheels                  & 65 mm (dia) rubber tyre wheels                        & 72            & 4             & 288            \\
			Strap belt              & 260 mm LiPo battery strap belt                        & 75            & 1             & 5              \\ \hline
		\end{tabular}%
	}
\end{table}

\begin{table}[htpb]
	\centering
	\caption{Vehicle Power Electronics BOM}
	\label{Table: Vehicle Power Electronics BOM}
	\resizebox{\textwidth}{!}{%
		\begin{tabular}{lllll}
			\hline
			\textbf{Component Name} & \textbf{Specifications}                      & \textbf{Rate (\rupee)} & \textbf{Qty.} & \textbf{Price (\rupee)} \\ \hline
			Battery pack            & Orange 5200 mAh 3S 40C/80C LiPo battery pack & 4149          & 1             & 4149           \\
			Battery charger         & 2S-3S balance LiPo battery charger           & 1180          & 1             & 1180           \\
			Voltage checker         & 1S-8S LiPo battery voltage checker           & 125           & 1             & 125            \\
			Master switch           & 12 mm ring light push button (blue)          & 170           & 1             & 170            \\
			Buck converter          & 10 A DC-DC CC CV buck converter              & 369           & 1             & 369            \\
			Motor driver            & 20 A motor driver with braking               & 590           & 1             & 590            \\
			Prototype board         & 6$\times$8 cm universal prototype board      & 51            & 1             & 51             \\
			PVC wires               & 3 m 24 AWG PVC wire                          & 49            & 2             & 98             \\
			Battery connector       & XT60 connector (male)                        & 35            & 1             & 35             \\
			DC jack                 & 2.1$\times$5.5 mm DC power jack (male)       & 25            & 1             & 25             \\
			Screw terminal          & 6 pin 5.08 mm pitch screw terminal block     & 33            & 1             & 33             \\
			Cooling fan             & 5 V DC 4020 cooling fan                      & 299           & 1             & 299            \\ \hline
		\end{tabular}%
	}
\end{table}

\begin{table}[htpb]
	\centering
	\caption{Vehicle Sensors BOM}
	\label{Table: Vehicle Sensors BOM}
	\resizebox{\textwidth}{!}{%
		\begin{tabular}{lllll}
			\hline
			\textbf{Component Name} & \textbf{Specifications}                          & \textbf{Rate (\rupee)} & \textbf{Qty.} & \textbf{Price (\rupee)} \\ \hline
			Throttle sensor         & Virtual sensory feedback (firmware)              & NA            & 1             & NA             \\
			Steering sensor         & Virtual sensory feedback (firmware)              & NA            & 1             & NA             \\
			Incremental encoders    & 16 PPR 2-channel hall-effect magnetic encoder    & NA            & 2             & NA             \\
			IPS                     & AprilTag marker                                  & 40            & 1             & 40             \\
			IMU                     & MPU9250 9-axis IMU                               & 428           & 1             & 428            \\
			LIDAR                   & RPLIDAR A1M8 (with mini-USB adapter)             & 7499          & 1             & 7499           \\
			Cameras                 & 8 MP Raspberry Pi Camera v2.1 (with CSI adapter) & 1999          & 2             & 3998           \\ \hline
		\end{tabular}%
	}
\end{table}

\begin{table}[htpb]
	\centering
	\caption{Vehicle Computational Resources BOM}
	\label{Table: Vehicle Computational Resources BOM}
	\resizebox{\textwidth}{!}{%
		\begin{tabular}{lllll}
			\hline
			\textbf{Component Name} & \textbf{Specifications}                & \textbf{Rate (\rupee)} & \textbf{Qty.} & \textbf{Price (\rupee)} \\ \hline
			Jetson Nano             & NVIDIA Jetson Nano Developer Kit - B01 & 9949          & 1             & 9949           \\
			Arduino Nano            & Arduino Nano Rev.3 development board   & 235           & 1             & 235            \\ \hline
		\end{tabular}%
	}
\end{table}

\begin{table}[htpb]
	\centering
	\caption{Vehicle Communication Systems BOM}
	\label{Table: Vehicle Communication Systems BOM}
	\resizebox{\textwidth}{!}{%
		\begin{tabular}{lllll}
			\hline
			\textbf{Component Name} & \textbf{Specifications}              & \textbf{Rate (\rupee)} & \textbf{Qty.} & \textbf{Price (\rupee)} \\ \hline
			LIDAR cable             & USB Type-A to Micro-USB Type-B cable & 59            & 1             & 59             \\
			Camera cables           & Camera serial interface (CSI) cable  & NA            & 1             & NA             \\
			MCU cable               & USB Type-A to Mini-USB Type-B cable  & 69            & 2             & 69             \\
			Wi-Fi adapter           & TP-Link USB Wi-Fi adapter            & 449           & 2             & 449            \\ \hline
		\end{tabular}%
	}
\end{table}

\begin{table}[htpb]
	\centering
	\caption{Vehicle Actuators BOM}
	\label{Table: Vehicle Actuators BOM}
	\resizebox{\textwidth}{!}{%
		\begin{tabular}{lllll}
			\hline
			\textbf{Component Name} & \textbf{Specifications}                    & \textbf{Rate (\rupee)} & \textbf{Qty.} & \textbf{Price (\rupee)} \\ \hline
			Drive actuators         & 6 V 120 RPM 120:1 TT motor (with encoders) & 739           & 2             & 1478           \\
			Steering actuator       & MG996R servo motor                         & 599           & 1             & 599            \\ \hline
		\end{tabular}%
	}
\end{table}

\begin{table}[htpb]
	\centering
	\caption{Vehicle Lights and Indicators BOM}
	\label{Table: Vehicle Lights and Indicators BOM}
	\resizebox{\textwidth}{!}{%
		\begin{tabular}{lllll}
			\hline
			\textbf{Component Name} & \textbf{Specifications}             & \textbf{Rate (\rupee)} & \textbf{Qty.} & \textbf{Price (\rupee)} \\ \hline
			LEDs                    & 5 mm (dia) LEDs (white, red, amber) & 3             & 12            & 36             \\
			DuPont cables           & M-F DuPont cables (10, 20 cm)       & 1.5           & 24            & 36             \\ \hline
		\end{tabular}%
	}
\end{table}

\begin{table}[htpb]
	\centering
	\caption{Infrastructural Setup BOM}
	\label{Table: Infrastructural Setup BOM}
	\resizebox{\textwidth}{!}{%
		\begin{tabular}{lllll}
			\hline
			\textbf{Component Name} & \textbf{Specifications}                       & \textbf{Rate (\rupee)} & \textbf{Qty.} & \textbf{Price (\rupee)} \\ \hline
			Map                     & Flex printing (10$\times$5 feet)              & 750           & 1             & 750            \\
			Obstructions            & 3 ply boxes (4$\times$4$\times$12 inches)     & 10            & 100           & 1000           \\
			AutoDRIVE Eye           & Overhead camera assembly and calibration grid & 359           & 1             & 359            \\ \hline
		\end{tabular}%
	}
\end{table}

\section{Software Tools}
\label{Section: Software BOM}

\begin{table}[htpb]
	\centering
	\caption{Software Tools BOM}
	\label{Table: Software Tools BOM}
	\resizebox{0.75\textwidth}{!}{%
		\begin{tabular}{ll}
			\hline
			\large{\textbf{Software Tool}}                    & \large{\textbf{License}} \\ \hline
			\multicolumn{2}{l}{\textbf{3D modelling}}                                    \\ \hline
			Dassault Syst\`{e}mes SolidWorks                  & Educational              \\
			Trimble SketchUp                                  & Educational              \\
			Autodesk 3DS Max                                  & Educational              \\ \hline
			\multicolumn{2}{l}{\textbf{Simulator development}}                           \\ \hline
			Unity                                             & Personal                 \\ \hline
			\multicolumn{2}{l}{\textbf{ADSS development}}                                \\ \hline
			Robot Operating System (ROS)                      & Open source              \\
			Python                                            & Open source              \\
			C++                                               & Open source              \\ \hline
			\multicolumn{2}{l}{\textbf{SCSS development}}                                \\ \hline
			Python                                            & Open source              \\
			HyperText Markup Language (HTML)                  & Open source              \\
			Cascading Style Sheets (CSS)                      & Open source              \\
			JavaScript (JS)                                   & Open source              \\
			Structured Query Language (SQL)                   & Open source              \\ \hline
			\multicolumn{2}{l}{\textbf{Documentation}}                                   \\ \hline
			Wondershare Filmora                               & Trial                    \\
			Microsoft Office                                  & Home and student edition \\
			MiKTeX                                            & Open source              \\ \hline
		\end{tabular}%
	}
\end{table}

\clearpage


\chapter{SUPPLEMENTAL MATERIAL}
\label{Appendix: B}

Considering the enormous span and complexity of this project, it is naturally not possible to express all the intricate details in the form of text, figures, tables and equations through the medium of this report. Consequently, this appendix points towards important supplemental material associated with this project.

\section{Source Files}
\label{Section: Source Files}
\begin{itemize}
	\sloppy
	\item \textbf{AutoDRIVE Parent Repository:} \url{https://github.com/Tinker-Twins/AutoDRIVE}
	\item \textbf{AutoDRIVE Testbed Branch:} \url{https://github.com/Tinker-Twins/AutoDRIVE/tree/AutoDRIVE-Testbed}
	\item \textbf{AutoDRIVE Simulator Branch:} \url{https://github.com/Tinker-Twins/AutoDRIVE/tree/AutoDRIVE-Simulator}
	\item \textbf{AutoDRIVE Devkit Branch:} \url{https://github.com/Tinker-Twins/AutoDRIVE/tree/AutoDRIVE-Devkit}
\end{itemize}

\section{Videos}
\label{Section: Videos}
\begin{itemize}
	\sloppy
	\item \textbf{AutoDRIVE Playlist [Updatable]:} \url{https://youtube.com/playlist?list=PLY45pkzWzH9_iRlOqmqvFdQPASwevTtbW}
	\item \textbf{AutoDRIVE Testbed:} \url{https://youtu.be/YFQzyfXV6Rw}
	\item \textbf{AutoDRIVE Simulator:} \url{https://youtu.be/i7R79jwnqlg}
	\item \textbf{Autonomous Parking:} \url{https://youtu.be/oBqIZZA0wkc}
	\item \textbf{Behavioural Cloning:} \url{https://youtu.be/rejpoogaXOE}
	\item \textbf{Intersection Traversal:} \url{https://youtu.be/AEFJbDzOpcM}
	\item \textbf{Smart City Management:} \url{https://youtu.be/fnxOpV1gFXo}
\end{itemize}

\clearpage

\end{document}